\documentclass[notitlepage]{report}
\usepackage[T1]{fontenc}
\usepackage[utf8]{inputenc}

\usepackage[a4paper, total={6in, 9in}]{geometry}

\date{Revised Feburary 21, 2025\footnote{Main changes: fixed some proof details for $k$-NN and fixed-radius NN regression and classification; note that this version does not have the publisher's formatting} \\[1em] Originally published on May 31, 2018 in \emph{Foundations and Trends in Machine Learning}}

\usepackage{etoolbox}
\makeatletter
\patchcmd{\@makechapterhead}{40\p@}{\chapheadbelowskip}{}{}
\makeatother
\newlength{\chapheadbelowskip}

\usepackage[authoryear]{natbib}
\usepackage{authblk}

\usepackage{mathpazo}
\usepackage[scaled=0.90]{helvet}
\usepackage[scaled=0.85]{beramono}

\usepackage[english]{babel}
\usepackage{float}
\usepackage{subfig}
\usepackage{amsmath}
\usepackage{amssymb}
\usepackage{amsthm}
\usepackage{enumitem}

\newcommand{\cf}{\emph{cf.}}
\newcommand{\ie}{\emph{i.e.}}
\newcommand{\eg}{\emph{e.g.}}

\newcommand{\row}{\mathrm{row}}
\newcommand{\col}{\mathrm{col}}
\newcommand{\proxy}{\mathrm{prx}}

\usepackage{mdframed}
\usepackage{nowidow}
\usepackage{hyperref}
\usepackage{microtype}

\newtheoremstyle{ftheorem}
  {-\topsep}
  {}
  {\normalfont}
  {}
  {\bfseries}
  {.}
  {.5em}
  {}
\newtheoremstyle{flemma}
  {-\topsep}
  {}
  {\normalfont}
  {}
  {\bfseries}
  {.}
  {.5em}
  {}
\newtheoremstyle{fcorollary}
  {-\topsep}
  {}
  {\normalfont}
  {}
  {\bfseries}
  {.}
  {.5em}
  {}
\newtheoremstyle{fproposition}
  {-\topsep}
  {}
  {\normalfont}
  {}
  {\bfseries}
  {.}
  {.5em}
  {}
\newtheoremstyle{fremark}
  {-\topsep}
  {}
  {\normalfont}
  {}
  {\bfseries}
  {.}
  {.5em}
  {}
\newtheoremstyle{fdefinition}
  {-\topsep}
  {}
  {\normalfont}
  {}
  {\bfseries}
  {.}
  {.5em}
  {}
\newtheoremstyle{fexample}
  {-\topsep}
  {}
  {\normalfont}
  {}
  {\bfseries}
  {.}
  {.5em}
  {}
\newtheoremstyle{fclaim}
  {-\topsep}
  {}
  {\normalfont}
  {}
  {\bfseries}
  {.}
  {.5em}
  {}
\newtheoremstyle{ffact}
  {-\topsep}
  {}
  {\normalfont}
  {}
  {\bfseries}
  {.}
  {.5em}
  {}
\newtheoremstyle{fassumption}
  {-\topsep}
  {}
  {\normalfont}
  {}
  {\bfseries}
  {.}
  {.5em}
  {}

\theoremstyle{ftheorem}
\theoremstyle{flemma}
\theoremstyle{fcorollary}
\theoremstyle{fproposition}
\theoremstyle{fremark}
\theoremstyle{fdefinition}
\theoremstyle{fexample}
\theoremstyle{fclaim}
\theoremstyle{ffact}
\theoremstyle{fassumption}

\newmdtheoremenv[backgroundcolor=gray!17,linewidth=6pt,linecolor=gray!20,innerleftmargin=5pt,innerrightmargin=5pt,innertopmargin=2.5pt,innerbottommargin=2.5pt]{ftheorem}{Theorem}[section]
\newmdtheoremenv[backgroundcolor=gray!17,linewidth=6pt,linecolor=gray!20,innerleftmargin=5pt,innerrightmargin=5pt,innertopmargin=2.5pt,innerbottommargin=2.5pt]{flemma}{Lemma}[section]
\newmdtheoremenv[backgroundcolor=gray!17,linewidth=6pt,linecolor=gray!20,innerleftmargin=5pt,innerrightmargin=5pt,innertopmargin=2.5pt,innerbottommargin=2.5pt,nobreak=true]{fcorollary}{Corollary}[section]
\newmdtheoremenv[backgroundcolor=gray!17,linewidth=6pt,linecolor=gray!20,innerleftmargin=5pt,innerrightmargin=5pt,innertopmargin=2.5pt,innerbottommargin=2.5pt]{fproposition}{Proposition}[section]
\newmdtheoremenv[backgroundcolor=gray!17,linewidth=6pt,linecolor=gray!20,innerleftmargin=5pt,innerrightmargin=5pt,innertopmargin=2.5pt,innerbottommargin=2.5pt]{fremark}{Remark}[section]
\newmdtheoremenv[backgroundcolor=gray!17,linewidth=6pt,linecolor=gray!20,innerleftmargin=5pt,innerrightmargin=5pt,innertopmargin=2.5pt,innerbottommargin=2.5pt]{fdefinition}{Definition}[section]
\newmdtheoremenv[backgroundcolor=gray!17,linewidth=6pt,linecolor=gray!20,innerleftmargin=5pt,innerrightmargin=5pt,innertopmargin=2.5pt,innerbottommargin=2.5pt]{fexample}{Example}[section]
\newmdtheoremenv[backgroundcolor=gray!17,linewidth=6pt,linecolor=gray!20,innerleftmargin=5pt,innerrightmargin=5pt,innertopmargin=2.5pt,innerbottommargin=2.5pt]{fclaim}{Claim}[section]
\newmdtheoremenv[backgroundcolor=gray!17,linewidth=6pt,linecolor=gray!20,innerleftmargin=5pt,innerrightmargin=5pt,innertopmargin=2.5pt,innerbottommargin=2.5pt]{ffact}{Fact}[section]
\newmdtheoremenv[backgroundcolor=gray!17,linewidth=6pt,linecolor=gray!20,innerleftmargin=5pt,innerrightmargin=5pt,innertopmargin=2.5pt,innerbottommargin=2.5pt,nobreak=true]{fassumption}{Assumption}[section]

\newcommand{\obsVar}{X}
\newcommand{\labelVar}{Y}

\newcommand{\permutationVar}{\xi}

\newcommand{\likableForUserTimeVar}{Z}

\usepackage{dsfont}
\newcommand{\E}{\mathbb{E}}
\renewcommand{\P}{\mathbb{P}}
\newcommand{\PP}{\mathcal{P}}

\newcommand{\bad}{\text{BAD}}
\newcommand{\BB}{\mathcal{B}}

\newcommand{\tc}{T_\text{learn}}

\newcommand{\ind}{\mathds{1}}
\newcommand{\HH}{\mathcal{H}}
\newcommand{\KK}{\mathcal{K}}
\newcommand{\NN}{\mathcal{N}}
\newcommand{\XX}{\mathcal{X}}

\newcommand{\assumpTechnical}{$\mathbf{A}_{\mathcal{X}, \rho, \mathbb{P}_X}^{\textsf{technical}}$}
\newcommand{\assumpDecay}{$\mathbf{A}_K^{\textsf{decay}(\tau)}$}
\newcommand{\assumpHolder}{$\mathbf{A}_{\eta}^{\textsf{H\"{o}lder}(C,\alpha)}$}
\newcommand{\assumpBesicovitch}{$\mathbf{A}_{\eta}^{\textsf{Besicovitch}}$}

\usepackage[algoruled,boxed,noend]{algorithm2e}

\usepackage{tikz}
\newcommand*\circled[1]{\tikz[baseline={([yshift=-1.5pt]char.base)}]{
            \node[shape=circle,draw,inner sep=0pt] (char) {#1};}}
\usepackage{graphicx}
\newcommand{\maxtimeshift}[1]{
              \scalebox{#1}{
                \circled{\ensuremath{\leftrightarrow}}}
              \,}
\newcommand{\timeadvance}[1]{
              \scalebox{#1}{
                \circled{\ensuremath{\leftarrow}}}
              \,}
\newcommand{\sepVar}{S}

\newenvironment{blockquote}{%
  \par%
  \medskip
  \leftskip=2.5em\rightskip=2.5em%
  \noindent\ignorespaces}{%
  \par\medskip}

\title{Explaining the Success of \\ Nearest Neighbor Methods in Prediction}

\author[1]{George H.~Chen}
\author[2]{Devavrat Shah}

\affil[1]{Carnegie Mellon University; georgechen@cmu.edu}
\affil[2]{Massachusetts Institute of Technology; devavrat@mit.edu}

\begin{document}

\sloppy

\maketitle

\begin{abstract}
Many modern methods for prediction leverage nearest neighbor search to find past training examples most similar to a test example, an idea that dates back in text to at least the 11th century and has stood the test of time. This monograph aims to explain the success of these methods, both in theory, for which we cover foundational nonasymptotic statistical guarantees on nearest-neighbor-based regression and classification, and in practice, for which we gather prominent methods for approximate nearest neighbor search that have been essential to scaling prediction systems reliant on nearest neighbor analysis to handle massive datasets.  Furthermore, we discuss connections to learning distances for use with nearest neighbor methods, including how random decision trees and ensemble methods learn nearest neighbor structure, as well as recent developments in crowdsourcing and graphons.

In terms of theory, our focus is on nonasymptotic statistical guarantees, which we state in the form of how many training data and what algorithm parameters ensure that a nearest neighbor prediction method achieves a user-specified error tolerance. We begin with the most general of such results for nearest neighbor and related kernel regression and classification in general metric spaces. In such settings in which we assume very little structure, what enables successful prediction is smoothness in the function being estimated for regression, and a low probability of landing near the decision boundary for classification.  In practice, these conditions could be difficult to verify empirically for a real dataset. We then cover recent theoretical guarantees on nearest neighbor prediction in the three case studies of time series forecasting, recommending products to people over time, and delineating human organs in medical images by looking at image patches.  In these case studies, clustering structure, which is easier to verify in data and more readily interpretable by practitioners, enables successful prediction.
\end{abstract}

\newpage

\tableofcontents

\newpage

\chapter{Introduction}
\label{chap:nn-survey-intro}

Things that appear similar are likely similar. For example, a baseball player's future performance can be predicted by comparing the player to other similar players~\citep{PECOTA}.  When forecasting election results for a U.S.~state, accounting for polling trends at similar states improves forecast accuracy~\citep{FiveThirtyEight}. In image editing, when removing an object from an image, one of the most successful ways to fill in the deleted pixels is by completing the missing pixels using image patches similar to the ones near the missing pixels~\citep{criminisi_2004}.  These are but a few examples of how finding similar instances or \textit{nearest neighbors} help produce predictions. Of course, this idea is hardly groundbreaking, with nearest neighbor classification already appearing as an explanation for visual object recognition in a medieval text \textit{Book of Optics} by acclaimed scholar Alhazen in the early 11th century.\footnote{A brief history of nearest neighbor classification and its appearance in Alhazen's \textit{Book of Optics} is given by~\citet{pellillo_investigates_alhazen}.  The exact completion date of \textit{Optics} is unknown.  \citet{al_khalili_2015} dates the work to be from years~1011 to~1021, coinciding with much of Alhazen's decade of imprisonment in Cairo, while \citet{smith_2001} claims a completion time between~1028 and~1038, closer to Alhazen's death circa~1040.} Despite their simplicity and age, nearest neighbor methods remain extremely popular,\footnote{Not only was the $k$-nearest neighbor method named as one of the top 10 algorithms in data mining \citep{wu_2008}, three of the other top 10 methods (AdaBoost, C4.5, and CART) have nearest neighbor interpretations. \nowidow} often used as a critical cog in a larger prediction machine. In fact, the machine can be biological, as there is now evidence that fruit flies' neural circuits execute approximate nearest neighbor in sensing odors as to come up with an appropriate behavioral response \citep{lsh_bio_evidence_2017}.

Although nearest neighbor classification dates back a millennium, analysis for when and why it works did not begin until far more recently, starting with a pair of unpublished technical reports by \citeauthor{fix_hodges_1951} (\citeyear{fix_hodges_1951,fix_hodges_1952}) on asymptotic convergence properties as well as a small dataset study, followed by the landmark result of \citet{cover_1967} that showed that $k$-nearest neighbors classification achieves an error rate that is at most twice the best error rate achievable. Decades later, Cover recollected how his paper with Hart came about:

\begin{blockquote}
Early in 1966 when I first began teaching at Stanford, a student, Peter Hart, walked into my office with an interesting problem. He said that Charles Cole and he were using a pattern classification scheme which, for lack of a better word, they described as the nearest neighbor procedure. This scheme assigned to an as yet unclassified observation the classification of the nearest neighbor. Were there any good theoretical properties of this procedure? \citep{cover_1982}
\end{blockquote}

\noindent
It would take some time for the term ``nearest neighbor'' to enter common parlance. However, the nearest neighbor procedure spread quickly across areas in computer science.  Not long after Cover and Hart's 1967 paper, Donald Knuth's third volume of \textit{The Art of Computer Programming} introduced nearest neighbor search as the \textit{post office problem} \citep{knuth_1973}, paving the beginnings of computational geometry.  In various coding theory contexts, \textit{maximum likelihood decoding} turns out to mean nearest neighbor classification \citep{coding_theory_book}.  Fast forwarding to present time, with the explosion in the availability of data in virtually all disciplines, architecting database systems that scale to this volume of data and that can efficiently find nearest neighbors has become a fundamental problem~\citep{db_book_2005}. Understanding when, why, and how well nearest neighbor prediction works now demands accounting for computational~costs.

\section{Explaining the Popularity of Nearest Neighbor Methods}

That nearest neighbor methods remain popular in practice largely has to do with their empirical success over the years. However, this explanation is perhaps overly simplistic. We highlight four aspects of nearest neighbor methods that we believe have been crucial to their continued popularity. First, the flexibility in choosing what ``near'' means in nearest neighbor prediction allows us to readily handle \textit{ad-hoc} distances, or to take advantage of existing representation and distance learning machinery such as deep neural networks or decision-tree-based ensemble learning approaches.  Second, the computational efficiency of numerous approximate nearest neighbor search procedures enables nearest neighbor prediction to scale to massive high-dimensional datasets common in modern applications. Third, nearest neighbor methods are nonparametric, making few modeling assumptions on data and instead letting the data more directly drive predictions.  Lastly, nearest neighbor methods are interpretable: they provide evidence for their predictions by exhibiting the nearest neighbors found.

\textit{Flexibility in defining similarity.}
Specifying what ``near'' means for a nearest neighbor method amounts to choosing a ``feature space'' in which data are represented (as ``feature vectors''), and a distance function to use within the feature space. For example, a common choice for the feature space and distance function are Euclidean space and Euclidean distance, respectively.  Of course, far more elaborate choices are possible and, in practice, often these are chosen in an \textit{ad-hoc} manner depending on the application.  For example, when working with time series, the distance function could involve a highly nonlinear time warp (to try to align two time series as well as possible before computing a simpler distance like Euclidean distance).  In choosing a ``good'' feature space (\ie, a good way to \textit{represent} data), features could be manually ``hand-engineered'' depending on the data modality (\eg, text, images, video, audio) or learned, for example, using deep neural networks (\eg,~\citealt[Chapter~15]{deep_learning_book}).  Meanwhile, sensor fusion is readily possible as features extracted from multiple sensors (\eg, different data modalities) can be concatenated to form a large feature vector.  Separately, the distance function itself can be learned, for example using Mahalanobis distance learning methods \citep{metric_learning} or Siamese networks \citep{siamese1,siamese2}.  In fact, decision trees and their use in ensemble methods such as random forests, AdaBoost, and gradient boosting can be shown to be weighted nearest neighbor methods that learn a distance function (we discuss this relationship toward the end of the monograph in Section~\ref{sec:decision-trees-ensemble-learning}, building on a previous observation made by \citealt{lin_random_forests}).  Thus, nearest neighbor methods actually mesh well with a number of existing representation and distance learning results.

\textit{Computational efficiency.}
Perhaps the aspect of nearest neighbor methods that has contributed the most to their popularity is their computational efficiency, which has enabled these methods to scale to massive datasets (``big data'').  Depending on the feature space and distance function chosen or learned by the practitioner, different fast approximate nearest neighbor search algorithms are available. These search algorithms, both for general high-dimensional feature spaces (\eg, \citealt{gionis_1999_original_lsh,datar_2004_good_intro_lsh,bawa_2005,CACM,ailon_2009,muja_flann_2009,boytsov_2013,dasgupta_2015,mathy_2015,andoni_2016}) and specialized to image patches (\eg, \citealt{patchmatch,ta_2014_optimized_patchmatch}), can rapidly determine which data points are close to each other while parallelizing across search queries. These methods often use locality-sensitive hashing~\citep{IndykMotwani}, which comes with a theoretical guarantee on approximation accuracy, or randomized trees (\eg, \citealt{bawa_2005,muja_flann_2009,dasgupta_2015,mathy_2015}), which quickly prune search spaces when the trees are sufficiently balanced. These randomized trees can even be efficiently constructed for streaming data using an arbitrary distance function~\citep{mathy_2015}.

\textit{Nonparametric.}
Roughly speaking, nearest neighbor methods being nonparametric means that they make very few assumptions on the underlying model for the data. This is a particularly attractive property since in a growing number of modern applications such as social networks, recommendation systems, healthcare decision support, and online education, we wish to analyze big data that we do not \textit{a priori} know the structure of.  A nonparametric approach sidesteps the question of explicitly positing or learning the structure underlying the data.  When we posit intricate structure for data, the structure may stray from reality or otherwise not account for the full palette of possibilities in what the data look like. When we learn structure, the computational overhead and amount of data needed may dwarf what is sufficient for tackling the prediction task at hand. Instead of positing or learning structure, nonparametric methods let the data more directly drive predictions.  However, being nonparametric doesn't mean that nearest neighbor methods have no parameters. We still have to choose a feature space and distance, and a poor choice of these could make prediction impossible.

\textit{Interpretability.}
Nearest neighbor methods naturally provide evidence for their decisions by exhibiting the nearest neighbors found in the data.  A practitioner can use the nearest neighbors found to diagnose whether the feature space and distance function chosen are adequate for the application of interest. For example, if on validation data, a nearest neighbor method is making incorrect predictions, we can look at the nearest neighbors of each validation data point to see why they tend to have incorrect labels. This often gives clues to the practitioner as to how to choose a better feature space or distance function. Alternatively, if the nearest neighbor method is producing accurate predictions, the nearest neighbors found tell us which training data points are driving the prediction for any particular validation or test point.  This interpretability is vital in applications such as healthcare that demand a high burden of proof before letting software influence potentially costly decisions that affect people's well-being.

\section{Nearest Neighbor Methods in Theory}

Although nearest neighbor methods for prediction have remained popular, only recently has a thorough theory been developed to characterize the error rate of these methods in fairly general settings. Roughly a millennium after the appearance of nearest neighbor classification in Alhazen's \textit{Book of Optics}, \citet{chaudhuri_dasgupta_2014} established arguably the most general performance guarantee to date, stating how many training data and how to choose the number of nearest neighbors to achieve a user-specified error tolerance, when the data reside in a metric space.\footnote{Within the same year a few months after Chaudhuri and Dasgupta's paper appeared on arXiv, Gadat \textit{et al.}~posted on arXiv the most general theory to date for nearest neighbor classification in the more restricted setting of finite dimensional spaces, which was finally published two years later in \textit{Annals of Statistics} \citep{gadat_2016}.} This flavor of result is ``nonasymptotic'' in that it can be phrased in a way that gives the probability of misclassification for \textit{any} training data set size; we do not need an asymptotic assumption that the amount of training data goes to infinity.  Chaudhuri and Dasgupta's result subsumes or matches classical results by \citet{fix_hodges_1951}, \citet{devroye_1994}, \citet{cerou_guyader_2006}, and \citet{audibert_2007}, while providing a perhaps more intuitive explanation for when nearest neighbor classification works, accounting for the metric used and the distribution from which the data are sampled.  Moreover, we show that their analysis can be translated to the regression setting, yielding theoretical guarantees that nearly match the best of existing regression results.

However, while the general theory for both nearest neighbor classification and regression has largely been fleshed out, a major criticism is that they do not give ``user-friendly'' error bounds that can readily be computed from available training data~\citep{kontorovich_weiss_2015}.  For example, Chaudhuri and Dasgupta's result for nearest neighbor classification depends on the probability of landing near the true decision boundary.  Meanwhile, nearest neighbor regression results depend on smoothness of the function being estimated, usually in terms of Lipschitz or more generally H\"{o}lder continuity parameters.  In practice, these quantities are typically difficult to estimate for a real dataset.  Unfortunately, this also makes the theory hard to use by practitioners, who often are interested in understanding how many training data they should acquire to achieve a certain level of accuracy, preferably in terms of interpretable application-specific structure rather than, for instance, H\"{o}lder continuity parameters (\eg, in healthcare, each training data point could correspond to a patient, and the cost of conducting a study may scale with the number of patients; being able to relate how many patients should be in the study in terms of specific disease or treatment quantities that clinicians can estimate would be beneficial).

Rather than providing results in as general a setting as possible, a recent trilogy of papers instead shows how clustering structure that is present in data enables enables nearest neighbor prediction to succeed at time series forecasting, recommending products to people, and finding human organs in medical images~\citep{georgehc_time_series_nips,georgehc_collaborative_filtering_nips, georgehc_image_segmentation_miccai}.  These papers establish nonasymptotic theoretical guarantees that trade off between the training data size and the prediction accuracy as a function of the number of clusters and the amount of noise present.  The theory here depends on the clusters being separated enough so that noise is unlikely to cause too many points to appear to come from a wrong cluster. Prediction succeeds when, for a test point, its nearest neighbors found in the training data are predominantly from the same cluster as the test point. That these theoretical guarantees are about clustering is appealing because clusters can often be estimated from data and interpreted by practitioners.

\section{The Scope of This Monograph}

This monograph aims to explain the success of nearest neighbor methods in prediction, covering both theory and practice. Our exposition intentionally strives to be as accessible as possible to theoreticians and practitioners alike.  As the number of prediction methods that rely on nearest neighbor analysis and the amount of literature studying these methods are both enormous, our coverage is carefully curated and inexhaustive.

On the theoretical side, our goal is to provide some of the most general \textit{nonasymptotic} results and give a flavor of the proof techniques involved. All key theoretical guarantees we cover are stated in the form of how many training data and what algorithm parameters ensure that a nearest neighbor prediction method achieves a user-specified error tolerance.

On the more practical side, we cover some examples of how nearest neighbor methods are used as part of a larger prediction system (recommending products to people in the problem of \textit{online collaborative filtering}, and delineating where a human organ is in medical images in the problem of \textit{patch-based image segmentation}). We also discuss a variety of approximate nearest neighbor search and related methods which have been pivotal to scaling nearest neighbor prediction to massive, even ever-growing datasets.

Our coverage is as follows, transitioning from theory to practice as we progress through the monograph:

\begin{itemize}

\item \textbf{Background (Chapter~\ref{chap:background}).}
We anchor notation and terminology used throughout the monograph.  Specifically we define the basic prediction tasks of classification and regression, and then present the three basic algorithms of $k$-nearest neighbor, fixed-radius near neighbor, and kernel regression. These regression methods can in turn be translated into classification methods.

\item \textbf{Regression (Chapter~\ref{chap:regression}).}
We present theoretical guarantees for $k$-nearest neighbor, fixed-radius near neighbor, and kernel regression where the data reside in a metric space. The proofs borrow heavily from the work by~\citet{chaudhuri_dasgupta_2014} with some influence from the work by \citet{gadat_2016}. These authors actually focus on classification, but proof ideas translate over to the regression setting.

\item \textbf{Classification (Chapter~\ref{chap:classification}).}
We show how the theoretical guarantees for regression can readily be converted to ones for classification.  However, it turns out that we can obtain classification guarantees using weaker conditions. We explain how \citet{chaudhuri_dasgupta_2014} achieve this for $k$-nearest neighbor classification and how the basic idea readily generalizes to fixed-radius near neighbor and kernel classification.

\item \textbf{Prediction Guarantees in Three Contemporary Applications (Chapter~\ref{chap:case-studies}).}
We present theoretical guarantees for nearest neighbor prediction in time series forecasting~\citep{georgehc_time_series_nips}, online collaborative filtering~\citep{georgehc_collaborative_filtering_nips}, and patch-based image segmentation~\citep{georgehc_image_segmentation_miccai}.  Despite these applications seeming disparate and unrelated, the theoretical guarantees for them turn out to be quite similar. In all three, clustering structure enables successful prediction.  We remark that the independence assumptions on training data and where clustering structure appears are both application-specific.

\item \textbf{Computation (Chapter~\ref{chap:computation}).}
We provide an overview of efficient data structures for exact and approximate nearest neighbor search that are used in practice. We focus on motifs these methods share rather than expounding on theoretical guarantees, which many of these methods lack.  Our starting point is the classical {\em k-d tree} data structure for exact nearest neighbor search \citep{kdtree}, which works extremely well for low-dimensional data but suffers from the ``curse of dimensionality'' due to an exponential dependence on dimension when executing a search query.  To handle exact high-dimensional nearest neighbor search, more recent approaches such as the \textit{cover tree} data structure exploit the idea that high-dimensional data often have low-dimensional structure \citep{beygelzimer_2006}. As such approaches can still be computationally expensive in practice, we turn toward approximate nearest neighbor search.  We describe {\em locality-sensitive hashing}~(LSH) \citep{IndykMotwani}, which forms the foundation of many approximate nearest neighbor search methods that come with theoretical guarantees.  We also discuss empirically successful approaches with partial or no theoretical guarantees: {\em random projection} or {\em partition trees} inspired by k-d trees, and the recently proposed {\em boundary forest}. 

\item \textbf{Adaptive Nearest Neighbors and Far Away Neighbors (Chapter~\ref{chap:discussion}).}
We end with remarks on distance learning with a focus on decision trees and various ensemble methods that turn out to be nearest neighbor methods, and then turn toward a new class of nearest neighbor methods that in some sense can take advantage of far away neighbors.

\end{itemize}
For readers seeking a more ``theory-forward'' exposition albeit without coverage of Chaudhuri and Dasgupta's classification and related regression results, there are recent books by \citet{devroye_2013} (on classification) and~\citet{biau_2015} (on nearest neighbor methods with sparse discussion on kernel regression and classification), and earlier books by \citet{gyorfi_book} (on nonparametric regression) and \mbox{\citet{tsybakov_2009}} (on nonparametric estimation). Unlike the other books mentioned, Tsybakov's regression coverage emphasizes fixed design (corresponding to the training feature vectors having a deterministic structure, such as being evenly spaced in a feature space), which is beyond the scope of this monograph.  As for theory on nearest neighbor search algorithms, there is a survey by \citet{clarkson_2006} that goes into substantially more detail than our overview in Chapter~\ref{chap:computation}. However, this survey does not cover a number of very recent advances in approximate nearest neighbor search that we discuss.

\chapter{Background}
\label{chap:background}

\begingroup
\allowdisplaybreaks

Given a data point (also called a feature vector) $X \in \mathcal{X}$ (for which we call $\mathcal{X}$ the feature space), we aim to predict its label $Y \in \mathbb{R}$.\footnote{For simplicity, we constrain the label to be real-valued in this monograph. In general, the space of labels could be higher dimensional and need not be numeric. Terminological remarks: (1) Some researchers only use the word ``label'' for when $Y$ takes on a discrete set of values, whereas we allow for labels to be continuous. (2) In statistics, the feature vector $X$ consists of entries called \textit{independent} (also, \textit{predictor} or \textit{experimental}) variables and its label~$Y$ is called the \textit{dependent} (also, \textit{response} or \textit{outcome}) variable.} We treat feature vector~$X$ as being sampled from a marginal distribution (which we also call the feature distribution)~$\mathbb{P}_X$. After sampling feature vector~$X$, its label $Y$ is sampled from a conditional distribution $\mathbb{P}_{Y \mid X}$.\footnote{An appropriate probability space is assumed to exist. Precise technical conditions are given in Section~\ref{sec:technicalities}.} To assist us in making predictions, we assume we have access to $n$ training data pairs $(X_1, Y_1), (X_2, Y_2), \dots, (X_n, Y_n) \in \mathcal{X} \times \mathbb{R}$ that are sampled i.i.d.~in the same manner as for $X$ and~$Y$, so the $i$-th training data point $X_i$ has label $Y_i$. We now describe the two basic prediction problems of regression and classification.

\section{Regression}
\label{sec:regression-setup}

Given an observed value of feature vector $X=x$, the label $Y$ for this feature vector is typically going to be noisy. Denoting the conditional expectation by
\[
\eta(x) \triangleq \mathbb{E}[Y \mid X=x],
\]
then the label $Y$ is going to be $\eta(x)$ with some noise added. As we generally cannot predict noise, the expected label $\eta(x)$ may seem like a reasonable guess for label $Y$. This motivates the problem of regression, which is to estimate (or ``learn'') the conditional expectation function $\eta : \mathcal{X} \rightarrow \mathbb{R}$ given training data $(X_1, Y_1), \dots, (X_n, Y_n) \in \mathcal{X} \times \mathbb{R}$. Once we have an estimate $\widehat{\eta}$ for $\eta$, then for any feature vector $x \in \mathcal{X}$ that we observe, we can predict its label to be $\widehat{\eta}(x)$. The function $\eta$ is also called the \textit{regression function}.

Let's make precise in what way regression function $\eta$ is a good guess for labels. Suppose we come up with a prediction function ${f:\mathcal{X}\rightarrow\mathbb{R}}$ where when we observe feature vector $X=x$, we predict its label~$Y$ to be $f(x)$. Then one way to measure the error is the expected squared difference between the true label $Y$ and the predicted label $f(x)$ conditioned on our observation $X=x$, \ie, $\mathbb{E}[(Y - f(x))^2 \mid X = x]$. It turns out that the lowest possible error is achieved by choosing $f = \eta$, so in this minimal expected squared error sense, regression function $\eta$ is optimal for predicting labels.

\begin{fproposition}
\label{prop:regression-function-minimizes-expected-square-error}
Let $f:\mathcal{X}\rightarrow\mathbb{R}$ be any measurable function that given a feature vector outputs a label. For observed feature vector $X=x$, the expected squared error for prediction $f(x)$ satisfies
\[
\mathbb{E}[(Y-f(x))^{2}\mid X=x]
\ge\mathbb{E}[(Y-\eta(x))^{2}\mid X=x].
\]
\end{fproposition}
In the inequality above, the right-hand side says what the lowest possible expected squared error is (in fact, since $\eta(x)=\mathbb{E}[Y\mid X=x]$, the right-hand side is the noise variance).  Equality is attained by choosing $f = \eta$.  Of course, we don't know what the regression function $\eta$ is. The regression problem is about estimating $\eta$ using training data.

Put another way, for a test point $X=x$, recall that its label $Y$ is generated as $\eta(x)$ plus unknown noise. In an expected squared prediction error sense, the best possible estimate for $Y$ is $\eta(x)$. However, even if we knew $\eta(x)$ perfectly, $Y$ is going to be off by precisely the unknown noise amount. The average squared value of this noise amount is the noise variance, which is the right-hand side of the above inequality.

\begin{proof}[Proof of Proposition~\ref{prop:regression-function-minimizes-expected-square-error}]
The proposition follows from a little bit of algebra and linearity of expectation:
\begin{align*}
&\mathbb{E}[(Y-f(x))^2\mid X=x] \\
&\!=\mathbb{E}\big[\big((Y-\eta(x))+(\eta(x)-f(x))\big)^2\,\big|\,X=x\big] \\
&\!=\mathbb{E}\big[
    (Y-\eta(x))^2
    + 2(Y-\eta(x))(\eta(x)-f(x))
    + (\eta(x)-f(x))^2\,\big|\,X=x\big] \\
&\!=\mathbb{E}[(Y-\eta(x))^2\mid X=x] \\
&\quad
  + 2(\underbrace{\mathbb{E}[Y\mid X=x]}_{\eta(x)}-\eta(x))(\eta(x)-f(x))
  + \underbrace{\mathbb{E}[(\eta(x)-f(x))^2\mid X=x]}_{\ge 0} \\
&\!\ge \mathbb{E}[(Y-\eta(x))^2\mid X=x].\tag*{\qedhere}
\end{align*}
\end{proof}

\subsection*{The Bias-Variance Decomposition and Tradeoff}

Now that we have motivated that regression function $\eta$ is worth estimating as it would minimize the expected squared prediction error, we present a way to analyze whether an estimate~$\widehat{\eta}$ for~$\eta$ is any good.  Building off our exposition above, the predictor $f$ will now be written as $\widehat{\eta}$ to emphasize that we are estimating $\eta$. Moreover, we now will use the fact that $\widehat{\eta}$ is estimated using $n$ random training data. Different realizations of training data result in a different function~$\widehat{\eta}$ being learned.  In particular, we can decompose the expected squared prediction error for any estimate~$\widehat{\eta}$ of~$\eta$ into three terms:
\begin{itemize}

\item the bias of the estimator~$\widehat{\eta}$, which measures error in terms of assumptions made by estimator~$\widehat{\eta}$ (for example, if $\eta$ is highly nonlinear, and $\widehat{\eta}$ is constrained to be a linear fit, then the bias will tend to be large)

\item the variance of the estimator~$\widehat{\eta}$, which measures how much the estimator changes when we change the training data (for example, if $\widehat{\eta}$ fits a degree-$n$ polynomial to the $n$ training data, then it will have high variance---changing any of the $n$ training data points could have a dramatic effect on the shape of $\widehat{\eta}$; contrast this to if $\widehat{\eta}$ fit a straight line instead)

\item ``irreducible'' noise variance (as explained by Proposition~\ref{prop:regression-function-minimizes-expected-square-error}, even if we knew the regression function~$\eta$ perfectly, the test point $X=x$ has label $Y$ that deviates from $\eta(x)$ by an unknown noise amount; the variance of this noise serves as a lower bound for the expected squared prediction error of any estimator)

\end{itemize}
We formally state the bias-variance decomposition shortly, which also makes it clear what the above three quantities are. Importantly, a good estimator~$\widehat{\eta}$ should simultaneously achieve low bias and low variance. Finding conditions that ensure these two quantities to be small is a recurring theme in both analyzing and designing prediction methods. This concept plays a critical role when we analyze nearest neighbor regression and classification methods in Chapters~\ref{chap:regression} and~\ref{chap:classification}.

We now state the bias-variance decomposition of the expected squared prediction error of estimator~$\widehat{\eta}$ at test point $X=x$. The derivation is short, so we provide it along the way.  We use ``$\mathbb{E}$'' to denote expectation over the training data $(X_1,Y_1),\dots,(X_n,Y_n)$ and test label~$Y$ (and not over $X$ since we condition on $X=x$), ``$\mathbb{E}_{Y\mid X=x}$'' to denote the conditional expectation of $Y$ given $X=x$, and ``$\mathbb{E}_n$'' to denote the expectation over the $n$ training data. Then the decomposition is
\begin{align*}
& \mathbb{E}[(Y-\widehat{\eta}(x))^{2}\mid X=x]\\
& =\mathbb{E}_{Y\mid X=x}\big[\mathbb{E}_n\big[(Y-\widehat{\eta}(x))^{2}\big]\big]\\
& =\mathbb{E}_{Y\mid X=x}\big[\mathbb{E}_n\big[\big((Y-\eta(x))-(\widehat{\eta}(x)-\eta(x))\big)^{2}\big]\big]\\
& \overset{\!\!(i)\!\!}{=}\mathbb{E}_{Y\mid X=x}\big[(Y-\eta(x))^{2}\big]+\mathbb{E}_n\big[(\widehat{\eta}(x)-\eta(x))^{2}\big]\\
& =\mathbb{E}_{Y\mid X=x}\big[(Y-\eta(x))^{2}\big]+\mathbb{E}_n\big[\big((\widehat{\eta}(x)-\mathbb{E}_n[\widehat{\eta}(x)])-(\eta(x)-\mathbb{E}_n[\widehat{\eta}(x)])\big)^{2}\big]\\
& \overset{\!\!(ii)\!\!}{=}\underbrace{\mathbb{E}_{Y\mid X=x}\big[(Y-\eta(x))^{2}\big]}_{\text{noise variance}}+\big(\!\!\!\!\underbrace{\mathbb{E}_n[\widehat{\eta}(x)]-\eta(x)}_{\text{bias of estimator }\widehat{\eta}\text{ at }x}\!\!\!\!\big)^{2}+\underbrace{\mathbb{E}_n\big[(\widehat{\eta}(x)-\mathbb{E}_n[\widehat{\eta}(x)])^{2}\big]}_{\text{variance of estimator }\widehat{\eta}\text{ at }x},
\end{align*}
where steps $(i)$ and $(ii)$ each involves expanding a squared quantity and noting that there is a term with expectation 0 that disappears (the same idea as described in the second-to-last line of the proof of Proposition~\ref{prop:regression-function-minimizes-expected-square-error}).  As a remark, the noise variance is precisely the right-hand side of the inequality in Proposition~\ref{prop:regression-function-minimizes-expected-square-error} using slightly different notation.

A key observation is that more complex estimators tend to have lower bias but higher variance. For example, consider when $\widehat{\eta}$ is a polynomial fit. If we use a higher-degree polynomial, we can more accurately predict $\eta$, but changes in the training data could have a more dramatic effect on the shape of~$\widehat{\eta}$. This phenomenon is referred to as the \textit{bias-variance tradeoff}.  As we discuss shortly when we introduce the different nearest neighbor regression and classification algorithms, choosing different algorithm parameters corresponds to changing an algorithm's bias and variance.

\section{Classification}
\label{sec:classification-setup}

In this monograph, we focus on binary classification, which has the same setup as regression for how the data are generated except that each label $Y$ takes on one of two values $\{0, 1\}$ (if the labels took on some two other values, which need not be numeric, we could always map them to 0 and 1).  Then the problem of classification is to use the training data $(X_1, Y_1), \dots, (X_n, Y_n) \in \mathcal{X} \times \{0, 1\}$ to estimate a function $\widehat{Y}$ that, given a feature vector $x \in \mathcal{X}$, outputs a predicted label $\widehat{Y}(x) \in \{0, 1\}$ (\ie, the function $\widehat{Y}$ ``classifies'' $x$ to be either of class 0 or of class 1).\footnote{Unlike in the regression problem, for the classification problem our goal is not to estimate the conditional expected label $\eta(x) = \mathbb{E}[Y \mid X=x]$. The reason for this is that often in classification (especially in the non-binary case when we have many labels), the underlying labels are not numeric and have no notion of an average, and thus talking about an average label lacks a meaningful interpretation, even if we relabel the non-numeric labels to be numeric.

Also, as a terminological remark, classification is some times also called \textit{discrimination} (\eg, \citealt{fix_hodges_1951}) or \textit{pattern recognition} (\eg, \citealt{devroye_2013}), although the latter is often used more generally than just for classification.} The function $\widehat{Y}$ is called a \textit{classifier}.

The best classifier in terms of minimizing probability of error turns out to be simple to describe. Given observed feature vector $X=x$, we compare the conditional probabilities ${\mathbb{P}(Y=0 \mid X=x)}$ and ${\mathbb{P}(Y=1 \mid X=x)}$; if the former is higher than the latter, we predict $X$ to have label~0 and otherwise we predict $X$ to have label~1. (If the probabilities are equal, we can actually break the tie arbitrarily, but for simplicity, let's say we break the tie in favor of label 1.) This prediction rule is called the \textit{Bayes classifier}:
\begin{align}
\widehat{Y}_{\text{Bayes}}(x)
&= \underset{y\in\{0,1\}}
            {\text{argmax}}~\mathbb{P}(Y=y\mid X=x)
   \label{eq:binary-Bayes-classifier-general-form} \\
&= \begin{cases}
   1 & \text{if }\mathbb{P}(Y=1 \mid X=x)\ge\mathbb{P}(Y=0 \mid X=x), \\
   0 & \text{otherwise}.
   \end{cases}
   \label{eq:binary-bayes-classifier-intermediate-form}
\end{align}
As an aside, if distributions $\mathbb{P}_Y$ and $\mathbb{P}_{Y \mid X}$ are both known, then from a Bayesian perspective, we can think of $\mathbb{P}_Y$ as a prior distribution for label~$Y$, $\mathbb{P}_{Y\mid X}$ as a likelihood model, and $\eta(x)=\mathbb{P}(Y = 1 \mid X=x)$ as the posterior probability of label~1 given observation $X=x$. Then the Bayes classifier yields a \textit{maximum a posteriori} (MAP) estimate of the label~$Y$.

The Bayes classifier minimizes the probability of misclassification.

\begin{fproposition}
\label{prop:bayes-classifier-minimizes-prob-error}
Let $f:\mathcal{X}\rightarrow\{0,1\}$ be any classifier that given a feature vector outputs a label. For observed feature vector $X=x$, the probability that the predicted label $f(x)$ is erroneous satisfies
\begin{equation*}
\mathbb{P}(Y \ne f(x) \mid X = x)
\ge \mathbb{P}(Y \ne \widehat{Y}_{\text{Bayes}}(x) \mid X = x).
\end{equation*}
In fact, the misclassification probability of classifier $f$ at the point $x$ exceeds that of the Bayes classifier $\widehat{Y}_{\text{Bayes}}$ by precisely
\begin{align*}
&\mathbb{P}(Y \ne f(x) \mid X = x)
- \mathbb{P}(Y \ne \widehat{Y}_{\text{Bayes}}(x) \mid X = x) \nonumber \\
&\quad= |2\eta(x) - 1| \ind\{f(x) \ne \widehat{Y}_{\text{Bayes}}(x)\},
\end{align*}
where $\ind\{\cdot\}$ is the indicator function ($\ind\{\mathcal{A}\}=1$ if statement $\mathcal{A}$ holds, and $\ind\{\mathcal{A}\}=0$ otherwise).
\end{fproposition}

Thus, to achieve the lowest possible misclassification probability, we want to choose $f = \widehat{Y}_{\text{Bayes}}$.  Unfortunately, similar to how in regression we do not actually know the regression function~$\eta$ (which minimizes expected squared error) and instead have to estimate it from training data, in classification we do not know the Bayes classifier $\widehat{Y}_{\text{Bayes}}$ (which minimizes probability of misclassification) and have to estimate it from training data.

In fact, in the binary classification setup with labels 0 and 1, the regression function~$\eta$ and the Bayes classifier $\widehat{Y}_{\text{Bayes}}$ are intricately related. To see this, note that in this binary classification setting, the regression function~$\eta$ tells us the conditional probability that an observed feature vector $x$ has label~1:
\[
\eta(x) = \mathbb{E}[Y \mid X = x] = \mathbb{P}(Y=1 \mid X=x).
\]
Thus, we have ${\mathbb{P}(Y=0 \mid X=x)} = 1 - \mathbb{P}(Y=1 \mid X=x) = 1 - \eta(x)$, and with some rearranging of terms, one can show that equation~\eqref{eq:binary-bayes-classifier-intermediate-form} can be written
\begin{equation}
\widehat{Y}_{\text{Bayes}}(x)
=
\begin{cases}
1 & \text{if }\eta(x) \ge \frac12, \\
0 & \text{otherwise}.
\end{cases}
\label{eq:bayes-classifier}
\end{equation}
In light of equation~\eqref{eq:bayes-classifier}, the Bayes classifier effectively knows where the decision boundary $\{x : \eta(x) = 1/2\}$ is! In practice, without assuming a known distribution governing $X$ and~$Y$, we know neither the regression function~$\eta$ nor where the decision boundary is. A natural approach is to first solve the regression problem by obtaining an estimate $\widehat{\eta}$ for $\eta$ and then plug in $\widehat{\eta}$ in place of $\eta$ in the above equation for the Bayes classifier. The resulting classifier is called a \textit{plug-in classifier}. Of course, if we are able to solve the regression problem accurately (\ie, we produce an estimate $\widehat{\eta}$ that very close to $\eta$), then we would also have a good classifier. The nearest neighbor and related kernel classifiers we encounter are all plug-in classifiers.\footnote{In this monograph, we focus on nearest neighbor and related kernel classifiers rather than general plug-in classifiers. We mention that there are known theoretical results for general plug-in classifiers, such as the work by \citet{audibert_2007}.}

We remark that regression can be thought of as a harder problem than classification. In regression, we care about estimating~$\eta$ accurately, whereas in classification, we only need to estimate the regions in which~$\eta$ exceeds a threshold $1/2$ accurately. For example, near a particular point~$x$, the function~$\eta$ could fluctuate wildly yet always be above $1/2$. In this scenario, estimating $\eta(x)$ from the labels of nearby training data points is challenging, but figuring out that $\eta(x) \ge 1/2$ is easy (since very likely the majority of nearby training points will have label at least $1/2$).  We make this intuition rigorous in Chapter~\ref{chap:classification}.

\begin{proof}[Proof of Proposition~\ref{prop:bayes-classifier-minimizes-prob-error}]
For any classifier $f:\mathcal{X}\rightarrow\{0,1\}$, we have
\begin{align}
& \mathbb{P}(Y\ne f(x)\mid X=x) \nonumber \\
& =1-\mathbb{P}(Y=f(x)\mid X=x) \nonumber \\
& =1-[\ind\{f(x)=1\}\mathbb{P}(Y=1\mid X=x)+\ind\{f(x)=0\}\mathbb{P}(Y=0\mid X=x)] \nonumber \\
& =1-[\ind\{f(x)=1\}\eta(x)+\ind\{f(x)=0\}(1-\eta(x))] \nonumber \\
& =1-\ind\{f(x)=1\}\eta(x)-\ind\{f(x)=0\}(1-\eta(x))].
\label{eq:pf-bayes-classifier-minimizes-prob-error-helper1}
\end{align}
The above equation holds even if we replace $f$ with $\widehat{Y}_{\text{Bayes}}$:
\begin{align}
&\mathbb{P}(Y\ne\widehat{Y}_{\text{Bayes}}(x)\mid X=x) \nonumber \\
&\quad =1-\ind\{\widehat{Y}_{\text{Bayes}}(x)=1\}\eta(x)-\ind\{\widehat{Y}_{\text{Bayes}}(x)=0\}(1-\eta(x)).
\label{eq:pf-bayes-classifier-minimizes-prob-error-helper2}
\end{align}
Hence, the difference in misclassification rate at point $x$ between classifiers $f$ and $\widehat{Y}_{\text{Bayes}}$ is given by the difference of equations~\eqref{eq:pf-bayes-classifier-minimizes-prob-error-helper1} and~\eqref{eq:pf-bayes-classifier-minimizes-prob-error-helper2}:
\begin{align}
& \mathbb{P}(Y\ne f(x)\mid X=x)-\mathbb{P}(Y\ne\widehat{Y}_{\text{Bayes}}(x)\mid X=x) \nonumber \\
& \quad=[1-\ind\{f(x)=1\}\eta(x)-\ind\{f(x)=0\}(1-\eta(x))] \nonumber \\
& \quad\quad-[1-\ind\{\widehat{Y}_{\text{Bayes}}(x)=1\}\eta(x)-\ind\{\widehat{Y}_{\text{Bayes}}(x)=0\}(1-\eta(x))] \nonumber \\
& \quad=\eta(x)(\ind\{\widehat{Y}_{\text{Bayes}}(x)=1\}-\ind\{f(x)=1\}) \nonumber \\
& \quad\quad+(1-\eta(x))(\ind\{\widehat{Y}_{\text{Bayes}}(x)=0\}-\ind\{f(x)=0\}).
\label{eq:pf-bayes-classifier-minimizes-prob-error-main-diff}
\end{align}
We can evaluate equation \eqref{eq:pf-bayes-classifier-minimizes-prob-error-main-diff} exhaustively for all possible cases:
\begin{itemize}

\item Case 1: $f(x)=\widehat{Y}_{\text{Bayes}}(x)$. Then equation~\eqref{eq:pf-bayes-classifier-minimizes-prob-error-main-diff} equals 0.

\item Case 2: $f(x)=0$ and $\widehat{Y}_{\text{Bayes}}(x)=1$. Then equation \eqref{eq:pf-bayes-classifier-minimizes-prob-error-main-diff} equals $\eta(x)+(1-\eta(x))(-1)=2\eta(x)-1$. From equation~\eqref{eq:bayes-classifier}, recall that $\widehat{Y}_{\text{Bayes}}(x)=1$ means that $\eta(x)\ge\frac{1}{2}$. In particular, $2\eta(x)-1$ is nonnegative.

\item Case 3: $f(x)=1$ and $\widehat{Y}_{\text{Bayes}}(x)=0$. Then equation \eqref{eq:pf-bayes-classifier-minimizes-prob-error-main-diff} equals $\eta(x)(-1)+(1-\eta(x))=1-2\eta(x)$. From equation~\eqref{eq:bayes-classifier}, $\widehat{Y}_{\text{Bayes}}(x)=0$ means that $\eta(x)<\frac{1}{2}$. Thus, $1-2\eta(x)$ is nonnegative.

\end{itemize}
Cases 2 and 3 can be combined to say that when $f(x)\ne\widehat{Y}_{\text{Bayes}}(x)$, equation~\eqref{eq:pf-bayes-classifier-minimizes-prob-error-main-diff} equals $|2\eta(x) - 1|$. Thus, we can write equation \eqref{eq:pf-bayes-classifier-minimizes-prob-error-main-diff} as
\begin{align}
& \mathbb{P}(Y\ne f(x)\mid X=x)-\mathbb{P}(Y\ne\widehat{Y}_{\text{Bayes}}(x)\mid X=x) \nonumber \\
& \quad=\begin{cases}
          0 & \text{if }\widehat{Y}_{\text{Bayes}}(x)=f(x),\\
          |2\eta(x)-1| & \text{if }\widehat{Y}_{\text{Bayes}}(x)\ne f(x),
        \end{cases} \nonumber \\
& \quad=|2\eta(x)-1|\ind\{\widehat{Y}_{\text{Bayes}}(x)\ne f(x)\}.
\label{eq:pf-bayes-classifier-minimizes-prob-error-main-diff2}
\end{align}
Since the right-hand side of equation \eqref{eq:pf-bayes-classifier-minimizes-prob-error-main-diff2} is nonnegative,
\[
\mathbb{P}(Y\ne f(x)\mid X=x)\ge\mathbb{P}(Y\ne\widehat{Y}_{\text{Bayes}}(x)\mid X=x). \qedhere
\]
\end{proof}

\section{Nearest Neighbor and Kernel Regression}
\label{sec:basic-methods}

We present regression methods before their classification variants since we will just plug in estimated regression functions into equation~\eqref{eq:bayes-classifier} to yield classifiers. In all methods we consider, we assume we have decided on a distance function $\rho:\mathcal{X}\times\mathcal{X}\rightarrow\mathbb{R}_+$ to use for measuring how far apart any two feature vectors are. In practice, distance function $\rho$ is often chosen by a practitioner in an \textit{ad-hoc} manner and it could, for instance, not be a proper metric (as is the case when we discuss time series forecasting in Chapter~\ref{chap:case-studies}).  We now present three nearest neighbor and related kernel methods.

\subsection{\texorpdfstring{$k$}{k}-Nearest Neighbor Regression}
\label{sec:k-NN-regression}

In $k$-nearest neighbor ($k$-NN) regression,\footnote{The $k$-NN method was first formalized in a technical report by \citet{fix_hodges_1951} for classification. We remark that Alhazen's \textit{Book of Optics} only describes \mbox{1-NN} classification rather than the general $k$-NN case; moreover, Alhazen's description has a ``reject'' option in case the test feature vector is too different from all the training feature vectors \citep{pellillo_investigates_alhazen}.  Moving beyond classification, \citet{watson_1964} mentions that Fix and Hodges' $k$-NN method can easily be used for regression as well.} to determine what the label should be for a point $x \in \mathcal{X}$, we find its $k$ nearest neighbors in the training data and average their labels.  To make this precise, we use the following notation.  Let $(X_{(i)}(x),Y_{(i)}(x))$ denote the $i$-th closest training data point among the training data $(X_{1},Y_{1}),\dots,(X_{n},Y_{n})$. Thus, the distance of each training data point to $x$ satisfies
\[
\rho(x,X_{(1)}(x))\le\rho(x,X_{(2)}(x))\le\cdots\le\rho(x,X_{(n)}(x)).
\]
The theory presented later requires that ties happen with probability~0.  In case the marginal distribution $\mathbb{P}_{X}$ allows for there to be ties with positive probability (\eg, the feature space $\mathcal{X}$ is discrete), we break ties randomly in the following manner.  For each training data point $X_i$, we sample a random variable $Z_i \sim \text{Uniform}[0,1]$ that can be thought of as a priority.  Then for the test feature vector $X$ that we are predicting the label for, whenever multiple training feature vectors are the same distance to $X$, we break ties by favoring lower values of $Z_i$.\footnote{This tie breaking mechanism was referred to as \textit{tie breaking by randomization} by \citet{devroye_1994}, who---on the way of establishing strong consistency of $k$-NN regression---compared three ways of breaking ties in nearest neighbor search.}

Then the $k$-NN estimate for the regression function $\eta$ at point $x \in \mathcal{X}$ is the average label of the $k$ nearest neighbors:
\[
\widehat{\eta}_{k\text{-NN}}(x)=\frac{1}{k}\sum_{i=1}^{k}Y_{(i)}(x),
\]
where we pre-specify the number of nearest neighbors $k \in \{1, 2, \dots, n\}$.

Intuitively, since the labels are noisy, to get a good estimate for the expected label $\eta(x)$, we should average over more labels by choosing a larger $k$. However, choosing a larger $k$ corresponds to using training points farther away from~$x$, which could have labels more dissimilar to $\eta(x)$. Thus, we want to choose $k$ to be in some sweet spot that is neither too small nor too large. This could be thought of in terms of the bias-variance tradeoff.  By choosing a smaller $k$, we obtain a more flexible regressor with lower bias but higher variance. For example, with $k=1$ and assuming no ties in distances between any pair of training points, then $k$-NN regression would have 100\% prediction accuracy on training data labels. This of course is likely an overfit to training data. On the opposite extreme, with $k=n$, then regardless of what the test point $x$ is that we wish to predict $\eta(x)$ for, the prediction is always just the average of all training labels. This corresponds to a regressor with high bias and low variance.  Choosing $k$ to be neither too small nor too large ensures both bias and variance are sufficiently low.

\subsection{Fixed-Radius Near Neighbor Regression}

Instead of looking at $k$ nearest neighbors, we can look at all nearest neighbors up to a threshold distance $h>0$ away. Finding these neighbors within radius $h$ is referred to as the \textit{fixed-radius near neighbor problem}.\footnote{Fixed-radius near neighbor search was used as part of a molecular visualization system by \citet{levinthal_1966}. An early survey on fixed-radius near neighbor methods was provided by \citet{bentley_1975}.} We refer to regression using these neighbors as \textit{fixed-radius NN regression}, which we denote as $\widehat{\eta}_{\text{NN}(h)}$:
\[
\widehat{\eta}_{\text{NN}(h)}(x)
= \begin{cases}
  {\displaystyle
   \frac{\sum_{i=1}^{n}\ind\{\rho(x, X_i) \le h\}Y_i}
        {\sum_{i=1}^{n}\ind\{\rho(x, X_i) \le h\}}}
   & \text{if }{\textstyle \sum_{i=1}^{n}\ind\{\rho(x, X_i) \le h\} > 0}, \\
   0 & \text{otherwise},
  \end{cases}
\]
where, as a reminder, $\ind\{\cdot\}$ is the indicator function that is~1 when its argument is true and~0 otherwise.

As with $k$-NN regression where we want to choose $k$ that is neither too small nor too large, with fixed-radius NN regression, we want to choose threshold distance $h$ that is neither too small nor too large. Smaller $h$ yields a regressor with lower bias and higher variance.  Note that fixed-radius NN regression has an issue that $k$-NN regression does not have to deal with: it could be that there are no training data found within distance $h$ of test point $x$. We could avoid this situation if the number of training data $n$ is sufficiently large so as to ensure some training data points land within distance $h$ of $x$.

\subsection{Kernel Regression}

Lastly, we have the case of kernel regression, for which we focus on the Nadaraya-Watson method proposed separately but within the same year by \citet{nadaraya_1964} and \citet{watson_1964}.  Here we have a kernel function $K:\mathbb{R}_{+}\rightarrow[0,1]$ that takes as input a ``normalized'' distance and outputs a similarity score between 0 and 1, where the distance is ``normalized'' because we divide the input distance by a bandwidth parameter $h > 0$. Specifically, the kernel regression estimate for $\eta(x)$ is denoted
\[
\widehat{\eta}_{K}(x;h)
= \begin{cases}
  {\displaystyle
   \frac{\sum_{i=1}^{n}K\big(\frac{\rho(x,X_i)}{h}\big)Y_i}
        {\sum_{i=1}^{n}K\big(\frac{\rho(x,X_i)}{h}\big)}}
  & \text{if }
    {\textstyle \sum_{i=1}^{n}K\big(\frac{\rho(x,X_{i})}{h}\big) > 0}, \\
  0 & \text{otherwise}.
  \end{cases}
\]
For example, fixed-radius NN regression corresponds to choosing $K(s)=\ind\{s\le1\}$, where bandwidth $h$ is the threshold distance. As a second example, a Gaussian kernel corresponds to $K(s)=\exp(-\frac{1}{2}s^{2})$ and the bandwidth $h$ corresponds to the standard deviation parameter of the Gaussian. A natural assumption that we make is that $K$ is monotonically decreasing, which is to say that two points being farther away implies them having smaller (or possibly the same) similarity score.

The kernel function $K$ affects how much each training point $X_i$ contributes to the final prediction through a weighted average. Ideally, for a test point $x$, training points with labels dissimilar to the expected label $\eta(x)$ should contribute little or no weight to the weighted average. Rather than down-weighting training points with labels dissimilar to the expected label $\eta(x)$, kernel function $K$ down-weights training points that are far from $x$. The hope is that training points close to $x$ do indeed have labels close to $\eta(x)$, and training points farther from $x$ have labels that can deviate more from $\eta(x)$. Thus, in addition to choosing bandwidth $h$ to be not too small and not too large as in fixed-radius NN regression, we now have the additional flexibility of choosing kernel function $K$, which should decay fast enough to reduce the impact of training points far from $x$ that may have labels dissimilar to $\eta(x)$.

\section{Nearest Neighbor and Kernel Classification}
\label{sec:basic-classifiers}

By plugging in each of the regression function estimates $\widehat{\eta}_{k\text{-NN}}$, $\widehat{\eta}_{\text{NN}(h)}$, and $\widehat{\eta}_{K}(\cdot; h)$ in place of $\eta$ in the optimal Bayes classifier equation~\eqref{eq:bayes-classifier}, we obtain the following plug-in classifiers corresponding to $k$-NN, fixed-radius NN, and kernel classifiers:
\begin{align*}
\widehat{Y}_{k\text{-NN}}(x)
&
=
\begin{cases}
1 & \text{if }{\displaystyle \frac{1}{k}\sum_{i=1}^{k}Y_{(i)}(x) \ge \frac12}, \\
0 & \text{otherwise},
\end{cases} \\
\widehat{Y}_{\text{NN}(h)}(x)
&
=
\begin{cases}
1 & \text{if }{\displaystyle \frac{\sum_{i=1}^{n}\ind\{\rho(x, X_i) \le h\}Y_i}
                                  {\sum_{i=1}^{n}\ind\{\rho(x, X_i) \le h\}} \ge \frac12} \\
& \quad\text{and denominator}\;{\textstyle \sum_{i=1}^{n}\ind\{\rho(x, X_i) \le h\} > 0}, \\
0 & \text{otherwise},
\end{cases} \\
\widehat{Y}_{K}(x; h)
&
=
\begin{cases}
1 & \text{if }{\displaystyle \frac{\sum_{i=1}^{n}K(\frac{\rho(x,X_{i})}{h})Y_{i}}
                                  {\sum_{i=1}^{n}K(\frac{\rho(x,X_{i})}{h})} \ge \frac12} \\
& \quad\text{and denominator}\;{\textstyle \sum_{i=1}^{n}K(\frac{\rho(x,X_{i})}{h}) > 0}, \\
0 & \text{otherwise}.
\end{cases}
\end{align*}
We remark that these classifiers have an election analogy and some times are referred to as running \textit{weighted majority voting}.  Specifically, each training point $X_i$ casts a vote for label $Y_i$. However, the election is ``biased'' in the sense that the voters do not get equal weight. For example in kernel classification, the $i$-th training point's vote gets weighted by a factor $K(\frac{\rho(x,X_i)}h)$. Thus the sum of all weighted votes for label 1 is $V_1(x; K, h) \triangleq \sum_{i=1}^n K(\frac{\rho(x,X_i)}h) \ind\{Y_i = 1\}$, and similarly the sum of all weighted votes for label 0 is $V_0(x; K, h) \triangleq \sum_{i=1}^n K(\frac{\rho(x,X_i)}h) \ind\{Y_i = 0\}$.  By rearranging terms, one can verify that the kernel classifier chooses the label with the majority of weighted votes, where we arbitrarily break the tie here in favor of label~1:
\[
\widehat{Y}_{K}(x; h)
=
\begin{cases}
1 & \text{if }V_1(x; K, h) \ge V_0(x; K, h), \\
0 & \text{otherwise}.
\end{cases}
\]
In $k$-NN classification, only the $k$ nearest training data to $x$ have equal positive weight and the rest of the points get weight 0.  In fixed-radius NN classification, every training point $X_i$ within distance $h$ of $x$ has equal positive weight, and all other training data points get weight~0.

\endgroup

\chapter{Theory on Regression}
\label{chap:regression}

\graphicspath{{figures/regression/}}

In this chapter, we present nonasymptotic theoretical guarantees for \mbox{$k$-NN}, fixed-radius NN, and kernel regression. Our analysis is heavily based on binary classification work by \citet{chaudhuri_dasgupta_2014}; their proof techniques easily transfer over to the regression setting. Specifically in analyzing expected regression error, we also borrow ideas from~\citet{audibert_2007} and~\citet{gadat_2016}.

We layer our exposition, beginning with a high-level overview of the nonasymptotic results covered in Section~\ref{sec:regression-theory-overview}, emphasizing key ideas in the analysis and giving a sense of how the results relate across the three regression methods. We then address technicalities in Section~\ref{sec:technicalities} needed to arrive at the precise statements of theoretical guarantees given for each of the methods in Sections~\ref{sec:k-NN-regression-theory}, \ref{sec:h-near-regression-theory}, and \ref{sec:kernel-regression-theory}.  As the final layer of detail, proofs are unveiled in Section~\ref{sec:regression-proofs}.

We end the chapter in Section~\ref{sec:automatic-k-h} with commentary on automatically selecting the number of nearest neighbors $k$ for $k$-NN regression, and the bandwidth~$h$ for fixed-radius NN and kernel regression. We point out existing theoretical guarantees for choosing $k$ and $h$ via cross-validation and data splitting. We also discuss adaptive approaches that choose different~$k$ and~$h$ depending on the test feature vector~$x$.

\section{Overview of Results}
\label{sec:regression-theory-overview}

In this section, we highlight key ideas and intuition in the analysis, glossing over the technical assumptions needed (specified in the next section). Our overview here largely motivates what the technicalities are. Because key ideas in the analysis recur across methods, we spend the most amount of time on $k$-NN regression results before drawing out similarities and differences to arrive at fixed-radius NN and kernel regression results.

\subsection{\texorpdfstring{$k$}{k}-NN Regression}
\label{sub:k-nn-regression-bad-events}

\begin{figure}
\centering
\includegraphics[scale=.8]{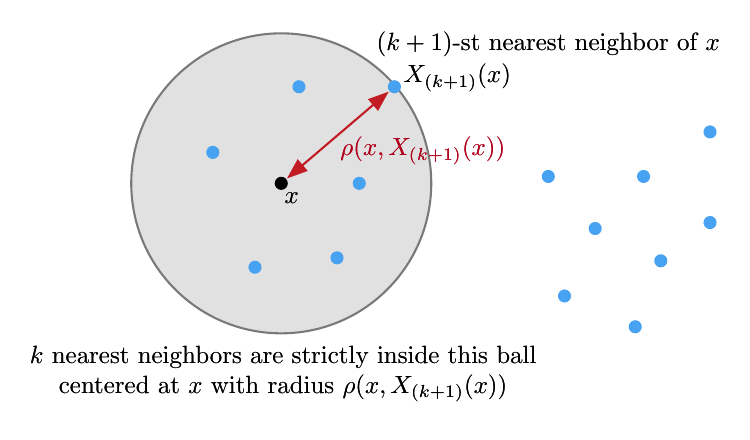}
\vspace{-1em}
\caption{Illustration to help with $k$-NN analysis ($k=6$ in this example): the blue points are training data, the test feature vector that we are making a prediction for is the black point $x$ and its $k$-th nearest neighbor $X_{(k+1)}(x)$ is on the boundary of the shaded ball, which has radius $\rho(x, X_{(k+1)}(x))$.
\label{fig:k-NN-helper}}
\end{figure}

For an observed feature vector $X=x$, recall that $k$-NN regression estimates expected label~$\eta(x)$ with the following estimate $\widehat{\eta}_{k\text{-NN}}(x)$:
\[
\widehat{\eta}_{k\text{-NN}}(x)=\frac{1}{k}\sum_{i=1}^{k}Y_{(i)}(x),
\]
where $(X_{(i)}(x),Y_{(i)}(x))$ denotes the $i$-th closest training data pair to point $x$ among the training data $(X_{1},Y_{1}),\dots,(X_{n},Y_{n})$, where ties happen with probability 0 (as ensured by random tie breaking as described in Section~\ref{sec:k-NN-regression}).  In particular, the $k$-NN estimate for $\eta(x)$ is the average label of the $k$ nearest neighbors.  We denote the distance function being used by $\rho$. For example, $\rho(x, X_{(k+1)}(x))$ refers to the distance between feature vector $x$ and its \mbox{$(k+1)$-st} nearest neighbor $X_{(k+1)}(x)$.  Pictorially, with blue points denoting training data and the single black point denoting the ``test'' feature vector $x$ that we aim to estimate the regression function's value at, then the $k$-NN estimate is the average label of the blue points strictly inside the shaded ball of Figure~\ref{fig:k-NN-helper}.

Since labels of training data are noisy, to get a good estimate for the expected label $\eta(x) = \mathbb{E}[Y \mid X=x]$, we should average over more values, \ie, we should choose $k$ to be large.  However, using a larger $k$ means that the average label is based on training points that could be farther away from~$x$ (in Figure~\ref{fig:k-NN-helper}, larger $k$ corresponds to growing the radius of the shaded ball each time to include the next closest training data point along the ball's boundary).  Training data farther from $x$ could have labels far from the expected label at point $x$ (\ie, $\eta(x)$), leading to a bad estimate.  We formalize these two bad events next and provide some intuition for how we can prevent each bad event from happening. We also relate them to the bias-variance tradeoff.

Let $\varepsilon>0$ and $\delta\in(0,1)$ be user-specified error and probability tolerances, respectively, where we shall guarantee that the regression estimate $\widehat{\eta}_{k\text{-NN}}(x)$ for $\eta(x)$ has error $|\widehat{\eta}_{k\text{-NN}}(x) - \eta(x)| \le \varepsilon$ with probability at least $1-\delta$.  Let $\mathbb{E}_{n|\widetilde{X}}$ denote the expectation over the training data $(X_1, Y_1), \dots, (X_n, Y_n)$ after conditioning on the \mbox{$(k+1)$-st} nearest neighbor $\widetilde{X} := X_{(k+1)}(x)$ (note that we are treating observed feature vector $x$ as fixed here). Why we condition on $\widetilde{X}$ is slightly technical and is discussed later in Section~\ref{sec:proof-k-NN-regression-rate-of-convergence}. Then we have:
\begin{itemize}

\item \textbf{Bad event \#1.}
We are not averaging over enough labels in estimating $\eta(x)$ to combat label noise. In other words, the number of nearest neighbors $k$ (which is the number of labels we average over) is too small.  As a result, the \mbox{$k$-NN} regression estimate $\widehat{\eta}_{k\text{-NN}}(x)$ is not close to the expectation $\mathbb{E}_{n|\widetilde{X}}[\widehat{\eta}_{k\text{-NN}}(x)]$, which we formalize as $| \widehat{\eta}_{k\text{-NN}}(x) - \mathbb{E}_{n|\widetilde{X}}[\widehat{\eta}_{k\text{-NN}}(x)] | > \frac{\varepsilon}2 $.

Naturally, to prevent this bad event from happening, we shall ask that $k$ be sufficiently large so that ${| \widehat{\eta}_{k\text{-NN}}(x) - \mathbb{E}_{n|\widetilde{X}}[\widehat{\eta}_{k\text{-NN}}(x)] | \le \frac{\varepsilon}2}$, which---in terms of the bias-variance tradeoff---controls the variance of \mbox{$k$-NN} regression to be small. To see this, note that we are ensuring ${(\widehat{\eta}_{k\text{-NN}}(x) - \mathbb{E}_{n|\widetilde{X}}[\widehat{\eta}_{k\text{-NN}}(x)])^2}$ to be small (at most $\frac{\varepsilon^2}4$), and the expectation of this quantity over randomness in the training data is the variance of estimator $\widehat{\eta}_{k\text{-NN}}$ at $x$.

\textit{Prevention:}
With a large enough number of nearest neighbors $k = \Omega(\frac1{\varepsilon^2}\log\frac1\delta)$, we can control the probability of this bad event to be at most $\frac{\delta}2$ (Lemma~\ref{lem:k-NN-rate-of-convergence-helper1}).

\item \textbf{Bad event \#2.}
Even if there were no label noise (\ie, the average label computed by $k$-NN is exactly $\mathbb{E}_{n|\widetilde{X}}[\widehat{\eta}_{k\text{-NN}}(x)]$), the average label is not close to $\eta(x)$, likely due to the average being computed using labels of nearest neighbors that are too far from~$x$. In other words, $k$ is too large! (For example, consider when $k=n$.) We formalize this bad event as ${| \mathbb{E}_{n|\widetilde{X}}[\widehat{\eta}_{k\text{-NN}}(x)] - \eta(x) |} > \frac{\varepsilon}2$, and we encode the notion that training data farther from $x$ could have labels farther from $\eta(x)$ using a smoothness condition on $\eta$ (specifically, H\"{o}lder continuity, defined shortly in Section~\ref{sec:technicalities}).

In terms of the bias-variance tradeoff, preventing this bad event means ensuring that ${| \mathbb{E}_{n|\widetilde{X}}[\widehat{\eta}_{k\text{-NN}}(x)] - \eta(x) |} \le \frac{\varepsilon}2$, where the left-hand side is precisely the absolute value of the $k$-NN estimator's bias at $x$. Thus, we control $k$-NN regression's bias to be small.

\textit{Idea behind prevention:}
Here, how regression function $\eta$ varies around the point $x$ is crucial.  For example, if $\eta$ is just a constant function across the entire feature space $\mathcal{X}$, then we can choose the number of nearest neighbors $k$ to be as large as the number of training points $n$. However, if within a small distance from $x$, $\eta$ already starts taking on very different values from $\eta(x)$, then we should choose $k$ to be small.

More formally, there is some critical distance $h^*$ (that depends on how $\eta$ varies around $x$, and how much error $\varepsilon$ we tolerate in the regression estimate) such that if all $k$ nearest neighbors found are within distance~$h^*$ from $x$, then we will indeed prevent this second bad event from happening (\ie, we successfully control $k$-NN regression's bias at $x$ to have absolute value at most $\frac{\varepsilon}2$).  If $\eta$ is ``smoother'' in that it changes value slowly as we move farther away from $x$, then $h^*$ is larger. If the desired regression error tolerance $\varepsilon$ is smaller, then $h^*$ is smaller.  In the extreme case where $\eta$ is a constant function, $h^*$ can be taken to be infinity, but otherwise $h^*$ will be some finite quantity.  The exact formula for critical distance $h^*$ will be given later and will depend on smoothness (H\"{o}lder continuity) parameters of $\eta$ and error tolerance~$\varepsilon$. In particular, $h^*$ does not depend on the number of nearest neighbors $k$ or the number of training data $n$.  Rather, with $h^*$ treated as a fixed number, we aim to choose $k$ small enough so that with high probability, the $k$ nearest neighbors are within distance $h^*$ of $x$.

Pictorially, critical distance $h^*$ tells us what the maximum radius of the shaded ball in Figure~\ref{fig:k-NN-helper} should be to prevent this second bad event from happening. The shaded ball has radius given by the distance from $x$ to its \mbox{$(k+1)$-st} nearest neighbor. We can shrink the radius of this shaded ball in two ways. First, of course, is to use a smaller $k$. The other option is to increase the number of training data $n$. In particular, by making $n$ larger, more training points are likely to land within distance $h^*$ of $x$.  Consequently, the $k$ nearest neighbors found will get closer to $x$, so the shaded ball shrinks in radius.

To ensure that the shaded ball has radius at most $h^*$, only shrinking $k$ (and not increasing $n$) may not be enough. For example, suppose that there is only a single training data point and we want to estimate $\eta(x)$ using $k$-NN regression with $k=1$. Then it is quite possible that we simply got unlucky and the single training data point is not within distance $h^*$ of $x$. We cannot possibly decrease $k$ any further, yet the shaded ball has radius exceeding~$h^*$. Thus, not only should $k$ be sufficiently small, $n$ should be sufficiently large.

Formally, we want enough training points to land in the ball centered at $x$ with radius $h^*$, which we denote as
\begin{equation}
\mathcal{B}_{x, h^*} \triangleq \{x' \in \mathcal{X} : \rho(x, x') \le h^*\}.
\label{eq:definition-of-ball}
\end{equation}
(In general, $\mathcal{B}_{c,r}$ denotes a ball centered at $c$ with radius~$r$.) The number of training data that fall into ball $\mathcal{B}_{x, h^*}$ scales with the probability that a feature vector sampled from $\mathbb{P}_X$ lands in the ball:
\[
\mathbb{P}_X(\mathcal{B}_{x, h^*})
= \mathbb{P}(\text{feature vector}\sim\mathbb{P}_X\text{~lands in~}
\mathcal{B}_{x, h^*}).
\]

We can now concisely state how we prevent this second bad event.

\textit{Prevention:}
With a small enough number of nearest neighbors $k=\mathcal{O}(n \mathbb{P}_X(\mathcal{B}_{x, h^*}))$ and large enough number of training data $n=\Omega(\frac1{\mathbb{P}_X(\mathcal{B}_{x, h^*})}\log\frac1\delta)$, we can control the probability of this second bad event to happen with probability at most $\frac\delta2$ (Lemma~\ref{lem:k-NN-rate-of-convergence-helper2-1}).

\end{itemize}
In summary, if we have a large enough number of training data $n=\Omega(\frac1{\mathbb{P}_X(\mathcal{B}_{x, h^*})}\log\frac1\delta)$ and choose a number of nearest neighbors $k$ neither too small nor too large, $\Theta(\frac1{\varepsilon^2}\log\frac1\delta) \le k \le \Theta(n\mathbb{P}_X(\mathcal{B}_{x, h^*}))$, then we can prevent both bad events from happening with a probability that we can control, which in turn ensures that the error in estimating $\eta(x)$ is small.  Concretely, with probability at least $1-\delta$, we ensure that $| \widehat{\eta}_{k\text{-NN}}(x) - \eta(x) | \le \varepsilon$.

To see why we have this guarantee, note that the probability that at least one of the bad events happen (\ie, the union of the bad events happens) is, by a union bound, at most $\frac\delta2 + \frac\delta2 = \delta$. This means that with probability at least $1 - \delta$, neither bad event happens, for which the error $| \widehat{\eta}_{k\text{-NN}}(x) - \eta(x) |$ in estimating $\eta(x)$ is guaranteed to be at most $\varepsilon$ since, with the help of the triangle inequality,
\begin{align}
&| \widehat{\eta}_{k\text{-NN}}(x) - \eta(x) | \nonumber \\
&\quad = \big|
     \big(
       \widehat{\eta}_{k\text{-NN}}(x) - \mathbb{E}_{n|\widetilde{X}}[\widehat{\eta}_{k\text{-NN}}(x)]
     \big)
     +
     \big(
       \mathbb{E}_{n|\widetilde{X}}[\widehat{\eta}_{k\text{-NN}}(x)] - \eta(x)
     \big)
   \big| \nonumber \\
&\quad \le 
   |
       \widehat{\eta}_{k\text{-NN}}(x) - \mathbb{E}_{n|\widetilde{X}}[\widehat{\eta}_{k\text{-NN}}(x)]
   |
     +
   |
       \mathbb{E}_{n|\widetilde{X}}[\widehat{\eta}_{k\text{-NN}}(x)] - \eta(x)
   | \nonumber \\
&\quad \le \frac{\varepsilon}2 + \frac{\varepsilon}2 \nonumber \\
&\quad = \varepsilon.
\label{eq:regression-high-level-triangle-inequality}
\end{align}
In Section~\ref{sec:k-NN-regression-theory}, we provide precise statements of theoretical guarantees for $k$-NN regression, first in estimating $\eta(x)$ for a specific observed feature vector~$x$ (Theorem~\ref{thm:k-NN-regression-rate-of-convergence}), which we sketched out an analysis for above, and then we explain how to account for randomness in sampling the observed feature vector $X=x$ from feature distribution~$\mathbb{P}_X$. This latter guarantee ensures small expected regression error $\mathbb{E}[|\widehat{\eta}_{k\text{-NN}}(X) - \eta(X)|]$, where the expectation is over both the randomness in the training data and in the test feature vector $X=x$ (Theorem~\ref{thm:k-NN-regression-expectation-rate-of-convergence}). We find that this guarantee is nearly the same as an existing result by \citet[Theorem~2]{kohler_2006}.

\begin{figure}
\centering
\includegraphics[scale=.78]{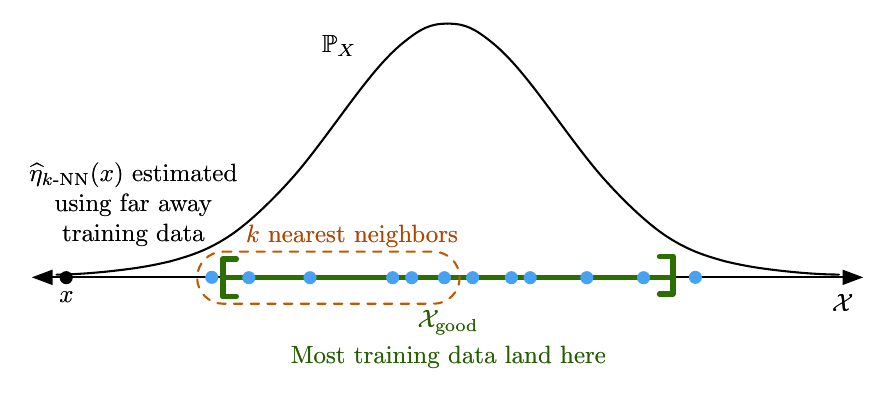}
\vspace{-2.7em}
\caption{Example where $k$-NN regression accuracy can be low ($k=6$ in this example): when the feature distribution $\mathbb{P}_X$ is univariate Gaussian, training data (blue points) are likely to land near the mean of the Gaussian, \eg, mostly within the green region labeled $\mathcal{X}_{\text{good}}$.  If we want to estimate $\eta(x)$ for~$x$ very far from the mean, then it is likely that its $k$ nearest neighbors (circled in orange) are not close to $x$, and unless $\eta$ is extremely smooth, then regression estimate $\widehat{\eta}_{k\text{-NN}}(x)$ will be inaccurate.
\label{fig:good-region-helper}}
\end{figure}

In obtaining this expected regression error guarantee, the challenge is that observed feature vector $X=x$ can land in a region with extremely low probability. When this happens, the nearest neighbors found are likely to be far away, and unless $\eta$ is extremely smooth, then the estimate $\widehat{\eta}_{k\text{-NN}}(x)$ for $\eta(x)$ is going to be inaccurate. An example of this phenomenon is shown for when feature distribution~$\mathbb{P}_X$ is a univariate Gaussian in Figure~\ref{fig:good-region-helper}.  As suggested by this figure, most training data land in a region (denoted as $\mathcal{X}_{\text{good}}$) of the feature space that, in some sense, has high probability. We examine a way to define this \textit{sufficient mass region} $\mathcal{X}_{\text{good}}$ that relates to the \textit{strong density assumption} by \citet{audibert_2007} and the \textit{strong minimal mass assumption} by \citet{gadat_2016}. After defining the sufficient mass region $\mathcal{X}_{\text{good}}$, then the derivation of the expected regression error bound basically says that if $X=x$ lands in $\mathcal{X}_{\text{good}}$, then with high probability we can control the expected error $\mathbb{E}[|\widehat{\eta}_{k\text{-NN}}(X) - \eta(X)|]\le\varepsilon$. Otherwise, we tolerate potentially disastrous regression error.

As a preview, splitting up the feature space into a ``good'' region and its complement, the ``bad'' region, is a recurring theme across all the nonasymptotic regression and classification results we cover that account for randomness in observed feature vector $X=x$. In classification, although the same good region as in regression can be used, we shall see that there is a different way to define the good region in which its complement, the bad region, corresponds precisely to the probability of landing near the decision boundary.

\begin{figure}
\centering
\vspace{-.25em}
\includegraphics[scale=0.55]{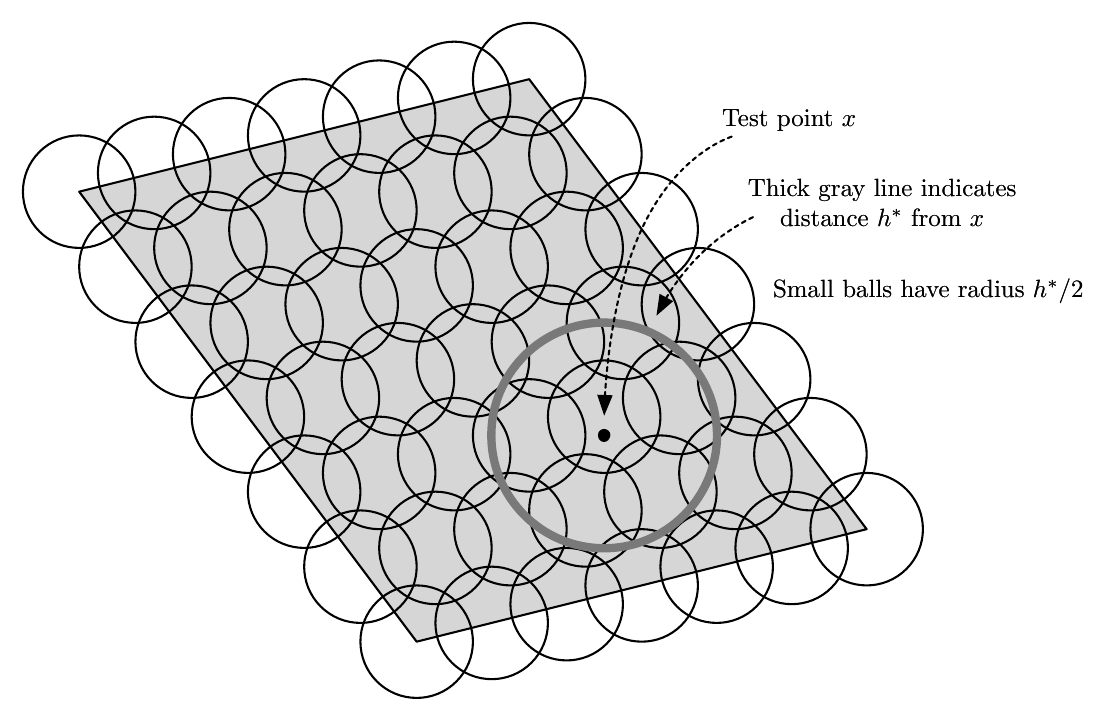}
\vspace{-.75em}
\caption{Diagram to help explain the alternative strategy for guaranteeing low expected regression error.  The shaded region is the feature space~$\mathcal{X}$, which is covered by small balls each with radius $h^*/2$. Test point~$x$ lands in one of these small balls.  Regardless of where~$x$ lands, the small ball that it lands in is contained in the ball with radius $h^*$ centered at $x$.  Enough training data should be collected so nearly every small ball has at least $k$ training points land in it.
}
\label{fig:k-NN-covering-helper}
\end{figure}

We also show an alternative proof technique in Section~\ref{sub:expected-regression-error-alt-approach} for establishing an expected $k$-NN regression error guarantee using topological properties of the feature space and distribution. This argument is stylistically different from using notions such as the sufficient mass region, strong density assumption, or the strong minimal mass assumption and is in some sense more general, although it comes at a cost: it asks for a larger number of training data than the other expected regression error guarantees we present. The proof idea is to show under fairly general conditions on the underlying feature space and distribution, a finite number of small balls with radius $h^*/2$ ``cover'' the whole feature space $\mathcal{X}$ (meaning that their union contains $\mathcal{X}$), and that there are enough training data that fall into nearly every one of these small balls. As a result, for any test feature vector $x$ sampled from the feature distribution, since it will belong to one of these balls, it is highly likely that it has enough nearby neighbors (at least $k$ neighbors within distance $h^*$).  A diagram illustrating this covering argument is shown in Figure~\ref{fig:k-NN-covering-helper}.  While this proof technique also works for fixed-radius NN and kernel regression, we only state the result for $k$-NN regression.

Lastly, we remark that all the main theorems in this chapter state how to choose the training dataset size $n$ and number of nearest neighbors $k$ to achieve a user-specified regression error tolerance $\varepsilon>0$. Thus,~$n$ and~$k$ can be thought of as functions of $\varepsilon$.  In Section~\ref{sub:how-to-translate-101} (at the very end of all the proofs of this chapter), we explain how to translate these theorems to instead state, for a fixed choice of $n$ and $k$, what error tolerance $\varepsilon$ can be achieved, \ie, $\varepsilon$ is a function of $n$ and $k$. As this translation is relatively straightforward (especially if one does the translation in big~O notation and ignores log factors), we only show how to do it for two of the $k$-NN regression guarantees (Theorems~\ref{thm:k-NN-regression-rate-of-convergence} and~\ref{thm:k-NN-regression-expectation-rate-of-convergence}), with the details spelled out (without big~O notation).  We refrain from presenting similar translations for all the other theorem and corollary statements in this chapter.

\subsection{Fixed-Radius NN Regression.}

\begin{figure}[t]
\centering
\vspace{-1em}
\includegraphics[scale=.8]{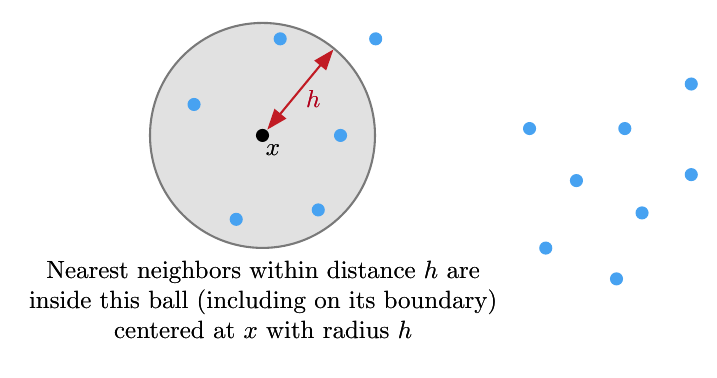}
\vspace{-1em}
\caption{Illustration to help with fixed-radius NN analysis: the blue points are training data, the test feature vector that we are making a prediction for is the black point $x$ and the nearest neighbors used are the ones inside the shaded ball, which has radius $h$.
\label{fig:h-near-helper}}
\end{figure}

Recall that the fixed-radius near neighbor estimate with threshold distance $h > 0$ is given by:
\[
\widehat{\eta}_{\text{NN}(h)}(x)
=
\begin{cases}
{\displaystyle
\frac{\sum_{i=1}^{n}\ind\{\rho(x, X_i) \le h\}Y_{i}}
     {N_{x,h}}}
& \text{if~}N_{x,h} > 0, \\
0 & \text{otherwise},
\end{cases}
\]
where
\[
N_{x,h} := \sum_{i=1}^{n}\ind\{\rho(x, X_i) \le h\}
\]
is the number of training points within distance $h$ of $x$.  Pictorially, the nearest neighbors whose labels are used in the averaging are precisely the training data (blue points) in the shaded ball (including on the ball's boundary) of Figure~\ref{fig:h-near-helper}.

Similar to the case of $k$-NN regression, to combat noise in estimating the expected label $\eta(x)$, we want to average over more values, \ie, set the threshold distance $h$ to be large (this is analogous to preventing $k$-NN regression's bad event \#1 in Section~\ref{sub:k-nn-regression-bad-events}).  But having $h$ be too large means that we may be averaging over the labels of points which are not close to~$\eta(x)$ (this is analogous to $k$-NN regression's bad event~\#2 in Section~\ref{sub:k-nn-regression-bad-events}).  However, unlike $k$-NN regression, fixed-radius NN regression has an additional bad event: if threshold distance $h$ is chosen to be too small, then there could be no nearest neighbors found, so no labels are averaged at all. Put another way, $k$-NN regression can adapt to regions of the feature space with sparse training data coverage by latching onto farther away training data whereas fixed-radius NN regression cannot.

Thus, we have the following bad events (and how to prevent them), where the first two are like the two $k$-NN regression bad events; as with our presentation of $k$-NN regression results, $\varepsilon>0$ and $\delta\in(0,1)$ are user-specified error and probability tolerances, and $\mathbb{E}_{n|N_{x,h}}$ denotes the expectation over the random training data, conditioned on random variable $N_{x,h}$ (unlike for the $k$-NN case, we do not condition on the \mbox{$(k+1)$-st} nearest neighbor):
\begin{itemize}

\item \textbf{Bad event \#1.}
The average label deviates too much from its expectation (conditioned on $N_{x,h}$): ${|\widehat{\eta}_{\text{NN}(h)}(x) - \mathbb{E}_{n|N_{x,h}}[\widehat{\eta}_{\text{NN}(h)}(x)]| > \frac{\varepsilon}2.}$ In terms of the bias-variance tradeoff, as with the $k$-NN regression case, preventing this bad event relates to controlling estimator variance to be small.

\textit{Prevention:}
Assuming that bad event \#3 below does not happen, then with a large enough number of training data $n=\Omega(\frac{1}{\mathbb{P}_X(\mathcal{B}_{x, h})\varepsilon^2}\log\frac1\delta)$, we can control this bad event to happen with probability at most~$\frac{\delta}2$ (Lemma~\ref{lem:radius-NN-helper2}).

\item \textbf{Bad event \#2.}
The true expected estimate deviates too much from the true value: $|\mathbb{E}_{n|N_{x,h}}[\widehat{\eta}_{\text{NN}(h)}(x)] - \eta(x)| > \frac{\varepsilon}2$.  In terms of the bias-variance tradeoff, as with the \mbox{$k$-NN} regression case, preventing this bad event directly controls estimator bias to be small.

\textit{Prevention:}
Assuming that bad event \#3 below does not happen, then this bad event deterministically does not happen so long as the threshold distance $h$ is chosen to be at most $h^*$, which depends on the smoothness of regression function $\eta$ (H\"{o}lder continuity parameters) and the user-specified error tolerance $\varepsilon$ (Lemma~\ref{lem:radius-NN-helper3}). In fact, $h^*$ is the same as in $k$-NN regression.

\item \textbf{Bad event \#3.}
The number of training data landing within distance $h$ of~$x$ is too small, specifically $\le \frac12 n \mathbb{P}_X(\mathcal{B}_{x, h})$.

\textit{Prevention:}
With the number of training data $n=\Omega(\frac1{\mathbb{P}_X(\mathcal{B}_{x, h})}\log\frac1\delta)$, we can control this bad event to happen with probability at most $\frac{\delta}2$ (Lemma~\ref{lem:radius-NN-helper1}).

\end{itemize}
Thus, with a large enough number of training data $n=\Omega(\frac{1}{\mathbb{P}_X(\mathcal{B}_{x, h})\varepsilon^2}\log\frac1\delta)$ and a threshold distance $h$ to be at most critical threshold $h^*$, we can prevent the bad events from happening with a probability that we can control, which then means that the error $|\widehat{\eta}_{\text{NN}(h)}(x) - \eta(x)|$ is at most $\varepsilon$, again using a triangle inequality argument similar to inequality~\eqref{eq:regression-high-level-triangle-inequality}.

The probability of any of the bad events happening this time is still derived with a union bound but has a few extra steps. Letting $\mathcal{E}_1$, $\mathcal{E}_2$, and $\mathcal{E}_3$ denote the three bad events, respectively, then
\begin{align*}
&\mathbb{P}(\mathcal{E}_1
             \text{~or~}
            \mathcal{E}_2
              \text{~or~}
            \mathcal{E}_3) \\
&=
\mathbb{P}\big([\mathcal{E}_1 \text{~or~} \mathcal{E}_2]
                \text{~and~} \mathcal{E}_3\big)
+
\mathbb{P}\big([\mathcal{E}_1 \text{~or~} \mathcal{E}_2]
                \text{~and~} [\text{not~}\mathcal{E}_3]\big) \\
&=
\underbrace{\mathbb{P}\big(\mathcal{E}_1 \text{~or~} \mathcal{E}_2 \mid
               			   \mathcal{E}_3\big)}_{\le 1}
\mathbb{P}(\mathcal{E}_3)
+
\underbrace{\mathbb{P}\big([\mathcal{E}_1 \text{~or~} \mathcal{E}_2]
           			       \mid \text{not~}\mathcal{E}_3\big)}_{
\substack{
\le \mathbb{P}(\mathcal{E}_1 \mid \text{not~}\mathcal{E}_3)
    + \mathbb{P}(\mathcal{E}_2 \mid \text{not~}\mathcal{E}_3) \\
\text{union bound}
}
}
\underbrace{\mathbb{P}(\text{not~}\mathcal{E}_3)}_{\le 1} \\
&\le \mathbb{P}(\mathcal{E}_3) +
\mathbb{P}(\mathcal{E}_1 \mid \text{not~}\mathcal{E}_3)
    + \mathbb{P}(\mathcal{E}_2 \mid \text{not~}\mathcal{E}_3).
\end{align*}
The prevention strategies ensure that $\mathbb{P}(\mathcal{E}_1 \mid \mathcal{E}_3) \le \frac{\delta}2$, $\mathbb{P}(\mathcal{E}_2 \mid \mathcal{E}_3) = 0$, and $\mathbb{P}(\mathcal{E}_3) \le \frac{\delta}2$, which plugged into the above inequality means that the probability of at least one bad event happening is at most $\delta$, \ie, the probability that none of them happen is at least $1-\delta$.

The fixed-radius NN regression guarantees (Theorem~\ref{thm:radius-NN-rate-of-convergence}, for a specific observed feature vector $X=x$, and Theorem~\ref{thm:radius-NN-expectation-rate-of-convergence}, accounting for randomness in sampling $X=x$ from $\mathbb{P}_X$) turn out to be almost identical to those of $k$-NN regression. The reason for this is that in both cases, the analysis ends up asking that the radius of the shaded ball (in either Figures~\ref{fig:k-NN-helper} or \ref{fig:h-near-helper}) be at most $h^*$.  In $k$-NN regression, we choose $k$ in a way where the distance from $x$ to its $(k+1)$-th nearest neighbor is less than $h^*$ whereas in fixed-radius NN regression, we directly set the threshold distance $h$ to be less than $h^*$. We find that the fixed-radius NN regression result (accounting for randomness in $X=x$) nearly matches an existing result by \citet[Theorem~5.2]{gyorfi_book}.

\subsection{Kernel Regression}
\label{sec:kernel-regression-theory-high-level}

Recall that for a given bandwidth $h > 0$ and kernel function $K:\mathbb{R}_{+}\rightarrow[0,1]$ that takes as input a normalized distance (\ie, distance divided by the bandwidth~$h$) and outputs a similarity score between 0 and 1, the kernel regression estimate for $\eta(x)$ is
\[
\widehat{\eta}_{K}(x;h)
= \begin{cases}
  {\displaystyle
   \frac{\sum_{i=1}^{n}K\big(\frac{\rho(x,X_i)}{h}\big)Y_i}
        {\sum_{i=1}^{n}K\big(\frac{\rho(x,X_i)}{h}\big)}}
  & \text{if }
    {\textstyle \sum_{i=1}^{n}K\big(\frac{\rho(x,X_{i})}{h}\big) > 0}, \\
  0 & \text{otherwise}.
  \end{cases}
\]
The analysis for kernel regression is more delicate than for $k$-NN and fixed-radius NN regression (despite the latter being a special case of kernel regression). The main difficulty is in the weighting of different training data. For example, the kernel weight $K(\frac{\rho(x,X_{i})}{h})$ for a training data point $X_{i}$ extremely far from~$x$ could still be positive, and such a faraway point $X_{i}$ could have label $Y_{i}$ that is nowhere close to $\eta(x)$. To limit the impact of far away points, the kernel $K$ needs to decay fast enough. Intuitively how fast the kernel~$K$ decays should relate to how fast $\eta$ changes.

The weighting also affects a collection of training points. In particular, in both $k$-NN and fixed-radius NN regression, we take an \textit{unweighted} average of nearest neighbors' labels, which substantially simplified analysis as the usual desired behavior occurs in which error can be reduced by averaging over more labels. In kernel regression, however, since we take a weighted average, even if we increase the number of labels we average over, in the worst case, all the weight could be allocated toward a point with a bad label estimate.

For simplicity, we only present a result for which kernel $K$ monotonically decreases and actually becomes 0 after some normalized distance $\tau > 0$ (\ie, $K(s) \le {\ind\{s \le \tau\}}$ for all normalized distances $s\ge0$), which completely eliminates the first issue mentioned above of extremely far away points possibly still contributing positive kernel weight. The second issue of weighted averaging remains, however.  Kernels that satisfy this monotonic decay and zeroing out condition include, for instance, the naive kernel $K(s) = \ind\{s \le 1\}$ (in this case $\tau = 1$) that yields fixed-radius NN regression, and a truncated Gaussian kernel $K(s)=\exp(-\frac{1}{2}s^{2})\ind\{s\le3\}$ that gives weight 0 for distances exceeding 3 standard deviations (in this case $\tau=3$).  Also, we assume that regression function $\eta$ satisfies a smoothness condition (once again, H\"{o}lder continuity).

The above monotonic decay and zeroing out assumption on the kernel is not disastrous. In practice, for massive training datasets (\ie, $n$ extremely large), kernel functions with infinite support (\ie, where $K(s)$ never becomes 0) are often avoided since in general, computing the estimate $\widehat{\eta}_K(x;h)$ for a single point $x$ would involve computing $n$ distances and corresponding kernel weights, which could be prohibitively expensive. For example, in medical image analysis, computing the distance between two 3D images may involve solving a nonlinear image alignment problem. Instead, approximate nearest neighbors are found and then only for these approximate nearest neighbors $X_i$ do we compute their kernel weights $K(\frac{\rho(x,X_{i})}{h})$. Separately, for kernels that have infinite support, after some normalized distance away, the kernel weight is well-approximated by 0 (and might even actually be represented as 0 on a machine due to numerical precision issues, \eg, a Gaussian kernel evaluated at a distance of 100 standard deviations).

As with the $k$-NN and fixed-radius NN regression results, to guarantee that kernel regression has low regression error, we appeal to the triangle inequality. We bound the estimation error of $\eta(x)$ in absolute value, although this time with a slight twist:
\begin{equation}
|\widehat{\eta}_K(x;h)-\eta(x)|
\le
  \Big|\widehat{\eta}_K(x;h)-\frac{A}{B}\Big|
  + \Big|\frac{A}{B}-\eta(x)\Big|,
\label{eq:kernel-regression-error-decomposition}
\end{equation}
where $A\triangleq\mathbb{E}_n[K(\frac{\rho(x,X)}{h})Y]$, $B\triangleq\mathbb{E}_n[K(\frac{\rho(x,X)}{h})]$, and $\mathbb{E}_n$ is the expectation over the $n$ random training data.  In our earlier analysis of $k$-NN and fixed-radius NN regression, we did not have a term like~$\frac{A}{B}$ and instead used the expected regression estimate (albeit conditioned on the \mbox{$(k+1)$-st} nearest neighbor for the $k$-NN estimate, or instead on the random variable $N_{x,h}$ for the fixed-radius NN estimate). For kernel regression, this expectation is cumbersome to work with due to the technical issue mentioned earlier of the average label being weighted.

As before, for a user-specified error tolerance $\varepsilon>0$ and probability tolerance $\delta\in(0,1)$, we ensure that with probability at least $1-\delta$, the error in estimate $\eta(x)$ is at most $\varepsilon$ by asking each error term in decomposition~\eqref{eq:kernel-regression-error-decomposition} to be at most $\varepsilon/2$. The bad events are as follows, where our sketch of the result assumes $K(1) > 0$ (which we can always get by rescaling the input to the kernel; the formal theorem statements later do not make this assumption):
\begin{itemize}
\item \textbf{Bad event \#1.}
$|\widehat{\eta}_K(x;h)-\frac{A}{B}| > \frac{\varepsilon}2$.

\textit{Prevention:}
With a large enough number of training data $n=\Omega(\frac1{\varepsilon^2 \mathbb{P}_X(\mathcal{B}_{x, h})^4} \log\frac1\delta)$, we can control the probability of this bad event to be at most $\delta$ (Lemma~\ref{lem:kernel-regression-rate-of-convergence-helper1}).

\item \textbf{Bad event \#2.}
$|\frac{A}{B} - \eta(x)| > \frac{\varepsilon}2$.

\textit{Prevention:}
If the bandwidth $h$ is at most some $h^*>0$ that depends on smoothness of regression function $\eta$, the desired error tolerance $\varepsilon$, and the cutoff normalized distance $\tau$, then this bad event deterministically does not happen (Lemma~\ref{lem:kernel-regression-error2}). Note that for the naive kernel corresponding to fixed-radius NN regression, the $h^*$ here is the same as in the $k$-NN and fixed-radius NN regression results.

\end{itemize}
In terms of the bias-variance tradeoff, depending on how well the ratio~$\frac{A}{B}$ approximates the expected regression estimate $\mathbb{E}_n[\widehat{\eta}_K(x;h)]$ (this approximation gets better with a larger training dataset size~$n$), then preventing bad events \#1 and \#2 relate to controlling estimator variance and bias, respectively, somewhat similar to the $k$-NN and fixed-radius NN regression cases.

Putting together the above two prevention strategies, we see that with enough training data $n=\Omega(\frac1{\varepsilon^2 \mathbb{P}_X(\mathcal{B}_{x, h})^4} \log\frac1\delta)$, then with probability at least $1-\delta$, we guarantee low regression error ${|\widehat{\eta}_K(x;h) - \eta(x)| \le \varepsilon}$.  Precise statements are given in Theorem~\ref{thm:kernel-regression-rate-of-convergence} for a specific observed feature vector~$x$, and Theorem~\ref{thm:kernel-regression-expected-error-rate-of-convergence}, which accounts for randomness in sampling $X=x$ from $\mathbb{P}_X$.

Specifically for fixed-radius NN regression, the theoretical guarantee using this general kernel result is weaker than the earlier guarantee specifically derived for fixed-radius NN regression in terms of how much training data is sufficient to ensure the same level of error: the kernel result asks for $n = \Omega(\frac{1}{\varepsilon^2 \mathbb{P}_X(\mathcal{B}_{x, h})^4}\log\frac1\delta)$ whereas the fixed-radius NN regression result asks for $n = \Omega(\frac{1}{\varepsilon^2 \mathbb{P}_X(\mathcal{B}_{x, h})}\log\frac1\delta)$.  Of course, on the flip side, the kernel regression result is more general.

We end our high-level overview of kernel regression guarantees by remarking that a series of results establish rate of convergence guarantees for kernel regression that handle more general kernel functions~\citep{krzyzak_1986,krzyzak_pawlak_1987}.  These results are a bit messier as they depend on the smallest solution to a nonlinear equation related to how the kernel function decays.

\section{Key Definitions, Technicalities, and Assumptions}
\label{sec:technicalities}

This section summarizes all the key definitions, technicalities, and assumptions that appear in the theorem statements of regression and classification guarantees (in both Chapters~\ref{chap:regression} and~\ref{chap:classification}).  Importantly, we adopt abbreviations for all the major assumptions. Because these assumptions repeatedly appear in the theorem statements, rather than restating the assumptions each time, the shorthand notation we adopt will keep the theorem statements a bit more concise.

The theory to be presented is fairly general as to handle a wide range of distributions governing feature vectors and labels. To support this level of generality, two key assumptions required by the guarantees will be technical and will ensure that some measure-theoretic arguments carry through. We bundle these two assumptions together in Assumption~\ref{assump:technical} below.
\begin{fassumption}[abbreviated \assumpTechnical] ~
\label{assump:technical}
\begin{itemize}

\item[(a)] The feature space $\mathcal{X}$ and distance $\rho$ form a separable metric space (Section~\ref{sec:technicality-separable-metric-space}).

\item[(b)] The feature distribution $\mathbb{P}_X$ is a Borel probability measure (Section~\ref{sec:technicality-borel-prob-measure}).

\end{itemize}
\end{fassumption}
In addition to providing mathematical grounding needed for the theory, these assumptions enable us to properly define what a feature vector being ``observable'' means (Section~\ref{sec:technicality-support}). This is important because we measure regression error only for such observable feature vectors.

Also as suggested by our overview of results, smoothness of the regression function $\eta$ comes into play, which we formalize through H\"{o}lder continuity (Section~\ref{sec:technicality-holder}). When we talk about regression guarantees at a specific observable feature vector $x$, we can get away with a weaker assumption than H\"{o}lder continuity, referred to as the Besicovitch condition (Section~\ref{sec:technicality-besicovitch}).

Lastly, as we have actually already described for kernel regression in Section~\ref{sec:kernel-regression-theory-high-level}, we work with kernel functions satisfying the following decay assumption.
\begin{fassumption}[abbreviated \assumpDecay]
Kernel function $K:\mathbb{R}_+\rightarrow[0,1]$ monotonically decreases and becomes 0 after some normalized distance $\tau>0$ (\ie, $K(s) \le \ind\{s \le \tau\}$ for all $s\ge0$).
\end{fassumption}

\subsection{The Feature Space and Distance Form a Separable Metric Space}
\label{sec:technicality-separable-metric-space}

We assume that the practitioner chooses feature space $\mathcal{X}$ and distance function $\rho$ so that $(\mathcal{X}, \rho)$ forms a metric space (meaning that $\rho$ satisfies the requirements of a metric). As suggested by our overview, we will often be reasoning about balls in the feature space, and working with a metric space ensures that these balls are properly defined.  For a technical reason to be described later in this section, we ask that the metric space be separable (meaning that it has a countable dense subset). Examples of metric spaces $(\mathcal{X}, \rho)$ that are separable include when $\mathcal{X}$ is any finite or countably infinite set (in which $\mathcal{X}$ itself is the countable dense subset), as well as when $\mathcal{X}$ is the Euclidean space $\mathbb{R}^d$ for any fixed dimension $d$ (in which the $d$-dimensional rational number grid $\mathbb{Q}^d$ is a countable dense subset).

As a preview, in Chapter~\ref{chap:case-studies} we encounter scenarios where distance function $\rho$ is not a metric (in time series forecasting) or cannot be computed exactly (in online collaborative filtering, where we only obtain noisy distances). Of course, in these settings, there is additional problem structure that enables nearest neighbor methods to still succeed.

\subsection{The Feature Distribution is a Borel Probability Measure}
\label{sec:technicality-borel-prob-measure}

Next, we assume that the feature distribution $\mathbb{P}_X$ is a Borel probability measure, which roughly speaking means that it assigns a probability to every possible ball (whether open or closed) in the metric space (and consequently also countable unions, countable intersections, and relative complements of these balls).  This is a desirable property since, as suggested by the outline of results in the previous section, we will reason about probabilities of feature vectors landing in different balls.

\subsection{Support of the Feature Distribution and Separability}
\label{sec:technicality-support}

Thus far, we have generally been careful in saying that for an \textit{observed} feature vector $x \in \mathcal{X}$, we want to estimate $\eta(x)$. The technical caveat for why the word ``observed'' appears is that the conditional expectation $\mathbb{E}[Y \mid X=x]$ as well as the conditional distribution of $Y$ given $X=x$ need to be well-defined. For instance, if $X$ is a discrete random variable and we observe feature vector $x$ with probability~0 (\ie, $\mathbb{P}(X=x) = 0$), then we cannot actually condition on $X=x$.  We aim to estimate $\eta(x)$ only for ``observable'' feature vectors $x$, which we formalize via the \textit{support} of distribution $\mathbb{P}_X$:
\[
\text{supp}(\mathbb{P}_X)
\triangleq \{ x \in \mathcal{X} :
\mathbb{P}_X(\mathcal{B}_{x, r}) > 0
\text{~for all~}r > 0 \},
\]
where as a reminder, the definition of the closed ball $\mathcal{B}_{x,r}$ is given in equation~\eqref{eq:definition-of-ball}. In words, any feature vector for which landing in any size ball around it has positive probability is in the support. This definition neatly handles when feature distribution $\mathbb{P}_X$ is, for instance, either discrete or continuous, \eg, if $\mathbb{P}_X$ is Bernoulli, then $\text{supp}(\mathbb{P}_X)=\{0, 1\}$, and if $\mathbb{P}_X$ is uniform over interval $[0,1]$, then $\text{supp}(\mathbb{P}_X)=[0, 1]$.

The support of $\mathbb{P}_X$ seemingly tells us which feature vectors are observable.  However, could it be that the probability of landing outside the support has positive probability, meaning that there are other feature vectors worth considering as observable? This is where the technical condition mentioned earlier of the metric space $(\mathcal{X}, \rho)$ being separable comes into play. Separability ensures that the support is all that matters: feature vectors land in the support with probability~1. \citet{cover_1967} provide a proof of this albeit embedded in another proof; a concise restatement and proof are given by~\citet[Lemma~23]{chaudhuri_dasgupta_2014}.

\subsection{Regression Function Smoothness Using H\"{o}lder Continuity}
\label{sec:technicality-holder}

Throughout our overview of results, we alluded to smoothness of the regression function~$\eta$. This is formalized using H\"{o}lder continuity.
\begin{fassumption}[abbreviated \assumpHolder]
Regression function~$\eta$ is H\"{o}lder continuous with parameters $C>0$ and $\alpha>0$ if
\begin{equation*}
|\eta(x) - \eta(x')|
\le C \rho(x, x')^\alpha\qquad\text{for all~}x,x'\in\mathcal{X}.
\end{equation*}
\end{fassumption}
We remark that Lipschitz continuity, which is also often used as a smoothness condition, corresponds to H\"{o}lder continuity with $\alpha=1$.

H\"{o}lder continuity tells us how much the regression function~$\eta$'s value can change in terms of how far apart two different feature vectors are.  For example, suppose $x$ is the observed feature vector of interest, and a training feature vector $x'$ is a distance $h = \rho(x, x')$ away.  If we want to ensure that the regression function values $\eta(x)$ and $\eta(x')$ are close (say, at most $\frac{\varepsilon}2$), then H\"{o}lder continuity tells us how large $h$ should be:
\[
|\eta(x) - \eta(x')| \le C h^\alpha \underbrace{\le}_{\text{want this to hold}} \frac{\varepsilon}2.
\]
So long as $h \le (\frac{\varepsilon}{2C})^{1/\alpha} \triangleq h^*$, then the desired second inequality above does indeed hold. In fact, as a preview, this $h^*$ turns out to be the $h^*$ in the earlier overview of $k$-NN and fixed-radius NN regression results.

H\"{o}lder continuity enforces a sort of ``uniform'' notion of smoothness in regression function~$\eta$ across the entire feature space~$\mathcal{X}$. If we only care about being able to predict $\eta(x)$ well for any specific observable feature vector~$x$, then we can get by with a milder notion of smoothness to be described next, referred to as the Besicovitch condition.  A major change is that for any $x$, the critical distance $h^*$ for which we want nearest neighbors of $x$ to be found will now depend on $x$, and we will only know that there exists such an $h^*$ rather than what its value should be as in the case of $\eta$ being H\"{o}lder continuous.

\subsection{Regression Function Smoothness Using the Besicovitch Condition}
\label{sec:technicality-besicovitch}

When focusing on predicting $\eta(x)$ at any specific observable feature vector $x$ (so we pick any $x\in\text{supp}(\mathbb{P}_X)$ and do not account for randomness in sampling $x$ from $\mathbb{P}_X$), whether the regression function $\eta$ has crazy behavior far away from $x$ is irrelevant.  In this case, instead of H\"{o}lder continuity, we can place a much milder condition on~$\eta$ called the Besicovitch condition.
\begin{fassumption}[abbreviated \assumpBesicovitch]
Regression function $\eta$ satisfies the Besicovitch condition if
\begin{equation*}
\lim_{r\downarrow0}\mathbb{E}[Y\mid X \in \mathcal{B}_{x,r}]=\eta(x)
\quad\text{for }x\text{ almost everywhere w.r.t.~}\mathbb{P}_X.
\end{equation*}
\end{fassumption}
This condition says that if we were to sample training feature vectors close to $x$ (up to radius~$r$), then their average label approaches $\eta(x)$ as we shrink the radius $r$.  These closeby training data would be the nearest neighbors found by $k$-NN regression.  One can check that H\"{o}lder continuity implies the Besicovitch condition.

We remark that there are different versions of this Besicovitch condition (\eg, a version in Euclidean space used for establishing nearest neighbor regression pointwise consistency is provided by \citet{devroye_1981}, and a general discussion of the Besicovitch condition in dimensions both finite and infinite is provided by \citet{cerou_guyader_2006}). This condition is also referred to as a differentiation condition as it is asking that Lebesgue differentiation holds (this terminology is used, for instance, by \citet{abraham_biau_cadre_2006} and \citet{chaudhuri_dasgupta_2014}).

\section{Theoretical Guarantees for \texorpdfstring{$k$}{k}-NN Regression}
\label{sec:k-NN-regression-theory}

We are ready to precisely state theoretical guarantees for $k$-NN regression. We first provide a guarantee for the regression estimation error at a specific observed feature vector $x \in \text{supp}(\mathbb{P}_X)$.  This is referred to as a \textit{pointwise} regression error since it focuses on a specific point $x$.  As a reminder, the recurring key assumptions made (\assumpTechnical, \assumpBesicovitch, \assumpHolder) are precisely stated in Section~\ref{sec:technicalities}.

\begin{ftheorem}
[$k$-NN regression pointwise error]
\label{thm:k-NN-regression-rate-of-convergence}
Under assumptions \assumpTechnical~and \assumpBesicovitch, let $x\in\text{supp}(\mathbb{P}_{X})$ be a feature vector, $\varepsilon>0$ be an error tolerance in estimating expected label $\eta(x)={\mathbb{E}[Y \mid X=x]}$, and $\delta\in(0,1)$ be a probability tolerance. Suppose that $Y\in[y_{\min},y_{\max}]$ for some constants $y_{\min}$ and $y_{\max}$.  There exists a threshold distance $h^{*}\in(0,\infty)$ such that for any smaller distance $h\in(0,h^{*})$, if the number of training points satisfies 
\begin{equation}
n\ge\frac{8}{\mathbb{P}_X( \mathcal{B}_{x, h} )}\log\frac{2}{\delta},
\label{eq:k-NN-regression-rate-of-convergence-constraint-on-n}
\end{equation}
and the number of nearest neighbors satisfies
\begin{equation}
\frac{2(y_{\max}-y_{\min})^{2}}{\varepsilon^{2}}\log\frac{4}{\delta}\le k\le\frac{1}{2}n\mathbb{P}_X(\mathcal{B}_{x,h}),
\label{eq:k-NN-regression-rate-of-convergence-constraint-on-k}
\end{equation}
then with probability at least $1-\delta$ over randomness in sampling the training data, $k$-NN regression at point $x$ has error
\[
|\widehat{\eta}_{k\text{-NN}}(x)-\eta(x)|\le\varepsilon.
\]
Furthermore, if the function $\eta$ satisfies assumption \assumpHolder, then we can take $h^{*}=(\frac{\varepsilon}{2C})^{1/\alpha}$ and the above guarantee holds for $h=h^{*}$ as well.
\end{ftheorem}
Let's interpret this theorem. Using any distance $h$ less than $h^*$ (or also equal to, when $\eta$ is H\"{o}lder continuous), the theorem gives sufficient conditions on how to choose the number of training data points $n$ and the number of nearest neighbors $k$. What is happening is that when the training data size $n$ is large enough, then we will with high probability see at least $k$ points land within distance $h$ from $x$, which means that the $k$ nearest neighbors of $x$ are certainly within distance $h$.  Then taking the average label value of these $k$ nearest neighbors, with high probability, this average value gets close to $\eta(x)=\mathbb{E}[Y\mid X=x]$, up to error tolerance~$\varepsilon$.

If we demand a very small error $\varepsilon$ in estimating $\eta(x)$, then depending on how the function $\eta$ fluctuates around the point $x$, we may need $h$ to be very small as well, which means that we need a lot more training data so as to see more training points land within distance $h$ from $x$. Specifically in the case when $\eta$ is H\"{o}lder continuous, we see that the largest $h$ that we provide a guarantee for is $h=(\frac{\varepsilon}{2C})^{1/\alpha}$.  For example, in the case that $\eta$ is Lipschitz ($\alpha=1$), we have $h=\mathcal{O}(\varepsilon)$.

Note that if error tolerance $\varepsilon$ is too small, then there could be no choice of number of nearest neighbors $k$ satisfying the condition~\eqref{eq:k-NN-regression-rate-of-convergence-constraint-on-k}. In particular, smaller $\varepsilon$ means that we should average over more values to arrive at a good approximation of $\eta$, \ie, the number of nearest neighbors $k$ should be larger.  But larger $k$ means that the $k$ nearest neighbors may reach beyond distance $h$ from $x$, latching onto training data points too far from $x$ that do not give us label values close to $\eta(x)$.

Fundamentally, how much $\eta$ fluctuates around the point $x$ significantly impacts our ability to estimate $\eta(x)$ accurately. Binary classification, as we will discuss more about later in Chapter~\ref{chap:classification}, is a much simpler problem because we only care about being able to correctly identify whether $\eta(x)$ is above or below $1/2$.  This suggests that somehow in the classification case, we can remove the H\"{o}lder continuity assumption.  What matters in binary classification is how $\eta$ fluctuates around the decision boundary $\{ x'\in\mathcal{X} : \eta(x') = 1/2 \}$.  For example, if test feature vector $x$ is near the decision boundary, and if $\eta$ fluctuates wildly around the decision boundary, then the nearest neighbors of $x$ may have labels that do not give us a good estimate for $\eta(x)$, making it hard to tell whether $\eta(x)$ is above or below 1/2.

Our analysis of $k$-NN regression makes it clear that we want to choose $k$ carefully so nearest neighbors found are within distance $h$ from $x$, for a range of possible $h$ that ensure that regression error can be at most $\varepsilon$. Here, using a larger $h$ (in the theorem of course; there is no parameter $h$ in the algorithm) means that to achieve the same error~$\varepsilon$, we can use a smaller number of training data $n$ (\cf, inequality~\eqref{eq:k-NN-regression-rate-of-convergence-constraint-on-n}) and a wider range of possible $k$ (since the upper bound in sandwich inequality~\eqref{eq:k-NN-regression-rate-of-convergence-constraint-on-k} increases).  However, we cannot in general make $h$ arbitrarily large, as we only provide a guarantee for $h$ being less than $h^*$ (or also equal to, when $\eta$ is H\"{o}lder continuous).

The analysis we present for $k$-NN regression reasons about training data points landing within distance $h$ from a test point $x$, and relates this distance $h$ with the chosen number of nearest neighbors $k$. Of course, we could sidestep this issue of relating $h$ and $k$ by just having the algorithm look at all nearest neighbors within a distance $h$, which precisely results in the fixed-radius NN regression.

\subsection{Toward Expected Regression Error: Partitioning the Feature Space Using the Sufficient Mass Region}
\label{sub:toward-expected-regression-error-partitioning-the-feature-space}

As mentioned in the overview of results, to turn the guarantee on pointwise regression error $|\widehat{\eta}(x) - \eta(x)|$ of Theorem~\ref{thm:k-NN-regression-rate-of-convergence} into one on the expected regression error ${\mathbb{E}[|\widehat{\eta}(X) - \eta(X)|]}$, where the expectation is over all training data and the test feature vector $X=x$ sampled from feature distribution~$\mathbb{P}_X$, the key idea is to partition the feature space into a high probability ``good'' region $\mathcal{X}_{\text{good}}$ (where we can choose $n$ and~$k$ to control regression error) and its complement $\mathcal{X}_{\text{good}}^c = \mathcal{X}\setminus\mathcal{X}_{\text{good}}$, the bad region (where we tolerate potentially terrible regression accuracy).  Formally, we define the \textit{sufficient mass region} $\mathcal{X}_{\text{good}}$ of feature distribution~$\mathbb{P}_X$ parameterized by constants $p_{\min} > 0$, $d > 0$, and $r^*>0$ to be
\begin{align*}
&\mathcal{X}_{\text{good}}(\mathbb{P}_X;p_{\min}, d, r^*) \\
&\quad\triangleq
   \big\{
     x \in \text{supp}(\mathbb{P}_X)
     \;:\;
     \mathbb{P}_X(\mathcal{B}_{x,r}) \ge p_{\min} r^d
     \;\text{for all }r\in(0,r^*]
   \big\}.
\end{align*}
The basic idea is that inside this region $\mathcal{X}_{\text{good}}$ (specifically with $r^* = (\frac{\varepsilon}{2C})^{1/\alpha}$), we can readily apply Theorem~\ref{thm:k-NN-regression-rate-of-convergence} (with the case where $\eta$ is H\"{o}lder continuous with parameters $C$ and $\alpha$), and outside this region, we just assume a worst-case regression error of $y_{\max} - y_{\min}$. In fact, we do not need $\eta$ to be H\"{o}lder continuous outside of the sufficient mass region $\mathcal{X}_{\text{good}}$ (plus a little bit of radius around this region due to some boundary effects). We remark that the sufficient mass region is a slight variant on the strong minimal mass assumption by \citet{gadat_2016}, which in turn is related to the strong density assumption by \citet{audibert_2007}.\footnote{The strong minimal mass assumption (SMMA) of \citet{gadat_2016} essentially asks that for every $x\in\text{supp}(\mathbb{P}_X)$, we have $\mathbb{P}_X(\mathcal{B}_{x,r}) \ge p_{\min}' p_X(x)r^d$, where $p_{\min}'>0$ is some constant and $p_X$ is the density of distribution $\mathbb{P}_X$ with respect to Lebesgue measure. The definition we introduced for the sufficient mass region does not need to hold over all of $\text{supp}(\mathbb{P}_X)$, nor does it depend on a density with respect to Lebesgue measure. If the entire support of $\mathbb{P}_X$ is inside a sufficient mass region, and $\mathbb{P}_X$ has a density $p_X$ with respective to Lebesgue measure that is lower-bounded by a positive constant, then it is straightforward to see that SMMA and the sufficient mass region are equivalent; in this case SMMA is also equivalent to the strong density assumption (SDA) by~\citet{audibert_2007} (\cf, \citealt[Proposition~3.1]{gadat_2016}), so all three conditions coincide.  We discuss the SDA more later.}

Let's build intuition for sufficient mass region $\mathcal{X}_{\text{good}}(\mathbb{P}_X;p_{\min}, d, r^*)$ of the feature space.  First off, this region is defined for any choice of $p_{\min} > 0$, $d > 0$, and $r^* > 0$. A poor choice of these parameters could mean that $\mathcal{X}_{\text{good}}(\mathbb{P}_X;p_{\min}, d, r^*)=\emptyset$, a fruitless region to work with.  We give a few examples of feature distributions and nonempty sufficient mass regions.

\begin{figure}
\centering
\subfloat[][The probability of landing in $\mathcal{B}_{x,r}$ is
$\frac1{b-a}\cdot2r$.\label{subfig:uniform-good-region-helper1}]{
\includegraphics[scale=.8,trim={0em 1.7em 0 3.6em},clip]{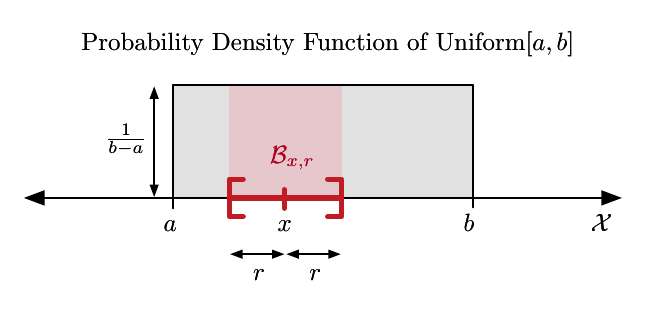}
} \\
\subfloat[][The probability of landing in $\mathcal{B}_{a,r}$ is
$\frac1{b-a}\cdot r$.\label{subfig:uniform-good-region-helper2}]{
\includegraphics[scale=.8,trim={0 1.7em 0 3.6em},clip]{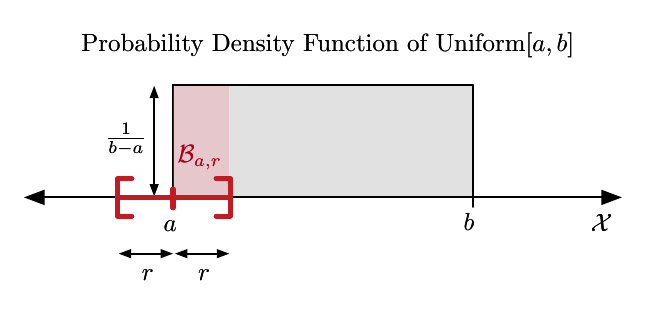}
\vspace{-3em}
}
\caption{Probability density function of $\mathbb{P}_X\sim\text{Uniform}[a,b]$, and examples of two different balls that a feature vector drawn from $\mathbb{P}_X$ can land in, one ball completely contained in $[a,b]$ (top), and one ball on an endpoint of $[a,b]$ (bottom).}
\label{fig:uniform-good-region-helper}
\end{figure}

\begin{fexample}[Uniform distribution]
Suppose the feature space is the real line $\mathcal{X} = \mathbb{R}$, the feature distribution is $\mathbb{P}_X \sim \text{Uniform}[a, b]$ for constants $a < b$, and the metric $\rho$ is the usual distance $\rho(x,x') = |x-x'|$. Then every part of $\text{supp}(\mathbb{P}_X)$ is equally likely, so there should be a sufficient mass region corresponding to the full support of the feature distribution. Formally, this means that we should be able to find $p_{\min} > 0$, $d > 0$, and $r^* > 0$ such that $\mathcal{X}_{\text{good}}(\mathbb{P}_X;p_{\min}, d, r^*) = [a, b]$. This of course turns out to be the case since for any $x \in \text{supp}(\mathbb{P}_X) = [a, b]$, and any $r \in (0, b-a]$,
\begin{equation}
\mathbb{P}_X(\mathcal{B}_{x,r}) \ge \frac{1}{b-a} r,
\label{eq:good-part-uniform-rv-example}
\end{equation}
meaning that $\mathcal{X}_{\text{good}}(\mathbb{P}_X; \frac{1}{b-a}, 1, b-a) = [a, b] = \text{supp}(\mathbb{P}_X)$.

Let's see why inequality~\eqref{eq:good-part-uniform-rv-example} holds.  The basic idea is to look at the probability of a feature vector sampled from $\mathbb{P}_X$ landing in balls of radius $r$ centered at different points in interval $[a,b]$.  Consider a ball $\mathcal{B}_{x,r}=[x-r,x+r]$ that is completely contained in interval $[a,b]$, as shown in Figure~\ref{fig:uniform-good-region-helper}\subref{subfig:uniform-good-region-helper1}. The probability of landing in this ball is the area of the shaded pink rectangle, $\frac1{b-a}\cdot2r$.

Pictorially, by thinking about sliding this ball $\mathcal{B}_{x,r}$ from left to right so that its center $x$ goes from endpoint $a$ to endpoint $b$, note that the probability of landing in the ball is lowest precisely when the center of the ball is at either $a$ or $b$, when exactly half of the ball is contained in interval $[a,b]$. As shown in Figure~\ref{fig:uniform-good-region-helper}\subref{subfig:uniform-good-region-helper2}, landing in the ball centered at the left endpoint~$a$ has probability ${\frac1{b-a}\cdot r}$. One can check that the probability of landing in ball centered at the right endpoint $b$ is the same.

By this pictorial argument, $\frac1{b-a}\cdot r$ is the lowest probability of landing in any of these balls, which almost fully justifies inequality~\eqref{eq:good-part-uniform-rv-example}.  The missing piece is how large the radius $r$ can be. By looking at Figure~\ref{fig:uniform-good-region-helper}\subref{subfig:uniform-good-region-helper2}, note that the largest $r$ such that the area of the shaded region is still $\frac{1}{b-a}\cdot r$ is $r=b-a$. One can check that for this choice of $r$, sliding the ball from $a$ to $b$, the probability of landing in the ball is always at least (and in fact in this case equal to) $\frac{1}{b-a}\cdot r$. Any larger $r$ makes this lower bound no longer hold.

We remark that inequality~\eqref{eq:good-part-uniform-rv-example} holding implies that $[a,b]$ is a subset of $\mathcal{X}_{\text{good}}(\mathbb{P}_X;\frac1{b-a},1,b-a)$.  Of course, since the sufficient mass region is defined to be a subset of $\text{supp}(\mathbb{P}_X)=[a,b]$, we get a sandwich relationship and can conclude that $\mathcal{X}_{\text{good}}(\mathbb{P}_X;\frac1{b-a},1,b-a) = [a,b] = \text{supp}(\mathbb{P}_X)$ in this case. The next example shows a setting in which this sandwich relationship does not hold.
\end{fexample}

\begin{figure}
\centering
\includegraphics[scale=.8,trim={0 1.7em 0 3.4em},clip]{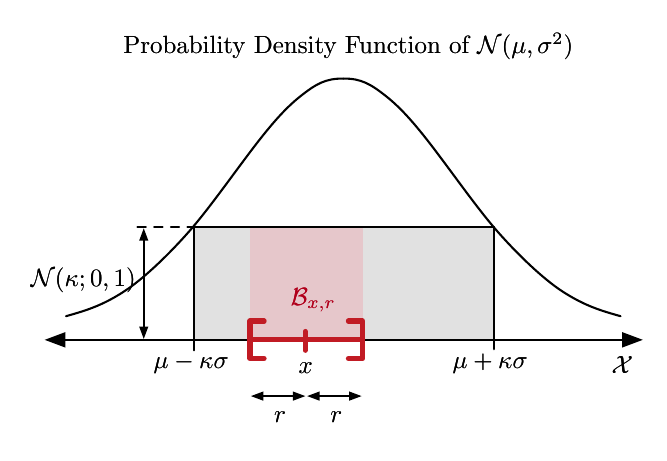}
\vspace{-.5em}
\caption{The probability density function of $\mathbb{P}_X \sim \mathcal{N}(\mu, \sigma^2)$ (outer unshaded envelope) is lower-bounded on the interval $[\mu - \kappa\sigma, \mu + \kappa\sigma]$ by a box function with height $\mathcal{N}(\mu + \kappa\sigma; \mu, \sigma^2) = \mathcal{N}(\kappa; 0, 1)$. The probability of landing in $\mathcal{B}_{x,r}$ is lower-bounded by the area of the pink rectangle.}
\label{fig:gauss-good-region-helper}
\end{figure}

\begin{fexample}[Gaussian distribution]
Suppose the feature space is the real line $\mathcal{X} = \mathbb{R}$, the feature distribution is $\mathbb{P}_X \sim \mathcal{N}(\mu, \sigma^2)$, and the metric $\rho$ is the usual distance $\rho(x,x') = |x-x'|$. Intuitively, due to the symmetry of the Gaussian distribution and that it strictly decreases at feature values farther away from the mean $\mu$, a sufficient mass region of the feature space should correspond to an interval centered at the mean $\mu$. It turns out that for any $\kappa>0$, the interval $[\mu - \kappa \sigma, \mu + \kappa \sigma]$ is contained in $\mathcal{X}_{\text{good}}(\mathbb{P}_X;\mathcal{N}(\kappa; 0, 1), 1, 2\kappa\sigma)$, where $\mathcal{N}(x; \mu, \sigma^2)$ is the probability density function of a $\mathcal{N}(\mu, \sigma^2)$ distribution evaluated at $x$ (so $\mathcal{N}(\kappa; 0, 1)$ is the standard Gaussian density evaluated at $\kappa$).

To see why this is the case, note that the Gaussian density of $\mathcal{N}(\mu, \sigma^2)$ within interval $[\mu - \kappa \sigma, \mu + \kappa \sigma]$ is lower-bounded by a ``box'' function that is precisely a uniform distribution's density except with height $\mathcal{N}(\mu + \kappa \sigma; \mu, \sigma^2) = \mathcal{N}(\kappa; 0, 1)$, as shown in Figure~\ref{fig:gauss-good-region-helper}. Our pictorial analysis from the uniform distribution case carries over to this Gaussian case since we can just reason about the shaded box in Figure~\ref{fig:gauss-good-region-helper}. Once again, we find a lower bound on $\mathbb{P}_X(\mathcal{B}_{x,r})$ by looking at the area of the shaded pink rectangle when the ball $\mathcal{B}_{x,r}$ is centered at either of the endpoints of interval $[\mu - \kappa \sigma, \mu + \kappa \sigma]$, and the maximum radius is the width of the gray shaded box. This yields the inequality
\[
\mathbb{P}_X(\mathcal{B}_{x,r})
\ge \mathcal{N}(\kappa; 0, 1) \cdot r
\quad\text{for all~}x\in[\mu - \kappa\sigma, \mu + \kappa\sigma],
r \in (0, 2\kappa\sigma],
\]
which implies that $[\mu - \kappa\sigma, \mu + \kappa\sigma] \subseteq \mathcal{X}_{\text{good}}(\mathbb{P}_X;\mathcal{N}(\kappa; 0, 1), 1, 2\kappa\sigma)$. Note that the sufficient mass region with those parameters is in fact larger than the interval $[\mu - \kappa\sigma, \mu + \kappa\sigma]$ as it includes a little bit more probability mass beyond either end of that interval.
\end{fexample}
A large generalization of the two examples above is also true.  Notice the pattern: in the uniform distribution example, we showed that the compact set $[a,b]$ that is contained in (and actually equal to) the support of the feature distribution $\mathbb{P}_X \sim \text{Uniform}[a, b]$ is inside a sufficient mass region (with appropriately chosen parameters).  A key fact used is that the density of $\mathbb{P}_X$ has a lower bound within the compact set $[a,b]$ (namely, $\frac{1}{b-a}$).  In the Gaussian distribution example, we showed that the compact set $[\mu - \kappa\sigma, \mu + \kappa\sigma]$ that is contained in the support of feature distribution $\mathbb{P}_X \sim \mathcal{N}(\mu, \sigma^2)$ is inside a sufficient mass region (again, with appropriately chosen parameters). Once again, we crucially used the fact that the density of $\mathbb{P}_X$ has a lower bound within the compact set $[\mu - \kappa\sigma, \mu + \kappa\sigma]$, namely $\mathcal{N}(\kappa; 0, 1)$. It turns out that for a wide variety of compact sets that are subsets of supports of feature distributions, we can readily say what sufficient mass regions $\mathcal{X}_{\text{good}}$ they fall in.

A technical wrinkle is that the geometry of the compact set matters, specifically at the boundary of the set. In both the uniform and Gaussian distribution examples, the lower bound depended on the boundary behavior, specifically that the ball $\mathcal{B}_{x,r}$ centered at an endpoint was only \textit{half} inside the compact set we cared about ($[a,b]$ and $[\mu-\kappa\sigma,\mu+\kappa\sigma]$). This factor of $\frac12$ changes depending on the shape of the compact set and the dimensionality the feature space is in. Without imposing a constraint on the geometry of the compact set, putting a handle on this factor could be problematic and the factor could also be 0 (\eg, the feature distribution is uniform over the unit square $[0,1]^2$ and the compact set chosen is a line segment within the square, which has volume~0).

Formally, for feature distribution $\mathbb{P}_X$ defined over a feature space~$\mathcal{X}$ that is a subset of Euclidean space $\mathbb{R}^d$, we say that a set $\mathcal{A} \subseteq \text{supp}(\mathbb{P}_X)$ is a \textit{strong density region} with parameters ${p_{\text{L}}>0}$, $p_{\text{U}}>0$, $c_0\in(0,1]$, and $r^*>0$ if:
\begin{itemize}[noitemsep,nolistsep]

\item The set $\mathcal{A}$ is compact.

\item The feature distribution $\mathbb{P}_X$ has a density $p_X$ with respect to Lebesgue measure $\lambda$ of $\mathbb{R}^d$.

\item The density $p_X$ has lower bound $p_{\text{L}}$ and upper bound $p_{\text{U}}$ within set $\mathcal{A}$:
\[
p_{\text{L}} \le p_X(x) \le p_{\text{U}}\qquad\text{for all~}x\in\mathcal{A}.
\]
We emphasize that lower bound parameter $p_{\text{L}}$ is strictly positive.

\item The set $\mathcal{A}$ has a boundary that is in the following sense ``regular'':
\[
\lambda( \mathcal{A} \cap \mathcal{B}_{x,r} ) \ge c_0 \lambda(\mathcal{B}_{x,r})
\qquad\text{for all~}x \in \mathcal{A}, r \le r^*.
\]

\end{itemize}
The last constraint is geometric.  While we ask it to hold for all $x\in\mathcal{A}$, whenever the ball $\mathcal{B}_{x,r}$ is strictly in the compact set $\mathcal{A}$, the inequality trivially holds for any $c_0\in(0,1]$, so what matters is the boundary geometry.  In the earlier uniform and Gaussian distribution examples, ${c_0 = \frac12}$ with the least amount of overlap happening when $\mathcal{B}_{x,r}$ was centered at an endpoint of the compact set (which was a one-dimensional closed interval).  As a much broader example, any convex polytope $\mathcal{A}\subseteq\text{supp}(\mathbb{P}_X)$ with a non-empty $d$-dimensional interior satisfies this regularity condition for some $r^*>0$ that relates to how far apart corners of the polytope are.

We remark that $\mathcal{A} = \text{supp}(\mathbb{P}_X)$ being a strong density region precisely corresponds to the strong density assumption of~\citet{audibert_2007} holding for feature distribution~$\mathbb{P}_X$. In this sense, the strong density region is a simple theoretical tool for partitioning the feature space into a piece that satisfies the strong density assumption (and falls in a sufficient mass region $\mathcal{X}_{\text{good}}$ for which pointwise regression will easily succeed in) and another piece that we give up on (and tolerate awful regression accuracy).

The following proposition makes it clear in what way our earlier definition of a sufficient mass region is a generalization of the strong density region.

\begin{fproposition}
\label{claim:good-region-big-case}
For any feature distribution $\mathbb{P}_X$ with support contained in $\mathbb{R}^d$, and for strong density region $\mathcal{A} \subseteq \text{supp}(\mathbb{P}_X)$ with parameters $p_{\text{L}}>0$, $p_{\text{U}}>0$, $c_0>0$, and $r^*>0$, then $\mathcal{A} \subseteq \mathcal{X}_{\text{good}}(\mathbb{P}_X;p_{\text{L}} c_0 v_d, d, r^*)$, where $v_d$ is the volume of a unit closed ball (as a reminder, this ball's shape depends on metric $\rho$).
\end{fproposition}
For example, in the earlier uniform and Gaussian distribution examples, $v_d = 2$, which is the diameter of a closed interval of radius 1.

\subsection{Expected Regression Error}\label{ssec:expected-reg-error}

Equipped with intuition for what the sufficient mass region $\mathcal{X}_{\text{good}}$ is through examples and connections to existing literature, we are ready to state a performance guarantee for $k$-NN regression accounting for randomness in the test feature vector $X = x$.  In what follows, we denote the ``bad'' region to be the complement of the sufficient mass region:
\begin{align*}
\mathcal{X}_{\text{bad}}(\mathbb{P}_X;p_{\min}, d, r^*)
&\triangleq
[\mathcal{X}_{\text{good}}(\mathbb{P}_X;p_{\min}, d, r^*)]^c \\
&= \mathcal{X} \setminus \mathcal{X}_{\text{good}}(\mathbb{P}_X;p_{\min}, d, r^*).
\end{align*}

\begin{ftheorem}
[$k$-NN regression expected error]
\label{thm:k-NN-regression-expectation-rate-of-convergence}
Let $\varepsilon > 0$ and $\delta \in (0,1)$. Under assumptions \assumpTechnical~and \assumpHolder, suppose that $Y\in[y_{\min},y_{\max}]$ for some constants $y_{\min}$ and $y_{\max}$. Let $p_{\min} > 0$ and $d > 0$. If the number of training points satisfies
\begin{equation*}
n \ge \Big( \frac{2C}{\varepsilon} \Big)^{d/\alpha}
      \frac{16}{p_{\min}} \log \frac{\sqrt{8}}{\delta}
\end{equation*}
and the number of nearest neighbors satisfies
\begin{equation*}
\frac{4(y_{\max}-y_{\min})^{2}}{\varepsilon^{2}}\log\frac{4}{\delta}
\le k
\le \frac{1}{2} n p_{\min} \Big( \frac{\varepsilon}{2C} \Big)^{d/\alpha},
\end{equation*}
then $k$-NN regression has expected error
\begin{align*}
&\mathbb{E}[| \widehat{\eta}_{k\text{-NN}}(X) - \eta(X) |] \\
&\le \varepsilon + (y_{\max} - y_{\min}) \big(\mathbb{P}_X(\mathcal{X}_{\text{bad}}(\mathbb{P}_X;p_{\min}, d, {\textstyle (\frac{\varepsilon}{2C})^{1/\alpha}})) + \delta\big).
\end{align*}
\end{ftheorem}
The high-level idea is exactly as we have already mentioned: if $X$ lands in the sufficient mass region, then we should expect error at worst $\varepsilon$ for the same reason as why Theorem~\ref{thm:k-NN-regression-rate-of-convergence} holds.  Specifically, we apply Theorem~\ref{thm:k-NN-regression-rate-of-convergence} with choice $h^* = (\frac{\varepsilon}{2C})^{1/\alpha}$. When $X$ lands outside the sufficient mass region, then the regression error could be terrible, where the worst it could be is $y_{\max} - y_{\min}$.

As a technical note, the conditions on the number of training data~$n$ and the number of nearest neighbors~$k$ differ slightly from those in Theorem~\ref{thm:k-NN-regression-rate-of-convergence}. We actually plug in $\frac{\delta^2}4$ in place of $\delta$ in Theorem~\ref{thm:k-NN-regression-rate-of-convergence}; this is explained in the proof (Section~\ref{sec:proof-k-NN-regression-expectation-rate-of-convergence}). Qualitatively, this change has little impact since the dependence on $\delta$ is logarithmic.

If the entire support of the feature distribution $\mathbb{P}_X$ is in a good region $\mathcal{X}_{\text{good}}$ with appropriately chosen parameters (so that the probability of landing in $\mathcal{X}_{\text{bad}}$ is 0), then we obtain the following result by plugging in $\varepsilon = \frac{\varepsilon'}{2}$ and $\delta = \frac{\varepsilon'}{2(y_{\max} - y_{\min})}$ for $\varepsilon'\in(0,2(y_{\max}-y_{\min}))$.  (The result would then be strictly in terms of~$\varepsilon'$, which we write simply as~$\varepsilon$ below.)

\begin{fcorollary}
\label{cor:k-NN-regression-simple-case}
Under assumptions \assumpTechnical~and \assumpHolder, suppose that $Y\in[y_{\min},y_{\max}]$ for some constants $y_{\min}$ and $y_{\max}$. Let $\varepsilon \in (0, 2(y_{\max} - y_{\min}))$ be an estimation error tolerance. If
\begin{align}
&\text{supp}(\mathbb{P}_X) \nonumber \\
&= \mathcal{X}_{\text{good}}(\mathbb{P}_X;p_{\min}, d, {\textstyle (\frac{\varepsilon}{4C})^{1/\alpha}})
\;\text{for some }p_{\min} > 0\text{ and }d > 0,
\label{eq:support-of-feature-dist-is-all-good}
\end{align}
the number of training points satisfies
\begin{equation}
n \ge \Big( \frac{4C}{\varepsilon} \Big)^{d/\alpha}
      \frac{16}{p_{\min}} \log \frac{\sqrt{32}(y_{\max} - y_{\min})}{\varepsilon},
\label{eq:k-NN-regression-simple-case-constraint-on-n}
\end{equation}
and the number of nearest neighbors satisfies
\begin{equation}
\frac{16(y_{\max}-y_{\min})^{2}}{\varepsilon^{2}}\log\frac{8(y_{\max} - y_{\min})}{\varepsilon}
\le k
\le \frac{1}{2} n p_{\min} \Big( \frac{\varepsilon}{4C} \Big)^{d/\alpha},
\label{eq:k-NN-regression-simple-case-constraint-on-k}
\end{equation}
then $k$-NN regression has expected error
\[
\mathbb{E}[| \widehat{\eta}_{k\text{-NN}}(X) - \eta(X) |]
\le \varepsilon.
\]
\end{fcorollary}
An implication of Proposition~\ref{claim:good-region-big-case} is that any feature distribution $\mathbb{P}_X$ with compact support in $\mathbb{R}^d$ and that has density (again, need not be with respect to Lebesgue measure) lower-bounded by a positive constant satisfies condition~\eqref{eq:support-of-feature-dist-is-all-good} provided that $\varepsilon>0$ is chosen to be sufficiently small. Thus, this wide class of feature distributions works with the above corollary.

We compare Corollary~\ref{cor:k-NN-regression-simple-case} to an existing result on $k$-NN regression in Euclidean feature spaces of sufficiently high dimensionality \citep[Theorem~2]{kohler_2006}.  This result is actually stated in terms of expected squared error $\mathbb{E}[(\widehat{\eta}_{k\text{-NN}}(X) - \eta(X))^2]$. With an application of Jensen's inequality and some algebra (as detailed in Section~\ref{sec:how-to-rephrase-kohler}), we can phrase it in terms of expected error $\mathbb{E}[|\widehat{\eta}_{k\text{-NN}}(X) - \eta(X)|]$.

\begin{ftheorem}[\citealt{kohler_2006}, Theorem~2 rephrased]
\label{thm:kohler-rephrased}
Let $\varepsilon > 0$ be an estimation error tolerance. Suppose that:
\begin{itemize}

\item The regression function $\eta$ is both bounded in absolute value by some constant $L\ge1$, and also H\"{o}lder continuous with parameters $C\ge1$ and $\alpha\in(0,1]$.

\item The feature space is in $\mathbb{R}^d$ for some $d > 2\alpha$, and the feature distribution $\mathbb{P}_X$ satisfies $\mathbb{E}[\|X\|^{\beta}] < \infty$ for some constant $\beta > \frac{2\alpha d}{d - 2\alpha}$, where $\|\cdot\|$ is the Euclidean norm.

\item The conditional variance of the label given a feature vector is bounded: $\sup_{x\in\mathbb{R}^d}\text{Var}(Y \mid X=x) \le \sigma^2$ for some $\sigma > 0$.

\end{itemize}
Then there exists a constant $c$ that depends on $d$, $\alpha$, $\beta$, $L$, $\sigma^2$, and $\mathbb{E}[\|X\|^\beta]$ such that if the number of training points satisfies
\begin{equation*}
n \ge \frac{c^{\frac{d}{2\alpha}+1} C^{\frac{d}{\alpha}}}{\varepsilon^{\frac{d}{\alpha}+2}},
\end{equation*}
and the number of nearest neighbors is set to
\[
k = \Bigg\lceil \frac{n^{\frac{2\alpha}{2\alpha+d}}}{C^{\frac{2d}{2\alpha+d}}} \Bigg\rceil,
\]
then $k$-NN regression has expected error
\[
\mathbb{E}[|\widehat{\eta}_{k\text{-NN}}(X) - \eta(X)|] \le \varepsilon.
\]
\end{ftheorem}
To put Corollary~\ref{cor:k-NN-regression-simple-case} and Theorem~2 of~\citet{kohler_2006} on equal footing for comparison, consider when the feature distribution~$\mathbb{P}_X$ satisfies the strong density assumption, $\eta$ is H\"{o}lder continuous with parameters $C\ge1$ and $\alpha\in(0,1]$, $d>2\alpha$, and label $Y$ is bounded in the interval $[y_{\min}, y_{\max}]$ for some constants $y_{\min}$ and $y_{\max}$.  These conditions readily meet those needed by both of the theoretical results.

Specifically for Theorem~2 of~\citet{kohler_2006}, we explain what $\beta$, $L$, and $\sigma^2$ can be taken to be.  The compactness of the feature distribution support means that $X$ is bounded, and so trivially $\mathbb{E}[\|X\|^\beta] < \infty$ for every $\beta > \frac{2\alpha d}{d - 2\alpha}$.  Since the label $Y$ is bounded in $[y_{\min}, y_{\max}]$, this implies that the conditional expected label $\eta(x) = \mathbb{E}[Y\mid X=x]$ is also bounded in $[y_{\min}, y_{max}]$, \ie, $L=\max\{|y_{\min}|, |y_{\max}|\}$. Moreover, one can show that for any $x \in \text{supp}(\mathbb{P}_X)$,
\[
\text{Var}(Y \mid X = x) \le \frac{(y_{\max} - y_{\min})^2}4 = \sigma^2.
\]
(In general, any random variable bounded in $[a,b]$ has variance at most $\frac{(b-a)^2}4$.)

Now let's see what the two different results ask for to ensure that $\mathbb{E}[|\widehat{\eta}(X) - \eta(X)|] \le \varepsilon$:
\begin{itemize}

\item Corollary~\ref{cor:k-NN-regression-simple-case}: We choose the number of nearest neighbors to be
\[
k = \Big\lfloor
      \frac12 n p_{\min} \Big(\frac{\varepsilon}{4C}\Big)^{d/\alpha}
    \Big\rfloor,
\]
in which case to ensure that both of the conditions~\eqref{eq:k-NN-regression-simple-case-constraint-on-n} and~\eqref{eq:k-NN-regression-simple-case-constraint-on-k} are met, we ask that
\begin{align*}
n &\ge \max\bigg\{
\Big( \frac{4C}{\varepsilon} \Big)^{d/\alpha}
      \frac{16}{p_{\min}} \log \frac{\sqrt{32}(y_{\max} - y_{\min})}{\varepsilon}, \\
&\qquad\qquad
\frac{2}{p_{\min}} \Big(\frac{4C}{\varepsilon}\Big)^{d/\alpha}
\bigg\lceil
\frac{16(y_{\max}-y_{\min})^{2}}{\varepsilon^{2}}\log\frac{8(y_{\max} - y_{\min})}{\varepsilon}
\bigg\rceil
\bigg\},
\end{align*}
where the second part of the maximum ensures that
\[
k \ge \lceil\frac{16(y_{\max}-y_{\min})^{2}}{\varepsilon^{2}}\log\frac{8(y_{\max} - y_{\min})}{\varepsilon}\rceil.
\]
In fact, this second part of the maximum is always larger than the first, so it suffices to ask that
\begin{align*}
n &\ge
\frac{2}{p_{\min}} \Big(\frac{4C}{\varepsilon}\Big)^{d/\alpha}
\bigg\lceil
\frac{16(y_{\max}-y_{\min})^{2}}{\varepsilon^{2}}\log\frac{8(y_{\max} - y_{\min})}{\varepsilon}
\bigg\rceil.
\end{align*}
Hence, we can choose $n = \Theta(\frac{C^{\frac{d}{\alpha}}}{\varepsilon^{\frac{d}{\alpha}+2}} \log \frac1{\varepsilon}\big)$, in which case $k = \Theta(n(\frac{\varepsilon}C)^{d/\alpha}) = \Theta(\frac1{\varepsilon^2}\log\frac1{\varepsilon})$. Note that we treat $p_{\min}$ and $(y_{\max} - y_{\min})^2 = \Theta(\sigma^2)$ as constants since in Theorem~2 of~\citet{kohler_2006}, they get absorbed into the constant $c$.

\item
Theorem~2 of~\citet{kohler_2006}: The smallest we could set the training data size $n$ for the theorem's result to apply is
\[
n = \frac{c^{\frac{d}{2\alpha}+1} C^{\frac{d}{\alpha}}}{\varepsilon^{\frac{d}{\alpha}+2}} = \Theta\bigg( \frac{C^{\frac{d}{\alpha}}}{\varepsilon^{\frac{d}{\alpha}+2}} \bigg),
\]
in which case the theorem says to choose
\[
k = \bigg\lceil C^{-\frac{2d}{2\alpha+d}}\bigg( \frac{c^{\frac{d}{2\alpha}+1} C^{\frac{d}{\alpha}}}{\varepsilon^{\frac{d}{\alpha}+2}} \bigg)^{\frac{2\alpha}{2\alpha+d}} \bigg\rceil = \Big\lceil \frac{c}{\varepsilon^2} \Big\rceil = \Theta\Big(\frac{1}{\varepsilon^2}\Big).
\]

\end{itemize}
Thus, up to a log factor, the two performance guarantees both ensure low expected error $\mathbb{E}[|\widehat{\eta}(X) - \eta(X)|] \le \varepsilon$ using training data size $n = \widetilde{\Theta}(\frac{C^{\frac{d}{\alpha}}}{\varepsilon^{\frac{d}{\alpha}+2}})$ and number of nearest neighbors $k = \widetilde{\Theta}(\frac{1}{\varepsilon^2})$.  For the choices of $n$ and $k$ above, we see that Corollary~\ref{cor:k-NN-regression-simple-case} uses slightly more training data and number of nearest neighbors by a factor of $\log\frac{1}{\varepsilon}$.

Alternatively, we can also phrase the result of Corollary~\ref{cor:k-NN-regression-simple-case} in terms of what expected regression error $\varepsilon$ is achievable for a given~$n$ and~$k$ (this is in line with how Kohler \textit{et al.}~originally phrase their Theorem~2, which we present in Section~\ref{sec:how-to-rephrase-kohler}).  Specifically, for sufficiently large $n$ and $k$ that are treated as fixed, we can actually show that Corollary~\ref{cor:k-NN-regression-simple-case} implies that we can achieve expected regression error
\[
\mathbb{E}[|\widehat{\eta}_{k\text{-NN}}(X)-\eta(X)|]\le \varepsilon = \max\bigg\{\widetilde{\Theta}\Big(\frac1{\sqrt{k}}\Big),~\widetilde{\Theta}\bigg( \Big(\frac{k}{n}\Big)^{\alpha/d} \bigg)\bigg\},
\]
derived using the approach described in Section~\ref{sub:how-to-translate-101}. Ignoring log factors, if we make the two terms in the maximization match (\ie, we set $\frac1{\sqrt{k}} \sim (\frac{k}{n})^{\alpha/d}$, which amounts to optimizing our choice of $k$), then we get that we should set $k \sim n^{\frac{2\alpha}{2\alpha+d}}$, which agrees with Kohler \textit{et al.}'s choice for $k$. Plugging this in for~$\varepsilon$, we get that the expected regression error scales as $n^{-\frac{\alpha}{2\alpha+d}}$, agreeing with Kohler \textit{et al.}'s Theorem~2 translated to be in terms of absolute value error (\cf, inequality \eqref{eq:kohler-abs-value-error}).

\subsection{Expected Regression Error: An Alternative Approach}
\label{sub:expected-regression-error-alt-approach}

In this section, we present an alternative approach to guaranteeing low expected $k$-NN regression error using a covering argument. We previously depicted this argument in Figure~\ref{fig:k-NN-covering-helper}.  We start by defining a relevant notion called the {\em covering number} for a probability space, which is going to tell us what the minimum number of balls is needed to cover up the feature space.
\begin{fdefinition}[($r$, $\delta$)-covering number]
Given $\delta \in (0, 1)$ and $r > 0$, let $N(\P_X, r, \delta) \geq 1$ be the smallest number of balls each of radius $r$, say $\{\BB_1,\dots, \BB_N\}$ with $N = N(\P_X, r, \delta)$, so that $\P_X(\cup_{i=1}^N \BB_i) \geq 1-\delta$. Here, each $\BB_i = \BB_{c_i, r}$ is ball of radius $r$ centered at $c_i \in \XX$. 
\end{fdefinition}
In principle, the covering number may not be finite. Our interest is in the setting where the covering number is finite for all $r > 0$ and $\delta \in (0,1)$. This turns out to hold if in addition to assumption~\assumpTechnical~of Section \ref{sec:technicalities}, the feature space $(\XX,\rho)$ is a compact metric space. Under these assumptions, $\XX$ is a compact Polish space. In this case, any probability measure $\P_X$ is {\em tight}. That is, for any $\delta > 0$, there exists a compact set $\KK_\delta$ such that $\P_X(\KK_\delta) \geq 1-\delta$. By compactness, the open cover of $\KK_\delta$ obtained by considering balls of radius $r$ at each of its points has a finite subcover. This leads to covering number $N(\P_X, r, \delta)$ being finite. 

Equipped with the above definition, we state the following result that builds on Theorem \ref{thm:k-NN-regression-rate-of-convergence}. 
\begin{ftheorem}
[$k$-NN expected regression error with covering number]
\label{thm:k-NN-regression-covering-number}
Suppose that assumptions \assumpTechnical~and \assumpHolder~hold, feature space $\XX$ is a compact Polish space, and $Y\in[y_{\min},y_{\max}]$ for some constants $y_{\min}$ and $y_{\max}$. Let $\varepsilon > 0$ and $\delta \in (0,1)$. Suppose we choose the number of nearest neighbors $k$ and the number of training data $n$ to satisfy
\begin{align}
k & = \bigg\lceil \frac{2 (y_{\max} - y_{\min})^2}{\varepsilon^2} \log \frac{4}{\delta}\bigg\rceil \label{eq:knnrcn.2}\\
n & \geq  \frac{N}{\delta} \max\Big\{2k,~ 8 \log \frac{2}{\delta}\Big\},\label{eq:knnrcn.3}
\end{align}
where $h^* = (\frac{\varepsilon}{2C})^{1/\alpha}$ and $N(\P_X, \frac{h^*}2, \delta)$ is the $(\frac{h^*}2, \delta)$-covering number for $\P_X$ as defined above. Then the expected regression error satisfies the bound
\begin{align}\label{eq:knnrcn.1}
\mathbb{E}[|\widehat{\eta}_{k\text{-NN}}(X) - \eta(X)|]  & \leq \varepsilon + 3 \delta (y_{\max} - y_{\min}).
\end{align}
\end{ftheorem}
Let's interpret this result. Consider a specific choice of $\delta$:
\[
\delta = \frac{\varepsilon}{y_{\max} - y_{\min}}.
\]
Then the above result states that for small enough $\varepsilon$, by choosing $k$ that scales as $\frac1{\varepsilon^{2}} \log \frac{1}{\varepsilon}$ (assuming $y_{\max} - y_{\min}$ constant), it is sufficient to have the number of training data $n$ scale as $N(\P_X, \frac{h^*}2, \varepsilon)\frac{k}{\delta}$ to obtain expected regression error scaling as $\varepsilon$.

To quickly compare this with the collection of results stated at the end of Section \ref{ssec:expected-reg-error}, consider a simple scenario where $\P_X$ is a uniform distribution over a unit-volume ball in ${\mathbb R}^d$. Then the corresponding covering number $N(\P_X, \frac{h^*}2, \varepsilon)$ scales as $\mathcal{O}((\frac1{h^*})^d)$, which  is $\mathcal{O}((\frac{C}{\varepsilon})^{\frac{d}{\alpha}})$. Therefore, Theorem~\ref{thm:k-NN-regression-covering-number} says that to achieve expected regression error at most $\varepsilon$, we can choose
\[ 
k = \Theta\Big(\frac1{\varepsilon^2} \log \frac{1}{\varepsilon}\Big), \quad \text{and} \quad 
n = \Theta\bigg(\frac{C^{\frac{d}{\alpha}}}{\varepsilon^{\frac{d}{\alpha}+3}}
\log \frac{1}{\varepsilon}\bigg).
\]
We can compare these choices to what Theorem~2 of~\citet{kohler_2006} asks for (which come from our analysis at the end of Section~\ref{ssec:expected-reg-error}):
\[
k = \Theta\Big(\frac{1}{\varepsilon^2}\Big), \quad \text{and} \quad
n = \Theta\bigg( \frac{C^{\frac{d}{\alpha}}}{\varepsilon^{\frac{d}{\alpha}+2}} \bigg).
\]
Recall that Corollary~\ref{cor:k-NN-regression-simple-case} asks for similar choices for $k$ and $n$ except each with an extra multiplicative $\log\frac1{\varepsilon}$ factor. Thus, across all three theoretical guarantees, we choose $k=\widetilde{\Theta}(\frac1{\varepsilon^2})$. However, the expected regression error guarantee using the covering argument asks for more training data $n$ by a multiplicative factor of $\frac1{\varepsilon}\log\frac1{\varepsilon}$ compared to Kohler \textit{et al.}'s result.

Put another way, Theorem~\ref{thm:k-NN-regression-covering-number} asks that to achieve expected regression error at most $\varepsilon$, the number of training data $n$ should scale as $\frac1{\varepsilon^3}\log\frac1{\varepsilon}$ times the covering number with radius $h^*/2$.  Basically, each ball that is covering the feature space (the unit-volume ball in this example) has roughly $\frac1{\varepsilon^3}\log\frac1{\varepsilon}$ training data points landing in it. Supposing that test feature vector $x$ lands in one such ball, Kohler \textit{et al.}'s result suggests that we only need to find $k=\Theta(\frac1{\varepsilon^2})$ training data points nearby. Thus, potentially we do not actually need $\Theta(\frac1{\varepsilon^3}\log\frac1{\varepsilon})$ training points landing in each ball and we could instead get away with $\Theta(\frac1{\varepsilon^2})$ training points.  The extra multiplicative factor of $\frac1{\varepsilon}\log\frac1{\varepsilon}$ may be a penalty of the proof technique being rather general.

\section{Theoretical Guarantees for Fixed-Radius NN Regression}
\label{sec:h-near-regression-theory}

We now state results for fixed-radius NN regression, first for pointwise error at a particular $x\in\text{supp}(\mathbb{P}_X)$ and then for expected error accounting for randomness in sampling $X=x$ from $\mathbb{P}_X$.  Turning the pointwise error guarantee to the expected error guarantee follows the exact same reasoning as for $k$-NN regression. 

\begin{ftheorem}
[Fixed-radius NN regression pointwise error]
\label{thm:radius-NN-rate-of-convergence}
Under assumptions \assumpTechnical~and \assumpBesicovitch, let $x\in\text{supp}(\mathbb{P}_{X})$ be a feature vector, $\varepsilon>0$ be an error tolerance in estimating expected label $\eta(x)=\mathbb{E}[Y \mid X=x]$, and $\delta\in(0,1)$ be a probability tolerance. Suppose that $Y\in[y_{\min},y_{\max}]$ for some constants $y_{\min}$ and $y_{\max}$.  Then there exists a distance $h^{*}\in(0,\infty)$ such that if our choice of threshold distance satisfies $h\in(0,h^{*})$, and the number of training points $n$ satisfies 
\begin{equation}
n\ge\max\bigg\{\frac{8}{\mathbb{P}_X(\mathcal{B}_{x,h})}\log\frac{2}{\delta},\;\,\frac{4(y_{\max}-y_{\min})^{2}}{\mathbb{P}_X(\mathcal{B}_{x,h})\varepsilon^{2}}\log\frac{4}{\delta}\bigg\},\label{eq:radius-NN-rate-of-convergence-constraint-on-n}
\end{equation}
then with probability at least $1-\delta$ over randomness in sampling the training data, fixed-radius NN regression with threshold distance~$h$ has error
\[
|\widehat{\eta}_{\text{NN}(h)}(x)-\eta(x)|\le\varepsilon.
\]
Furthermore, if the function $\eta$ satisfies assumption \assumpHolder, then we can take $h^{*}=(\frac{\varepsilon}{2C})^{1/\alpha}$ and the above guarantee holds for $h=h^{*}$ as well.
\end{ftheorem}
Note that Theorems~\ref{thm:k-NN-regression-rate-of-convergence} and~\ref{thm:radius-NN-rate-of-convergence} are quite similar. In fact, in Theorem~\ref{thm:k-NN-regression-rate-of-convergence}, if we take the left-most and right-most sides of the constraint on the number of nearest neighbors $k$ (\cf, sandwich inequality~\eqref{eq:k-NN-regression-rate-of-convergence-constraint-on-k}), rearranging terms yields the constraint $n\ge\frac{4(y_{\max}-y_{\min})^{2}}{\mathbb{P}_X(\mathcal{B}_{x,h})\varepsilon^{2}}\log\frac{4}{\delta}$, which is what shows up in inequality~\eqref{eq:radius-NN-rate-of-convergence-constraint-on-n} of Theorem~\ref{thm:radius-NN-rate-of-convergence}. Thus, we remove the dependency on $k$ but otherwise the result and intuition for the analysis are quite similar.

We now turn to the expected error guarantee.

\begin{ftheorem}
[Fixed-radius NN regression expected error]
\label{thm:radius-NN-expectation-rate-of-convergence}
Let $\varepsilon>0$ and $\delta\in(0,1)$.  Under assumptions \assumpTechnical~and \assumpHolder, suppose that $Y\in[y_{\min},y_{\max}]$ for some constants $y_{\min}$ and $y_{\max}$.  Let $p_{\min} > 0$ and $d > 0$. If the threshold distance satisfies $h\in(0, (\frac{\varepsilon}{2C})^{1/\alpha}]$, and the number of training points satisfies
\begin{equation*}
n \ge \frac{8}{p_{\min}h^d}
      \max\Big\{
        2 \log \frac{\sqrt{8}}{\delta},
        \frac{(y_{\max} - y_{\min})^2}
              {\varepsilon^2} \log \frac{4}{\delta}
      \Big\},
\end{equation*}
then fixed-radius NN regression with threshold distance $h$ has \mbox{expected} error
\begin{align*}
&\mathbb{E}[| \widehat{\eta}_{\text{NN}(h)}(X) - \eta(X) |] \\
&\le \varepsilon + \max\{|y_{\min}|,|y_{\max}|,y_{\max} - y_{\min}\} \\
&\qquad\quad\times \big(\mathbb{P}_X(\mathcal{X}_{\text{bad}}(\mathbb{P}_X;p_{\min}, d, {\textstyle (\frac{\varepsilon}{2C})^{1/\alpha}})) + \delta\big).
\end{align*}
\end{ftheorem}
Note that the worst-case fixed-radius NN regression error is
\[
\max\{|y_{\min}|,|y_{\max}|,y_{\max} - y_{\min}\}.
\]
For $k$-NN regression, the worst-case error was $y_{\max} - y_{\min}$. The reason why the terms $|y_{\min}|$ and $|y_{\max}|$ appear for fixed-radius NN regression is that when test point $X$ has no neighbors within distance $h$, the estimate for $\eta(X)$ is 0, which could be off by at most $\max\{|y_{\min}|, |y_{\max}|\}$.  Otherwise, if at least one neighbor is found, then the worst-case error is the same as in $k$-NN regression: $y_{\max} - y_{\min}$.  Of course, in practice a better solution is to report that there are no neighbors within distance $h$. We have fixed-radius NN regession output 0 in this case to be consistent with the presentation of kernel regression in \citet{gyorfi_book}.

Next, as with $k$-NN regression, we can specialize the expected regression error performance guarantee to the case where the sufficient mass region constitutes the entire support of the feature distribution~$\mathbb{P}_X$ (which we explained earlier holds for the case when $\mathbb{P}_X$ satisfies the strong density assumption of~\citet{audibert_2007}).

\begin{fcorollary}
\label{cor:radius-NN-regression-simple-case}
Under assumptions \assumpTechnical~and \assumpHolder, suppose that $Y\in[y_{\min},y_{\max}]$ for some constants $y_{\min}$ and $y_{\max}$.  Let $\varepsilon \in (0, 2\max\{|y_{\min}|,|y_{\max}|,y_{\max} - y_{\min}\})$ be an estimation error tolerance. If the support of the feature distribution satisfies
\begin{align*}
&\text{supp}(\mathbb{P}_X) \\
&= \mathcal{X}_{\text{good}}(\mathbb{P}_X;p_{\min}, d, {\textstyle (\frac{\varepsilon}{4C})^{1/\alpha}})
\;\text{for some }p_{\min} > 0\text{ and }d > 0,
\end{align*}
the threshold distance satisfies $h\in(0, (\frac{\varepsilon}{4C})^{1/\alpha}]$, and the number of training points satisfies
\begin{align}
n &\ge \frac{8}{p_{\min}h^d}
      \max\Big\{
        2 \log \frac{\sqrt{32}\max\{|y_{\min}|,|y_{\max}|,y_{\max} - y_{\min}\}}{\varepsilon}, \nonumber \\
&\qquad\quad
        \frac{4(y_{\max} - y_{\min})^2}
              {\varepsilon^2} \log \frac{8\max\{|y_{\min}|,|y_{\max}|,y_{\max} - y_{\min}\}}{\varepsilon}
      \Big\},
\label{eq:radius-NN-regression-simple-case-constraint-on-n}
\end{align}
then fixed-radius NN regression with threshold distance $h$ has expected error
\[
\mathbb{E}[| \widehat{\eta}_{\text{NN}(h)}(X) - \eta(X) |]
\le \varepsilon.
\]
\end{fcorollary}
We can compare this corollary with an existing result on fixed-radius NN regression \citep[Theorem~5.2]{gyorfi_book}, which is originally stated in terms of expected squared error $\mathbb{E}[(\widehat{\eta}_{\text{NN}(h)}(X) - \eta(X))^2]$, but we rephrase it in terms of expected error $\mathbb{E}[|\widehat{\eta}_{\text{NN}(h)}(X) - \eta(X)|]$. We omit the details of how the rephrasing is done here since it is the same way as how we converted Theorem~2 of~\citet{kohler_2006} into Theorem~\ref{thm:kohler-rephrased} from earlier (\cf, Section~\ref{sec:how-to-rephrase-kohler}).

\begin{ftheorem}[\citet{gyorfi_book}, Theorem~5.2 rephrased]
Let $\varepsilon > 0$ be an estimation error tolerance. Suppose that the feature distribution has compact support $\mathcal{S^*} \subset \mathbb{R}^d$, ${\text{Var}(Y \mid X=x)} \le \sigma^2$ for $x \in \mathbb{R}^d$ and some $\sigma > 0$, and regression function $\eta$ is Lipschitz continuous with parameter $C > 0$ (\ie, $\eta$ is H\"{o}lder continuous with parameters $C$ and $\alpha=1$).  Then there exist constants $c' > 0$ and $c'' > 0$ such that if the number of training points satisfies
\begin{equation*}
n \ge (c'')^{\frac{d+2}2} C^d \cdot \frac{\sigma^2 + \sup_{z\in S^*} |\eta(z)|^2}{\varepsilon^{d+2}},
\end{equation*}
and the threshold distance is set to
\[
h = c'\Big(\frac{\sigma^2 + \sup_{z\in S^*} |\eta(z)|^2}{C^2}\Big)^{\frac1{d+2}}n^{-\frac1{d+2}},
\]
then fixed-radius NN regression with threshold distance $h$ has expected error
\[
\mathbb{E}[|\widehat{\eta}_{\text{NN}(h)}(X) - \eta(X)|] \le \varepsilon.
\]
\end{ftheorem}
To compare Corollary~\ref{cor:radius-NN-regression-simple-case} and Theorem~5.2 of~\citet{gyorfi_book}, we consider when the feature distribution $\mathbb{P}_X$ satisfies the strong density assumption, regression function $\eta$ is H\"{o}lder continuous with parameters $C > 0$ and $\alpha = 1$ (so that $\eta$ is Lipschitz continuous), and label $Y$ is bounded in the interval $[y_{\min}, y_{\max}]$ for some constants $y_{\min}$ and $y_{\max}$. In this case, since $Y$ is bounded in this way,
\[
\text{Var}(Y \mid X = x) \le \frac{(y_{\max} - y_{\min})^2}4,
\]
so Theorem~5.2 of~\citet{gyorfi_book} applies with $\sigma = \frac{y_{\max} - y_{\min}}2$.

For the comparison here, it helps to define the following quantity:
\[
\sigma_{\max}^2 \triangleq \sup_{z\in S^*} |\eta(z)|^2 = \max\{y_{\min}^2, y_{\max}^2\}.
\]
Then note that
\[
4\sigma^2 = (y_{\max} - y_{\min})^2 \le (2 \max\{|y_{\min}|,|y_{\max}|\})^2 = 4\sigma_{\max}^2,
\]
where the inequality holds because the largest the distance could be between $y_{\max}$ and $y_{\min}$ is if we took the distance between $\max\{|y_{\min}|,|y_{\max}|\}$ and $-\max\{|y_{\min}|,|y_{\max}|\}$.  In particular, $\sigma^2 = \mathcal{O}(\sigma_{\max}^2)$.

Now let's look at what the two different theoretical guarantees ask for to ensure that $\mathbb{E}[|\widehat{\eta}_{\text{NN}(h)}(X) - \eta(X)|] \le \varepsilon$:
\begin{itemize}

\item Corollary~\ref{cor:radius-NN-regression-simple-case}: We choose the threshold distance to be
\[
h=\frac{\varepsilon}{4C} = \Theta\Big(\frac{\varepsilon}{C}\Big),
\]
and to ensure that condition~\eqref{eq:radius-NN-regression-simple-case-constraint-on-n} is met, we ask that
\begin{align*}
n &= \bigg\lceil
\frac{8}{p_{\min} (\frac{\varepsilon}{4C})^d}
\Big(2 + \frac{4(y_{\max} - y_{\min})^2}{\varepsilon^2}\Big) \\
&\quad\;\,\times
\log \frac{8\max\{|y_{\min}|,|y_{\max}|,y_{\max} - y_{\min}\}}{\varepsilon}
\bigg\rceil \\
& = \Theta\Big(\frac{C^d\sigma^2}{\varepsilon^{d+2}} \log \frac{\sigma_{\max}}{\varepsilon} \Big).
\end{align*}

\item Theorem~5.2 of~\citet{gyorfi_book}: Let's set the number of training data as small as possible for the theorem's result to still apply:
\begin{align*}
n &= (c'')^{\frac{d+2}2} C^d \cdot \frac{\sigma^2 + \sup_{z\in S^*} |\eta(z)|^2}{\varepsilon^{d+2}} \\
  &= (c'')^{\frac{d+2}2} C^d \cdot \frac{\sigma^2 + \sigma_{\max}^2}{\varepsilon^{d+2}} \\
  &= \Theta\Big( \frac{C^d \sigma_{\max}^2}{\varepsilon^{d+2}} \Big).
\end{align*}
Then the choice of threshold distance is
\begin{align*}
h &= c'\Big(\frac{\sigma^2 + \sup_{z\in S^*} |\eta(z)|^2}{C^2}\Big)^{\frac1{d+2}}\frac1{\big[(c'')^{\frac{d+2}2} C^d \cdot \frac{\sigma^2 + \sup_{z\in S^*} |\eta(z)|^2}{\varepsilon^{d+2}}\big]^{\frac1{d+2}}} \\
&= \frac{c'}{\sqrt{c''} C} \varepsilon \\
&= \Theta\Big(\frac{\varepsilon}{C}\Big).
\end{align*}

\end{itemize}
Both results guarantee low expected error ${\mathbb{E}[|\widehat{\eta}(X) - \eta(X)|]} \le \varepsilon$ using training data size $n = \widetilde{\mathcal{O}}(\frac{C^d \sigma_{\max}^2}{\varepsilon^{d+2}})$ and threshold distance $h = \Theta(\frac{\varepsilon}{C})$. Corollary~\ref{cor:radius-NN-regression-simple-case} uses slightly more training data by a multiplicative factor of $\log\frac{\sigma_{\max}}{\varepsilon}$, and outside of the log, it depends on $\sigma^2$ instead of $\sigma_{\max}^2$ as needed by Theorem~5.2 of~\citet{gyorfi_book}.

\section{Theoretical Guarantees for Kernel Regression}
\label{sec:kernel-regression-theory}

We now discuss the general case of kernel regression.  We assume that $K$ monotonically decreases and actually becomes 0 after some normalized distance $\tau>h$ (\ie, $K(s) \le {\ind\{s \le \tau\}}$).  We assume the regression function $\eta$ to be H\"{o}lder continuous.

\begin{ftheorem}
[Kernel regression pointwise error]
\label{thm:kernel-regression-rate-of-convergence}
Under assumptions \assumpTechnical, \assumpDecay, and \assumpHolder, let $x\in\text{supp}(\mathbb{P}_{X})$ be a feature vector, $\varepsilon>0$ be an error tolerance in estimating expected label $\eta(x)={\mathbb{E}[Y\mid X=x]}$, and $\delta\in(0,1)$ be a probability tolerance. Suppose that $Y\in[y_{\min},y_{\max}]$ for some constants $y_{\min}$ and $y_{\max}$.  Let $\phi>0$ be any normalized distance for which $K(\phi)>0$. If the chosen bandwidth satisfies
\[
h \le \frac{1}{\tau}\Big(\frac{\varepsilon}{2C}\Big)^{1/\alpha},
\]
and the number of training data satisfies
\begin{align*}
&n
 \ge
   \max
     \Big\{
       \frac{2\log\frac{4}{\delta}}
           {[K(\phi)
             \mathbb{P}_X(\mathcal{B}_{x,\phi h})
            ]^2}, \\
&\qquad\qquad\;\;
      \frac{8[\max\{|y_{\min}|,|y_{\max}|\}+(y_{\max}-y_{\min})]^2\log\frac{4}{\delta}}
           {\varepsilon^2[K(\phi)
                         \mathbb{P}_X(\mathcal{B}_{x,\phi h})
                         ]^4}\Big\},
\end{align*}
then with probability at least $1-\delta$, kernel regression with kernel~$K$ and bandwidth~$h$ has error
\[
|\widehat{\eta}_{K}(x;h)-\eta(x)|\le\varepsilon.
\]
\end{ftheorem}
For example, fixed-radius NN regression estimate $\widehat{\eta}_{\text{NN}(h)}$ with threshold distance $h$ satisfies the conditions of Theorem~\ref{thm:kernel-regression-rate-of-convergence} with $\phi=1$, $K(\phi)=1$, and $\tau=1$. Thus, Theorem~\ref{thm:kernel-regression-rate-of-convergence} says that with the number of training data $n$ satisfying
\begin{align*}
&n\ge\max\Big\{\frac{2\log\frac{4}{\delta}}{[\mathbb{P}_X(\mathcal{B}_{x,h})]^{2}}, \\
&\qquad\qquad\;\;\frac{8[\max\{|y_{\min}|,|y_{\max}|\}+(y_{\max}-y_{\min})]^{2}\log\frac{4}{\delta}}{\varepsilon^{2}[\mathbb{P}_X(\mathcal{B}_{x,h})]^{4}}\Big\},
\end{align*}
and threshold distance $h$ satisfying
\[
h \le \Big(\frac{\varepsilon}{2C}\Big)^{1/\alpha},
\]
then $|\widehat{\eta}_{\text{NN}(h)}(x)-\eta(x)|\le\varepsilon$ with probability at least $1-\delta$. Ignoring constant factors, this result requires more training data (specifically because of the dependence on $\mathbb{P}_X(\mathcal{B}_{x,h})$ to the second and fourth powers) and is thus a weaker performance guarantee than Theorem~\ref{thm:radius-NN-rate-of-convergence}. However, both theorems place the same constraint on the choice for threshold distance: $h=\mathcal{O}\big((\frac{\varepsilon}{2C})^{1/\alpha}\big)$.

\begin{ftheorem}
[Kernel regression expected error]
\label{thm:kernel-regression-expected-error-rate-of-convergence}
Let $\varepsilon > 0$ and $\delta\in(0,1)$.  Under assumptions \assumpTechnical, \assumpDecay, and \assumpHolder, suppose that $Y\in[y_{\min},y_{\max}]$ for some constants $y_{\min}$ and $y_{\max}$.  Let $p_{\min} > 0$ and $d > 0$. Let $\phi>0$ be any normalized distance for which $K(\phi)>0$. If the chosen bandwidth satisfies $h \le \frac{1}{\tau}(\frac{\varepsilon}{2C})^{1/\alpha}$, and the number of training data satisfies satisfies 
\begin{align*}
&n
 \ge
   \max
     \Big\{
       \frac{4\log\frac{4}{\delta}}
           {[K(\phi) p_{\min} \phi^d h^d]^2}, \\
&\qquad\qquad\;\;
      \frac{16[\max\{|y_{\min}|,|y_{\max}|\}+(y_{\max}-y_{\min})]^2\log\frac{4}{\delta}}
           {\varepsilon^2[K(\phi) p_{\min} \phi^d h^d]^4}\Big\},
\end{align*}
then kernel regression with kernel $K$ and bandwidth $h$ has expected error
\begin{align*}
&\mathbb{E}[| \widehat{\eta}_{K}(X;h) - \eta(X) |] \\
&\le \varepsilon + \max\{|y_{\min}|,|y_{\max}|,y_{\max} - y_{\min}\} \\
&\qquad\quad\times\big(\mathbb{P}_X(\mathcal{X}_{\text{bad}}(\mathbb{P}_X;p_{\min}, d, {\textstyle \frac1{\tau} (\frac{\varepsilon}{2C})^{1/\alpha}})) + \delta\big).
\end{align*}
\end{ftheorem}

\section{Proofs}
\label{sec:regression-proofs}

We now present proofs of the main nonasymptotic theoretical guarantees presented as well as a few auxiliary results.

\subsection{Proof of Theorem~\ref{thm:k-NN-regression-rate-of-convergence}}
\label{sec:proof-k-NN-regression-rate-of-convergence}

The proof is a slight variation on Chaudhuri and Dasgupta's $k$-NN classification result \citep[proof of Theorem~1]{chaudhuri_dasgupta_2014}.  In this section, we abbreviate $\widehat{\eta}$ to mean $\widehat{\eta}_{k\text{-NN}}$. Importantly, we first present a proof of the theorem for the case where ties happen with probability~0 as is, \textit{without any form of tie breaking}; the precise statement of this no-ties guarantee is in Lemma~\ref{lem:k-NN-rate-of-convergence-general-no-ties} below.  At the very end of this section, we discuss how to modify the proof to handle the random tie breaking described in Section~\ref{sec:k-NN-regression}, yielding a generalization of Theorem~\ref{thm:k-NN-regression-rate-of-convergence} that is given by Lemma~\ref{lem:k-NN-rate-of-convergence-general}.

As suggested by the overview of results in Section~\ref{sec:regression-theory-overview}, in the $k$-NN regression case, the $k$ nearest neighbors are strictly inside the shaded ball of Figure~\ref{fig:k-NN-helper}, and because we want to reason about being strictly inside a ball and not on the boundary of it, we occasionally reason about open balls.  We denote an open ball centered at $c\in\mathcal{X}$ with radius $r>0$ by
\[
\mathcal{B}_{c,r}^o \triangleq \{ x'\in\mathcal{X} : \rho(c, x') \le r\}.
\]
We begin by proving the following general lemma.
\begin{flemma}
\label{lem:k-NN-rate-of-convergence-general-no-ties}
Under assumption \assumpTechnical, suppose that in finding nearest neighbors, ties happen with probability~0 without any form of tie breaking.  Let $x\in\text{supp}(\mathbb{P}_{X})$ be a feature vector, and $\eta(x)=\mathbb{E}[Y\mid X=x]\in\mathbb{R}$ be the expected label value for $x$. Let $\varepsilon>0$ be an error tolerance in estimating $\eta(x)$, and $\delta\in(0,1)$ be a probability tolerance. Suppose that $Y\in[y_{\min},y_{\max}]$ for some constants $y_{\min}$ and $y_{\max}$, and
\[
\lim_{r\downarrow0}\mathbb{E}[Y\mid X\in \mathcal{B}_{x,r}^o]=\eta(x).
\]
(Note that this is not quite the same as the Besicovitch condition; this difference will be reconciled when we talk about handling ties.) Let $\gamma\in(0,1)$. Then there exists a threshold distance $h^{*}\in(0,\infty)$ such that for any smaller distance $h\in(0,h^{*})$ and with the number of nearest neighbors satisfying $k\le(1-\gamma)n\mathbb{P}_{X}(\mathcal{B}_{x,h})$, then with probability at least
\[
1-2\exp\Big(-\frac{k\varepsilon^{2}}{2(y_{\max}-y_{\min})^{2}}\Big)-\exp\Big(-\frac{\gamma^{2}n\mathbb{P}_{X}(\mathcal{B}_{x,h})}{2}\Big),
\]
we have
\[
|\widehat{\eta}(x)-\eta(x)|\le\varepsilon.
\]
Furthermore, if the function $\eta$ is H\"{o}lder continuous with parameters $C>0$ and $\alpha>0$, then we can take $h^{*}=(\frac{\varepsilon}{2C})^{1/\alpha}$ and the above guarantee holds for $h=h^{*}$ as well.
\end{flemma}
\subsubsection{Proof of Lemma~\ref{lem:k-NN-rate-of-convergence-general-no-ties}}
Fix $x\in\text{supp}(\mathbb{P}_{X})$ (the derivation often involves looking at the probability of $X$ landing in a ball centered at $x$; with $x\in\text{supp}(\mathbb{P}_{X})$, we can rest assured that such probabilities are strictly positive). Let $\varepsilon>0$. We upper-bound the error $|\widehat{\eta}(x)-\eta(x)|$ with the triangle inequality:
\begin{align*}
|\widehat{\eta}(x)-\eta(x)| & =|(\widehat{\eta}(x)-\mathbb{E}_{n|\widetilde{X}}[\widehat{\eta}(x)])+(\mathbb{E}_{n|\widetilde{X}}[\widehat{\eta}(x)]-\eta(x))|\\
 & \le|\widehat{\eta}(x)-\mathbb{E}_{n|\widetilde{X}}[\widehat{\eta}(x)]|+|\mathbb{E}_{n|\widetilde{X}}[\widehat{\eta}(x)]-\eta(x)|,
\end{align*}
where $\mathbb{E}_{n|\widetilde{X}}[\widehat{\eta}(x)]$ is an expectation over training points $(X_{1},Y_{1})$, $\dots$, $(X_{n},Y_{n})$, conditioned on the \mbox{$(k+1)$-st} nearest neighbor $\widetilde{X} := X_{(k+1)}(x)$ (so that $\mathbb{E}_{n|\widetilde{X}}[\widehat{\eta}(x)]$ is a function of the random variable $\widetilde{X}$). The proof proceeds by showing that, with high probability, each of the two right-hand side terms is upper-bounded by~$\frac{\varepsilon}{2}$.

Let's first show when, with high probability, ${|\widehat{\eta}(x)-\mathbb{E}_{n|\widetilde{X}}[\widehat{\eta}(x)]|\le\frac{\varepsilon}{2}}$.
\begin{flemma}[\citet{chaudhuri_dasgupta_2014}, Lemma~10 slightly reworded]
\label{lem:k-NN-rate-of-convergence-helper1}
When ties happen with probability~0,
\begin{align*}
\mathbb{P}_n\Big(|\widehat{\eta}(x)-\mathbb{E}_{n|\widetilde{X}}[\widehat{\eta}(x)]|\ge\frac{\varepsilon}{2}\Big)& \le2\exp\Big(-\frac{k\varepsilon^{2}}{2(y_{\max}-y_{\min})^{2}}\Big),
\end{align*}
where $\mathbb{P}_n$ denotes the probability distribution over sampling the $n$ training data.
\end{flemma}
In particular, with probability at least $1-2\exp\big(-\frac{k\varepsilon^{2}}{2(y_{\max}-y_{\min})^{2}}\big)$, we have $|\widehat{\eta}(x)-\mathbb{E}_{n|\widetilde{X}}[\widehat{\eta}(x)]|\le\frac{\varepsilon}{2}$.
\begin{proof}
The randomness in the problem can be described as follows:
\begin{enumerate}

\item Sample a feature vector $\widetilde{X}\in\mathcal{X}$ from the marginal distribution of the \mbox{$(k+1)$-st} nearest neighbor of $x$.

\item Sample $k$ feature vectors i.i.d.~from $\mathbb{P}_X$ conditioned on landing in the ball $\mathcal{B}_{x,\rho(x,\widetilde{X})}^o$.

\item Sample $n-k-1$ feature vectors i.i.d.~from $\mathbb{P}_X$ conditioned on landing in $\mathcal{X}\setminus \mathcal{B}_{x,\rho(x,\widetilde{X})}^o$.

\item Randomly permute the $n$ feature vectors sampled.

\item For each feature vector $X_i$ generated, sample its label $Y_i$ based on conditional distribution $\mathbb{P}_{Y\mid X=X_{i}}$.

\end{enumerate}
Then the points sampled in step~2 are precisely the $k$ nearest neighbors of $x$, and their $Y$ values are i.i.d.~with expectation $\mathbb{E}_{n|\widetilde{X}}[\widehat{\eta}(x)]={\mathbb{E}\big[Y\,\big|\,X\in \mathcal{B}_{x,\rho(x,\widetilde{X})}^o\big]}$.

Finally, since the $Y$ values for the $k$ nearest neighbors are i.i.d.~and are each bounded between $y_{\min}$ and $y_{\max}$, Hoeffding's inequality yields
\[
\mathbb{E}_{n|\widetilde{X}}\Big[\ind\big\{|\widehat{\eta}(x)-\mathbb{E}_{n|\widetilde{X}}[\widehat{\eta}(x)]|\ge\frac{\varepsilon}{2}\big\}\Big]\le2\exp\Big(-\frac{k\varepsilon^{2}}{2 (y_{\max}-y_{\min})^{2}}\Big).
\]
Applying $\mathbb{E}_{\widetilde{X}}$ to both sides and noting that $\mathbb{E}_{\widetilde{X}}\mathbb{E}_{n|\widetilde{X}}=\mathbb{E}_n$, we get
\[
\underbrace{\mathbb{E}_n\Big[\ind\big\{|\widehat{\eta}(x)-\mathbb{E}_{n|\widetilde{X}}[\widehat{\eta}(x)]|\ge\frac{\varepsilon}{2}\big\}\Big]}_{=\mathbb{P}_n\big(|\widehat{\eta}(x)-\mathbb{E}_{n|\widetilde{X}}[\widehat{\eta}(x)]|\ge\frac{\varepsilon}{2}\big)}\le2\exp\Big(-\frac{k\varepsilon^{2}}{2 (y_{\max}-y_{\min})^{2}}\Big).
\]
This finishes the proof.
\end{proof}
The description of randomness in Lemma~\ref{lem:k-NN-rate-of-convergence-helper1}'s proof is carefully crafted. We briefly remark on ties and why the \mbox{$(k+1)$-st} nearest neighbor matters:
\begin{itemize}

\item
\textit{Why ties happening with probability~0 matters:}
since there are no ties, the $k$ nearest neighbors are unambiguous, and the conditional distributions in steps~2 and~3 do indeed properly distinguish between the distribution of the $k$ nearest neighbors and the other training data.

To see why ties are problematic, consider the case when some of the $k$ nearest neighbors are tied with the \mbox{$(k+1)$-st} nearest neighbor.  Then the ball $\mathcal{B}_{x,\rho(x,\widetilde{X})}^o$ necessarily excludes training data among the $k$ nearest neighbors that are tied to the \mbox{$(k+1)$-st} nearest neighbor, even though the goal of constructing this open ball is to have it exactly contain the $k$ nearest neighbors!

\item
\textit{Why we are using an open ball with radius to the \mbox{$(k+1)$-st} nearest neighbor and not a closed ball to the $k$-th nearest neighbor:}
If instead step~1 samples the $k$-th nearest neighbor, and step~2 samples $k-1$ feature vectors from a closed ball with radius going to the $k$-th nearest neighbor, then we would only have control over the $k-1$ nearest neighbors' feature vectors being i.i.d.~(after conditioning on $\widetilde{X}$) and not all $k$.

\end{itemize}
Next we show when, with high probability, $|\mathbb{E}_{n|\widetilde{X}}[\widehat{\eta}(x)]-\eta(x)|\le\frac{\varepsilon}{2}$. As discussed in the proof of Lemma~\ref{lem:k-NN-rate-of-convergence-helper1},
\[
\mathbb{E}_{n|\widetilde{X}}[\widehat{\eta}(x)]=\mathbb{E}\big[Y\,\big|\,X\in \mathcal{B}_{x,\rho(x,X_{(k+1)}(x))}^o\big].
\]
Suppose that we could show that there exists some $h>0$ such that
\begin{equation}
|\mathbb{E}[Y\mid X\in \mathcal{B}_{x,r}^o]-\eta(x)|\le\frac{\varepsilon}{2}\qquad\text{for all }r\in(0,h].\label{eq:lem-k-NN-rate-of-convergence-helper2-helper1}
\end{equation}
Then provided that $\rho(x,X_{(k+1)}(x))\le h$, then
\begin{align}
|\mathbb{E}_{n|\widetilde{X}}[\widehat{\eta}(x)]-\eta(x)| &=|\mathbb{E}\big[Y\,\big|\,X\in \mathcal{B}_{{\scriptstyle x, \underbrace{\rho(x,X_{(k+1)}(x))}_{\le h}}}^o\big]-\eta(x)| \nonumber \\
(\text{inequality~}\eqref{eq:lem-k-NN-rate-of-convergence-helper2-helper1}) & \le\frac{\varepsilon}{2}, \label{eq:k-NN-no-tie-helper}
\end{align}
in which case we would be done.

Before establishing the existence of $h$, we first show that for any distance $r>0$, with high probability we can ensure that ${\rho(x,X_{(k+1)}(x))\le r}$.  Thus, once we do show that $h$ exists, we also know that we can ensure that $\rho(x,X_{(k+1)})\le h$ with high probability.
\begin{flemma}[Slight variant on Lemma~9 of~\citet{chaudhuri_dasgupta_2014}]
\label{lem:k-NN-rate-of-convergence-helper2-1}Let $r>0$ and $\gamma\in(0,1)$.
For positive integer $k\le{(1-\gamma)n\mathbb{P}_{X}(\mathcal{B}_{x,r})}$,
\[
\mathbb{P}_n\big(\rho(x,X_{(k+1)}(x))\ge r\big)\le\exp\Big(-\frac{\gamma^{2}n\mathbb{P}_{X}(\mathcal{B}_{x,r})}{2}\Big),
\]
where $\mathbb{P}_n$ is the probability distribution over sampling the $n$ training data.
\end{flemma}
Thus, inequality $\rho(x,X_{(k+1)}(x))\le r$ holds with probability at least ${1-\exp\big(-\frac{\gamma^{2}n\mathbb{P}_{X}(\mathcal{B}_{x,r})}{2}\big)}$.
\begin{proof}
Fix $r>0$ and $\gamma\in(0,1)$. Let $N_{x,r}$ be the number of training points that land in the closed ball $\mathcal{B}_{x,r}$. In particular, note that $N_{x,r}\sim\text{Binomial}\big(n,\mathbb{P}_{X}(\mathcal{B}_{x,r})\big)$.  Then by a Chernoff bound for the binomial distribution, for any positive integer $k\le(1-\gamma)n\mathbb{P}_{X}(\mathcal{B}_{x,r})$, we have
\begin{align}
\mathbb{P}_n(N_{x,r}\le k) & \le\exp\bigg(-\frac{\big(n\mathbb{P}_{X}(\mathcal{B}_{x,r})-k\big)^{2}}{2n\mathbb{P}_{X}(\mathcal{B}_{x,r})}\bigg)\nonumber \\
 & \le\exp\bigg(-\frac{\big(n\mathbb{P}_{X}(\mathcal{B}_{x,r})-(1-\gamma)n\mathbb{P}_{X}(\mathcal{B}_{x,r})\big)^{2}}{2n\mathbb{P}_{X}(\mathcal{B}_{x,r})}\bigg)\nonumber \\
 & =\exp\Big(-\frac{\gamma^{2}n\mathbb{P}_{X}(\mathcal{B}_{x,r})}{2}\Big).\label{eq:lem-k-NN-rate-of-convergence-helper2-helper2}
\end{align}
Observe that if the number of training points $N_{x,r}$ that land in $\mathcal{B}_{x,r}$ is at most $k$, then it means that the \mbox{$(k+1)$-st} nearest-neighbor $X_{(k+1)}(x)$ must be at least a distance $r$ away from $x$. Hence, the event $\{N_{x,r}\le k\}$ is a superset of the event $\{\rho(x,X_{(k+1)}(x))\ge r\}$ since the former event happening implies that the latter event happens.  This superset relation combined with inequality~\eqref{eq:lem-k-NN-rate-of-convergence-helper2-helper2} yields
\[
\mathbb{P}_n\big(\rho(x,X_{(k+1)}(x))\ge r\big)\le\mathbb{P}_n(N_{x,r}\le k)\le\exp\Big(-\frac{\gamma^{2}n\mathbb{P}_{X}(\mathcal{B}_{x,r})}{2}\Big).\qedhere
\]
\end{proof}
Now we show which distance $h$ ensures that inequality~\eqref{eq:lem-k-NN-rate-of-convergence-helper2-helper1} holds. When we only know that
\[
\lim_{r\downarrow0}\mathbb{E}[Y\mid X\in \mathcal{B}_{x,r}^o]=\eta(x),
\]
then the definition of a limit implies that there exists $h^{*}>0$ (that depends on $x$ and $\varepsilon$) such that
\[
|\mathbb{E}[Y\mid X\in \mathcal{B}_{x,h}^o]-\eta(x)|<\frac{\varepsilon}{2}\qquad\text{for all }h\in(0,h^{*}),
\]
\ie, inequality~\eqref{eq:lem-k-NN-rate-of-convergence-helper2-helper1} holds, and so indeed we have $|\mathbb{E}_{n|\widetilde{X}}[\widehat{\eta}(x)]-\eta(x)|\le\frac{\varepsilon}{2}$ as shown earlier in inequality~\eqref{eq:k-NN-no-tie-helper}.

If we know that $\eta$ is H\"{o}lder continuous with parameters $C$ and $\alpha$, then for $h^{*}=(\frac{\varepsilon}{2C})^{1/\alpha}$ and $h\in(0,h^{*}]$, we have
\begin{align}
|\mathbb{E}[Y\mid X\in \mathcal{B}_{x,h}^o]-\eta(x)|
& =\bigg|\frac{\int_{\widetilde{x}\in \mathcal{B}_{x,h}^o}\mathbb{E}[Y\mid X=\widetilde{x}]d\mathbb{P}_{X}(\widetilde{x})}{\mathbb{P}_{X}(\mathcal{B}_{x,h}^o)}-\eta(x)\bigg|\nonumber \\
& =\bigg|\frac{\int_{\widetilde{x}\in \mathcal{B}_{x,h}^o}\eta(\widetilde{x})d\mathbb{P}_{X}(\widetilde{x})}{\mathbb{P}_{X}(\mathcal{B}_{x,h}^o)}-\eta(x)\bigg|\nonumber \\
& =\bigg|\frac{\int_{\widetilde{x}\in \mathcal{B}_{x,h}^o}(\eta(\widetilde{x})-\eta(x))d\mathbb{P}_{X}(\widetilde{x})}{\mathbb{P}_{X}(\mathcal{B}_{x,h}^o)}\bigg|\nonumber \\
(\text{Jensen's inequality}) & \le\frac{\int_{\widetilde{x}\in \mathcal{B}_{x,h}^o}|\eta(\widetilde{x})-\eta(x)|d\mathbb{P}_{X}(\widetilde{x})}{\mathbb{P}_{X}(\mathcal{B}_{x,h}^o)}\nonumber \\
& \le\frac{\int_{\widetilde{x}\in \mathcal{B}_{x,h}}\sup_{x'\in \mathcal{B}_{x,h}^o}|\eta(x')-\eta(x)|d\mathbb{P}_{X}(\widetilde{x})}{\mathbb{P}_{X}(\mathcal{B}_{x,h}^o)}\nonumber \\
& =\Big(\sup_{x'\in \mathcal{B}_{x,h}^o}|\eta(x')-\eta(x)|\Big)\underbrace{\frac{\int_{\widetilde{x}\in \mathcal{B}_{x,h}^o}d\mathbb{P}_{X}(\widetilde{x})}{\mathbb{P}_{X}(\mathcal{B}_{x,h}^o)}}_{1}\nonumber \\
(\text{H\" older continuity}) & \le Ch^{\alpha}\nonumber \\
& \le\frac{\varepsilon}{2}.\label{eq:k-NN-rate-of-convergence-holder}
\end{align}
Thus, inequality~\eqref{eq:lem-k-NN-rate-of-convergence-helper2-helper1} holds, and so $|\mathbb{E}_{n|\widetilde{X}}[\widehat{\eta}(x)]-\eta(x)|\le\frac{\varepsilon}{2}$.

Putting together all the pieces and specifically also union-bounding over the bad events of Lemmas~\ref{lem:k-NN-rate-of-convergence-helper1} and~\ref{lem:k-NN-rate-of-convergence-helper2-1} (plugging in $r=h$ for Lemma~\ref{lem:k-NN-rate-of-convergence-helper2-1}), we obtain Lemma~\ref{lem:k-NN-rate-of-convergence-general-no-ties}.

\subsubsection{Handling Ties}

By how open (and closed) balls are defined for metric spaces, they cannot separate out tied points. The resolution is to use a different definition of an open ball that makes use of the random tie breaking procedure's generated priority random variables $Z_1,\dots,Z_n\overset{\text{i.i.d.}}{\sim}\text{Uniform}[0,1]$ (\cf, Section~\ref{sec:k-NN-regression}). Specifically, for center $c\in\mathcal{X}$, radius $r>0$, and priority $z\in[0,1]$, we define the open ball
\begin{align*}
&\mathcal{B}_{c,r,z}^o \\
&\triangleq
\{ (x', z') \in \mathcal{X}\times[0,1]
   : \text{either~}\rho(c,x') < r\text{~or~}(\rho(c,x') = r\text{~and~}z'<z) \}.
\end{align*}
The following modifications enable the proof above to carry through with random tie breaking:
\begin{itemize}

\item
\textit{Modifications to the description of randomness in Lemma~\ref{lem:k-NN-rate-of-convergence-helper1}:}
In step~1 of the description of randomness above, we sample $(\widetilde{X}, \widetilde{Z}) \in \mathcal{X}\times[0,1]$ from the marginal distribution of the \mbox{$(k+1)$-st} nearest neighbor of $x$, where $\widetilde{X}$ is the feature vector and its random priority is $\widetilde{Z}$. In steps~2 and~3, we use the ball $\mathcal{B}_{x,\rho(x,\widetilde{X}),\widetilde{Z}}^o$ instead of $\mathcal{B}_{x,\rho(x,\widetilde{X})}^o$. Also, in steps~2,~3, and~5, we sample both feature vectors and priorities. Lastly, the $k$-NN regression estimate has expectation $\mathbb{E}_{n|\widetilde{X}}[\widehat{\eta}(x)]={\mathbb{E}\big[Y\,\big|\,X\in \mathcal{B}_{x,\rho(x,\widetilde{X}),\widetilde{Z}}^o\big]}$ instead of $\mathbb{E}_{n|\widetilde{X}}[\widehat{\eta}(x)]=\mathbb{E}\big[Y\,\big|\,X\in \mathcal{B}_{x,\rho(x,\widetilde{X})}^o\big]$.

\item
\textit{Enforcing $|\mathbb{E}_{n|\widetilde{X}}[\widehat{\eta}(x)]-\eta(x)| \le \frac{\varepsilon}2$:}
Given the change in the description of randomness to account for random tie breaking, we now have
\[
\mathbb{E}_{n|\widetilde{X}}[\widehat{\eta}(x)]=\mathbb{E}\big[Y\,\big|\,X\in \mathcal{B}_{x,\rho(x,X_{(k+1)}(x)),Z_{(k+1)}(x)}^o\big],
\]
where $Z_{(k+1)}(x)$ is the priority random variable of the \mbox{$(k+1)$-st} nearest neighbor to $x$.  Thus, we modify the derivation of inequality~\eqref{eq:k-NN-no-tie-helper} by replacing $\mathcal{B}_{x,\rho(x,X_{(k+1)}(x))}^o$ with $\mathcal{B}_{x,\rho(x,X_{(k+1)}(x)),Z_{(k+1)}(x)}^o$.  However, now the problem is that inequality~\eqref{eq:lem-k-NN-rate-of-convergence-helper2-helper1} no longer implies the modified inequality~\eqref{eq:k-NN-no-tie-helper}. The fix is to ask for inequality~\eqref{eq:lem-k-NN-rate-of-convergence-helper2-helper1} to hold using closed balls instead of open balls:
\begin{equation}
|\mathbb{E}[Y\mid X\in \mathcal{B}_{x,r}]-\eta(x)|\le\frac{\varepsilon}{2}\qquad\text{for all }r\in(0,h].
\label{eq:lem-k-NN-rate-of-convergence-helper2-helper1-variant}
\end{equation}
In fact, this closed ball version implies the open ball version of inequality~\eqref{eq:lem-k-NN-rate-of-convergence-helper2-helper1} using a similar argument as in Lemma~25 of~\citet{chaudhuri_dasgupta_2014}.  Then inequalities~\eqref{eq:lem-k-NN-rate-of-convergence-helper2-helper1} and \eqref{eq:lem-k-NN-rate-of-convergence-helper2-helper1-variant} both holding implies that
\[
|\mathbb{E}[Y\mid X\in \mathcal{B}_{x,r,z}^o]-\eta(x)|\le\frac{\varepsilon}{2}\qquad\text{for all }z\in[0,1],r\in(0,h],
\]
which now does imply inequality~\eqref{eq:k-NN-no-tie-helper}, modified to use the ball $\mathcal{B}_{x,\rho(x,X_{(k+1)}(x)),Z_{(k+1)}(x)}^o$. The reason why having both inequalities~\eqref{eq:lem-k-NN-rate-of-convergence-helper2-helper1} and~\eqref{eq:lem-k-NN-rate-of-convergence-helper2-helper1-variant} holding (again, the latter implies the former) resolves the issue is that $\mathbb{E}[Y \mid X \in \mathcal{B}_{x,r,z}^o]$ is a convex combination of ${\mathbb{E}[Y \mid X \in \mathcal{B}_{x,r}^o]}$ and ${\mathbb{E}[Y \mid X \in \mathcal{B}_{x,r}]}$ (\cf, \citealt[Lemma~24]{chaudhuri_dasgupta_2014}). Thus when ${\mathbb{E}[Y \mid X \in \mathcal{B}_{x,r}^o]}$ and ${\mathbb{E}[Y \mid X \in \mathcal{B}_{x,r}]}$ are each within~$\frac{\varepsilon}2$ of~$\eta(x)$, then any convex combination of these two conditional expectations is also within~$\frac{\varepsilon}2$ of~$\eta(x)$.

As for ensuring that inequality~\eqref{eq:lem-k-NN-rate-of-convergence-helper2-helper1-variant} holds, we ask that
\[
\lim_{r\downarrow0}\mathbb{E}[Y\mid X \in \mathcal{B}_{x,r}]
=\eta(x).
\]
Then the definition of a limit implies that there exists $h^{*}>0$ (that depends on $x$ and $\varepsilon$) such that for all $h\in(0,h^*)$, we have
\[
|\mathbb{E}[Y\mid X\in \mathcal{B}_{x,h}]-\eta(x)|<\frac{\varepsilon}{2}.
\]
Alternatively we could assume the stronger assumption of H\"{o}lder continuity (note that the derivation of inequality~\eqref{eq:k-NN-rate-of-convergence-holder} works if open ball~$\mathcal{B}_{x,h}^o$ is replaced by closed ball $\mathcal{B}_{x,h}$).

\end{itemize}
The above modifications lead to the following lemma, which is actually a more general statement than Theorem~\ref{thm:k-NN-regression-rate-of-convergence}.  As with our exposition of $k$-NN regression earlier, we implicitly assume random tie breaking occurs if there are ties without explicitly stating it in the lemma.
\begin{flemma}
\label{lem:k-NN-rate-of-convergence-general}
Under assumptions \assumpTechnical~and \assumpBesicovitch, let $x\in\text{supp}(\mathbb{P}_{X})$ be a feature vector, and $\eta(x)=\mathbb{E}[Y\mid X=x]\in\mathbb{R}$ be the expected label value for $x$. Let $\varepsilon>0$ be an error tolerance in estimating $\eta(x)$, and $\delta\in(0,1)$ be a probability tolerance. Suppose that $Y\in[y_{\min},y_{\max}]$ for some constants $y_{\min}$ and $y_{\max}$.  Let $\gamma\in(0,1)$. Then there exists a threshold distance $h^{*}\in(0,\infty)$ such that for any smaller distance $h\in(0,h^{*})$ and with the number of nearest neighbors satisfying $k\le(1-\gamma)n\mathbb{P}_{X}(\mathcal{B}_{x,h})$, then with probability at least
\[
1-2\exp\Big(-\frac{k\varepsilon^{2}}{2(y_{\max}-y_{\min})^{2}}\Big)-\exp\Big(-\frac{\gamma^{2}n\mathbb{P}_{X}(\mathcal{B}_{x,h})}{2}\Big),
\]
we have
\[
|\widehat{\eta}(x)-\eta(x)|\le\varepsilon.
\]
Furthermore, if the function $\eta$ satisfies assumption \assumpHolder, then we can take $h^{*}=(\frac{\varepsilon}{2C})^{1/\alpha}$ and the above guarantee holds for $h=h^{*}$ as well.
\end{flemma}
Theorem~\ref{thm:k-NN-regression-rate-of-convergence} follows as a corollary of Lemma~\ref{lem:k-NN-rate-of-convergence-general}, where we set $\gamma=\frac{1}{2}$ and note that each of the two bad event probabilities can be upper-bounded by $\frac{\delta}{2}$ with conditions on $n$ and $k$:
\begin{align*}
n\ge\frac{8}{\mathbb{P}_{X}(\mathcal{B}_{x,h})}\log\frac{2}{\delta}\;\, & \Rightarrow\;\,\exp\Big(-\frac{\gamma^{2}n\mathbb{P}_{X}(\mathcal{B}_{x,h})}{2}\Big)\le\frac{\delta}{2},\\
k\ge\frac{2(y_{\max}-y_{\min})^{2}}{\varepsilon^{2}}\log\frac{4}{\delta}\;\, & \Rightarrow\;\,2\exp\Big(-\frac{k\varepsilon^{2}}{2(y_{\max}-y_{\min})^{2}}\Big)\le\frac{\delta}{2}.\tag*{\qed}
\end{align*}

\subsection{Proof of Theorem~\ref{thm:k-NN-regression-expectation-rate-of-convergence}}
\label{sec:proof-k-NN-regression-expectation-rate-of-convergence}

We use the same ideas as in the ending of the proof of Theorem~1 in~\cite{chaudhuri_dasgupta_2014}.  In this section, we abbreviate~$\widehat{\eta}$ to mean $\widehat{\eta}_{k\text{-NN}}$.  We use $\mathbb{E}_{X}$ to denote the expectation over the test point $X$, $\mathbb{E}_n$ to denote the expectation over training data $(X_{1},Y_{1}),\dots,(X_{n},Y_{n})$, and $\mathbb{P}_n$ to denote the probability distribution over sampling the $n$ training data.  Also, throughout this proof, denote $h^{*}\triangleq(\frac{\varepsilon}{2C})^{1/\alpha}$.

By the law of total expectation,
\begin{align}
& \mathbb{E}[|\widehat{\eta}(X)-\eta(X)|] \nonumber \\
& =\underbrace{\mathbb{E}\big[|\widehat{\eta}(X)-\eta(X)|\;\big|\;|\widehat{\eta}(X)-\eta(X)|<\varepsilon\big]}_{\le\varepsilon}\underbrace{\mathbb{P}(|\widehat{\eta}(X)-\eta(X)|<\varepsilon)}_{\le1} \nonumber \\
& \quad+\underbrace{\mathbb{E}\big[|\widehat{\eta}(X)-\eta(X)|\;\big|\;|\widehat{\eta}(X)-\eta(X)|\ge\varepsilon\big]}_{\le y_{\max}-y_{\min}\text{ (worst case regression error)}}\mathbb{P}(|\widehat{\eta}(X)-\eta(X)|\ge\varepsilon) \nonumber \\
& \le\varepsilon+(y_{\max}-y_{\min})\mathbb{P}(|\widehat{\eta}(X)-\eta(X)|\ge\varepsilon). \label{eq:k-NN-expected-error-proof-helper1}
\end{align}
We next upper-bound $\mathbb{P}(|\widehat{\eta}(X)-\eta(X)|\ge\varepsilon)$. By the law of total probability,
\begin{align}
& \mathbb{P}(|\widehat{\eta}(X)-\eta(X)|\ge\varepsilon) \nonumber \\
& =\underbrace{\mathbb{P}\big(|\widehat{\eta}(X)-\eta(X)|\ge\varepsilon\;\big|\;X\in\mathcal{X}_{\text{bad}}(\mathbb{P}_X;p_{\min},d,h^{*})\big)}_{\le1} \nonumber \\
& \qquad\;\times\mathbb{P}\big(X\in\mathcal{X}_{\text{bad}}(\mathbb{P}_X;p_{\min},d,h^{*})\big) \nonumber \\
& \quad+\mathbb{P}\big(|\widehat{\eta}(X)-\eta(X)|\ge\varepsilon\;\big|\;X\in\mathcal{X}_{\text{good}}(\mathbb{P}_X;p_{\min},d,h^{*})\big) \nonumber \\
& \qquad\;\times\underbrace{\mathbb{P}\big(X\in\mathcal{X}_{\text{good}}(\mathbb{P}_X;p_{\min},d,h^{*})\big)}_{\le1} \nonumber \\
& \le\mathbb{P}\big(X\in\mathcal{X}_{\text{bad}}(\mathbb{P}_X;p_{\min},d,h^{*})\big) \nonumber \\
& \quad+\mathbb{P}\big(|\widehat{\eta}(X)-\eta(X)|\ge\varepsilon\;\big|\;X\in\mathcal{X}_{\text{good}}(\mathbb{P}_X;p_{\min},d,h^{*})\big). \label{eq:k-NN-expected-error-proof-helper2}
\end{align}
Most of the proof is on upper-bounding the second term on the right-hand side. To keep the notation from getting cluttered, henceforth, we assume we are conditioning on $X\in\mathcal{X}_{\text{good}}(\mathbb{P}_X;p_{\min},d,h^{*})$ without writing it into every probability and expectation term, \eg, we write the second term on the right-hand side above as ${\mathbb{P}(|\widehat{\eta}(X)-\eta(X)|\ge\varepsilon)}$.  Then with a specific choice of random variable $\Xi$ to be specified shortly:
\begin{align}
& \mathbb{P}(|\widehat{\eta}(X)-\eta(X)|\ge\varepsilon) \nonumber \\
& =\mathbb{E}_n\big[\mathbb{E}_{X}[\ind\{|\widehat{\eta}(X)-\eta(X)|\ge\varepsilon\}]\big] \nonumber \\
& =\mathbb{E}_n\Big[\mathbb{E}_{X}[\ind\{|\widehat{\eta}(X)-\eta(X)|\ge\varepsilon\}]\;\Big|\;\Xi<\frac{\delta}{2}\Big]\underbrace{\mathbb{P}_{n}\Big(\Xi<\frac{\delta}{2}\Big)}_{\le1} \nonumber \\
& \quad+\underbrace{\mathbb{E}_n\Big[\mathbb{E}_{X}[\ind\{|\widehat{\eta}(X)-\eta(X)|\ge\varepsilon\}]\;\Big|\;\Xi\ge\frac{\delta}{2}\Big]}_{\le1}\mathbb{P}_{n}\Big(\Xi\ge\frac{\delta}{2}\Big) \nonumber \\
& \le\mathbb{E}_n\Big[\mathbb{E}_{X}[\ind\{|\widehat{\eta}(X)-\eta(X)|\ge\varepsilon\}]\;\Big|\;\Xi<\frac{\delta}{2}\Big]+\mathbb{P}_{n}\Big(\Xi\ge\frac{\delta}{2}\Big). \label{eq:k-NN-expected-error-proof-helper3}
\end{align}
Let's explain what $\Xi$ is. First, note that for any $x\in\mathcal{X}_{\text{good}}(\mathbb{P}_X;p_{\min},d,h^{*})$,
\begin{align}
&\ind\big\{|\widehat{\eta}(x)-\eta(x)|\ge\varepsilon\big\} \nonumber \\
&\quad\le\underbrace{\ind\big\{|\widehat{\eta}(x)-\mathbb{E}_{n|\widetilde{X}}[\widehat{\eta}(x)]|\ge\frac{\varepsilon}{2}\big\}+\ind\big\{\rho(x,x_{(k+1)}(x))\ge h^{*}\big\}}_{\triangleq\text{BAD}(x,x_{1},\dots,x_{n},y_{1},\dots,y_{n})},
\label{eq:k-NN-expected-error-proof-helper4}
\end{align}
where $(x_{1},y_{1}),\dots,(x_{n},y_{n})$ are a specific instantiation of training data that determines $\widehat{\eta}$, and $x_{(k+1)}(x)$ is the \mbox{$(k+1)$-st} nearest training point to~$x$.  To see why this inequality is true, it suffices to show that if the right-hand side is 0, then so is the left-hand side. Note that $\text{BAD}(x,x_{1},\dots,x_{n},y_{1},\dots,y_{n})=0$ if and only if ${|\widehat{\eta}(x)-\mathbb{E}_{n|\widetilde{X}}[\widehat{\eta}(x)]|}<\frac{\varepsilon}{2}$ and $\rho(x,x_{(k+1)}(x))<h^{*}$; the latter implies that ${|\mathbb{E}_{n|\widetilde{X}}[\widehat{\eta}(x)]-\eta(x)|}\le\frac{\varepsilon}{2}$ as we already discussed in the proof of Theorem~\ref{thm:k-NN-regression-rate-of-convergence}. Hence, by the triangle inequality $|\widehat{\eta}(x)-\eta(x)|\le|\widehat{\eta}(x)-\mathbb{E}_{n|\widetilde{X}}[\widehat{\eta}(x)]|+|\mathbb{E}_{n|\widetilde{X}}[\widehat{\eta}(x)]-\eta(x)|<\frac{\varepsilon}{2}+\frac{\varepsilon}{2}=\varepsilon$, \ie, the left-hand side of the above inequality is also~0.

Then we define random variable $\Xi$ as
\[
\Xi\triangleq\mathbb{E}_{X}[\text{BAD}(X,X_{1},\dots,X_{n},Y_{1},\dots,Y_{n})],
\]
which is a function of the random training data $(X_{1},Y_{1}),\dots,(X_{n},Y_{n})$; the expectation is only over $X$ and not over the random training data.

We now show an upper bound for $\mathbb{P}_{n}(\Xi\ge\frac{\delta}{2})$, which is the second term in the right-hand side of~\eqref{eq:k-NN-expected-error-proof-helper3}. Given our choice of training data size $n$ and number of nearest neighbors $k$, then by Lemmas~\ref{lem:k-NN-rate-of-convergence-helper1} and~\ref{lem:k-NN-rate-of-convergence-helper2-1}, for $x\in\mathcal{X}_{\text{good}}(\mathbb{P}_X;p_{\min},d,h^{*})$,
\begin{align}
&\mathbb{E}_n\big[\text{BAD}(x,X_{1},\dots,X_{n},Y_{1},\dots,Y_{n})\big] \nonumber\\
&=\mathbb{E}_n\Big[\ind\big\{|\widehat{\eta}(x)-\mathbb{E}_{n|\widetilde{X}}[\widehat{\eta}(x)]|\ge\frac{\varepsilon}{2}\big\}+\ind\big\{\rho(x,X_{(k+1)}(x))\ge h^{*}\big\}\Big] \nonumber\\
&=\mathbb{P}_n\Big(|\widehat{\eta}(x)-\mathbb{E}_{n|\widetilde{X}}[\widehat{\eta}(x)]|\ge\frac{\varepsilon}{2}\Big)+\mathbb{P}_n\big(\rho(x,X_{(k+1)}(x))\ge h^{*}\big) \nonumber\\
&\le\frac{\delta^{2}}{4},
\label{eq:k-NN-expected-error-proof-helper5}
\end{align}
where the last step is due to the following:
\begin{itemize}

\item
Using Lemma~\ref{lem:k-NN-rate-of-convergence-helper1},
\begin{align*}
&k\ge\frac{4(y_{\max}-y_{\min})^{2}}{\varepsilon^{2}}\log\frac{4}{\delta}\;\, \\
&\qquad\Rightarrow\;\,\mathbb{P}_n\Big(|\widehat{\eta}(x)-\mathbb{E}_{n|\widetilde{X}}[\widehat{\eta}(x)]|\ge\frac{\varepsilon}{2}\Big)\le\frac{\delta^{2}}{8}.
\end{align*}

\item
Using Lemma~\ref{lem:k-NN-rate-of-convergence-helper2-1} (with $r=h^*$ and $\gamma = 1/2$) and noting that $\mathbb{P}_X(\mathcal{B}_{x,r}) \ge p_{\min} (h^*)^d$ for $x \in \mathcal{X}_{\text{good}}(\mathbb{P}_X;p_{\min}, d, (\frac{\varepsilon}{2C})^{1/\alpha})$,
\begin{gather*}
n\ge\frac{16}{p_{\min}(h^*)^d}\log\frac{\sqrt{8}}{\delta} \ge \frac{16}{\mathbb{P}_X(\mathcal{B}_{x,h^*})}\log\frac{\sqrt{8}}{\delta}
\quad\; \\ \Rightarrow\;\,\mathbb{P}_n\big(\rho(x,X_{(k+1)}(x))\ge h^*\big)\le\frac{\delta^{2}}{8}.
\end{gather*}
Note that Lemma~\ref{lem:k-NN-rate-of-convergence-helper2-1} (with $r=h^*$ and $\gamma=1/2$) further requires $k \le \frac12 n \mathbb{P}_X(\mathcal{B}_{x,h^*})$, which is satisfied since we ask for the more stringent condition $k \le \frac12 n p_{\min} (h^*)^d$.

\end{itemize}
Then by Markov's inequality,
\begin{align}
\mathbb{P}_{n}\big(\Xi\ge\frac{\delta}{2}\big) & \le\frac{\mathbb{E}_n[\Xi]}{\delta/2} \nonumber \\
& =\frac{\mathbb{E}_n[\mathbb{E}_{X}[\text{BAD}(X,X_{1},\dots,X_{n},Y_{1},\dots,Y_{n})]]}{\delta/2} \nonumber \\
& =\frac{\mathbb{E}_{X}[\mathbb{E}_n[\text{BAD}(X,X_{1},\dots,X_{n},Y_{1},\dots,Y_{n})]]}{\delta/2} \nonumber \\
(\text{by inequality~\eqref{eq:k-NN-expected-error-proof-helper5}}) & \le\frac{\mathbb{E}_{X}[\delta^2/4]}{\delta/2} \nonumber \\
& =\frac{\delta}{2}. \label{eq:k-NN-expected-error-proof-helper6}
\end{align}
Thus, combining inequalities \eqref{eq:k-NN-expected-error-proof-helper3}, \eqref{eq:k-NN-expected-error-proof-helper4}, and \eqref{eq:k-NN-expected-error-proof-helper6}, we have
\begin{align}
& \mathbb{P}(|\widehat{\eta}(X)-\eta(X)|\ge\varepsilon) \nonumber \\
& \le\mathbb{E}_n\Big[\mathbb{E}_{X}[\underbrace{\ind\{|\widehat{\eta}(X)-\eta(X)|\ge\varepsilon\}}_{\le\text{BAD}(X,X_{1},\dots,X_{n},Y_{1},\dots,Y_{n})}]\;\Big|\;\Xi<\frac{\delta}{2}\Big]+\underbrace{\mathbb{P}_{n}\Big(\Xi\ge\frac{\delta}{2}\Big)}_{\le\frac{\delta}{2}} \nonumber \\
& \le\mathbb{E}_n\Big[\underbrace{\mathbb{E}_{X}[\text{BAD}(X,X_{1},\dots,X_{n},Y_{1},\dots,Y_{n})]}_{=\Xi}\;\Big|\;\Xi<\frac{\delta}{2}\Big]+\frac{\delta}{2} \nonumber \\
& \le\frac{\delta}{2}+\frac{\delta}{2} \nonumber \\
& =\delta. \label{eq:k-NN-expected-error-proof-helper7}
\end{align}
Again, we are conditioning on the event $X\in\mathcal{X}_{\text{good}}(\mathbb{P}_X;p_{\min},d,h^{*})$, \ie, we have ${\mathbb{P}\big(|\widehat{\eta}(X)-\eta(X)|\ge\varepsilon \mid X\in\mathcal{X}_{\text{good}}(\mathbb{P}_X;p_{\min},d,h^{*})\big)} \le \delta$, which combined with inequalities \eqref{eq:k-NN-expected-error-proof-helper1} and \eqref{eq:k-NN-expected-error-proof-helper2} yields
\begin{align*}
& \mathbb{E}[|\widehat{\eta}(X)-\eta(X)|]\\
& \le\varepsilon+(y_{\max}-y_{\min})\big(\mathbb{P}\big(X\in\mathcal{X}_{\text{bad}}(\mathbb{P}_X;p_{\min},d,h^{*})\big)\\
& \qquad\qquad\qquad\quad+\mathbb{P}\big(|\widehat{\eta}(X)-\eta(X)|\ge\varepsilon\;\big|\;X\in\mathcal{X}_{\text{good}}(\mathbb{P}_X;p_{\min},d,h^{*})\big)\big)\\
& \le\varepsilon+(y_{\max}-y_{\min})\big(\mathbb{P}\big(X\in\mathcal{X}_{\text{bad}}(\mathbb{P}_X;p_{\min},d,h^{*})\big)+\delta\big). \tag*{\qed}
\end{align*}

\subsection{Proof of Proposition~\ref{claim:good-region-big-case}}
\label{sec:good-region-big-case-pf}

The proof is nearly identical to part of the proof presented of Proposition~3.1~of \citet{gadat_2016} (which shows a setting in which the strong minimal mass assumption \citep{gadat_2016} is equivalent to the strong density assumption \citep{audibert_2007}).

Under the strong density region of set $\mathcal{A} \subseteq \text{supp}(\mathbb{P}_X)$, for $x\in\mathcal{A}$ and $r \in (0,r^*]$,
\[
\mathbb{P}_X(\mathcal{B}_{x,r})
= \int_{\mathcal{B}_{x,r}}
	p_X(t)dt
\ge p_{\text{L}} \lambda(\mathcal{B}_{x,r} \cap \text{supp}(\mathbb{P}_X))
\ge p_{\text{L}} c_0 \underbrace{\lambda(\mathcal{B}_{x,r})}_{v_d r^d},
\]
where $v_d$ is the volume of a $d$-dimensional unit closed ball (such as~$\mathcal{B}_{x,1}$); note that $v_d$ depends on the choice of metric $\rho$. The above inequality implies that $\mathcal{A} \subseteq \mathcal{X}_{\text{good}}(\mathbb{P}_X;p_{\text{L}} c_0 v_d, d, r^*)$.

\subsection{Rephrasing of Theorem~2 of~\texorpdfstring{\citet{kohler_2006}}{Kohler et al.~[2006]}}
\label{sec:how-to-rephrase-kohler}

We reproduce Theorem~2 of~\citet{kohler_2006} in its original form (except using notation from this monograph) and explain how we rephrase it to obtain the version we presented in Theorem~\ref{thm:kohler-rephrased}.

\begin{ftheorem}[\citet{kohler_2006}, Theorem~2]
Let $\varepsilon > 0$ be an estimation error tolerance. Suppose that:
\begin{itemize}

\item The regression function $\eta$ is both bounded in absolute value by some constant $L\ge1$, and also H\"{o}lder continuous with parameters $C\ge1$ and $\alpha\in(0,1]$.

\item The feature space is in $\mathbb{R}^d$ for some $d > 2\alpha$, and the feature distribution $\mathbb{P}_X$ satisfies $\mathbb{E}[\|X\|^{\beta}] < \infty$ for some constant $\beta > \frac{2\alpha d}{d - 2\alpha}$, where $\|\cdot\|$ is the Euclidean norm.

\item The conditional variance of the label given a feature vector is bounded: $\sup_{x\in\mathbb{R}^d}\text{Var}(Y \mid X=x) \le \sigma^2$ for some $\sigma > 0$.

\end{itemize}
Then there exists a constant $c$ that depends on $d$, $\alpha$, $\beta$, $L$, $\sigma^2$, and $\mathbb{E}[\|X\|^\beta]$ such that if the number of nearest neighbors is set to
\[
k = \Bigg\lceil \frac{n^{\frac{2\alpha}{2\alpha+d}}}{C^{\frac{2d}{2\alpha+d}}} \Bigg\rceil,
\]
then
\begin{equation}
\mathbb{E}[(\widehat{\eta}_{k\text{-NN}}(X) - \eta(X))^2]
\le c \cdot \frac{C^{\frac{2d}{2\alpha+d}}}{n^{\frac{2\alpha}{2\alpha+d}}}.
\label{eq:kohler-expected-squared-error}
\end{equation}
\end{ftheorem}
Converting a guarantee in terms of squared error to absolute value error is straightforward with Jensen's inequality:
\[
\mathbb{E}[(\widehat{\eta}_{k\text{-NN}}(X) - \eta(X))^2]
\ge (\mathbb{E}[|\widehat{\eta}_{k\text{-NN}}(X) - \eta(X)|])^2.
\]
Thus, inequality~\eqref{eq:kohler-expected-squared-error} implies that
\begin{equation}
\mathbb{E}[|\widehat{\eta}_{k\text{-NN}}(X) - \eta(X)|]
\le
 \sqrt{\mathbb{E}[(\widehat{\eta}_{k\text{-NN}}(X) - \eta(X))^2]}
\le
 \sqrt{ c \cdot \frac{C^{\frac{2d}{2\alpha+d}}}{n^{\frac{2\alpha}{2\alpha+d}}} },
\label{eq:kohler-abs-value-error}
\end{equation}
which is at most $\varepsilon$ when
\[
n\ge\frac{c^{\frac{d}{2\alpha}+1}C^{\frac{d}{\alpha}}}{\varepsilon^{\frac{d}{\alpha}+2}}.
\]
This yields the statement of Theorem~\ref{thm:kohler-rephrased}.

\subsection{Proof of Theorem \ref{thm:k-NN-regression-covering-number}}

The proof will follow primarily from Theorem \ref{thm:k-NN-regression-rate-of-convergence}. Recall that we had established that the critical distance in which we want the $k$ nearest neighbors to be found from a test point is $h^* = (\frac{\varepsilon}{2C})^{1/\alpha}$, using H\"{o}lder continuity of regression function $\eta$. Under the assumption that $\XX$ is a compact Polish space, it follows that $\P_X$ has a finite  covering number for any $\delta \in (0,1)$ and $r > 0$. For a given $\delta \in (0,1)$, let $N = N(\P_X, \frac{h^*}2, \delta)$ be the $(\frac{h^*}2, \delta)$-covering number of $\P_X$. Let the corresponding $N$ balls be denoted as $\BB_1,\dots, \BB_N$ with $\BB_i \subset \XX$ for $i \in \{1,\dots,N\}$.  By how the covering number is defined,
\begin{align*}
\P_X\bigg(\bigcup_{i=1}^N \BB_i\bigg) & \geq 1-\delta.
\end{align*}
As stated in the theorem statement, we have 
\begin{align*}
k & = \bigg\lceil \frac{2 (y_{\max} - y_{\min})^2}{\varepsilon^2} \log \frac{4}{\delta}\bigg\rceil. \\
n & \geq  \frac{N}{\delta} \max\Big\{2k,~8 \log \frac{2}{\delta}\Big\}.
\end{align*}
We define the set of {\em bad} indices from $1,\dots, N$ as 
\begin{align*}
\bad & = \bigg\{ i \in \{1,\dots,N\} : \P_X(\BB_i) < \frac{1}{n} \max\Big\{2k,~8 \log \frac{2}{\delta}\Big\} \bigg\}.
\end{align*}
This set is carefully constructed based on conditions \eqref{eq:k-NN-regression-rate-of-convergence-constraint-on-n} and \eqref{eq:k-NN-regression-rate-of-convergence-constraint-on-k} of Theorem \ref{thm:k-NN-regression-rate-of-convergence}.

Next, we define sets $E_1, E_2, E_3 \subseteq \text{supp}(\mathbb{P}_X)$ as follows:
\begin{align*}
E_1 & = \bigcup_{i \in \bad} \BB_i, \\
E_2 & = \bigg(\bigcup_{i =1}^N \BB_i\bigg) \backslash E_1, \\
E_3 & = \text{supp}(\P_X) \backslash (E_1 \cup E_2).
\end{align*} 
Since $E_1 \cup E_2 = \cup_{i=1}^N \BB_i$, by the definition of a $(\frac{h^*}2, \delta)$-covering number, we have $\P_X(E_3) \leq \delta$. Using the definition of the set $\bad$ and the fact that $|\bad| \leq N$, we have
\begin{align*}
\P_X(E_1) & = \P_X\bigg(\bigcup_{i \in \bad} \BB_i\bigg) \\
& \leq \sum_{i \in \bad} \P_X(\BB_i) \\
& \leq |\bad| \Big(\max_{i \in \bad} \P_X(\BB_i)\Big) \\
& \leq \frac{N}{n} \max\Big\{2k,~8 \log \frac{2}{\delta}\Big\}.
\end{align*}
This gives us $\P_X(E_1) \leq \delta$. Since $\text{supp}(\P_X) = E_1 \cup E_2 \cup E_3$, and since $\P_X(E_1) \leq \delta$ and $\P_X(E_3) \leq \delta$, we have that $\P_X(E_2) \geq 1-2\delta$. Now for any $x \in \text{supp}(\P_X)$, if $x \in E_1 \cup E_3$, then a trivial upper bound on the regression error is $y_{\max} - y_{\min}$. On the other hand, if $x \in E_2$, we will argue that conditions \eqref{eq:k-NN-regression-rate-of-convergence-constraint-on-n} and \eqref{eq:k-NN-regression-rate-of-convergence-constraint-on-k} of Theorem \ref{thm:k-NN-regression-rate-of-convergence} are satisfied with $h = h^*$. To see this, note that for any $x \in E_2$, there exists $i \in \{1,\dots,N\}$ such that $i \notin \bad$ and $x \in \BB_i$. Since $\BB_i$ is ball of radius $h^*/2$ and is in $ \BB_i$, it follows that $\BB_i \subset \BB_{x, h^*}$. Therefore, 
\begin{align*}
\P_X(\BB_{x, h^*}) & \geq \P_X(\BB_i)
 \geq \frac{1}{n} \max\Big\{2k,~ 8 \log \frac{2}{\delta}\Big\}.
\end{align*}
This immediately implies inequality \eqref{eq:k-NN-regression-rate-of-convergence-constraint-on-n}. The choice of $k$ and the above implies inequality \eqref{eq:k-NN-regression-rate-of-convergence-constraint-on-k}. Therefore, Theorem \ref{thm:k-NN-regression-rate-of-convergence} implies that for any $x \in E_2$ and using our choices of $k$ and $n$, with probability at least $1-\delta$, 
\[
|\widehat{\eta}_{k\text{-NN}}(x)-\eta(x)|\le\varepsilon.
\]
When the above does not hold, we can bound the error by $y_{\max} - y_{\min}$. Putting everything together, we conclude that 
\[
\mathbb{E}[|\widehat{\eta}_{k\text{-NN}}(X) - \eta(X)|] \leq \varepsilon + 3 \delta (y_{\max} - y_{\min}).
\]

\subsection{Proof of Theorem~\ref{thm:radius-NN-rate-of-convergence}}

In this section, we abbreviate $\widehat{\eta}$ to mean $\widehat{\eta}_{\text{NN}(h)}$. Let $x\in\text{supp}(\mathbb{P}_{X})$, $h>0$, and $\varepsilon>0$. Note that
\[
\widehat{\eta}(x)=\frac{1}{N_{x,h}}\sum_{i=1}^{N_{x,h}}A_{i},
\]
where $N_{x,h}\sim\text{Binomial}\big(n,\mathbb{P}_{X}(\mathcal{B}_{x,h})\big)$ is the number of points that land in the ball $\mathcal{B}_{x,h}$, and $A_{1},A_{2},\dots$ are i.i.d.~with distribution given by that of $Y$ conditioned on $X\in \mathcal{B}_{x,h}$; for the corner case where $N_{x,h}=0$, we use the convention that $0/0=0$. We first establish that $N_{x,h}\ge(1-\gamma)n\mathbb{P}_{X}(\mathcal{B}_{x,h})$ (which is strictly positive, so that $N_{x,h}\ge1$) with high probability.
\begin{flemma}
\label{lem:radius-NN-helper1}For any $\gamma\in(0,1)$, we have
\[
\mathbb{P}\big(N_{x,h}\le(1-\gamma)n\mathbb{P}_{X}(\mathcal{B}_{x,h})\big)\le\exp\Big(-\frac{\gamma^{2}n\mathbb{P}_{X}(\mathcal{B}_{x,h})}{2}\Big).
\]
\end{flemma}
\begin{proof}
As we have already shown in inequality \eqref{eq:lem-k-NN-rate-of-convergence-helper2-helper2}, for any nonnegative integer $k\le(1-\gamma)n\mathbb{P}_{X}(\mathcal{B}_{x,h})$,
\begin{align*}
\mathbb{P}(N_{x,h}\le k)  & \le
 \exp\Big(-\frac{\gamma^{2}n\mathbb{P}_{X}(\mathcal{B}_{x,h})}{2}\Big).
\end{align*}
The event $\{N_{x,h}\le k\}$ is a superset of event ${\{N_{x,h}\le(1-\gamma)n\mathbb{P}_{X}(\mathcal{B}_{x,h})\}}$ as the former implies the latter, so
\begin{align*}
&\mathbb{P}\big(N_{x,h}\le(1-\gamma)n\mathbb{P}_{X}(\mathcal{B}_{x,h})\big) \\
&\quad\le\mathbb{P}(N_{x,h}\le k)\le\exp\Big(-\frac{\gamma^{2}n\mathbb{P}_{X}(\mathcal{B}_{x,h})}{2}\Big). \tag*{\qedhere}
\end{align*}
\end{proof}
Thus, with probability at least $1-\exp\big(-\frac{\gamma^{2}n\mathbb{P}_{X}(\mathcal{B}_{x,h})}{2}\big)$, we have $N_{x,h}>(1-\gamma)n\mathbb{P}_{X}(\mathcal{B}_{x,h})$, which is strictly positive given that $x\in\text{supp}(\mathbb{P}_{X})$ (which implies $\mathbb{P}_{X}(\mathcal{B}_{x,h})>0$). Henceforth we condition on this event happening.

Similar to how we prove the $k$-NN regression pointwise error guarantee (Theorem~\ref{thm:k-NN-regression-rate-of-convergence}), we note that the triangle inequality yields
\[
|\widehat{\eta}(x)-\eta(x)|\le|\widehat{\eta}(x)-\mathbb{E}_{n|N_{x,h}}[\widehat{\eta}(x)]|+|\mathbb{E}_{n|N_{x,h}}[\widehat{\eta}(x)]-\eta(x)|,
\]
where again we bound each of the right-hand terms by $\frac{\varepsilon}{2}$ with high probability.

We now show that $|\widehat{\eta}(x)-\mathbb{E}_{n|N_{x,h}}[\widehat{\eta}(x)]|\le\frac{\varepsilon}{2}$ with high probability.
\begin{flemma}
\label{lem:radius-NN-helper2}
We have
\begin{align*}
&\mathbb{P}_n\Big(
  |\widehat{\eta}(x)-\mathbb{E}_{n|N_{x,h}}[\widehat{\eta}(x)]|\ge\frac{\varepsilon}{2}
 ~\Big|~
 N_{x,h}>(1-\gamma)n\mathbb{P}_{X}(\mathcal{B}_{x,h})
\Big) \\
&\quad\le2\exp\Big(-\frac{(1-\gamma)n\mathbb{P}(\mathcal{B}_{x,h})\varepsilon^{2}}{2(y_{\max}-y_{\min})^{2}}\Big).
\end{align*}
\end{flemma}
\begin{proof}
Recall that conditioned on random variable $N_{x,h}$, we have $\widehat{\eta}(x)=\frac{1}{N_{x,h}}\sum_{i=1}^{N_{x,h}}A_{i}$ with $A_{1},\dots,A_{N_{x,h}}\overset{\text{i.i.d.}}{\sim}\mathbb{P}_{Y\mid X\in \mathcal{B}_{x,h}}$. Then by Hoeffding's inequality and the constraint that $N_{x,h}>(1-\gamma)n\mathbb{P}_X(\mathcal{B}_{x,h})$, we have
\begin{align*}
&\mathbb{E}_{n|N_{x,h}}\Big[\ind\big\{|\widehat{\eta}(x)-\mathbb{E}_{n|N_{x,h}}[\widehat{\eta}(x)]|\ge\frac{\varepsilon}{2}\big\}\Big] \\
&\quad\le2\exp\Big(-\frac{N_{x,h}\varepsilon^{2}}{2(y_{\max}-y_{\min})^{2}}\Big) \\
&\quad\le2\exp\Big(-\frac{(1-\gamma)n\mathbb{P}_X(\mathcal{B}_{x,h})\varepsilon^{2}}{2(y_{\max}-y_{\min})^{2}}\Big).
\end{align*}
Applying $\mathbb{E}_{N_{x,h}}$ to both sides, where the expectation is over the distribution of $N_{x,h}$ (conditioned to satisfy $N_{x,h}>(1-\gamma)n\mathbb{P}_X(\mathcal{B}_{x,h})$) yields the claim.
\end{proof}

We now show that $|\mathbb{E}_{n|N_{x,h}}[\widehat{\eta}(x)]-\eta(x)|\le\frac{\varepsilon}{2}$ with an appropriate choice of threshold distance $h$.
\begin{flemma}
\label{lem:radius-NN-helper3}
If the number of training points landing in $\mathcal{B}_{x,h}$ satisfies $N_{x,h}>(1-\gamma)n\mathbb{P}_{X}(\mathcal{B}_{x,h})$, then under the condition that
\[
\lim_{r\downarrow0}\mathbb{E}[Y\mid X\in \mathcal{B}_{x,r}]=\eta(x),
\]
there exists $h^{*}\in(0,\infty)$ such that so long as our choice of threshold distance satisfies $h<h^{*}$, then
\[
|\mathbb{E}_{n|N_{x,h}}[\widehat{\eta}(x)]-\eta(x)|\le\frac{\varepsilon}{2}.
\]
Furthermore, if the function $\eta$ is H\"{o}lder continuous with parameters $C>0$ and $\alpha>0$, then we can take $h^{*}=(\frac{\varepsilon}{2C})^{1/\alpha}$ and the above guarantee holds for $h=h^{*}$ as well.
\end{flemma}
\begin{proof}
Recall that conditioned on $N_{x,h}$ (which is at least one since we assume that $N_{x,h}>(1-\gamma)n\mathbb{P}_{X}(\mathcal{B}_{x,h})$), we have $\widehat{\eta}(x)=\frac{1}{N_{x,h}}\sum_{i=1}^{N_{x,h}}A_{i}$ with $A_{1},\dots,A_{N_{x,h}}\overset{\text{i.i.d.}}{\sim}\mathbb{P}_{Y\mid X\in \mathcal{B}_{x,h}}$. Then
\begin{align*}
\mathbb{E}_{n|N_{x,h}}[\widehat{\eta}(x)]
&=\frac{1}{N_{x,h}}\sum_{i=1}^{N_{x,h}}\mathbb{E}[A_{i}] \\
&=\frac{1}{N_{x,h}}\sum_{i=1}^{k}\mathbb{E}[Y\mid X\in \mathcal{B}_{x,h}] \\
&=\mathbb{E}[Y\mid X\in \mathcal{B}_{x,h}].
\end{align*}
Let's look at the case when we know that
\[
\lim_{r\downarrow0}\mathbb{E}[Y\mid X\in \mathcal{B}_{x,r}]=\eta(x).
\]
This implies that there exists $h^{*}>0$ (that depends on $x$ and $\varepsilon$) such that
\[
|\mathbb{E}[Y\mid X\in \mathcal{B}_{x,r}]-\eta(x)|<\frac{\varepsilon}{2}\qquad\text{for all }r\in(0,h^{*}).
\]
Hence, provided that $h\in(0,h^{*})$,
\[
|\mathbb{E}_{n|N_{x,h}}[\widehat{\eta}(x)]-\eta(x)|=|\mathbb{E}[Y\mid X\in \mathcal{B}_{x,h}]-\eta(x)|\le\frac{\varepsilon}{2}.
\]
If we know that $\eta$ is H\"{o}lder continuous with parameters $C$ and $\alpha$, then by a similar argument as we used in showing inequality~\eqref{eq:k-NN-rate-of-convergence-holder},
\begin{align*}
|\mathbb{E}_{n|N_{x,h}}[\widehat{\eta}(x)]-\eta(x)| & =|\mathbb{E}[Y\mid X\in \mathcal{B}_{x,h}]-\eta(x)|\\
 & \le\sup_{x'\in \mathcal{B}_{x,h}}|\eta(x')-\eta(x)|\\
 & \le Ch^{\alpha},
\end{align*}
which is at most $\frac{\varepsilon}{2}$ when $h\le(\frac{\varepsilon}{2C})^{1/\alpha}$, \ie, in this case we can take $h^{*}=(\frac{\varepsilon}{2C})^{1/\alpha}$.
\end{proof}
Putting together Lemmas~\ref{lem:radius-NN-helper1},~\ref{lem:radius-NN-helper2}, and~\ref{lem:radius-NN-helper3} (and union-bounding over the bad events of Lemmas~\ref{lem:radius-NN-helper1} and~\ref{lem:radius-NN-helper2}), we obtain the following lemma that is more general than Theorem~\ref{thm:radius-NN-rate-of-convergence}.
\begin{flemma}
Under assumptions \assumpTechnical~and \assumpBesicovitch, let $x\in\text{supp}(\mathbb{P}_{X})$ be a feature vector, $\varepsilon>0$ be an error tolerance in estimating $\eta(x)$, and $\delta\in(0,1)$ be a probability tolerance. Suppose that $Y\in[y_{\min},y_{\max}]$ for some constants $y_{\min}$ and $y_{\max}$.  Then there exists a distance $h^{*}\in(0,\infty)$ such that if our choice of threshold distance satisfies $h\in(0,h^{*})$, with probability at least
\[
1-\exp\Big(-\frac{\gamma^{2}n\mathbb{P}_{X}(\mathcal{B}_{x,h})}{2}\Big)-2\exp\Big(-\frac{(1-\gamma)n\mathbb{P}_X(\mathcal{B}_{x,h})\varepsilon^{2}}{2(y_{\max}-y_{\min})^{2}}\Big),
\]
we have $|\widehat{\eta}(x)-\eta(x)|\le\varepsilon$.

Furthermore, if the function $\eta$ satisfies assumption \assumpHolder, then we can take $h^{*}=(\frac{\varepsilon}{2C})^{1/\alpha}$ and the above guarantee holds for $h=h^{*}$ as well.
\end{flemma}
We obtain Theorem~\ref{thm:radius-NN-rate-of-convergence} by setting $\gamma=\frac12$ and noting that
\begin{align*}
n\ge\frac{8}{\mathbb{P}_{X}(\mathcal{B}_{x,h})}\log\frac{2}{\delta}\;\, & \Rightarrow\;\,\exp\Big(-\frac{n\mathbb{P}_{X}(\mathcal{B}_{x,h})}{8}\Big)\le\frac{\delta}{2},\\
n\ge\frac{4(y_{\max}-y_{\min})^{2}}{\mathbb{P}_X(\mathcal{B}_{x,h})\varepsilon^{2}}\log\frac{4}{\delta}\;\, & \Rightarrow\;\,2\exp\Big(-\frac{n\mathbb{P}_X(\mathcal{B}_{x,h})\varepsilon^{2}}{4(y_{\max}-y_{\min})^{2}}\Big)\le\frac{\delta}{2}. \tag*{\qed}
\end{align*}

\subsection{Proof of Theorem~\ref{thm:radius-NN-expectation-rate-of-convergence}}
\label{sec:proof-radius-NN-expectation-rate-of-convergence}

The proof is nearly identical to that of Theorem~\ref{thm:k-NN-regression-expectation-rate-of-convergence} (Section~\ref{sec:proof-k-NN-regression-expectation-rate-of-convergence}).  The main change is to replace inequality~\eqref{eq:k-NN-expected-error-proof-helper4} with
\begin{align*}
&\ind\{\widehat{\eta}_{\text{NN}(h)}(x) - \eta(x)| \ge \varepsilon\} \\
&\le \ind\{ N_{x,h} \le \frac12 n \mathbb{P}_X(\mathcal{B}_{x,h}) \}
     + \ind\{ |\widehat{\eta}_{\text{NN}(h)}(x) - \mathbb{E}_{n|N_{x,h}}[\widehat{\eta}_{\text{NN}(h)}(x)]| \ge \frac{\varepsilon}2 \},
\end{align*}
where random variable $N_{x,h}$ is the number of training points that land in the ball $\mathcal{B}_{x,h}$. We define the right-hand side of the above inequality as the new function $\text{BAD}$ that replaces the one in equation~\eqref{eq:k-NN-expected-error-proof-helper4}. For a fixed test point $x$, the expectation (over randomness in training data) of BAD can be upper-bounded by $\frac{\delta^2}4$ in a similar manner as in inequality~\eqref{eq:k-NN-expected-error-proof-helper5} by now using Lemmas~\ref{lem:radius-NN-helper1} and~\ref{lem:radius-NN-helper2} (both with $\gamma = 1/2$), along with the fact that $x \in \mathcal{X}_{\text{good}}(\mathbb{P}_X;p_{\min}, d, (\frac{\varepsilon}{2C})^{1/\alpha})$ (the threshold distance $h$ and the number of training data $n$ to use to ensure this $\frac{\delta^2}4$ bound are in the theorem statement).  The rest of the proof is the same with the notable exception that the worst-case regression error is $\max\{|y_{\min}|, |y_{\max}|, y_{\max} - y_{\min}\}$ instead of merely $y_{\max} - y_{\min}$, which was the case for $k$-NN regression.

\subsection{Proof of Theorem~\ref{thm:kernel-regression-rate-of-convergence}}

In this section, we abbreviate $\widehat{\eta}$ to mean $\widehat{\eta}_{K}(\cdot;h)$ for the given bandwidth $h>0$. Fix $x\in\mathcal{X}$. Note that
\[
\widehat{\eta}(x)=\frac{\frac{1}{n}\sum_{i=1}^{n}K(\frac{\rho(x,X_{i})}{h})Y_{i}}{\frac{1}{n}\sum_{i=1}^{n}K(\frac{\rho(x,X_{i})}{h})}.
\]
The numerator is the average of $n$ i.i.d.~terms each bounded between $y_{\min}$ and $y_{\max}$, and with expectation $A\triangleq\mathbb{E}[K(\frac{\rho(x,X)}{h})Y]$.  The denominator is the average of $n$ i.i.d.~terms each bounded between 0 and 1, and with expectation $B\triangleq\mathbb{E}[K(\frac{\rho(x,X)}{h})]$.  Thus, as $n$ grows large, $\widehat{\eta}(x)$ approaches $\frac{A}{B}=\frac{\mathbb{E}[K(\frac{\rho(x,X)}{h})Y]}{\mathbb{E}[K(\frac{\rho(x,X)}{h})]}$.

We decompose the regression error using the triangle inequality:
\begin{equation}
|\widehat{\eta}(x)-\eta(x)|\le\Big|\widehat{\eta}(x)-\frac{A}{B}\Big|+\Big|\frac{A}{B}-\eta(x)\Big|.
\label{eq:kernel-regression-triangle-ineq}
\end{equation}
The rest of the proof is on upper-bounding each of the two right-hand side terms by $\frac{\varepsilon}{2}$ with high probability.
\begin{flemma}
\label{lem:kernel-regression-rate-of-convergence-helper1}
Let $A\triangleq\mathbb{E}[K(\frac{\rho(x,X)}{h})Y]$ and $B\triangleq\mathbb{E}[K(\frac{\rho(x,X)}{h})]$. Let $\delta \in (0,1)$. With $\phi>0$ such that $K(\phi)>0$, and the number of training points satisfies
\begin{align}
n
&\ge
  \max\Big\{
    \frac{2\log\frac{4}{\delta}}{[K(\phi)\mathbb{P}_{X}(\mathcal{B}_{x,\phi h})]^{2}}, \nonumber \\
&\qquad\qquad\frac{8[\max\{|y_{\min}|,|y_{\max}|\}+(y_{\max}-y_{\min})]^{2}\log\frac{4}{\delta}}{\varepsilon^{2}[K(\phi)\mathbb{P}_{X}(\mathcal{B}_{x,\phi h})]^{4}}\Big\}.
\label{eq:kernel-regression-more-interpretable-error1-constraint-on-n}
\end{align}
then with probability at least $1-\delta$,
\[
\Big|\widehat{\eta}(x)-\frac{A}{B}\Big|\le\frac{\varepsilon}{2}.
\]
\end{flemma}
\begin{proof}
By Hoeffding's inequality,
\begin{align*}
\mathbb{P}\bigg(\bigg|\frac{1}{n}\sum_{i=1}^{n}K\Big(\frac{\rho(x,X_{i})}{h}\Big)Y_{i}-A\bigg|\ge (y_{\max} - y_{\min})\sqrt{\frac{\log\frac{4}{\delta}}{2n}} \bigg) & \le \frac{\delta}2, \\
\mathbb{P}\bigg(\bigg|\frac{1}{n}\sum_{i=1}^{n}K\Big(\frac{\rho(x,X_{i})}{h}\Big)-B\bigg|\ge \sqrt{\frac{\log\frac{4}{\delta}}{2n}} \bigg) & \le \frac{\delta}2.
\end{align*}
Thus, union bounding over these two bad events, we have, with probability at least $1-\delta$,
\[
\bigg|
  \frac{1}{n}
    \sum_{i=1}^n
      K\Big(\frac{\rho(x,X_{i})}{h}\Big) Y_{i}
  - A
\bigg|
\le (y_{\max} - y_{\min})\sqrt{\frac{\log\frac{4}{\delta}}{2n}}
\triangleq \varepsilon_1,
\]
and
\[
\bigg|
  \frac{1}{n}
    \sum_{i=1}^n
      K\Big(\frac{\rho(x,X_{i})}{h}\Big)
  - B
\bigg|
\le \sqrt{\frac{\log\frac{4}{\delta}}{2n}}
\triangleq \varepsilon_2.
\]
With both of these bounds holding, then
\[
\widehat{\eta}(x) =\frac{A\pm\varepsilon_{1}}{B\pm\varepsilon_{2}}.
\]
As we show next, by our assumption that $n$ is large enough, it turns out that $B-\varepsilon_{2}>0$. To see this, note that we have $\phi>0$ such that $K(\phi)>0$, and so $B$ satisfies the lower bound
\begin{align}
B & =\mathbb{E}\Big[K\Big(\frac{\rho(x,X)}{h}\Big)\Big] \nonumber \\
& =\mathbb{E}\Big[K\Big(\frac{\rho(x,X)}{h}\Big)\;\Big|\;\frac{\rho(x,X)}{h}\le\phi\Big]\mathbb{P}(\rho(x,X)\le\phi h) \nonumber \\
& \quad+\mathbb{E}\Big[K\Big(\frac{\rho(x,X)}{h}\Big)\;\Big|\;\frac{\rho(x,X)}{h}>\phi\Big]\mathbb{P}(\rho(x,X)>\phi h) \nonumber \\
& \ge\underbrace{\mathbb{E}\Big[K\Big(\frac{\rho(x,X)}{h}\Big)\;\Big|\;\frac{\rho(x,X)}{h}\le\phi\Big]}_{\ge K(\phi)\text{ since }K\text{ monotonically decreases}}\underbrace{\mathbb{P}(\rho(x,X)\le\phi h)}_{\mathbb{P}_{X}(\mathcal{B}_{x,\phi h})} \nonumber \\
& \ge\underbrace{K(\phi)}_{>0\text{ by assumption}}\times\underbrace{\mathbb{P}_{X}(\mathcal{B}_{x,\phi h})}_{>0\text{ since }x\in\text{supp}(\mathbb{P}_{X})}.
\label{eq:expected-kernel-weight-lower-bound}
\end{align}
Meanwhile, using condition~\eqref{eq:kernel-regression-more-interpretable-error1-constraint-on-n} on the training data size $n$, we have $n \le \frac{2\log\frac{4}{\delta}}{[K(\phi)\mathbb{P}_{X}(\mathcal{B}_{x,\phi h})]^{2}}$, which means that
\[
\varepsilon_2
=\sqrt{\frac{\log\frac4{\delta}}{2n}}\le\frac{1}{2}K(\phi)\mathbb{P}_{X}(\mathcal{B}_{x,\phi h})\le\frac{1}{2}B.
\]
Hence,
\[
B-\varepsilon_{2}\ge\frac{1}{2}B,
\]
which, combined with the observations that $|A| \le \max\{|y_{\min}|,|y_{\max}|\}$ and $B \le 1$, gives
\begin{align}
& \Big|\widehat{\eta}(x)-\frac{A}{B}\Big| \nonumber \\
& \le\max\Big\{\Big|\frac{A+\varepsilon_{1}}{B+\varepsilon_{2}}-\frac{A}{B}\Big|,\Big|\frac{A+\varepsilon_{1}}{B-\varepsilon_{2}}-\frac{A}{B}\Big|,\Big|\frac{A-\varepsilon_{1}}{B+\varepsilon_{2}}-\frac{A}{B}\Big|,\Big|\frac{A-\varepsilon_{1}}{B-\varepsilon_{2}}-\frac{A}{B}\Big|\Big\} \nonumber \\
&= \max\Big\{\Big|\frac{-\varepsilon_2 A+\varepsilon_1 B}{B(B+\varepsilon_2)}\Big|,\Big|\frac{\varepsilon_2 A+\varepsilon_1 B}{B(B-\varepsilon_2)}\Big|,\Big|\frac{-\varepsilon_2 A-\varepsilon_1 B}{B(B+\varepsilon_2)}\Big|,\Big|\frac{\varepsilon_2 A-\varepsilon_1 B}{B(B-\varepsilon_2)}\Big|\Big\} \nonumber \\
& \le\frac{\varepsilon_{2}|A|+\varepsilon_{1}B}{B(B-\varepsilon_{2})} \nonumber \\
& \le\frac{\varepsilon_{2}\max\{|y_{\min}|,|y_{\max}|\}+\varepsilon_{1}}{B\cdot\frac{1}{2}B} \nonumber \\
& =\frac{2(\max\{|y_{\min}|,|y_{\max}|\}+(y_{\max}-y_{\min}))}{B^{2}}\sqrt{\frac{\log\frac{4}{\delta}}{2n}}.
\label{eq:kernel-regression-rate-of-convergence-estimator-variation-proxy-helper}
\end{align}
Putting together inequalities~\eqref{eq:kernel-regression-rate-of-convergence-estimator-variation-proxy-helper} and~\eqref{eq:expected-kernel-weight-lower-bound} yields:
\[
\Big|\widehat{\eta}(x)-\frac{A}{B}\Big|
\le \frac{2(\max\{|y_{\min}|,|y_{\max}|\}+(y_{\max}-y_{\min}))}{[K(\phi)\mathbb{P}_X(\mathcal{B}_{x,\phi h})]^2}\sqrt{\frac{\log\frac{4}{\delta}}{2n}},
\]
which is at most $\frac{\varepsilon}{2}$ by our assumption that
\[
n \ge \frac{8[\max\{|y_{\min}|,|y_{\max}|\}+(y_{\max}-y_{\min})]^{2}\log\frac{4}{\delta}}{\varepsilon^{2}[K(\phi)\mathbb{P}_{X}(\mathcal{B}_{x,\phi h})]^{4}}.\qedhere
\]
\end{proof}
Now we show how to choose the bandwidth $h$ to get $\Big|\frac{A}{B}-\eta(x)\Big|\le\frac{\varepsilon}{2}$.

\begin{flemma}
\label{lem:kernel-regression-error2}
Denote $A\triangleq\mathbb{E}[K(\frac{\rho(x,X)}{h})Y]$ and $B=\mathbb{E}[K(\frac{\rho(x,X)}{h})]$.  Under assumption \assumpDecay, if $h \in (0, \frac{1}{\tau}(\frac{\varepsilon}{2C})^{1/\alpha}]$, then
\[
\Big|\frac{A}{B}-\eta(x)\Big|\le\frac{\varepsilon}{2}.
\]
\end{flemma}
\begin{proof}
Note that
\[
\mathbb{E}\Big[\frac{K(\frac{\rho(x,X)}{h})}{B}\Big]=\frac{\mathbb{E}[K(\frac{\rho(x,X)}{h})]}{B}=\frac{B}{B}=1.
\]
Let random variable $S=\frac{\rho(x,X)}{h}$ be the distance between $x$ and $X$ normalized by bandwidth $h$. We have
\begin{align}
\Big|\frac{A}{B}-\eta(x)\Big| & =\Big|\mathbb{E}\Big[\frac{K(\frac{\rho(x,X)}{h})}{B}Y\Big]-\mathbb{E}\Big[\frac{K(\frac{\rho(x,X)}{h})}{B}\Big]\eta(x)\Big| \nonumber \\
& =\Big|\mathbb{E}\Big[\frac{K(\frac{\rho(x,X)}{h})}{B}(Y-\eta(x))\Big]\Big| \nonumber \\
& =\bigg|\mathbb{E}_{S}\bigg[\mathbb{E}\Big[\frac{K(S)}{B}(Y-\eta(x))\;\Big|\;S\Big]\bigg]\bigg|\nonumber \\
& =\Big|\mathbb{E}_{S}\Big[\frac{K(S)}{B}\mathbb{E}[Y-\eta(x)\mid S]\Big]\Big| \nonumber \\
& =\Big|\mathbb{E}_{S}\Big[\frac{K(S)}{B}\big(\mathbb{E}[Y\mid S]-\eta(x)\big)\Big]\Big| \nonumber \\
(\text{Jensen's inequality}) & \le\mathbb{E}_{S}\Big[\Big|\frac{K(S)}{B}\big(\mathbb{E}[Y\mid S]-\eta(x)\big)\Big|\Big] \nonumber \\
(\text{since }K(s)\ge0\text{ for all }s\ge0) & =\mathbb{E}_{S}\Big[\frac{K(S)}{B}\big|\mathbb{E}[Y\mid S]-\eta(x)\big|\Big] \nonumber \\
& =\mathbb{E}_{S}\Big[\frac{K(S)}{B}\big|\mathbb{E}[Y\mid\rho(x,X)=hS]-\eta(x)\big|\Big] \nonumber \\
(\text{H\" older continuity}) & \le\mathbb{E}_{S}\Big[\frac{K(S)}{B}C(hS)^{\alpha}\Big] \nonumber \\
& = C h^\alpha \mathbb{E}_{S}\Big[\frac{K(S)}{B}S^{\alpha}\Big].
\label{eq:kernel-regression-rate-of-convergence-big-string-of-inequalities}
\end{align}
Since $K(S) = 0$ whenever $S > \tau$, then $K(S)S^\alpha \le K(S)\tau^\alpha$, and so
\begin{equation}
\mathbb{E}_S\Big[\frac{K(S)}{B}S^\alpha\Big]
\le \mathbb{E}_S\Big[\frac{K(S)}{B}\tau^\alpha\Big]
= \tau^\alpha \mathbb{E}_S\Big[\frac{K(S)}{B}\Big]
= \tau^\alpha \frac{\mathbb{E}_S[K(S)]}{B}
= 1,
\label{eq:kernel-regression-truncated-kernel-key-insight}
\end{equation}
where the last equality uses the fact that $\mathbb{E}_S[K(S)] = \mathbb{E}[K(\frac{\rho(x, X)}{h})] = B$.  Combining inequalities~\eqref{eq:kernel-regression-rate-of-convergence-big-string-of-inequalities} and~\eqref{eq:kernel-regression-truncated-kernel-key-insight},
\[
\Big|\frac{A}{B}-\eta(x)\Big|
\le C h^\alpha \mathbb{E}_S\Big[\frac{K(S)}{B}S^{\alpha}\Big]
\le C h^\alpha \tau^\alpha,
\]
which is at most $\frac{\varepsilon}{2}$ when
\[
h \le \frac{1}{\tau}\Big(\frac{\varepsilon}{2C}\Big)^{1/\alpha}. \qedhere
\]
\end{proof}
Combining Lemmas~\ref{lem:kernel-regression-rate-of-convergence-helper1} and~\ref{lem:kernel-regression-error2} yields Theorem~\ref{thm:kernel-regression-rate-of-convergence}. \qed

\subsection{Proof of Theorem~\ref{thm:kernel-regression-expected-error-rate-of-convergence}}

As with the proof of Theorem~\ref{thm:radius-NN-expectation-rate-of-convergence}, the proof here is nearly identical to that of Theorem~\ref{thm:k-NN-regression-expectation-rate-of-convergence} (Section~\ref{sec:proof-k-NN-regression-expectation-rate-of-convergence}).  The main change is to replace inequality~\eqref{eq:k-NN-expected-error-proof-helper4} with
\begin{align*}
&\ind\{\widehat{\eta}_{K}(x;h) - \eta(x)| \ge \varepsilon\} \\
&\le \ind\Bigg\{
\bigg|\frac{1}{n}\sum_{i=1}^{n}{\textstyle K\Big(\frac{\rho(x,X_{i})}{h}\Big)}Y_{i}-\mathbb{E}[{\textstyle K(\frac{\rho(x,X)}{h})}Y]\bigg|\ge (y_{\max} - y_{\min})\sqrt{\frac{\log\frac{4}{\delta}}{n}} \Bigg\} \\
&\quad
+
\ind\Bigg\{
\bigg|\frac{1}{n}\sum_{i=1}^{n}{\textstyle K\Big(\frac{\rho(x,X_{i})}{h}\Big)}-\mathbb{E}[{\textstyle K(\frac{\rho(x,X)}{h})}]\bigg|\ge \sqrt{\frac{\log\frac{4}{\delta}}{n}}
\Bigg\},
\end{align*}
where we define the right-hand side of the above inequality as the new function $\text{BAD}$ that replaces the one in equation~\eqref{eq:k-NN-expected-error-proof-helper4}. For a fixed test point $x$, the expectation (over randomness in training data) of BAD can be upper-bounded by $\frac{\delta^2}4$ by using Lemma~\ref{lem:kernel-regression-rate-of-convergence-helper1} (in the statement of this lemma, we substitute $\frac{\delta^2}4$ in place of $\delta$), along with the fact that $x \in \mathcal{X}_{\text{good}}(\mathbb{P}_X;p_{\min}, d, \frac1{\tau}(\frac{\varepsilon}{2C})^{1/\alpha})$ (the bandwidth $h$ and the number of training data $n$ to use to ensure that this $\frac{\delta^2}4$ bound holds are in the theorem statement).  As with fixed-radius NN regression, the worst-case regression error is $\max\{|y_{\min}|, |y_{\max}|, y_{\max} - y_{\min}\}$.

\subsection{How to Modify Theorems to Say What Error is Achievable}
\label{sub:how-to-translate-101}

The main theorem statements in this chapter specify how we should choose the training data set size~$n$ and the number of nearest neighbors~$k$ (or the bandwidth $h$ in the case of fixed-radius NN and kernel regression) to achieve a user-specified error tolerance~$\varepsilon$. In other words, $\varepsilon$ is treated as fixed, and we explain how to choose $n$ and $k$. We can actually translate the theorem statements to instead say what error $\varepsilon$ can be achieved, with $n$ and $k$ treated as fixed, albeit still with some restrictions on how $n$ and $k$ can be chosen up front.  In this section, we show how to do this translation for the guarantees on $k$-NN regression pointwise error (Theorem~\ref{thm:k-NN-regression-rate-of-convergence}) and expected error (Theorem~\ref{thm:k-NN-regression-expectation-rate-of-convergence}).  We do not do this translation for other theorems, although from our steps below, it should be clear that the translation can be done.

The translations rely on the following lemma involving the \mbox{Lambert}~$W$
function.

\begin{flemma}
\label{lem:lambert}
Let $W$ be the Lambert $W$ function. For any $z\in(0,b)$, $a>0$, $b>0$, and $c>0$, if
\[
z\ge b\exp\bigg(-\frac{1}{c}W\Big(\frac{cb^{c}}{a}\Big)\bigg),
\]
then
\[
z^{c}\ge a\log\frac{b}{z}\,.
\]
\end{flemma}
\begin{proof}
Suppose that $z\ge b\exp\big(-\frac{1}{c}W(\frac{cb^{c}}{a})\big)$. By rearranging terms to isolate $W$ on one side, we get
\[
W\Big(\frac{cb^{c}}{a}\Big)\ge c\log\frac{b}{z}.
\]
Note that $W^{-1}(s)=se^{s}$ for $s>0$ is a monotonically increasing function. Hence, applying $W^{-1}$ to both sides of the above inequality,
\[
\frac{cb^{c}}{a}\ge\Big(c\log\frac{b}{z}\Big)\exp\Big(c\log\frac{b}{z}\Big)=c\Big(\frac{b}{z}\Big)^{c}\log\frac{b}{z}.
\]
Another rearrangement of terms yields the claim.
\end{proof}

\vspace{0.3em}
\noindent
\textbf{Translating Theorem~\ref{thm:k-NN-regression-rate-of-convergence}.}
We aim to write the theorem statement to no longer use either $\varepsilon$ or $\delta$. To simplify matters, we constrain
\[
\delta=\frac{\varepsilon}{y_{\max}-y_{\min}}.
\]
Since the theorem requires $\delta\in(0,1)$, we thus will require
\[
\varepsilon\in(0,y_{\max}-y_{\min}).
\]
The theorem asks that we choose $n$ and $k$ so that
\[
n\overset{(i)}{\ge}\frac{8}{\mathbb{P}_{X}(\mathcal{B}_{x,h})}\log\frac{2}{\delta},
\]
and
\[
\frac{2(y_{\max}-y_{\min})^{2}}{\varepsilon^{2}}\log\frac{4}{\delta}\overset{(ii)}{\le}k\overset{(iii)}{\le}\frac{1}{2}n\mathbb{P}_{X}(\mathcal{B}_{x,h}).
\]
Using our choice of $\delta=\frac{\varepsilon}{y_{\max}-y_{\min}}$, by rearranging inequality $(i)$, we get
\begin{equation}
\varepsilon\ge2(y_{\max}-y_{\min})\exp\Big(-\frac{n\mathbb{P}_{X}(\mathcal{B}_{x,h})}{8}\Big).\label{eq:kNN-ptwise-translator-helper1}
\end{equation}
Plugging in $\delta=\frac{\varepsilon}{y_{\max}-y_{\min}}$ into inequality $(ii)$, and rearranging terms, we get
\[
\varepsilon^2 \ge \frac{2(y_{\max}-y_{\min})^{2}}{k}\log\frac{4(y_{\max}-y_{\min})}{\varepsilon}.
\]
Applying Lemma~\ref{lem:lambert} (and recalling that we require $\varepsilon<y_{\max}-y_{\min}$), we enforce the inequality above to hold by asking that
\begin{equation}
\varepsilon\ge4(y_{\max}-y_{\min})\exp\Big(-\frac{1}{2}W(16k)\Big),\label{eq:kNN-ptwise-translator-helper2}
\end{equation}
where $W$ is the Lambert $W$ function. Then Theorem 2.1 of~\citet{hoorfar_2008} yields
\[
W(16k)\ge\log(16k)-\log\log(16k).
\]
Thus, to ensure that inequality \eqref{eq:kNN-ptwise-translator-helper2} is satisfied, it suffices to ask that
\begin{align}
\varepsilon & \ge4(y_{\max}-y_{\min})\exp\Big(-\frac{1}{2}\big(\log(16k)-\log\log(16k)\big)\Big)\nonumber \\
 & =(y_{\max}-y_{\min})\sqrt{\frac{\log(16k)}{k}},\label{eq:kNN-ptwise-translator-helper3}
\end{align}
where the last line follows from a bit of algebra.

Putting together the two sufficient conditions on $\varepsilon$ given by inequalities \eqref{eq:kNN-ptwise-translator-helper1} and \eqref{eq:kNN-ptwise-translator-helper3}, we see that we can set
\begin{equation}
\varepsilon=\max\bigg\{2(y_{\max}-y_{\min})\exp\Big(-\frac{n\mathbb{P}_{X}(\mathcal{B}_{x,h})}{8}\Big),\;(y_{\max}-y_{\min})\sqrt{\frac{\log(16k)}{k}}\bigg\}.\label{eq:kNN-ptwise-translator-eps}
\end{equation}
However, recalling that we must have $\varepsilon<y_{\max}-y_{\min}$, then we impose some sufficient conditions on $n$ and $k$ (so that each of the two terms in the maximization above is at most $y_{\max}-y_{\min}$):
\[
n>\frac{8\log2}{\mathbb{P}_{X}(\mathcal{B}_{x,h})},\quad\text{and}\quad k\ge5.
\]
As a reminder, we still have to account for inequality $(iii)$. In summary, we can ask that
\[
n>\frac{8\log2}{\mathbb{P}_{X}(\mathcal{B}_{x,h})},
\quad\text{and}\quad 5\le k\le \frac{1}{2}n\mathbb{P}_{X}(\mathcal{B}_{x,h})
\]
to ensure that the conditions in Theorem \ref{thm:k-NN-regression-rate-of-convergence} hold. Then with probability at least $1-\delta=1-\frac{\varepsilon}{y_{\max}-y_{\min}}$ over randomness in the training data,
\[
|\widehat{\eta}_{k\text{-NN}}(x)-\eta(x)|\le\varepsilon,
\]
where $\varepsilon$ is given in equation \eqref{eq:kNN-ptwise-translator-eps}. In other words, treating test point $x$ as fixed and taking the expectation $\mathbb{E}_n$ over randomness in the training data,
\begin{align*}
\mathbb{E}_n[|\widehat{\eta}_{k\text{-NN}}(x)-\eta(x)|] & \le\varepsilon+\delta\times\!\!\!\!\!\!\!\!\!\!\!\!\!\!\!\!\underbrace{(y_{\max}-y_{\min})}_{\text{w.p.~}\delta\text{ we assume a worst-case error}}\\
& =\varepsilon+\frac{\varepsilon}{y_{\max}-y_{\min}}\cdot(y_{\max}-y_{\min})\\
& =2\varepsilon.
\end{align*}
Thus, the translated theorem is as follows.
\begin{ftheorem}
[$k$-NN regression pointwise error, alternative form]
Under assumptions \assumpTechnical~and \assumpBesicovitch, let $x\in\text{supp}(\mathbb{P}_{X})$ be a feature vector. Suppose that $Y\in[y_{\min},y_{\max}]$ for some constants $y_{\min}$ and $y_{\max}$.  There exists a threshold distance $h^{*}\in(0,\infty)$ such that for any smaller distance $h\in(0,h^{*})$, if the number of training points and the number of nearest neighbors satisfy
\begin{equation*}
n > \frac{8\log2}{\mathbb{P}_{X}(\mathcal{B}_{x,h})},
\quad\text{and}\quad 5\le k\le\frac{1}{2}n\mathbb{P}_{X}(\mathcal{B}_{x,h}),
\end{equation*}
then with probability at least $1-\frac{\varepsilon_{k,n}}{y_{\max}-y_{\min}}$, $k$-NN regression at point~$x$ has error
\[
|\widehat{\eta}_{k\text{-NN}}(x)-\eta(x)| \le \varepsilon_{k,n},
\]
where
\begin{equation*}
\varepsilon_{k,n}
=(y_{\max}-y_{\min})\max\bigg\{2\exp\Big(-\frac{n\mathbb{P}_{X}(\mathcal{B}_{x,h})}{8}\Big),\;\sqrt{\frac{\log(16k)}{k}}\bigg\}.
\end{equation*}
Put another way, the expected regression error over randomness in the training data at point $x$ is
\[
\mathbb{E}_n[|\widehat{\eta}_{k\text{-NN}}(x) - \eta(x)|]
\le 2\varepsilon_{k,n}.
\]
Furthermore, if the function $\eta$ satisfies assumption \assumpHolder, then we can take $h^{*}=(\frac{\varepsilon}{2C})^{1/\alpha}$ and the above guarantee holds for $h=h^{*}$ as well.
\end{ftheorem}

\vspace{0.3em}
\noindent
\textbf{Translating Theorem~\ref{thm:k-NN-regression-expectation-rate-of-convergence}.}
The translation strategy is the same as what we did above for Theorem~\ref{thm:k-NN-regression-rate-of-convergence}.  Theorem~\ref{thm:k-NN-regression-expectation-rate-of-convergence} asks that we choose $n$ and $k$ so that
\[
n\overset{(i)}{\ge}\Big(\frac{2C}{\varepsilon}\Big)^{d/\alpha}\frac{16}{p_{\min}}\log\frac{\sqrt{8}}{\delta},
\]
and
\[
\frac{4(y_{\max}-y_{\min})^{2}}{\varepsilon^{2}}\log\frac{4}{\delta}\overset{(ii)}{\le}k\overset{(iii)}{\le}\frac{1}{2}np_{\min}\Big(\frac{\varepsilon}{2C}\Big)^{d/\alpha}.
\]
Again, we choose $\delta=\frac{\varepsilon}{y_{\max}-y_{\min}}$, so the theorem requires $\varepsilon\in(0,y_{\max}-y_{\min})$. To handle inequality $(i)$, we plug in $\delta=\frac{\varepsilon}{y_{\max}-y_{\min}}$ and rearrange terms to get
\[
\varepsilon^{d/\alpha}\ge\frac{16(2C)^{d/\alpha}}{np_{\min}}\log\frac{\sqrt{8}(y_{\max}-y_{\min})}{\varepsilon}.
\]
Applying Lemma \ref{lem:lambert} (and noting that we require $\varepsilon<y_{\max}-y_{\min}$), the above inequality holds if
\begin{align}
\varepsilon & \ge
\sqrt{8}(y_{\max}-y_{\min})\exp\Bigg\{-\frac{\alpha}{d}W\bigg(\underbrace{\frac{np_{\min}d}{16\alpha}\Big[\frac{\sqrt{8}(y_{\max}-y_{\min})}{2C}\Big]^{d/\alpha}}_{\triangleq\xi}\bigg)\Bigg\}.\label{eq:kNN-expected-translator-helper3}
\end{align}
If $\xi\ge e$, which we can get by asking that
\begin{equation}
n\ge\frac{16\alpha e}{p_{\min}d}\Big[\frac{2C}{\sqrt{8}(y_{\max}-y_{\min})}\Big]^{d/\alpha},\label{eq:kNN-expected-translator-helper5}
\end{equation}
then Theorem 2.1 of \citet{hoorfar_2008} implies that
\[
W(\xi)\ge\log\xi-\log\log\xi.
\]
In other words, to ensure that inequality (\ref{eq:kNN-expected-translator-helper3}) holds, we can ask that
\begin{align}
\varepsilon & \ge\sqrt{8}(y_{\max}-y_{\min})\exp\Big(-\frac{\alpha}{d}(\log\xi-\log\log\xi)\Big)\nonumber \\
 & =\sqrt{8}(y_{\max}-y_{\min})\Big(\frac{\log\xi}{\xi}\Big)^{\alpha/d}\nonumber \\
 & =\sqrt{8}(y_{\max}-y_{\min})\Bigg(\frac{\log\big(\frac{np_{\min}d}{16\alpha}\Big[\frac{\sqrt{8}(y_{\max}-y_{\min})}{2C}\Big]^{d/\alpha}\big)}{\frac{np_{\min}d}{16\alpha}\Big[\frac{\sqrt{8}(y_{\max}-y_{\min})}{2C}\Big]^{d/\alpha}}\Bigg)^{\alpha/d}\nonumber \\
 & =2C\Bigg(\frac{16\alpha\log\big(\frac{np_{\min}d}{16\alpha}\Big[\frac{\sqrt{8}(y_{\max}-y_{\min})}{2C}\Big]^{d/\alpha}\big)}{np_{\min}d}\Bigg)^{\alpha/d}.\label{eq:kNN-expected-translator-helper4}
\end{align}
Next, by the same reasoning as how we translated Theorem~\ref{thm:k-NN-regression-rate-of-convergence}, inequality $(ii)$ can be satisfied by asking that
\begin{equation}
\varepsilon\ge(y_{\max}-y_{\min})\sqrt{\frac{\log(16k)}{k}}.\label{eq:kNN-expected-translator-helper1}
\end{equation}
Finally, rearranging inequality $(iii)$, we get
\begin{equation}
\varepsilon\ge2C\Big(\frac{2k}{np_{\min}}\Big)^{\alpha/d}.\label{eq:kNN-expected-translator-helper2}
\end{equation}
Thus, by combining inequalities (\ref{eq:kNN-expected-translator-helper4}), (\ref{eq:kNN-expected-translator-helper1}), and (\ref{eq:kNN-expected-translator-helper2}), we see that we can set
\begin{align*}
\varepsilon & =\max\Bigg\{2C\Bigg(\frac{16\alpha\log\big(\frac{np_{\min}d}{16\alpha}\Big[\frac{\sqrt{8}(y_{\max}-y_{\min})}{2C}\Big]^{d/\alpha}\big)}{np_{\min}d}\Bigg)^{\alpha/d},\\
& \quad\qquad\quad(y_{\max}-y_{\min})\sqrt{\frac{\log(16k)}{k}}, \;
2C\Big(\frac{2k}{np_{\min}}\Big)^{\alpha/d}\Bigg\}.
\end{align*}
Remembering that we require $\varepsilon<y_{\max}-y_{\min}$, we secure sufficient conditions to guarantee each argument in the maximization is at most $y_{\max}-y_{\min}$:
\begin{align*}
n & >\Big(\frac{2C}{y_{\max}-y_{\min}}\Big)^{d/\alpha}\frac{16\alpha\log\big(\frac{np_{\min}d}{16\alpha}\Big[\frac{\sqrt{8}(y_{\max}-y_{\min})}{2C}\Big]^{d/\alpha}\big)}{p_{\min}d},\\
k & \ge5,\\
k & <\frac{np_{\min}}{2}\Big(\frac{y_{\max}-y_{\min}}{2C}\Big)^{d/\alpha}.
\end{align*}
We can combine these requirements with requirement (\ref{eq:kNN-expected-translator-helper5}) to get the final translated theorem statement.
\begin{ftheorem}
[$k$-NN regression expected error, alternative form]
Under assumptions \assumpTechnical~and \assumpHolder, suppose that $Y\in[y_{\min},y_{\max}]$ for some constants $y_{\min}$ and $y_{\max}$.  Let $p_{\min} > 0$ and $d > 0$.  If the number of training points satisfies
\begin{align*}
n & >\Big(\frac{2C}{y_{\max}-y_{\min}}\Big)^{d/\alpha}\frac{16\alpha}{p_{\min}d} \\
& \quad \times \max\Big\{\log\Big(\frac{np_{\min}d}{16\alpha}\Big[\frac{\sqrt{8}(y_{\max}-y_{\min})}{2C}\Big]^{d/\alpha}\Big), \;
\frac{e}{(\sqrt{8})^{d/\alpha}}\Big\},
\end{align*}
and the number of nearest neighbors satisfies
\begin{align*}
5 & \le k<\frac{np_{\min}}{2}\Big(\frac{y_{\max}-y_{\min}}{2C}\Big)^{d/\alpha},
\end{align*}
then $k$-NN regression has expected error
\begin{align*}
&\mathbb{E}[| \widehat{\eta}_{k\text{-NN}}(X) - \eta(X) |] \\
& \le 
\max\Bigg\{2C\Bigg(\frac{16\alpha\log\big(\frac{np_{\min}d}{16\alpha}\Big[\frac{\sqrt{8}(y_{\max}-y_{\min})}{2C}\Big]^{d/\alpha}\big)}{np_{\min}d}\Bigg)^{\alpha/d},\\
&\quad\qquad\quad(y_{\max}-y_{\min})\sqrt{\frac{\log(16k)}{k}}, \; 2C\Big(\frac{2k}{np_{\min}}\Big)^{\alpha/d}\Bigg\}.
\end{align*}
\end{ftheorem}

\section{Choice of Number of Nearest Neighbors and Kernel Bandwidth}
\label{sec:automatic-k-h}

In practice, the number of nearest neighbors $k$ and the kernel bandwidth~$h$ (which corresponds to the threshold distance for fixed-radius NN regression) are usually selected via cross-validation or more simply data splitting.  For $k$-NN regression, theoretical results are available for these two ways of choosing $k$ \citep[Theorem~8.2 for cross-validation, Corollary~7.3 for data splitting]{gyorfi_book}.  Similar results are available for choosing threshold distance~$h$ in fixed-radius NN regression \citep[Theorem~8.1 for cross-validation, Corollary~7.1 for data splitting]{gyorfi_book}.

A promising direction is to adaptively choose the number of nearest neighbors~$k$ or the kernel bandwidth~$h$ depending on the test feature vector~$x\in\mathcal{X}$.  In this case, we write $k$ as $k(x)$ and $h$ as $h(x)$ to emphasize their dependence on $x$.  For weighted $k$-NN regression, \citet{samory_2011} provides a simple way to choose $k(x)$ that achieves a near-optimal error rate.  For a given $x$, the method finds the value of~$k$ that balances variance and squared bias of the regression estimate at~$x$.  Meanwhile, a similar strategy turns out to be possible for kernel regression.  \citet{samory_2013} show how to select bandwidth $h(x)$ by controlling local surrogates of bias and variance of the regression estimate at~$x$.  Their resulting adaptive kernel regression method adapts to both unknown local dimension and unknown local H\"{o}lder continuity parameters.  Importantly, these results for adaptively choosing $k(x)$ and $h(x)$ are quite general and handle when feature space $\mathcal{X}$ and distance function $\rho$ form a general metric space. As a preview, toward the end of this monograph in Section~\ref{sec:decision-trees-ensemble-learning}, we show how decision trees and various ensemble learning methods can be viewed as adaptively choosing $k(x)$ for a test feature vector $x$.

Kpotufe and Garg's adaptive kernel regression result actually falls under a bigger umbrella of automatic bandwidth selection methods based on the seminal line of work by Lepski \textit{et al.}~\citep{lepski_et_al_1997,lepski_spokoiny_1997,lepski_levit_1998}. A key idea in this line of work is to monitor how the regression estimate at a point $x$ changes as we vary the bandwidth $h$. A good choice of bandwidth $h(x)$ should satisfy some stability criterion.  We point out that specifically for Euclidean feature spaces~$\mathcal{X}$, Goldenschluger and Lepski have also worked on adaptive bandwidth selection for regression, density estimation, and other problems (\eg, \citealt{goldenshluger_lepski_2008,goldenshluger_lepski_2009,goldenshluger_lepski_2011,goldenshluger_lepski_2013,goldenshluger_lepski_2014}).

More recently, \citet{anava_2016} propose solving an optimization problem to adaptively choose what $k$ to use for $x$ in an approach called \mbox{$k^*$-NN}. This optimization problem makes the bias-variance tradeoff explicit and is efficient to solve. We provide a brief overview of the algorithm and show that in a simple scenario, \mbox{$k^*$-NN} agrees with the regression theory we have established in this chapter. In particular, \mbox{$k^*$-NN} finds a number of nearest neighbors $k^*$ that satisfies the sandwich inequality for $k$ in Theorem \ref{thm:k-NN-regression-rate-of-convergence}. Thus, this algorithm does what the theory suggests in terms of choosing $k$, but in an entirely data driven manner!

\subsubsection{$k^*$-NN}

\begin{figure}
\centering
\includegraphics[scale=0.8]{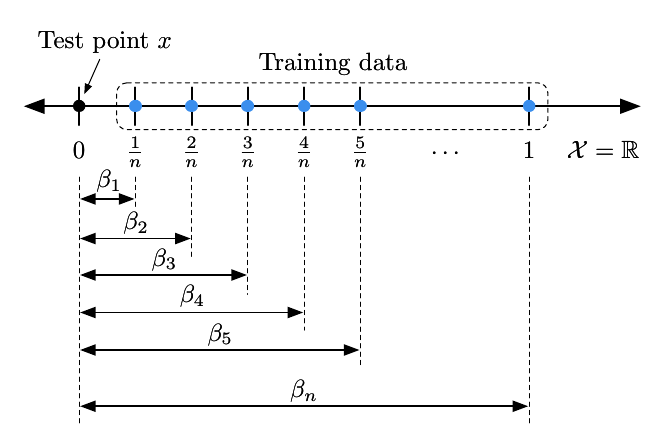}
\vspace{-1em}
\caption{Illustration of where the data are in the $k^*$-NN toy example we analyze.}
\label{fig:k-star-helper}
\end{figure}

Recall that for a given test feature vector $x\in\mathcal{X}$, we aim to estimate $\eta(x)$. We denote $(X_{(i)}(x),Y_{(i)}(x))$ to be the $i$-th closest training data point to $x$ among the training data $(X_{1},Y_{1}),\dots,(X_{n},Y_{n})$, The $k$-NN regression estimate is the average label of the $k$ nearest neighbors found for $x$:
\[
\widehat{\eta}_{k\text{-NN}}(x)=\frac{1}{k}\sum_{i=1}^{k}Y_{(i)}(x).
\]
In general, we could instead think of finding what weight $\alpha_i \in [0,1]$  we should assign to the label of the $i$-th point to produce an estimate $\widehat{\eta}(x)$. That is, 
\[
\widehat{\eta}(x) = \sum_{i=1}^n \alpha_i Y_{(i)}(x), ~~\sum_{i=1}^n \alpha_i = 1, ~~\alpha_i \geq 0 \quad\text{for all} ~i \in \{1,2,\dots,n\}.
\]
To minimize noise or variance, we would like to involve more neighbors so as to average across more labels. This can be captured by choosing $\alpha = [\alpha_i] \in [0,1]^n$ so that $\| \alpha\|_2$ is small, where we still have the constraint that the nonnegative weights sum to~1, \ie, $\|\alpha\|_1 = 1$. On the other hand, to reduce error induced by using far away points, we would like only nearby neighbors to have larger weight (recall from our discussion in Section~\ref{sub:k-nn-regression-bad-events} that using neighbors closer to $x$ aims to reduce bias of the \mbox{$k$-NN} regression estimate at $x$). This can be captured by minimizing $\alpha^T \beta = \sum_{i=1}^n \alpha_i \beta_i$, where $\beta_i = \rho(x, X_{(i)}(x))$. Combining these two pieces, \citet{anava_2016} proposed selecting $\alpha$ to be the solution of the following optimization problem:
\[
\text{minimize}~  \|\alpha\|_2 +  \alpha^T \beta, ~~\text{such~that~} \sum_{i=1}^n \alpha_i = 1, ~ \alpha \in [0,1]^n.
\]
We now provide an illustrative toy example in which we can roughly compute $k^*$. We can then compare $k^*$ with the condition on what $k$ should be according to Theorem~\ref{thm:k-NN-regression-rate-of-convergence}.  Specifically, consider when the feature space is simply the real line $\mathcal{X}=\mathbb{R}$, the test feature is at the origin $x=0$, and the $n$ training points are laid out regularly away from $x$ so that $\beta_i = \frac{i}{n}$ for $i \in \{1, \dots, n\}$.  This is an idealized version of the scenario where $\P_X\sim\text{Uniform}[0,1]$.  This setup is depicted in Figure~\ref{fig:k-star-helper}, where the black point is the test point $x$ and the blue points are the training data.  Meanwhile, we suppose that the regression function $\eta$ satisfies assumption \assumpHolder~with constants $C=1$ and $\alpha=1$ (so $\eta$ is Lipschitz), and $y_{\max} - y_{\min} = 1$.

We compute what $k^*$ roughly equals.  Let $\alpha^*$ be a solution of the above optimization problem.  By looking at its Lagrangian dual formulation, \citet[Theorem 3.1]{anava_2016} argue that there exists $k^* \in \{1,\dots,n\}$ for which the $k^*$ nearest neighbors have positive weights and all the other training data have 0 weight. Formally, $\alpha^*_i > 0$ for $i \leq k^*$, and $\alpha^*_i = 0$ for $i > k^*$.  Furthermore, for some $\lambda \in [\beta_{k^*}, \beta_{k^*+1})$, we have $\alpha^*_i \propto (\lambda - \beta_i)$ for all $i \leq k^*$.  To determine $k^*$ and the weights $\alpha^*$, using equation (3) of \citet{anava_2016}, we have that $\sum_{i=1}^{k^*} (\lambda-\beta_i)^2 = 1$. Given that $\lambda \in  [\beta_{k^*}, \beta_{k^*+1})$, it follows that 
\begin{align}
\sum_{i=1}^{k^*}(\beta_{k^*+1}-\beta_i)^2 & \overset{(i)}{\geq} \underbrace{\sum_{i=1}^{k^*} (\lambda-\beta_i)^2}_{=1}
\overset{(ii)}{\geq} \sum_{i=1}^{k^*}(\beta_{k^*}-\beta_i)^2. \label{eq:ks.1}
\end{align}
Recalling that $\beta_i = \rho(x,X_{(i)}(x))$, inequality $(i)$ of \eqref{eq:ks.1} can be written as
\[
1\le\sum_{i=1}^{k^{*}}\Big(\frac{k^{*}+1}{n}-\frac{i}{n}\Big)^{2}=\frac{1}{n^{2}}\cdot\frac{1}{6}k^{*}\big(2(k^{*})^{2}+3k^{*}+1\big).
\]
Considering when $n$ and $k^*$ are sufficiently large, then this inequality implies that $k^* \gtrsim n^{2/3}$. Similarly, inequality $(ii)$ of \eqref{eq:ks.1} can be written as
\[
1\ge\sum_{i=1}^{k^{*}}\Big(\frac{k^{*}}{n}-\frac{i}{n}\Big)^{2}=\frac{1}{n^{2}}\sum_{i=1}^{k^{*}}(k^{*}-i)^{2}=\frac{1}{n^{2}}\cdot\frac{1}{6}k^{*}\big(2(k^{*})^{2}-3k^{*}+1\big).
\]
For large enough $n$ and $k^*$, then this inequality implies that $k^* \lesssim n^{2/3}$. We conclude that for sufficiently large $k^*$ and $n$, we have $k^* \sim n^{2/3}$.

Next, we compute what the number of nearest neighbors $k$ should be according to Theorem \ref{thm:k-NN-regression-rate-of-convergence}.  Since regression function $\eta$ is H\"{o}lder continuous with parameters $C = 1$ and $\alpha = 1$, then to achieve $\varepsilon$ error in regression estimation, we want the nearest neighbors whose labels we average over to be within critical distance $h^* \sim \varepsilon$.  Next, suppose that both inequalities in sandwich inequality \eqref{eq:k-NN-regression-rate-of-convergence-constraint-on-k} actually hold with equality, which along with our assumption that $y_{\max}-y_{\min}=1$ means that $\frac1{\varepsilon^2} \sim n \P_X(\BB_{x, h^*})$. Since $\P_X\sim\text{Uniform}[0,1]$, and $h^* \sim \varepsilon$, we get that $\frac1{\varepsilon^2} \sim n \times \varepsilon$. That is, $\varepsilon \sim n^{-1/3}$. Thus, by looking at the right-hand side of sandwich inequality~\eqref{eq:k-NN-regression-rate-of-convergence-constraint-on-k} that we have turned into an equality, we see that $k \sim n h^*$ which is $n^{2/3}$---just like what $k^*$-NN algorithm chooses for $k^*$. In this case, Theorem \ref{thm:k-NN-regression-rate-of-convergence} says that the regression error is bounded above by $\varepsilon \sim n^{-1/3}$.

That the $k^*$-NN algorithm chooses the number of nearest neighbors to match the requirement of Theorem \ref{thm:k-NN-regression-rate-of-convergence} in this toy example is reassuring.  We remark that the $k^*$-NN algorithm should be modified so that $\beta_i =  \rho(x, X_{(i)}(x))^\alpha$ if we believe that the regression function is H\"{o}lder continuous with parameter $\alpha$ that may not be $1$, \ie, not Lipschitz as considered by \citet{anava_2016}.

\chapter{Theory on Classification}
\label{chap:classification}

We now turn to binary classification. As a reminder, the setup here is that we have i.i.d.~training data $(X_{1},Y_{1}),(X_{2},Y_{2}),\dots,(X_{n},Y_{n})$, where as with the regression setup, the feature vectors $X_{i}$'s reside in feature space $\mathcal{X}$, and distances between feature vectors are computed via a user-specified function $\rho$.  However, now the labels $Y_{i}$'s are either~0 or~1 (and if not, we can always map the two outcomes to being~0 or~1). Given a new feature vector $X=x\in\mathcal{X}$, we want to predict its associated label $Y$. As before, we can think of the data generation process as first sampling feature vector $x$ from the feature distribution $\mathbb{P}_{X}$, and then sampling the label from a conditional distribution $\mathbb{P}_{Y\mid X=x}$.

Then the Bayes classifier, which is optimal in terms of minimizing probability of error (Proposition~\ref{prop:bayes-classifier-minimizes-prob-error}), predicts the label of $x$ to be~1 precisely if the regression estimate $\eta(x)=\mathbb{E}[Y\mid X=x]={\mathbb{P}(Y=1\mid X=x)}$ exceeds~1/2:
\begin{equation}
\widehat{Y}_{\text{Bayes}}(x)
=
\begin{cases}
1 & \text{if }\eta(x) \ge \frac12, \\
0 & \text{otherwise}.
\end{cases}
\tag{equation~\eqref{eq:bayes-classifier}, reproduced}
\end{equation}
We break the tie at $\eta(x)=\frac12$ in favor of label~1.

Recall that $k$-NN, fixed-radius NN, and kernel classification are plug-in classifiers in that they each produce an estimate $\widehat{\eta}$ that plugs into the above equation, replacing the unknown $\eta$.  Thus, in analyzing the performance of these three methods, we primarily examine when the estimate $\widehat{\eta}(x)$ ``agrees'' with $\eta(x)$ in that they are both greater than 1/2, or they are both less than 1/2. If there is agreement, then it means that a plug-in classifier using $\widehat{\eta}$ in place of~$\eta$ classifies the point~$x$ the same way as the optimal Bayes classifier.

That classification only cares about whether~$\eta$ is above or below threshold~1/2 suggests it to be an easier problem than regression.  For example, within a region of the feature space~$\mathcal{X}$ for which~$\eta$ consistently is above~1/2, we can even have~$\eta$ not be smooth, which can make regression difficult whereas classification can be easy.  In this chapter, we first show in Section~\ref{sec:regression-to-classification-guarantees} how to convert the regression guarantees from Chapter~\ref{chap:regression} into classification guarantees, which rely on regression having low error as a first step.  This requirement is, of course, overly conservative.  We then explain in Section~\ref{sec:classification-guarantees-under-weaker-assumptions} how \citet{chaudhuri_dasgupta_2014} prove a general nonasymptotic guarantee for $k$-NN classification under weaker assumptions.  We show how their proof ideas readily extend to establishing analogous guarantees for fixed-radius NN and kernel classification. We remark that Chaudhuri and Dasgupta's nonasymptotic $k$-NN classification guarantee readily recovers various existing asymptotic $k$-NN classification consistency results~\citep{fix_hodges_1951,stone_1977,devroye_1994,cerou_guyader_2006}.

All guarantees we present for $k$-NN, fixed-radius NN, and kernel classification say how to choose algorithm parameters to ensure low probability of disagreeing with the optimal Bayes classifier.  A major component of this disagreement probability (and in fact, the single dominant component in the case of $k$-NN classification without requiring regression to have low error first) is the probability of feature vector~$X$ landing near the decision boundary $\{ x : \eta(x) = 1/2\}$.  The intuition here is that when a feature vector~$x$ is near the decision boundary, the conditional probabilities $\eta(x)=\mathbb{P}(Y=1\mid X=x)$ and $\mathbb{P}(Y=0\mid X=x)$ are both close to~1/2. As both labels have almost equal probability, predicting the correct label becomes challenging.

To address the difficulty of classification near the decision boundary, an earlier approach to establishing classification guarantees is to assume so-called \textit{margin bounds}, which ask that the probability of landing near the decision boundary decays in a particular fashion~\citep{mammen_1999,tsybakov_2004,audibert_2007}.\footnote{Being near the decision boundary is also called being in the ``margin'' between the two label classes~0 and~1. A ``large margin'' refers to the probability of landing near the decision boundary being small, effectively making the classification problem easier.  One way to parse this terminology is to think of two Gaussians with the same variance, corresponding to the two label classes~0 and~1. The decision boundary is precisely at the average of their means. If they are farther apart, then the ``margin'' between them is larger, and also the probability of landing near (\eg, up to distance~$\varepsilon$ away) the decision boundary drops.} The $k$-NN classification guarantee by~\citet{chaudhuri_dasgupta_2014} turns out to be tight in that by adding both a smoothness condition (H\"{o}lder continuity, \ie, assumption~\assumpHolder~from Section~\ref{sec:technicalities}) and a margin bound condition, $k$-NN classification has an error upper bound that matches an existing lower bound established by~\citet{audibert_2007}. In the final secton of this chapter, we discuss this tightness result, which is established by~\citet[Theorem~7]{chaudhuri_dasgupta_2014}. We also mention a key result by \citet[Theorem~4.1]{gadat_2016} that cleanly delineates regimes in which even if we assume smoothness and a margin bound condition, it is impossible for $k$-NN classification to be uniformly consistent.

We briefly point out some results regarding automatically choosing the number of nearest neighbors $k$ that we do not discuss later in the chapter.  \citet{hall_2008} looked at how to select $k$ through analyzing two concrete models of data generation, and they suggest a practical approach that uses bootstrap sampling and computing some empirical error rates on training data to choose $k$. This choice of $k$ is non-adaptive in that it is not a function of the test feature vector $x\in\mathcal{X}$.  \citet{gadat_2016} suggest adaptively choosing $k=k(x)$ depending on the test feature vector $x$. The idea here is similar to adaptively choosing $k(x)$ for $k$-NN regression (see Section~\ref{sec:automatic-k-h}).  As a reminder, we show in Section~\ref{sec:decision-trees-ensemble-learning} how decision trees and various ensemble learning methods are nearest neighbor methods that adaptively choose $k(x)$; these methods readily handle classification.

\section{Converting Guarantees from Regression to Classification}
\label{sec:regression-to-classification-guarantees}

For any plug-in classifer that uses $\widehat{\eta}$ in place of $\eta$, the classifier agrees with the optimal Bayes classifier on the label for~$x$ if $\widehat{\eta}(x)$ and $\eta(x)$ are both at least~1/2, or both less than~1/2. For example, if $\eta(x)=0.7$, then to get agreement between the two classifiers, we want $\widehat{\eta}(x)\ge1/2$.  One way to get this is to have the regression error at~$x$ satisfy ${|\widehat{\eta}(x) - \eta(x)|} \le 0.2$. In particular, we do not need the regression error at~$x$ to be arbitrarily small; it suffices to have it be small enough to ensure that $\widehat{\eta}(x)$ and $\eta(x)$ are on the same side of threshold~1/2. We formalize this observation in the lemma below, which provides a turnkey solution for converting theoretical guarantees for regression into guarantees for classification.
\begin{flemma}
\label{lem:regression-to-classification-convertor}
Let $x\in\text{supp}(\mathbb{P}_{X})$, $\widehat{\eta}(x)$ be any estimate for $\eta(x)$, and $\varepsilon\in(0,\frac12)$ be a user-specified error tolerance.  If one can ensure that the regression error at $x$ is low enough, namely
\[
|\widehat{\eta}(x)-\eta(x)|\le\varepsilon,
\]
and the conditional probability of label~$Y=1$ given $X=x$ is sufficiently far away from the decision boundary~1/2, namely $|\eta(x)-\frac12| > \varepsilon$, then a plug-in classifier using $\widehat{\eta}$ in place of $\eta$ is guaranteed to make the same classification as the Bayes classifier.
\end{flemma}
Of course, we can ensure that the regression error at~$x$ is at most $\varepsilon$ using the earlier nearest neighbor and kernel regression guarantees. Thus, on top of the conditions each regression guarantee already assumes, to turn it into a classification guarantee, we now further ask that the test point~$x$ be in the region
\begin{align*}
\mathcal{X}_{\text{far-from-boundary}}(\varepsilon)
&\triangleq
 \Big\{ x \in \text{supp}(\mathbb{P}_X) : \Big|\eta(x) - \frac12\Big| > \varepsilon \Big\}.
\end{align*}
We denote the complement of $\mathcal{X}_{\text{far-from-boundary}}$ as
\begin{align*}
\mathcal{X}_{\text{close-to-boundary}}(\varepsilon)
&\triangleq
[\mathcal{X}_{\text{far-from-boundary}}(\varepsilon)]^c \\
&=
 \Big\{ x \in \mathcal{X} : x \notin \text{supp}(\mathbb{P}_X)
        \text{~or~}\Big|\eta(x) - \frac12\Big| \le \varepsilon \Big\}.
\end{align*}
Immediately, we can produce the following results on \mbox{$k$-NN}, fixed-radius NN, and kernel classification from their corresponding regression results.  As a reminder, the recurring key assumptions made (\assumpTechnical, \assumpBesicovitch, \assumpHolder) are precisely stated in Section~\ref{sec:technicalities}.

\begin{ftheorem}
[$k$-NN classification guarantees based on corresponding regression guarantees (Theorems~\ref{thm:k-NN-regression-rate-of-convergence} and~\ref{thm:k-NN-regression-expectation-rate-of-convergence})]
Under assumptions \assumpTechnical~and \assumpBesicovitch, let $\varepsilon\in(0,\frac{1}{2})$ be an error tolerance in estimating $\eta$, and $\delta\in(0,1)$ be a probability tolerance.
\begin{itemize}

\item[(a)] (Pointwise agreement with the Bayes classifier)
For any feature vector $x\in\mathcal{X}_{\text{far-from-boundary}}(\varepsilon)$, there exists a threshold distance $h^{*}\in(0,\infty)$ such that for any smaller distance $h\in(0,h^{*})$, if the number of training points satisfies
\[
n\ge\frac{8}{\mathbb{P}_X(\mathcal{B}_{x,h})}\log\frac{2}{\delta},
\]
and the number of nearest neighbors satisfies
\[
\frac{2}{\varepsilon^{2}}\log\frac{4}{\delta}\le k\le\frac{1}{2}n\mathbb{P}_X(\mathcal{B}_{x,h}),
\]
then with probability at least $1-\delta$ over randomness in samping the training data, $k$-NN classification at point $x$ agrees with the Bayes classifier, \ie, $\widehat{Y}_{k\text{-NN}}(x)=\widehat{Y}_{\text{Bayes}}(x)$.

Furthermore, if the function $\eta$ satisfies assumption \assumpHolder, then we can take $h^{*}=(\frac{\varepsilon}{2C})^{1/\alpha}$ and the above guarantee holds for $h=h^{*}$ as well.

\item[(b)] (Probability of agreement with the Bayes classifier)
Suppose that $\eta$ satisfies assumption \assumpHolder.  Let $p_{\min}>0$ and $d>0$. If the number of training points satisfies
\[
n\ge\Big(\frac{2C}{\varepsilon}\Big)^{d/\alpha}\frac{16}{p_{\min}}\log\frac{\sqrt{8}}{\delta},
\]
and the number of nearest neighbors satisfies
\[
\frac{4}{\varepsilon^{2}}\log\frac{4}{\delta}\le k\le\frac{1}{2}np_{\min}\Big(\frac{\varepsilon}{2C}\Big)^{d/\alpha},
\]
then $k$-NN classification has probability of disagreeing with the Bayes classifier
\begin{align*}
& \mathbb{P}(\widehat{Y}_{k\text{-NN}}(X)\ne\widehat{Y}_{\text{Bayes}}(X))\\
& \le\mathbb{P}\big(X\in\mathcal{X}_{\text{close-to-boundary}}(\varepsilon)\cup\mathcal{X}_{\text{bad}}(p_{\min},d,{\textstyle (\frac{\varepsilon}{2C})^{1/\alpha}})\big)+\delta.
\end{align*}
\end{itemize}
\end{ftheorem}
\begin{ftheorem}
[Fixed-radius NN classification based on corresponding regression guarantees (Theorems~\ref{thm:radius-NN-rate-of-convergence} and~\ref{thm:radius-NN-expectation-rate-of-convergence})]
Under assumptions \assumpTechnical~and \assumpBesicovitch, let $\varepsilon\in(0,\frac{1}{2})$ be an error tolerance in estimating $\eta$, and $\delta\in(0,1)$ be a probability tolerance.
\begin{itemize}

\item[(a)] (Pointwise agreement with the Bayes classifier)
For any feature vector $x\in\mathcal{X}_{\text{far-from-boundary}}(\varepsilon)$, there exists a distance $h^{*}\in(0,\infty)$ such that if our choice of threshold distance satisfies $h\in(0,h^{*})$, and the number of training points satisfies
\[
n\ge\frac{4}{\mathbb{P}_X(\mathcal{B}_{x,h})\varepsilon^{2}}\log\frac{4}{\delta},
\]
then with probability at least $1-\delta$ over randomness in sampling the training data, fixed-radius NN classification with threshold distance $h$ agrees with the Bayes classifier, \ie, $\widehat{Y}_{\text{NN}(h)}(x)=\widehat{Y}_{\text{Bayes}}(x)$.

Furthermore, if the function $\eta$ satisfies assumption \assumpHolder, then we can take $h^{*}=(\frac{\varepsilon}{2C})^{1/\alpha}$ and the above guarantee holds for $h=h^{*}$ as well.

\item[(b)] (Probability of agreement with the Bayes classifier)
Suppose that $\eta$ satisfies assumption \assumpHolder.  Let $p_{\min}>0$ and $d>0$. If the threshold distance satisfies $h\in(0,(\frac{\varepsilon}{2C})^{1/\alpha}]$, and the number of training points satisfies
\[
n\ge\frac{8}{p_{\min}h^{d}\varepsilon^{2}}\log\frac{4}{\delta},
\]
then fixed-radius NN classification with threshold distance~$h$ has probability of disagreeing with the Bayes classifier
\begin{align*}
& \mathbb{P}(\widehat{Y}_{\text{NN}(h)}(X)\ne\widehat{Y}_{\text{Bayes}}(X))\\
& \le\mathbb{P}\big(X\in\mathcal{X}_{\text{close-to-boundary}}(\varepsilon)\cup\mathcal{X}_{\text{bad}}(p_{\min},d,{\textstyle (\frac{\varepsilon}{2C})^{1/\alpha}})\big)+\delta.
\end{align*}
\end{itemize}
\end{ftheorem}
\begin{ftheorem}
[Kernel classification guarantees based on corresponding regression guarantees (Theorems~\ref{thm:kernel-regression-rate-of-convergence} and~\ref{thm:kernel-regression-expected-error-rate-of-convergence})]
Under assumptions \assumpTechnical, \assumpDecay, and \assumpHolder, let $\varepsilon\in(0,\frac{1}{2})$ be an error tolerance in estimating $\eta$, and $\delta\in(0,1)$ be a probability tolerance.  Let $\phi>0$ be any constant for which $K(\phi)>0$. In what follows, suppose the bandwidth satisfies $h\in(0,\frac{1}{\tau}(\frac{\varepsilon}{2C})^{1/\alpha}]$.
\begin{itemize}

\item[(a)] (Pointwise agreement with the Bayes classifier)
For any feature vector $x\in\mathcal{X}_{\text{far-from-boundary}}(\varepsilon)$, if the number of training data satisfies
\[
n\ge\frac{32\log\frac{4}{\delta}}{\varepsilon^{2}[K(\phi)\mathbb{P}_X(\mathcal{B}_{x, \phi h})]^{4}},
\]
then with probability at least $1-\delta$ over randomness in the training data, kernel classification with kernel $K$ and bandwidth $h$ agrees with the Bayes classifier, \ie, $\widehat{Y}_K(x;h) = \widehat{Y}_{\text{Bayes}}(x)$.

\item[(b)] (Probability of agreement with the Bayes classifier)
Let $p_{\min}>0$ and $d>0$. If the number of training data satisfies
\[
n\ge\frac{64\log\frac{4}{\delta}}{\varepsilon^{2}[K(\phi)p_{\min}\phi^{d}h^{d}]^{4}},
\]
then kernel classification with kernel $K$ and bandwidth $h$ has probability of disagreeing with the Bayes classifier
\begin{align*}
& \mathbb{P}(\widehat{Y}_{K}(X;h)\ne\widehat{Y}_{\text{Bayes}}(X))\\
& \le\mathbb{P}\big(X\in\mathcal{X}_{\text{close-to-boundary}}(\varepsilon)\cup\mathcal{X}_{\text{bad}}(p_{\min},d,{\textstyle (\frac{\varepsilon}{2C})^{1/\alpha}})\big)+\delta.
\end{align*}
\end{itemize}
\end{ftheorem}

\section{Guarantees Under Weaker Assumptions}
\label{sec:classification-guarantees-under-weaker-assumptions}

We now establish guarantees for $k$-NN, fixed-radius NN, and kernel classification under weaker assumptions than their regression counterparts. Our exposition focuses on the $k$-NN case as the proof ideas for the latter two methods are nearly the same.  All the guarantees to follow will still depend on assumption \assumpTechnical~(defined in Section~\ref{sec:technicalities}).  The crux of the analysis looks at where smoothness of the regression function~$\eta$ mattered for regression and why we can drop this condition. The resulting guarantees will, however, require partitioning the feature space~$\mathcal{X}$ differently in terms of what the ``good'' and ``bad'' regions are. This partitioning differs between the three methods. Per method, its ``good'' region reflects where the method works well.

Importantly, the ``good'' region for $k$-NN classification accounts for the method being ``adaptive'': if a feature vector~$x$ lands in an extremely low probability region of the feature space~$\mathcal{X}$, then the $k$ nearest neighbors found will tend to be far away from~$x$, whereas if~$x$ lands in an high probability region of the feature space, then the $k$ nearest neighbors found will tend to be close to~$x$. Put another way, how far the $k$ nearest training points to~$x$ are depends on the probability mass around~$x$ according to feature distribution~$\mathbb{P}_X$.

In sharp contrast, fixed-radius NN classification, and thus also kernel classification (which includes fixed-radius NN classification as a special case), lack the adaptivity of $k$-NN classification to the feature distribution~$\mathbb{P}_X$. For example, if a feature vector~$x$ lands in an extremely low probability region of the feature space~$\mathcal{X}$, then quite possibly no training points are within threshold distance~$h$ of point~$x$.  When this happens, fixed-radius NN classification effectively just guesses a label without using label information from any of the training data. As a result, fixed-radius NN and kernel classification are more brittle than $k$-NN classification, and their ``good'' regions consist of points that not only are far away enough from the decision boundary but also in parts of the feature space with high enough probability (so that we are likely to see enough training data within distance~$h$ away). Furthermore, the good region for kernel classification also depends critically on what the kernel function~$K$ is, using fewer assumptions on~$K$ than earlier kernel regression results (Theorems~\ref{thm:kernel-regression-rate-of-convergence} and~\ref{thm:kernel-regression-expected-error-rate-of-convergence}).

\subsection{\texorpdfstring{$k$}{k}-NN Classification}

Before diving into the analysis, we give an overview of the main \mbox{$k$-NN} classification result proved in this section.  As previewed in this chapter's introduction, there is a region near (and that contains) the true decision boundary $\{x:\eta(x)=1/2\}$ for which classification is difficult. We call this this region $\mathcal{X}_{\text{bad-for-}k\text{-NN}}(p_{\text{min-mass}},\Delta)$ (formally defined via equations~\eqref{eq:good-region-for-k-NN-classification} and~\eqref{eq:bad-region-for-k-NN-classification} later), where $p_{\text{min-mass}}\in(0,1]$ is a parameter that specifies how closeby we want the $k$ nearest neighbors found to be to a test point (higher means we can look farther away), and $\Delta\in(0,1/2]$ is a parameter saying how confident we want a prediction to be (higher means more confident). Note that decreasing either $p_{\text{min-mass}}$ or $\Delta$ shrinks the size of $\mathcal{X}_{\text{bad-for-}k\text{-NN}}$.  In particular, with $p_{\text{min-mass}}\rightarrow0$ and $\Delta\rightarrow0$, then $\mathbb{P}\big(X\in\mathcal{X}_{\text{bad-for-}k\text{-NN}}(p_{\text{min-mass}},\Delta)\big)\rightarrow0$.  Then the probability that the $k$-NN and Bayes classifiers disagree can be controlled to be arbitrarily small.

\begin{ftheorem}[Informal statement of Corollary~\ref{eq:k-NN-classification-main-result-more-like-chaudhuri-dasgupta-theorem1}, which is a minor variation on Theorem~1 of~\citet{chaudhuri_dasgupta_2014}]
Let $\delta\in(0,1)$ be a user-specified error tolerance. If the number of nearest neighbors satisfies 
\[
\Theta\Big(\frac{1}{\delta}\Big)<k\le n\cdot\bigg(1-\Theta\bigg(\sqrt{\frac{1}{k}\log\frac{1}{\delta}}\,\,\bigg)\bigg),
\]
then
\[
\mathbb{P}(\widehat{Y}_{k\text{-NN}}(X)\ne\widehat{Y}_{\text{Bayes}}(X))\le\delta+\mathbb{P}\big(X\in\mathcal{X}_{\text{bad-for-}k\text{-NN}}(p_{\text{min-mass}},\Delta)\big),
\]
where
\[
p_{\text{min-mass}}
=\frac{k}{n}\cdot\frac{1}{1-\Theta(\sqrt{\frac{1}{k}\log\frac{1}{\delta}}\,\,)},
\qquad
\Delta=\Theta\Big(\sqrt{\frac{1}{k}\log\frac1\delta}\Big).
\]
\end{ftheorem}
In particular, by having $\delta\rightarrow0$, $n\rightarrow\infty$, $k\rightarrow\infty$, and ${\frac{k}{n}\rightarrow0}$, we can have both $p_{\text{min-mass}}\rightarrow0$ and $\Delta\rightarrow0$, meaning that
\[
{\mathbb{P}\big(X\in\mathcal{X}_{\text{bad-for-}k\text{-NN}}(p_{\text{min-mass}},\Delta)\big)}\rightarrow0.
\] 
Hence, the probability that the $k$-NN and Bayes classifiers disagree goes to 0, and so the probability of misclassification becomes the same for the two classifiers, \ie, ${\mathbb{P}(\widehat{Y}_{k\text{-NN}}(X) \ne Y)} \rightarrow {\mathbb{P}(\widehat{Y}_{\text{Bayes}}(X) \ne Y)}$.  In fact, by using a refinement of this argument that assumes that the Besicovitch condition \assumpBesicovitch~from Section~\ref{sec:technicalities} holds, \citet[Theorem~2]{chaudhuri_dasgupta_2014} establish weak and strong consistency of $k$-NN classification under a more general setting than previously done by \citet{fix_hodges_1951}, \citet{stone_1977}, and \citet{devroye_1994}.  We remark that weak but not strong consistency of nearest neighbor classification under similarly general conditions (specifically assumption \assumpTechnical~and a version of assumption~\assumpBesicovitch) has previously been established by~\citet{cerou_guyader_2006}.

\subsubsection*{Establishing Chaudhuri and Dasgupta's $k$-NN Classification
Result}

The main source of where smoothness of~$\eta$ mattered for regression and where it is no longer needed in classification is as follows.  For $k$-NN, fixed-radius NN, and kernel regression, a recurring step in deriving their pointwise guarantees is to control the regression error $|\widehat{\eta}(x) - \eta(x)|$ via the triangle inequality. For example, for $k$-NN regression, we used inequality~\eqref{eq:regression-high-level-triangle-inequality}, observing that
\begin{equation}
| \widehat{\eta}_{k\text{-NN}}(x) - \eta(x) |
\le 
    |
        \widehat{\eta}_{k\text{-NN}}(x) - \mathbb{E}_{n|\widetilde{X}}[\widehat{\eta}_{k\text{-NN}}(x)]
    |
      +
    |
        \mathbb{E}_{n|\widetilde{X}}[\widehat{\eta}_{k\text{-NN}}(x)] - \eta(x)
    |, \label{eq:regression-high-level-triangle-inequality-snippet}
\end{equation}
where $\mathbb{E}_{n|\widetilde{X}}$ is the expectation over the training data $(X_1, Y_1), \dots, (X_n, Y_n)$, conditioned on the \mbox{$(k+1)$-st} nearest neighbor $\widetilde{X}:=X_{(k+1)}(x)$. The right-hand side features a decomposition similar to but not the same as the bias-variance decomposition (which would examine expected squared error instead of absolute error), where the first term gauges variability in the $k$ samples from their expected mean, and the latter term is precisely the absolute value of the bias of estimator~$\widehat{\eta}_{k\text{-NN}}$.  With~$k$ large enough, the ``variability'' term can be controlled to be small with high probability. Meanwhile, controlling the ``bias'' term can be done by asking that enough training data are close to test point~$x$ within a critical distance that depends on the smoothness of~$\eta$ (\eg, smoother means that the nearest neighbors found can be farther from~$x$).

For classification, we do not actually need to control the ``bias'' term $| \mathbb{E}_{n|\widetilde{X}}[\widehat{\eta}_{k\text{-NN}}(x)] - \eta(x) |$ to be small! In particular, we still want $\mathbb{E}_{n|\widetilde{X}}[\widehat{\eta}_{k\text{-NN}}(x)]$ and $\eta(x)$ to ``agree'', but now agreement just means that they are either both greater than 1/2, or both less than 1/2.  For example, if $\eta(x)>1/2$, then it suffices to ask that $\mathbb{E}_{n|\widetilde{X}}[\widehat{\eta}_{k\text{-NN}}(x)]>1/2$.  In fact, if $\mathbb{E}_{n|\widetilde{X}}[\widehat{\eta}_{k\text{-NN}}(x)] \ge \frac12 + \Delta$ for some $\Delta \in (0,1/2]$, then so long as the ``variability'' term satisfies $| \widehat{\eta}_{k\text{-NN}}(x) - \mathbb{E}_{n|\widetilde{X}}[\widehat{\eta}_{k\text{-NN}}(x)] | < \Delta$ (which ensures that $\widehat{\eta}_{k\text{-NN}}(x) > 1/2$), we have $k$-NN and the optimal Bayes classifiers agreeing on the label for test point~$x$.  A similar statement holds when $\eta(x) < 1/2$. We summarize this observation as follows.

\begin{flemma}
\label{lem:k-NN-classifier-key-observation}
Let $\Delta\in(0,1/2]$. For a test feature vector~$x \in \mathcal{X}$, we have $\widehat{Y}_{k\text{-NN}}(x) = \widehat{Y}_{\text{Bayes}}(x)$ if the following two sufficient conditions hold:
\begin{itemize}
\item
$| \widehat{\eta}_{k\text{-NN}}(x) - \mathbb{E}_{n|\widetilde{X}}[\widehat{\eta}_{k\text{-NN}}(x)] | < \Delta$.

\item
Test point~$x$ lands in the region
\begin{align*}
&\mathcal{X}_{\text{good-for-}k\text{-NN}}^*(\Delta) \\
&\triangleq
 \Big\{ x\in\text{supp}(\mathbb{P}_X) :
    \eta(x) > \frac12,\,
    \mathbb{E}_{n|\widetilde{X}}[\widehat{\eta}_{k\text{-NN}}(x)] \ge \frac12 + \Delta
 \Big\} \\
&\quad
 \cup
 \Big\{ x\in\text{supp}(\mathbb{P}_X) :
    \eta(x) < \frac12,\,
    \mathbb{E}_{n|\widetilde{X}}[\widehat{\eta}_{k\text{-NN}}(x)] \le \frac12 - \Delta
 \Big\}.
\end{align*}
\end{itemize}
\end{flemma}
We now look at when these two conditions hold with high probability, which involves reusing lemmas encountered for regression. In fact, very little additional work is needed.

To begin, note that in proving $k$-NN regression's pointwise performance guarantee, we actually already saw how to ensure that the first condition holds with high probability for large enough $k$.

\begin{flemma}[Lemma~\ref{lem:k-NN-rate-of-convergence-helper1} with choice $\varepsilon=2\Delta$ and accounting for tie breaking as discussed at the end of Section~\ref{sec:proof-k-NN-regression-rate-of-convergence}, which actually then becomes Lemma~10 of~\citet{chaudhuri_dasgupta_2014}]
\label{lem:k-NN-classifier-ensure-k-large-enough}
Let $\Delta\in(0,1/2]$.  Under assumption \assumpTechnical, we have
\begin{align*}
\mathbb{P}_n\Big(|\widehat{\eta}_{k\text{-NN}}(x)-\mathbb{E}_{n|\widetilde{X}}[\widehat{\eta}_{k\text{-NN}}(x)]|\ge\Delta\Big)
&\le2\exp(-2 k \Delta^2),
\end{align*}
where $\mathbb{P}_n$ denotes the probability distribution over sampling the $n$ training data.
\end{flemma}
Thus, we have $|\widehat{\eta}_{k\text{-NN}}(x)-\mathbb{E}_{n|\widetilde{X}}[\widehat{\eta}_{k\text{-NN}}(x)]| < \Delta$ with probability at least $1-2\exp(-2 k \Delta^2)$.

As our discussion above suggests, the second condition that $x$ lands in the region $\mathcal{X}_{\text{good-for-}k\text{-NN}}(\Delta)$ is about $\mathbb{E}_{n|\widetilde{X}}[\widehat{\eta}_{k\text{-NN}}(x)]$ and $\eta(x)$ both being on the same side of threshold 1/2. To ensure that this happens, recall that for regression, to make $\mathbb{E}_{n|\widetilde{X}}[\widehat{\eta}_{k\text{-NN}}(x)]$ close to $\eta(x)$, we asked that the \mbox{$(k+1)$-st} nearest training point to~$x$ be at most some critical distance away from~$x$ that depended on the smoothness of~$\eta$. For classification, while we do not assume $\eta$ to be smooth, in making $\mathbb{E}_{n|\widetilde{X}}[\widehat{\eta}_{k\text{-NN}}(x)]$ agree with~$\eta(x)$, we again appeal to some notion of the $k$ nearest neighbors found being sufficiently close to~$x$. But how close?

As mentioned previously, the key observation is that $k$-NN classification is adaptive in that how far the $k$ nearest neighbors to~$x$ are depends on the probability mass surrounding~$x$ in feature distribution~$\mathbb{P}_X$. Higher probability of landing near~$x$ corresponds to the $k$ nearest neighbors of~$x$ tending to be closer to~$x$.  Thus, \citet{chaudhuri_dasgupta_2014} suggest using an adaptive distance where for a given test point~$x$, we ask that the $k$ nearest training points to $x$ be within the following distance that depends on~$x$ and also a probability parameter $p_{\text{min-mass}}\in(0,1]$:
\[
r_{p_{\text{min-mass}}}(x)
= \inf\{ r\ge0 \;:\; \mathbb{P}_X(\mathcal{B}_{x,r}) \ge p_{\text{min-mass}}\}.
\]
This is the smallest distance $r$ we have to go from $x$ before ${\mathbb{P}(X \in \mathcal{B}_{x,r})} \ge p_{\text{min-mass}}$.  Later when we discuss fixed-radius NN classification, we instead use a fixed threshold distance.

With the above intuition in mind, we follow Chaudhuri and Dasgupta's derivation and define the following good region
\begin{align}
&\mathcal{X}_{\text{good-for-}k\text{-NN}}(p_{\text{min-mass}}, \Delta) \nonumber \\
&\triangleq
 \Big\{ x\in\text{supp}(\mathbb{P}_X) :
    \eta(x) > \frac12, \nonumber \\
&\qquad
    \mathbb{E}[Y \mid X\in\mathcal{B}_{x,r}] \ge \frac12 + \Delta
    \text{~for all~}r\in[0, r_{p_{\text{min-mass}}}(x)]
 \Big\} \nonumber \\
&\quad\;
 \cup
 \Big\{ x\in\text{supp}(\mathbb{P}_X) :
    \eta(x) < \frac12, \nonumber \\
&\;\qquad\;\,\,
    \mathbb{E}[Y \mid X\in\mathcal{B}_{x,r}] \le \frac12 - \Delta
    \text{~for all~}r\in[0, r_{p_{\text{min-mass}}}(x)]
 \Big\},
 \label{eq:good-region-for-k-NN-classification}
\end{align}
which consists of two pieces: the part of the feature space where if a test point~$x$ lands there, we can be reasonably confident (dependent on~$\Delta$) that $k$-NN classification (with the $k$ nearest neighbors found being up to a distance~$r_{p_{\text{min-mass}}}(x)$ away from~$x$) agrees with the Bayes classifier on~$x$ having label~1, and also the analogous part where we can be reasonably confident that $k$-NN classification agrees with the Bayes classifier on~$x$ having label~0.  This interpretation of the above good region results from the following observation that links region $\mathcal{X}_{\text{good-for-}k\text{-NN}}(p_{\text{min-mass}}, \Delta)$ to the earlier good region $\mathcal{X}_{\text{good-for-}k\text{-NN}}^*(\Delta)$ from Lemma~\ref{lem:k-NN-classifier-key-observation}.

\begin{flemma}
\label{lem:k-NN-classifier-relate-good-regions}
Let $p_{\text{min-mass}}\in(0,1]$ and $\Delta\in(0,1/2]$.  Under assumption \assumpTechnical, let $x \in \mathcal{X}_{\text{good-for-}k\text{-NN}}(p_{\text{min-mass}}, \Delta)$.  Suppose that the distance between $x$ and its \mbox{$(k+1)$-st} nearest training point is at most $r_{p_{\text{min-mass}}}(x)$ (notationally, $\rho(x,X_{(k+1)}(x)) \le r_{p_{\text{min-mass}}}(x)$), meaning that the $k$ nearest neighbors of~$x$ are within distance $r_{p_{\text{min-mass}}}(x)$ of~$x$. Then $x \in \mathcal{X}_{\text{good-for-}k\text{-NN}}^*(\Delta)$.
\end{flemma}
Thus, so long as we can ensure that $\rho(x,X_{(k+1)}(x)) \le r_{p_{\text{min-mass}}}(x)$, then~$x$ being in $\mathcal{X}_{\text{good-for-}k\text{-NN}}(p_{\text{min-mass}}, \Delta)$ means that the second condition in Lemma~\ref{lem:k-NN-classifier-key-observation} is satisfied.  We have actually already previously shown when $\rho(x,X_{(k+1)}(x)) \le r_{p_{\text{min-mass}}}(x)$ with high probability! We recall this result after presenting the proof of Lemma~\ref{lem:k-NN-classifier-relate-good-regions}.

\begin{proof}[Proof of Lemma~\ref{lem:k-NN-classifier-relate-good-regions}]
Suppose $x \in \mathcal{X}_{\text{good-for-}k\text{-NN}}(p_{\text{min-mass}}, \Delta)$ and its \mbox{$(k+1)$-st} nearest neighbor are at most a distance $r_{p_{\text{min-mass}}}(x)$ apart.  This means that
\[
\mathbb{E}_{n|\widetilde{X}}[\widehat{\eta}_{k\text{-NN}}(x)]
=
\mathbb{E}[Y \mid X \in \mathcal{B}_{x,\underbrace{\scriptstyle \rho(x, X_{(k+1)}(x))}_{\le r_{p_{\text{min-mass}}}(x)}}].
\]
If the right-hand side is at least $\frac12 + \Delta$, then so is the left-hand side. Similarly, if the right-hand side is at most $\frac12 - \Delta$, then so is the left-hand side. Hence, $x$ is also inside $\mathcal{X}_{\text{good-for-}k\text{-NN}}^*(\Delta)$.
\end{proof}

\begin{flemma}[Lemma~\ref{lem:k-NN-rate-of-convergence-helper2-1} with choice $r=r_{p_{\text{min-mass}}}(x)$, which actually then becomes Lemma~9 of~\citet{chaudhuri_dasgupta_2014}]
\label{lem:k-NN-classifier-ensure-close-enough-neighbors}
Let $\gamma\in(0,1)$. We use $\mathbb{P}_n$ to denote the probability distribution over sampling the $n$ training data.  For positive integer $k\le(1-\gamma)np_{\text{min-mass}}$,
\begin{align*}
\mathbb{P}_n\big(\rho(x,X_{(k+1)}(x))\ge r_{p_{\text{min-mass}}}(x) \big)
&\le\exp\Big(-\frac{\gamma^{2}n p_{\text{min-mass}}}{2}\Big) \\
&\le\exp\Big(-\frac{\gamma^2 k}2\Big).
\end{align*}
(The second inequality follows since $k \le {(1-\gamma)np_{\text{min-mass}}} \le np_{\text{min-mass}}$.)
\end{flemma}
Hence, we have $\rho(x, X_{(k+1)}(x)) \le r_{p_{\text{min-mass}}}(x)$ with probability at least $1-\exp(-\gamma^2 k/2)$.

Putting together the pieces, for any $x\in\mathcal{X}_{\text{good-for-}k\text{-NN}}(p_{\text{min-mass}}, \Delta)$, by union bounding over the bad events of Lemmas~\ref{lem:k-NN-classifier-ensure-k-large-enough} and~\ref{lem:k-NN-classifier-ensure-close-enough-neighbors}, then with probability at least $1-2\exp(-2k\Delta^2)-\exp(-\gamma^2 k/2)$, neither bad event happens, which combined with Lemmas~\ref{lem:k-NN-classifier-key-observation} and~\ref{lem:k-NN-classifier-relate-good-regions}, implies that $k$-NN classification agrees with the optimal Bayes classifier in classifying test point~$x$. As a corner case, note that if $\Delta=1/2$ (corresponding to no noise), then we actually do not have to worry about the bad event of Lemma~\ref{lem:k-NN-classifier-ensure-k-large-enough} since conditioned on the bad event of Lemma~\ref{lem:k-NN-classifier-ensure-close-enough-neighbors} not happening, the $k$ nearest neighbors are all close enough to~$x$ and actually all have the same label.  We thus obtain the following nonasymptotic pointwise error guarantee for $k$-NN classification, which we restate shortly to make consistent with both our presentation of nearest neighbor and kernel regression guarantees (\ie, what algorithm parameters ensure a user-specified probability of error~$\delta$).

\begin{flemma}[$k$-NN classification pointwise agreement with the Bayes classifier]
Let $p_{\text{min-mass}}\in(0,1]$, $\Delta\in(0,1/2]$, and $\gamma\in(0,1)$.  Under assumption \assumpTechnical, suppose that
\[
x\in\mathcal{X}_{\text{good-for-}k\text{-NN}}(p_{\text{min-mass}},\Delta) \text{~and~} 
k \le (1-\gamma)np_{\text{min-mass}}.
\]
We consider two cases:
\begin{itemize}

\item If $\Delta\in(0,1/2)$, then with probability at least
\begin{equation}
1 - 2\exp(-2k\Delta^2) - \exp\Big(-\frac{\gamma^2 k}2\Big),
\label{eq:k-NN-classify-pointwise-prob-success-Delta-less-than-half}
\end{equation}
we have $\widehat{Y}_{k\text{-NN}}(x) = \widehat{Y}_{\text{Bayes}}(x)$.

\item If $\Delta=1/2$, then with probability at least
\begin{equation}
1 - \exp\Big(-\frac{\gamma^2 k}2\Big),
\label{eq:k-NN-classify-pointwise-prob-success-Delta-equal-half}
\end{equation}
we have $\widehat{Y}_{k\text{-NN}}(x) = \widehat{Y}_{\text{Bayes}}(x)$.
\end{itemize}
\end{flemma}
We reword this lemma in terms of controlling a probability of error $\delta\in(0,1)$.

\begin{ftheorem}[$k$-NN classification pointwise agreement with the Bayes classifier, slightly reworded]
Let $\delta\in(0,1)$, $p_{\text{min-mass}}\in(0,1]$, $\Delta\in(0,1/2]$, and $\gamma\in(0,1)$. Under assumption \assumpTechnical, suppose that $x\in\mathcal{X}_{\text{good-for-}k\text{-NN}}(p_{\text{min-mass}},\Delta)$, and either of the following holds:
\begin{itemize}

\item $\Delta\in(0,1/2)$ and the number of nearest neighbors satisfies
\[
\max\Big\{\frac{1}{2\Delta^{2}}\log\frac{4}{\delta},\frac{2}{\gamma^{2}}\log\frac{2}{\delta}\Big\} \le k \le (1-\gamma)np_{\text{min-mass}}.
\]
(Thus, each of the two bad events in success probability \eqref{eq:k-NN-classify-pointwise-prob-success-Delta-less-than-half} is controlled to have probability at most $\delta/2$.)

\item $\Delta=1/2$ and the number of nearest neighbors satisfies
\[
\frac{2}{\gamma^{2}}\log\frac{1}{\delta} \le k \le (1-\gamma)np_{\text{min-mass}}.
\]
(Thus, the bad event in success probability \eqref{eq:k-NN-classify-pointwise-prob-success-Delta-equal-half} is controlled to have probability at most~$\delta$.)

\end{itemize}
Then with probability at least $1-\delta$ over randomness in sampling the training data, we have $\widehat{Y}_{k\text{-NN}}(x) = \widehat{Y}_{\text{Bayes}}(x)$.
\end{ftheorem}
To go from a pointwise error guarantee to a guarantee accounting for randomness in sampling $X=x\sim\mathbb{P}_X$, the idea is identical to what was done for regression guarantees. For the same technical reason as in proving the $k$-NN regression expected error guarantee (Section~\ref{sec:proof-k-NN-regression-expectation-rate-of-convergence}), we replace $\delta$ with $\delta^2/4$; the proof to follow re-explains where this comes from. Overall, we obtain the result below, where we define the ``bad'' region as the complement of the good region:
\begin{equation}
\mathcal{X}_{\text{bad-for-}k\text{-NN}}(p_{\text{min-mass}},\Delta)
\triangleq
[\mathcal{X}_{\text{good-for-}k\text{-NN}}(p_{\text{min-mass}},\Delta)]^c.
\label{eq:bad-region-for-k-NN-classification}
\end{equation}
This bad region corresponds to the decision boundary plus a band around it of width dependent on~$p_{\text{min-mass}}$ (again, this relates to how close we ask the $k$ nearest neighbors to be to a test point) and $\Delta$ (how confident we want $k$-NN classifier to be in agreeing with the Bayes classifier).
\begin{ftheorem}[$k$-NN classification probability of agreement with the Bayes classifier]
\label{thm:k-NN-classifier-weaker-assumptions-main-result}
Let $\delta\in(0,1)$, $p_{\text{min-mass}}\in(0,1]$, $\Delta\in(0,1/2]$, and $\gamma\in(0,1)$.  Under assumption \assumpTechnical, suppose that either of the following holds:
\begin{itemize}

\item $\Delta\in(0,1/2)$ and the number of nearest neighbors satisfies
\begin{equation}
\max\Big\{\frac{1}{\Delta^2}\log\frac{4}{\delta},\frac{4}{\gamma^2}\log\frac{\sqrt{8}}{\delta}\Big\} \le k \le (1-\gamma)np_{\text{min-mass}}.
\label{eq:k-NN-classifier-prob-agreement-guarantee-constraint-on-k-Delta-less-than-half}
\end{equation}

\item $\Delta=1/2$ and the number of nearest neighbors satisfies
\begin{equation}
\frac{4}{\gamma^{2}}\log\frac{2}{\delta} \le k \le (1-\gamma)np_{\text{min-mass}}.
\label{eq:k-NN-classifier-prob-agreement-guarantee-constraint-on-k-Delta-equal-half}
\end{equation}

\end{itemize}
Then with probability at least $1-\delta-\mathbb{P}_X(\mathcal{X}_{\text{bad-for-}k\text{-NN}}(p_{\text{min-mass}},\Delta))$ over randomness in sampling the training data and in sampling $X\sim\mathbb{P}_X$, we have $\widehat{Y}_{k\text{-NN}}(X) = \widehat{Y}_{\text{Bayes}}(X)$.
\end{ftheorem}
Concretely, to ensure that sandwich inequality~\eqref{eq:k-NN-classifier-prob-agreement-guarantee-constraint-on-k-Delta-less-than-half} holds, we can choose $\Delta$, $\gamma$, and $p_{\text{min-mass}}$ so that the two inequalities within the sandwich inequality become equalities, unless of course doing so would require any of these to be outside its tolerated range of values.  With this approach in mind, a valid choice of parameters is as follows.

\begin{fcorollary}[Minor variation on Theorem~1 of~\citet{chaudhuri_dasgupta_2014}]
\label{eq:k-NN-classification-main-result-more-like-chaudhuri-dasgupta-theorem1}
Let $\delta\in(0,1)$.  Under assumption \assumpTechnical, suppose that the number of nearest neighbors satisfies
\[
4\log\frac{\sqrt{8}}{\delta} < k \le n\cdot(1-\gamma),
\]
and we choose the following constants
\[
\Delta = \min\bigg\{ \frac12, \, \sqrt{\frac1k\log\frac4\delta} \bigg\},\;\,
\gamma = \sqrt{\frac4k\log\frac{\sqrt{8}}{\delta}},\;\,\]
\[
p_{\text{min-mass}} = \frac{k}{n\cdot(1-\gamma)}.
\]
Then with probability at least $1-\delta-\mathbb{P}_X(\mathcal{X}_{\text{bad-for-}k\text{-NN}}(p_{\text{min-mass}},\Delta))$ over randomness in sampling the training data and in sampling $X\sim\mathbb{P}_X$, we have $\widehat{Y}_{k\text{-NN}}(X) = \widehat{Y}_{\text{Bayes}}(X)$.
\end{fcorollary}
One can check that the above choices for $\Delta$, $\gamma$, and $p_{\text{min-mass}}$ work regardless of whether $\Delta=1/2$ in Theorem~\ref{thm:k-NN-classifier-weaker-assumptions-main-result}, so both sandwich inequalities~\eqref{eq:k-NN-classifier-prob-agreement-guarantee-constraint-on-k-Delta-less-than-half} and~\eqref{eq:k-NN-classifier-prob-agreement-guarantee-constraint-on-k-Delta-equal-half} hold. Note that the constraint on the number of nearest neighbors $k$ ensures that $\gamma<1$ and $p_{\text{min-mass}}\le1$.

\begin{proof}[Proof of Theorem~\ref{thm:k-NN-classifier-weaker-assumptions-main-result}]
The proof is similar to that of Theorem~\ref{thm:k-NN-regression-expectation-rate-of-convergence} (Section~\ref{sec:proof-k-NN-regression-expectation-rate-of-convergence}).  We consider when $\Delta\in(0,1/2)$.  First note that Lemmas~\ref{lem:k-NN-classifier-key-observation} and~\ref{lem:k-NN-classifier-relate-good-regions} together imply that
\begin{align}
\ind\{\widehat{Y}_{k\text{-NN}}(x)
      \ne \widehat{Y}_{\text{Bayes}}(x)\}
&\le
\ind\{x \notin \mathcal{X}_{\text{good-for-}k\text{-NN}}(p_{\text{min-mass}},\Delta)\} \nonumber \\
&\quad
+
\ind\{\rho(x,X_{(k+1)}(x)) > r_{p_{\text{min-mass}}}(x)\} \nonumber \\
&\quad
+
\ind\{|\widehat{\eta}_{k\text{-NN}}(x)
      - \mathbb{E}_{n|\widetilde{X}}[\widehat{\eta}_{k\text{-NN}}(x)]| \ge \Delta\}.
\label{eq:chaudhuri-dasgupta-lemma-8}
\end{align}
In particular, the right-hand side being~0 means that the conditions for Lemma~\ref{lem:k-NN-classifier-key-observation} are met (with the help of Lemma~\ref{lem:k-NN-classifier-relate-good-regions}) and thus $\widehat{Y}_{k\text{-NN}}(x)=\widehat{Y}_{\text{Bayes}}(x)$, \ie, the left-hand side is also~0.

Next, taking the expectation of both sides with respect to the random training data (denoted $\mathbb{E}_n$) and the test feature vector~$X$ (denoted $\mathbb{E}_X$), we get
\begin{align}
& \mathbb{P}(\widehat{Y}_{k\text{-NN}}(X)\ne\widehat{Y}_{\text{Bayes}}(X)) \nonumber \\
&\quad =\mathbb{E}_n\big[\mathbb{E}_{X}[\ind\{\widehat{Y}_{k\text{-NN}}(X)\ne\widehat{Y}_{\text{Bayes}}(X)\}]\big] \nonumber \\
&\quad \le\mathbb{E}_n\big[\mathbb{E}_{X}[\ind\{X\notin\mathcal{X}_{\text{good-for-}k\text{-NN}}(p_{\text{min-mass}},\Delta)\} \nonumber \\
&\quad \qquad\qquad\;+\ind\{\rho(X,X_{(k+1)}(X)>r_{p_{\text{min-mass}}}(X)\} \nonumber \\
&\quad \qquad\qquad\;+\ind\{|\widehat{\eta}_{k\text{-NN}}(X)-\mathbb{E}_{n|\widetilde{X}}[\widehat{\eta}_{k\text{-NN}}(X)]|\ge\Delta\}]\big] \nonumber \\
&\quad =\mathbb{P}(X\notin\mathcal{X}_{\text{good-for-}k\text{-NN}}(p_{\text{min-mass}},\Delta))+\mathbb{E}_n[\Xi], \label{eq:k-NN-classifier-weaker-assumptions-pf-helper}
\end{align}
where for a specific test point~$x$ and choice of training data $(x_1,y_1)$, $\dots$, $(x_n,y_n)$, we define
\begin{align*}
\text{BAD}(x,x_{1},\dots,x_{n},y_{1},\dots,y_{n})
&\triangleq\ind\{\rho(x,x_{(k+1)}(x)>r_{p_{\text{min-mass}}}(x)\} \\
&\quad
 +\ind\{|\widehat{\eta}_{k\text{-NN}}(x)-\mathbb{E}_{n|\widetilde{X}}[\widehat{\eta}_{k\text{-NN}}(x)]|\ge\Delta\}], \\
\Xi & \triangleq\mathbb{E}_{X}[\text{BAD}(X,X_{1},\dots,X_{n},Y_{1},\dots,Y_{n})].
\end{align*}
Note that $\Xi$ is a function of random training data $(X_1,Y_1),\dots,(X_n,Y_n)$ since it only uses an expectation over the test point~$X$.

We finish the proof by showing that $\mathbb{E}_n[\Xi] \le \delta$ in inequality~\eqref{eq:k-NN-classifier-weaker-assumptions-pf-helper}. Specifically, by controlling each of the bad events of Lemmas~\ref{lem:k-NN-classifier-ensure-k-large-enough} and~\ref{lem:k-NN-classifier-ensure-close-enough-neighbors} to have probability at most $\delta^2/8$, we ensure that
\[
\mathbb{E}_n[\text{BAD}(X,X_1,\dots,X_n,Y_1,\dots,Y_n)]\le\frac{\delta^2}8 + \frac{\delta^2}8=\frac{\delta^2}4.
\]
Then
\begin{align*}
\mathbb{E}_n[\Xi] & =\underbrace{\mathbb{E}_n[\Xi\mid\Xi<\frac{\delta}{2}]}_{\le\frac{\delta}{2}}\underbrace{\mathbb{P}_{n}(\Xi<\frac{\delta}{2})}_{\le1}+\underbrace{\mathbb{E}_n[\Xi\mid\Xi\ge\frac{\delta}{2}]}_{\le1}\mathbb{P}_{n}(\Xi\ge\frac{\delta}{2})\\
& \le\frac{\delta}{2}+\mathbb{P}_{n}(\Xi\ge\frac{\delta}{2})\\
(\text{Markov ineq.}) & \le\frac{\delta}{2}+\frac{\mathbb{E}_n[\Xi]}{\delta/2}\\
& =\frac{\delta}{2}+\frac{\mathbb{E}_n\big[\mathbb{E}_{X}[\text{BAD}(X,X_{1},\dots,X_{n},Y_{1},\dots,Y_{n})]\big]}{\delta/2}\\
& =\frac{\delta}{2}+\frac{\mathbb{E}_{X}\big[\mathbb{E}_n[\text{BAD}(X,X_{1},\dots,X_{n},Y_{1},\dots,Y_{n})]\big]}{\delta/2}\\
& \le\frac{\delta}{2}+\frac{\mathbb{E}_{X}\big[\delta^{2}/4\big]}{\delta/2}\\
& =\frac{\delta}{2}+\frac{\delta}{2}\\
& =\delta.
\end{align*}
Finally, we remark that if $\Delta=1/2$, then we do not have to worry about the condition $|\widehat{\eta}(x) - \mathbb{E}_{n|\widetilde{X}}[\widehat{\eta}_{k\text{-NN}}(x)]| < \Delta$ holding. This means that the function BAD can be defined to only have its first term, and so we only have to control the probability of the bad event in Lemma~\ref{lem:k-NN-classifier-ensure-close-enough-neighbors} to be at most $\delta^2/4$.
\end{proof}

\subsection{Fixed-Radius NN Classification}

We state an analogous result for fixed-radius NN classification with threshold distance~$h$. The proof ideas are nearly identical to the $k$-NN classification case and begin with the observation that inequality~\eqref{eq:regression-high-level-triangle-inequality-snippet} and Lemma~\ref{lem:k-NN-classifier-key-observation} both hold with $\widehat{\eta}_{k\text{-NN}}$ replaced with $\widehat{\eta}_{\text{NN}(h)}$, and $\mathbb{E}_{n|\widetilde{X}}$ replaced with $\mathbb{E}_{n|N_{x,h}}$, where $N_{x,h}$ is the number of training points within distance $h$ of $x$. In particular, we have the following lemma.
\begin{flemma}
\label{lem:NN-h-classifier-key-observation}
Let $\Delta\in(0,1/2]$.
For a test feature vector~$x \in \mathcal{X}$, we have
$\widehat{Y}_{\text{NN}(h)}(x) = \widehat{Y}_{\text{Bayes}}(x)$ if the
following two sufficient conditions hold:
\begin{itemize}

\item $| \widehat{\eta}_{\text{NN}(h)}(x) - \mathbb{E}_{n|N_{x,h}}[\widehat{\eta}_{\text{NN}(h)}(x)] | < \Delta$.

\item Test point~$x$ lands in the region
\begin{align*}
&\mathcal{X}_{\text{good-for-NN}(h)}^*(\Delta) \\
&\triangleq
 \Big\{ x\in\text{supp}(\mathbb{P}_X) :
    \eta(x) > \frac12,\,
    \mathbb{E}_{n|N_{x,h}}[\widehat{\eta}_{\text{NN}(h)}(x)] \ge \frac12 + \Delta
 \Big\} \\
&\quad
 \cup
 \Big\{ x\in\text{supp}(\mathbb{P}_X) :
    \eta(x) < \frac12,\,
    \mathbb{E}_{n|N_{x,h}}[\widehat{\eta}_{\text{NN}(h)}(x)] \le \frac12 - \Delta
 \Big\}.
\end{align*}
\end{itemize}
\end{flemma}
Once again, we ensure that these two conditions hold with high probability.  However, the main change is that instead of using the good region $\mathcal{X}_{\text{good-for-}k\text{-NN}}(p_{\text{min-mass}},\Delta)$, we now use the following:
\begin{align*}
&\mathcal{X}_{\text{good-for-NN}(h)}(p_{\text{min-mass}}, \Delta) \\
&\quad\triangleq
 \big\{ x\in\text{supp}(\mathbb{P}_X) : \\
&\quad\quad\quad
    \eta(x) > \frac12,\;
    \mathbb{E}[Y \mid X\in\mathcal{B}_{x,h}] \ge \frac12 + \Delta,\;
    \mathbb{P}_X(\mathcal{B}_{x,h}) \ge p_{\text{min-mass}}
 \big\} \\
&\quad\quad\;\cup
 \big\{ x\in\text{supp}(\mathbb{P}_X) : \\
&\quad\quad\quad\quad
    \eta(x) < \frac12,\;
    \mathbb{E}[Y \mid X\in\mathcal{B}_{x,h}] \le \frac12 - \Delta,\;
    \mathbb{P}_X(\mathcal{B}_{x,h}) \ge p_{\text{min-mass}}
 \big\}.
\end{align*}
The crucial difference between this region and the one used for $k$-NN classification is that here we only look at regions of the feature space for which a ball of radius $h$ has sufficiently high probability $p_{\text{min-mass}}$. This constraint ensures that if~$x$ lands in region $\mathcal{X}_{\text{good-for-NN}(h)}(p_{\text{min-mass}}, \Delta)$, then enough nearest neighbors will likely be found within threshold distance~$h$, where the average label of these nearest neighbors is on the correct side of threshold~1/2.

Thus, the high-level idea is that if $x \in \mathcal{X}_{\text{good-for-NN}(h)}(p_{\text{min-mass}}, \Delta)$, then $\widehat{Y}_{\text{NN}(h)}(x)= \widehat{Y}_{\text{Bayes}}(x)$ when:
\begin{itemize}

\item There are enough nearest neighbors within distance~$h$ of~$x$.  (We use Lemma~\ref{lem:radius-NN-helper1}, noting that $\mathbb{P}_X(\mathcal{B}_{x,h})\ge p_{\text{min-mass}}$ when $x \in \mathcal{X}_{\text{good-for-NN}(h)}(p_{\text{min-mass}}, \Delta)$.)

\item The average label of the nearest neighbors found is close enough to the expected mean: $|\widehat{\eta}_{\text{NN}(h)}(x) - \mathbb{E}_{n|N_{x,h}}[\widehat{\eta}_{\text{NN}(h)}(x)]| < \Delta$. (We use Lemma~\ref{lem:radius-NN-helper2} with $\varepsilon=2\Delta$ and noting that $\mathbb{P}_X(\mathcal{B}_{x,h})\ge p_{\text{min-mass}}$ when $x \in \mathcal{X}_{\text{good-for-NN}(h)}(p_{\text{min-mass}}, \Delta)$.) As before, we can drop this condition when $\Delta=1/2$.

\end{itemize}
We can readily turn this reasoning into a pointwise guarantee as well as one accounting for randomness in sampling $X=x \sim \mathbb{P}_X$.
\begin{ftheorem}[Fixed-radius NN classification pointwise agreement with the Bayes classifier]
Let $\delta\in(0,1)$, $p_{\text{min-mass}}\in(0,1]$, $\Delta\in(0,1/2]$, and $\gamma\in(0,1)$.  Under assumption \assumpTechnical, suppose that $x\in\mathcal{X}_{\text{good-for-NN}(h)}(p_{\text{min-mass}},\Delta)$, and either of the following holds:
\begin{itemize}

\item $\Delta\in(0,1/2)$ and the number of training data satisfies
\[
n \ge
\max\Big\{
\frac1{2(1-\gamma)p_{\text{min-mass}}\Delta^2} \log\frac4\delta,\,
\frac2{\gamma^2 p_{\text{min-mass}}}\log\frac2\delta
\Big\}.
\]

\item $\Delta=1/2$ and the number of training data satisfies
\[
n \ge
\frac2{\gamma^2 p_{\text{min-mass}}}\log\frac1\delta.
\]

\end{itemize}
Then with probability at least $1-\delta$ over randomness in sampling the training data, we have $\widehat{Y}_{\text{NN}(h)}(x) = \widehat{Y}_{\text{Bayes}}(x)$.
\end{ftheorem}

\begin{ftheorem}[Fixed-radius NN classification probability of agreement with the Bayes classifier]
Let $\delta\in(0,1)$, $p_{\text{min-mass}}\in(0,1]$, $\Delta\in(0,1/2]$, and $\gamma\in(0,1)$.  Under assumption \assumpTechnical, suppose that either of the following holds:
\begin{itemize}

\item $\Delta\in(0,1/2)$ and the number of training data satisfies
\[
n \ge
\max\Big\{
\frac1{(1-\gamma)p_{\text{min-mass}}\Delta^2} \log\frac4\delta,\,
\frac4{\gamma^2 p_{\text{min-mass}}}\log\frac{\sqrt{8}}\delta
\Big\}.
\]

\item $\Delta=1/2$ and the number of training data satisfies
\[
n \ge
\frac4{\gamma^2 p_{\text{min-mass}}}\log\frac2\delta.
\]

\end{itemize}
Then with probability at least
\[
1-\delta-\mathbb{P}_X([\mathcal{X}_{\text{good-for-NN}(h)}(p_{\text{min-mass}},\Delta)]^c)
\]
over randomness in sampling the training data and in sampling $X\sim\mathbb{P}_X$, we have $\widehat{Y}_{\text{NN}(h)}(X) = \widehat{Y}_{\text{Bayes}}(X)$.
\end{ftheorem}

\subsection{Kernel Classification}

Lastly, we provide a guarantee for kernel classification, which ends up dropping not only the smoothness requirement on~$\eta$ but also some of the decay assumptions on the kernel function~$K$ that appeared in kernel regression guarantees (Theorems~\ref{thm:kernel-regression-rate-of-convergence} and~\ref{thm:kernel-regression-expected-error-rate-of-convergence}).  The proof idea is once again nearly the same as before. Note that the triangle inequality we worked off of for kernel regression is slightly different from the ones for $k$-NN and fixed-radius NN regression:
\begin{equation}
|\widehat{\eta}_K(x; h)-\eta(x)|
\le
\Big|\widehat{\eta}_K(x;h)-\frac{A}{B}\Big|+\Big|\frac{A}{B}-\eta(x)\Big|,
\tag{inequality~\eqref{eq:kernel-regression-triangle-ineq},
reproduced}
\end{equation} \vspace{-.5em} \\
\noindent
where $A\triangleq\mathbb{E}[K(\frac{\rho(x,X)}{h})Y]$, and $B\triangleq\mathbb{E}[K(\frac{\rho(x,X)}{h})]$. However, the observation we made for $k$-NN classification to obtain Lemma~\ref{lem:k-NN-classifier-key-observation} still works.

\begin{flemma}
\label{lem:kernel-classifier-key-observation}
Let $\Delta\in(0,1/2]$. Denote $A\triangleq\mathbb{E}[K(\frac{\rho(x,X)}{h})Y]$, and $B\triangleq\mathbb{E}[K(\frac{\rho(x,X)}{h})]$.  For a test feature vector~$x \in \mathcal{X}$, we have $\widehat{Y}_{K}(x;h) = \widehat{Y}_{\text{Bayes}}(x)$ if the following two sufficient conditions hold:
\begin{itemize}

\item $ | \widehat{\eta}_{K}(x;h) - \frac{A}{B} | < \Delta$.

\item Test point~$x$ lands in the region
\begin{align*}
&\mathcal{X}_{\text{good-for-kernel}}^*(\Delta;K,h) \\
&\triangleq
 \Big\{ x\in\text{supp}(\mathbb{P}_X) :
    \eta(x) > \frac12,\,
    \frac{A}{B} \ge \frac12 + \Delta
 \Big\} \\
&\quad
 \cup
 \Big\{ x\in\text{supp}(\mathbb{P}_X) :
    \eta(x) < \frac12,\,
    \frac{A}{B} \le \frac12 - \Delta
 \Big\}.
\end{align*}

\end{itemize}
\end{flemma}
Lemma~\ref{lem:kernel-regression-rate-of-convergence-helper1} with choice $\varepsilon=2\Delta$ shows when the first condition holds with high probability. To ask that the second condition holds, we assume $x$ lands in the following good region that clearly is a subset of $\mathcal{X}_{\text{good-for-kernel}}^*(\Delta;K,h)$:
\begin{align*}
&\mathcal{X}_{\text{good-for-kernel}}(p_{\text{min-mass}},\Delta,\phi;K,h) \\
&\triangleq
 \Big\{ x\in\text{supp}(\mathbb{P}_X) :
    \eta(x) > \frac12,\,
    \frac{A}{B} \ge \frac12 + \Delta,\,
    \mathbb{P}_X(\mathcal{B}_{x,\phi h})\ge p_{\text{min-mass}}
 \Big\} \\
&\quad
 \cup
 \Big\{ x\in\text{supp}(\mathbb{P}_X) :
    \eta(x) < \frac12,\,
    \frac{A}{B} \le \frac12 - \Delta,\,
    \mathbb{P}_X(\mathcal{B}_{x,\phi h})\ge p_{\text{min-mass}}
 \Big\},
\end{align*}
where $\phi>0$ is any constant for which $K(\phi) > 0$.  Similar to the good region $\mathcal{X}_{\text{good-for-NN}(h)}$ for fixed-radius NN classification, here we also ask that points inside the good region will likely have enough nearest neighbors found within distance~$h$ via the constraint $\mathbb{P}_X(\mathcal{B}_{x,h}) \ge p_{\text{min-mass}}$.

We can once again produce a pointwise guarantee and a guarantee accounting for randomness in sampling $X=x\sim\mathbb{P}_X$.
\begin{ftheorem}
[Kernel classification pointwise agreement with the Bayes classifier]
Let $\phi>0$ be any constant such that~${K(\phi)>0}$.  Under assumption \assumpTechnical, suppose that 
\[
x\in\mathcal{X}_{\text{good-for-kernel}}(p_{\text{min-mass}},\Delta,\phi;K,h).
\]
Let the number of training data satisfies
\begin{align*}
n
&\ge
  \max\Big\{
    \frac{2\log\frac{4}{\delta}}{[K(\phi)p_{\text{min-mass}}]^{2}},\;
\frac{8\log\frac{4}{\delta}}{\Delta^2[K(\phi)p_{\text{min-mass}}]^{4}}\Big\}.
\end{align*}
Then with probability at least $1-\delta$ over randomness in sampling the training data, we have $\widehat{Y}_K(x;h) = \widehat{Y}_{\text{Bayes}}(x)$.
\end{ftheorem}

\begin{ftheorem}
[Kernel classification probability of agreement with the Bayes classifier]
Let $\phi>0$ be any constant such that~${K(\phi)>0}$.  Under assumption \assumpTechnical, suppose that the number of training data satisfies
\begin{align*}
n
&\ge
  \max\Big\{
    \frac{4\log\frac{4}{\delta}}{[K(\phi)p_{\text{min-mass}}]^{2}},\;
\frac{16\log\frac{4}{\delta}}{\Delta^2[K(\phi)p_{\text{min-mass}}]^{4}}\Big\}.
\end{align*}
Then with probability at least $1-\delta$ over randomness in sampling the training data and the test point~$X\sim\mathbb{P}_X$, we have $\widehat{Y}_K(X;h) = \widehat{Y}_{\text{Bayes}}(X)$.
\end{ftheorem}

\section{Guarantees Under a Margin Bound Condition}
\label{sec:classification-margin-conditions}

In all our guarantees for classification, the probability of a classifier disagreeing with the optimal Bayes classifier depended on the probability of landing near the decision boundary. An alternative to procuring classification guarantees is to assume that this probability is small.  A series of papers develop theory for what is called a margin bound \citep{mammen_1999,tsybakov_2004,audibert_2007}. Formally, these authors assume the following condition
\begin{equation}
\mathbb{P}\Big(\Big| \eta(X) - \frac{1}{2} \Big| \le s\Big)
\le C_\text{margin} s^\varphi,
\label{eq:mammen-tsybakov-first}
\end{equation}
for some finite $C_{\text{margin}} > 0$, $\varphi > 0$, and all $0 < s \le s^*$ for some $s^* \le 1/2$. In other words, the probability that feature vector~$X$ lands within distance $s$ of the decision boundary decays as a polynomial in~$s$ with degree~$\varphi$. In particular, faster decay (\ie, smaller $C_{\text{margin}}$ and higher degree $\varphi$) means that landing near the decision boundary (\ie, in the margin) has lower probability, so the classification problem should be easier.

\citet[Theorem~7]{chaudhuri_dasgupta_2014} show that if the feature space $\mathcal{X}=\mathbb{R}^d$, the metric $\rho$ is the Euclidean norm, the regression function $\eta$ satisfies assumption \assumpHolder~ and margin condition~\eqref{eq:mammen-tsybakov-first} with $s^*=1/2$, and that feature space~$\mathcal{X}$ satisfies a technical condition of containing a fraction of every ball centered in it, then
\[
\mathbb{P}(\widehat{Y}_{k\text{-NN}}(X)\ne Y)
- \mathbb{P}(\widehat{Y}_{\text{Bayes}}(X)\ne Y)
\le \mathcal{O}(n^{-\alpha (\varphi + 1)/(2\alpha + d)}),
\]
which matches the lower bound of~\citet[Theorem~3.5]{audibert_2007}. That this result follows from the general $k$-NN classification guarantee (Theorem~\ref{thm:k-NN-classifier-weaker-assumptions-main-result}) suggests the general $k$-NN classification guarantee to be reasonably tight.

In fact, more is true if we assume both H\"{o}lder continuity and a margin condition to hold for regression function~$\eta$.  If we further assume feature distribution $\mathbb{P}_X$ to satisfy the strong minimal mass assumption by \citet{gadat_2016} (this assumption is discussed in Section~\ref{sub:toward-expected-regression-error-partitioning-the-feature-space}), then for $k$-NN classification to be uniformly consistent, it must be that the probability of landing in low probability regions of the feature space~$\mathcal{X}$ must decay sufficiently fast (precisely what this means is given by \citealt[Assumption A4]{gadat_2016}).  Moreover, with $\eta$ satisfying both H\"{o}lder continuity and a margin condition, even if the probability of landing in low probability regions of the feature space decays fast, $\mathbb{P}_X$ must satisfy the strong minimal mass assumption for $k$-NN classification to be uniformly consistent. This pair of results is provided by \citet[Theorem~4.1]{gadat_2016}.\footnote{As a technical remark, \citet{gadat_2016} use a slightly different version of margin condition~\eqref{eq:mammen-tsybakov-first} that requires the bound in \eqref{eq:mammen-tsybakov-first} to hold for all $s>0$, and not just for small enough $s$.}

While asking that the probability of landing near the decision boundary be small intuitively makes sense, for a variety of problems, determining the constants $C_\text{margin}$, $\varphi$, and $s^*$ in margin condition~\eqref{eq:mammen-tsybakov-first} is not straightforward.  To sidestep this task of determining these constants, in the next chapter, we explore how clustering structure enables nearest neighbor prediction to succeed in three contemporary applications.  Naturally, clustering structure essentially implies a sufficiently large margin between clusters, \ie, the probability of landing in a boundary region between two clusters is small.

We shall see that indeed clustering structure does relate to the margin condition~\eqref{eq:mammen-tsybakov-first}.  In Section~\ref{sec:clustering-and-margin-conditions}, we compute the constants $C_\text{margin}$, $\varphi$, and $s^*$ in a toy example in which the two label classes $Y=0$ and $Y=1$ correspond precisely to two different univariate Gaussian distributions. Specifically, suppose data are generated from classes $Y=0$ and $Y=1$ with equal probability. If $Y=1$, then we assume feature vector~$X$ to be sampled from univariate Gaussian distribution $\mathcal{N}(\mu, \sigma^2)$. Otherwise, $X$ is assumed to be sampled from $\mathcal{N}(-\mu, \sigma^2)$.  In this toy example, the margin condition holds with $\varphi=1$, $C_{\text{margin}}=\frac{16\mu}{3\sigma\sqrt{2\pi}}\exp(-\frac{\mu^2}{4\sigma^2})$, and $s^*=\min\{\frac14, \frac{3\mu^2}{16\sigma^2}\}$.  We suspect though that computing these constants in general even under a clustering assumption cannot easily be done analytically.

\chapter{Prediction Guarantees in Three Contemporary Applications}
\label{chap:case-studies}

\newcommand{\numClusters}{r}
\newcommand{\numBigClusters}{R}
\newcommand{\dsmmNumClusters}{r}
\newcommand{\dsmmNumClustersMax}{r}
\newcommand{\ratioVar}{R}

\graphicspath{{figures/}}

Our coverage of general results in regression and classification shows a seemingly complete picture of the theory. However, a variety of assumptions are made along the way that are not easy to check in practice. For example, given a real dataset, how would we check whether the regression function $\eta$ is H\"{o}lder continuous, or to measure the probability of landing in an ``effective boundary''? These are not straightforward questions to answer. In the classification setting, for instance, it seems unrealistic to know where the effective boundary would even be---it is precisely a region around the true boundary, which is what training a classifier aims to find.

Instead of trying to estimate H\"{o}lder continuity parameters of the regression function or trying to find the effective boundary or even characterize how, in the classification setting, the regression function behaves around the boundary, perhaps a more realistic question to ask is whether we can come up with other sufficient conditions on the distribution between feature vector $X$ and label $Y$ that enable us to procure theoretical performance guarantees---conditions that more readily relate to structural properties present in different applications.

Toward this goal, we examine how clustering structure enables nearest neighbor prediction to succeed in three contemporary applications with rather different structure, not only in what the data look like but also what the prediction task at hand is. These illustrative examples also showcase how nearest neighbor classification appears as part of more elaborate prediction systems.  The applications we cover are as follows, where we begin with the one most closely resembling the vanilla binary classification setup described in Section~\ref{sec:classification-setup}:

\begin{itemize}

\item \textbf{Time series forecasting (Section~\ref{chap:time-series-classification}).}
To forecast whether a news topic will go viral, we can compare its Tweet activity to those of past news topics that we know to have gone viral as well as those that did not. Each news topic is represented by a discrete time series (tracking Tweet activity) with a single label for whether the topic ever goes viral or not, \ie, the ``feature vector'' of the news topic is a time series. The training data are $n$ news topics for which we know their time series $X_1, \dots, X_n$ and labels $Y_1, \dots, Y_n$; the pairs $(X_1,Y_1),\dots, (X_n,Y_n)$ are modeled as i.i.d. In particular, time is indexed by integer indices so that $X_i(t)$ is the value of the $i$-th training time series at integer time $t\in\mathbb{Z}$.  For a test news topic, the goal is to predict its label $Y$ (\ie, forecast whether it will ever go viral) given only the values of its time series $X$ observed at time steps $1,2,\dots,T$ (so we observe $X(1), X(2), \dots, X(T)$ but no other values of $X$; in contrast, for training time series we assume we have access to their values for all time indices).  Ideally we want to have high classification accuracy when $T$ is small, corresponding to making an accurate forecast as early as possible.

This time series forecasting setup differs from the vanilla binary classification setup in two key ways. First, for the test time series, we only observe up to $T$ values rather than the full time series (note that we still assume access to complete time series in the training data).  Next, there is temporal structure that we can take advantage of. Specifically, in comparing two time series to see how well they match up, we first align them, which for simplicity we do by shifting one of them in time.  In practice, mechanisms more elaborate than time shifts are used to align time series, such as dynamic time warping \citep{sakoe_chiba_1978}.

Clustering structure appears since it turns out that there are only a few ways in which a news topic goes viral (Figure~\ref{fig:clusters}). This motivates a probabilistic model in which different time series appear as noisy versions of $\numClusters$ canonical time series that act as cluster centers, where each canonical time series (and thus each cluster) shares the same label.  We present theoretical guarantees for 1-NN and kernel time series classifiers under this probabilistic model.

\begin{figure}[t]
\centering
\includegraphics[scale=.6, clip=true, trim=3em 1.5em 3em 2em]{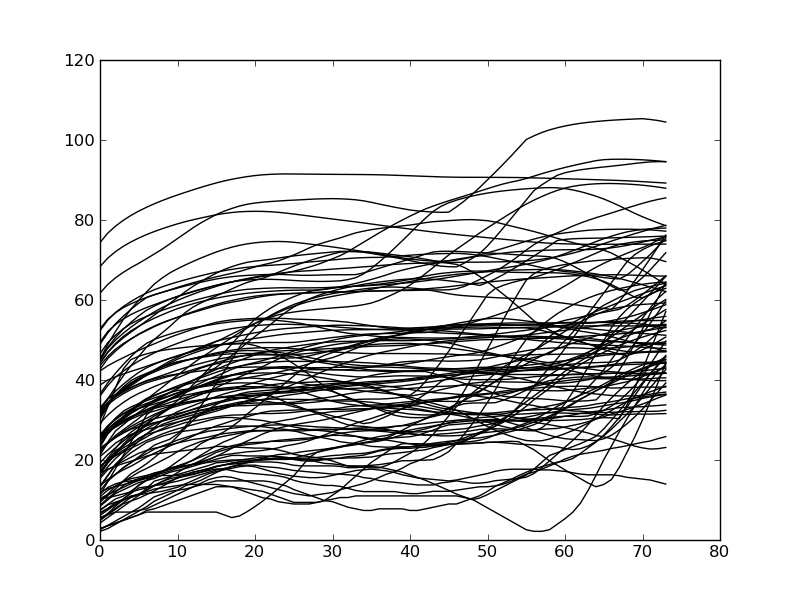}
\includegraphics[scale=.6, clip=true, trim=3em 1.5em 3em 2em]{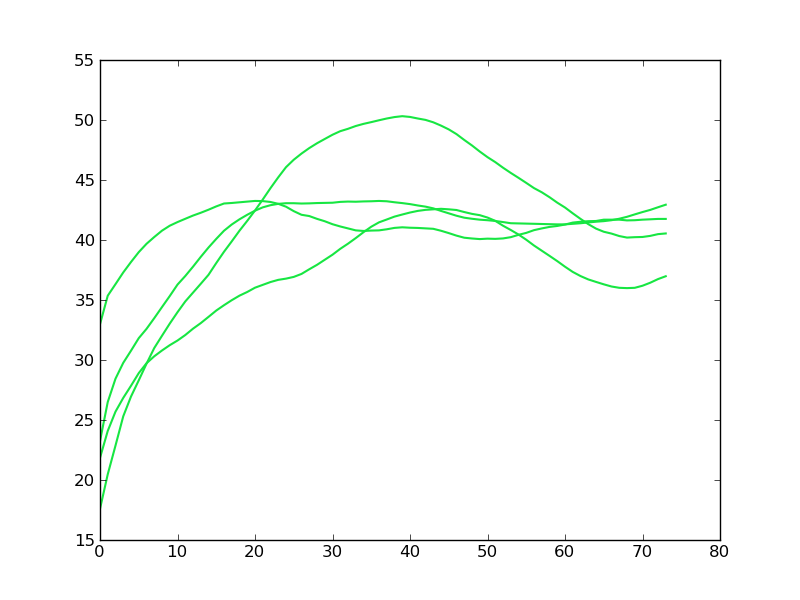}
\includegraphics[scale=.6, clip=true, trim=3em 1.5em 3em 2em]{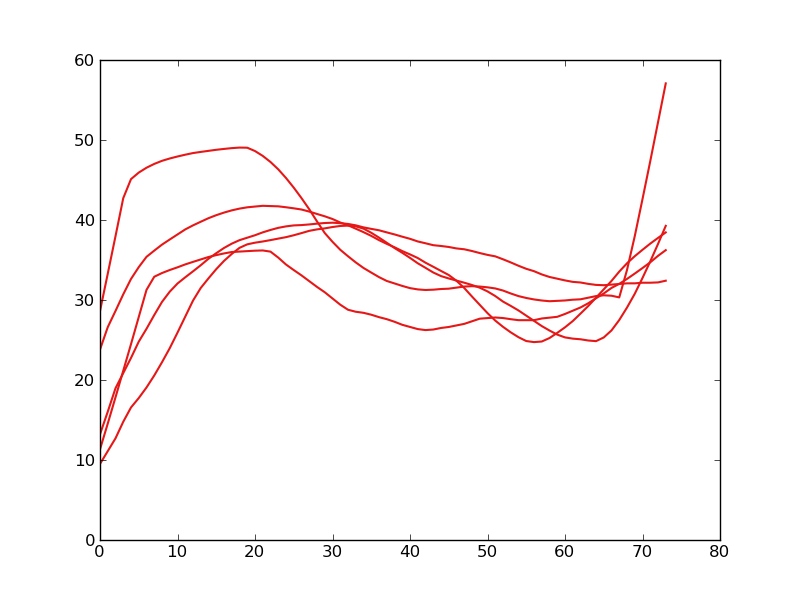}
\hspace{.5in}~~~ \\
\includegraphics[scale=.6, clip=true, trim=3em 1.5em 3em 2em]{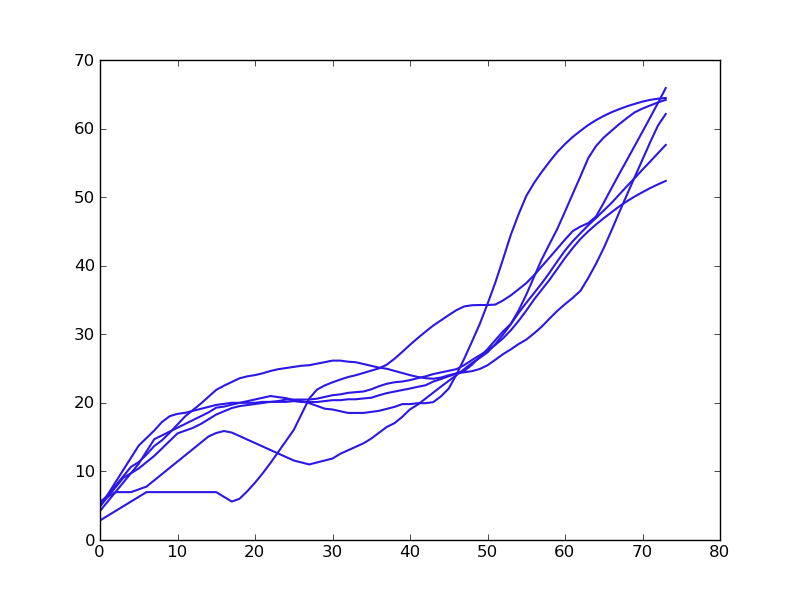}
\includegraphics[scale=.6, clip=true, trim=3em 1.5em 3em 2em]{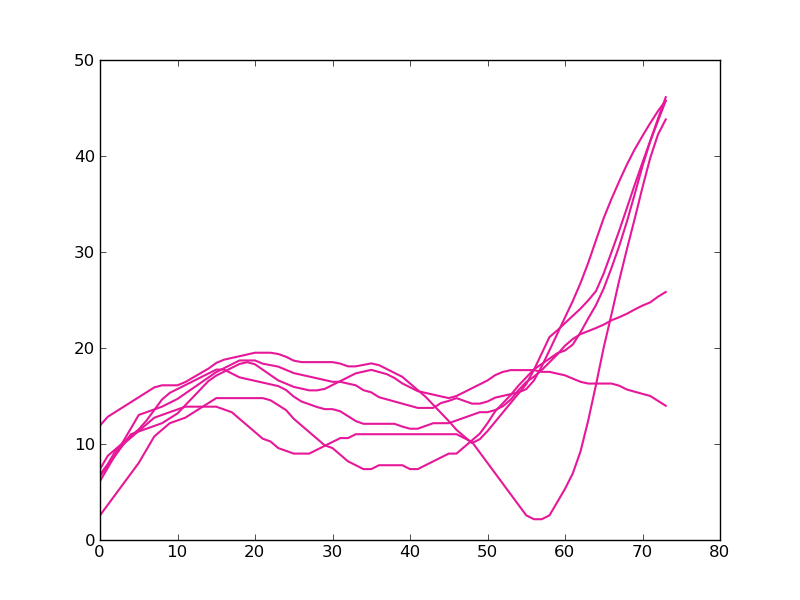}
\includegraphics[scale=.6, clip=true, trim=3em 1.5em 3em 2em]{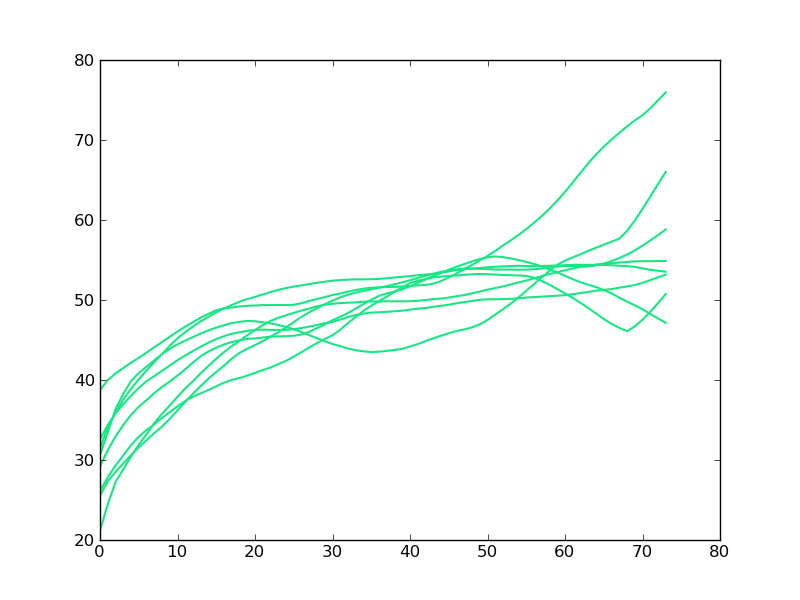}
\includegraphics[width=.6in]{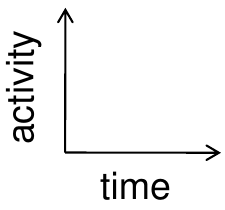}
\caption[How news topics go viral on Twitter]
{How news topics go viral on Twitter (figure source: \citealt{georgehc_time_series_nips}). The top left shows some time series of activity leading up to a news topic going viral. These time series superimposed look like clutter, but we can separate them into different clusters, as shown in the next five plots. Each cluster represents a ``way'' that a news topic becomes viral.}
\label{fig:clusters}
\end{figure}

\item \textbf{Online collaborative filtering (Section~\ref{chap:online-collaborative-filtering}).}
In a recommendation system that has both massive numbers of users and of items such as Netflix, we can recommend to each user an item that the user has not already consumed by looking at what similar users like, \eg, if Alice is similar to Bob, and Bob likes oranges, then perhaps Alice likes oranges too. The goal is to recommend, over time, as many items to users that they find likable. To come up with recommendations, at each time step we predict what users will find likable. This setup is called collaborative filtering since we filter the vast amount of items in finding recommendations for a user, and we do so by looking at which users are similar, which can be thought of as collaboration.\footnote{As a terminological remark, this setup uses what is called user-user collaborative filtering to emphasize that we look at which users are similar. There is also item-item collaborative filtering that instead makes recommendations based on which items are similar, \eg, if apples are similar to pears, and Alice likes apples, then perhaps Alice likes pears too.} The setup is online in that we continuously recommend items and receive revealed user ratings as feedback.

Online collaborative filtering has a component of forecasting over time but is in some ways considerably more complex than the earlier time series forecasting setup.  Previously, each news topic was associated with a single label of whether it ever goes viral or not. Here, there are many more labels to classify. Specifically, each user-item pair has a label corresponding to whether the user likes the item or not.  In addition to there being more labels, the recommendation system is choosing the training data. In particular, at any time step, the training data comprise of the ratings that users have revealed for items so far.  What the system recommends to users now affects what ratings users reveal, which make up the training data at the next time step! Moreover, what the system learns about the users from their revealed ratings affect its future recommendations, so an exploration-exploitation tradeoff naturally arises.

Once again, clustering structure appears, this time in real movie ratings data.  In particular, similar users form clusters and similar items form clusters as well, such that for a ratings matrix where rows index users and columns index items, we can reorder the rows and columns to make it apparent that user and item clusters appear (Figure~\ref{fig:bctf}).  For simplicity, we focus our exposition on the situation where there are $\numClusters$ user clusters, and we do not model the item clusters. The theoretical analysis is for an online recommendation algorithm that uses fixed-radius NN classification to identify likable items.

\begin{figure}
\centering
\includegraphics[width=9cm]{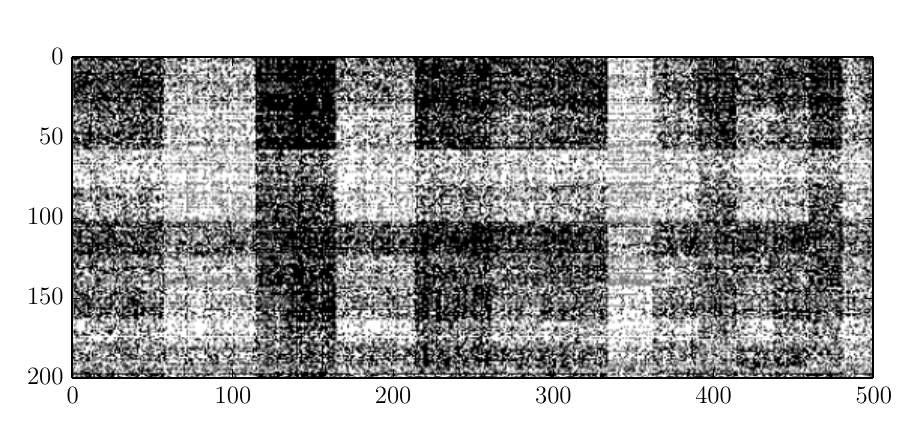}
\caption{Top users by top movies ratings matrix with rows and columns reordered to show clustering (block structure) of users and items for the MovieLens 10M dataset (figure source: \citealt{georgehc_collaborative_filtering_nips}).}
\label{fig:bctf}
\end{figure}

\item \textbf{Patch-based image segmentation (Section~\ref{chap:image-segmentation}).}
Given a medical image such as a CT scan of a patient's abdomen, to delineate where an organ is in an image, we can compare patches of the image to patches in a training database of images for which we know where the organ is. The goal is to classify, for every pixel in the test image, whether it is part of the organ of interest~or~not.

In a departure from the previous two applications, patch-based image segmentation looks at prediction over space rather than over time. Here, image structure is key: nearby pixels and patches are similar. That there is local smoothness in an image also means that the training data are not independent as image patches centered at nearby pixels overlap.

Clustering structure appears once more.  As it turns out, image patches for naturally occurring images can be very accurately modeled by Gaussian mixture models \citep{epll,patch_gmm}! In particular, plausible image patches appear to, for example, have a single homogeneous texture or be around where an edge appears. We can model image patches as being generated from (up to) $\numClusters$ canonical image patches. We present theoretical guarantees for for 1-NN and kernel classifiers that figure out whether each pixel is part of the organ of interest.

\end{itemize}
\begin{figure}
\includegraphics[scale=.8]{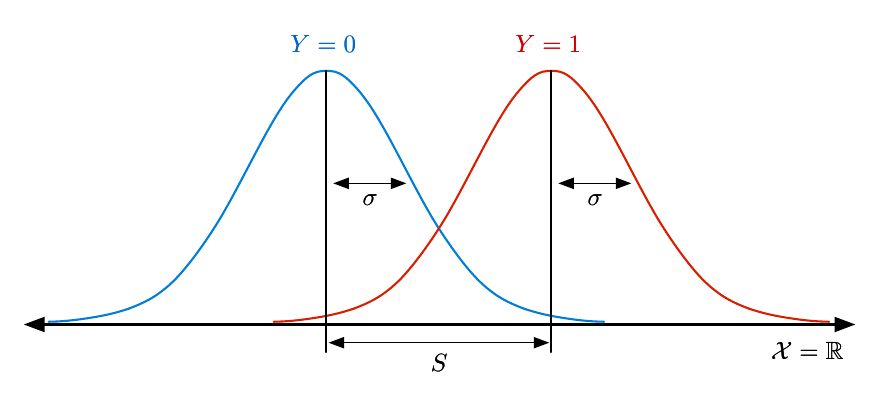}
\vspace{-3em}
\caption{An illustrative example highlighting one of the recurring themes of this chapter: separation between clusters should be large enough to overcome noise. In this toy example, the feature distribution $\mathbb{P}_X$ is a univariate Gaussian mixture model with $\numClusters=2$ equiprobable clusters corresponding to the two label classes $Y=0$ (blue) and $Y=1$ (red). The probability density functions are overlaid for each of the two clusters. In this example, the two Gaussians both have standard deviation $\sigma$, and their means are separated by a distance $S$. The larger the separation $S$ is compared to the noise level $\sigma$, the easier the classification problem becomes.}
\label{fig:case-studies-high-level}
\end{figure}

\noindent
Despite the structural differences between these three applications, the theory developed for each is of the same flavor: nearest neighbor classification can achieve arbitrarily high accuracy so long as the clusters are sufficiently separated and the training data include examples from all the clusters. Let's illustrate these two ideas with a simple toy example using 1-NN classification.

Consider when the feature distribution is a univariate Gaussian mixture model with $\numClusters=2$ clusters as shown in Figure \ref{fig:case-studies-high-level}. One cluster corresponds to label $Y=0$ (blue) and the other to label $Y=1$ (red).  The separation between the two cluster centers is $S$, and the noise level is given by the standard deviation $\sigma$ of the two Gaussians.  First off, if we do not have at least one training example for every cluster, then this means that all the training data have the same label. A test feature vector with the other label will get misclassified by a 1-NN classifier. Next, supposing that there is at least one training example per cluster, 1-NN classification has a better chance of succeeding when separation $S$ is large compared to the noise level $\sigma$. With $S\gg \sigma$, a test feature vector $x\sim\mathbb{P}_X$ will have a nearest neighbor training example that highly likely originates from the same cluster as $x$. As a result, 1-NN classification is likely to be correct.  However, if $S$ is much smaller than $\sigma$, then the training example closest to $x$ could easily be generated from a different cluster than $x$, resulting in a misclassification by 1-NN classification.

We can extend this basic intuition to the case when there are $\numClusters\ge2$ clusters, and where the noise models are not just Gaussian.  The basic strategy to establishing the theoretical guarantees of this chapter is as follows.  We model the data to originate from $\numClusters$ clusters, where within each cluster, the data have the same label (note that even for $\numClusters>2$, we constrain the number of labels to still be 2, so multiple clusters can share the same label). Despite not knowing what the clusters are, 1-NN classification can succeed if the following two sufficient conditions happen:
\begin{enumerate}

\item If the number of training data $n$ is large enough, then there will with high probability be a training example from each of the $\numClusters$ clusters. In particular, across all three applications, having $n=\widetilde{\Theta}(\numClusters)$ is sufficient.

\item For a given test data point $X$, we want its nearest neighbor within the training data to originate from the same cluster as~$X$ (which would mean that they have the same label, so 1-NN classification would correctly label $x$). We can ensure this if the clusters are well-separated, enough so that noise does not easily make a training data point from one cluster appear to come from another.

\end{enumerate}
Some tweaks on this general idea are needed to extend the analysis to online collaborative filtering and patch-based image segmentation.  In filling in the details for all three applications, application-specific structure appears.  In time series forecasting, how much of the test time series is observed and time shifts matter. In online collaborative filtering, the numbers of users and of items matter.  In patch-based image segmentation, image structure matters in the form of a local ``jigsaw condition'' that pieces together nearby patches. The dependence of the theoretical results on these structural elements is appealing due to their ease of interpretability within their respective applications. Of course, the clusters also readily lend themselves to interpretation.

\vspace{.3em}
\noindent
\textbf{Connection to classification results of Chapter~\ref{chap:classification}.}
The guarantees presented in the previous chapter on $k$-NN, fixed-radius NN, and kernel classification were stated in terms of the decision boundary, where classification is difficult.  One of the conditions we saw was a margin bound condition \eqref{eq:mammen-tsybakov-first} that says how fast the probability of landing near the decision boundary decays.  In many problem setups, coming up with an analytical expression that characterizes the probability of landing near the decision boundary is not straightforward.

In this chapter, we instead assume that data are generated as perturbed versions of cluster centers, and there is some upper bound on the noise level $\sigma$.  The noise level $\sigma$ controls how fast the probability of a data point being far from its cluster center decays.  We will typically not know exactly where the decision boundary is, aside from it being somewhere between clusters of opposite labels.  However, if the cluster centers are sufficiently far apart, then the probability of landing near the decision boundary will be low since it means being far away from the cluster centers.  In this way, clustering structure can be viewed as a sufficient condition that implicitly ensures that the probability of landing near the decision boundary has fast decay.  In Section~\ref{sec:clustering-and-margin-conditions}, for a specific Gaussian mixture model with $\numClusters=2$ clusters in Figure~\ref{fig:case-studies-high-level} (namely when the two cluster centers are at~$-\mu$ and~$\mu$ for $\mu>0$), we directly relate clustering structure to margin bound condition~\eqref{eq:mammen-tsybakov-first}.

\vspace{.3em}
\noindent
\textbf{Relaxing some assumptions on the distance function used.}
As an important technical remark, clustering structure also enables us to surmount distance-related technical hurdles absent in our coverage of regression and classification results in Chapters~\ref{chap:regression} and~\ref{chap:classification}.  In time series forecasting, the distance we examine for comparing time series that allows for time shifts is not a metric, which was assumed for all the results in Chapters~\ref{chap:regression} and~\ref{chap:classification}. In online collaborative filtering, the distance between users can be viewed as being noisy rather than deterministic, which also deviates from our previous coverage.  Despite these challenges, clustering structure enables us to establish nonasymptotic theoretical guarantees for nearest neighbor prediction.  Thus, while the presence of clusters is a strong (and, for various datasets, realistic) assumption, it allows us to relax other requirements in our earlier theoretical analysis.

\vspace{.3em}
\noindent
\textbf{Chapter outline.}
In the next sections (Sections \ref{chap:time-series-classification}, \ref{chap:online-collaborative-filtering}, \ref{chap:image-segmentation}), we provide an overview of the theoretical and empirical results for these contemporary three applications, closely following the thesis by \citet{georgehc_thesis} that subsumes and extends prior work by \citet{georgehc_time_series_nips}, \citet{georgehc_collaborative_filtering_nips}, and \citet{georgehc_image_segmentation_miccai}.  Full proofs are omitted but we point out where they can be found in Chen's thesis~\citep{georgehc_thesis}. Each of these three sections begins with a rather informal overview before doing a deeper dive.

The theory presented in this chapter has a key limitation: the proof techniques use a worst-case analysis that make them unable to readily extend to $k$-NN classification for $k>1$. This turns out to be related to dealing with training examples that are outliers or adversarial. The basic idea is that when the number of training data $n$ approaches $\infty$, we are going to see more and more outliers: points from a cluster that appear like they are from another.  A 1-NN approach is not robust to such outliers. The clunky theoretical workaround is to randomly subsample the training data to be of size $\widetilde{\Theta}(\numClusters)$, where $\numClusters$ is the number of clusters. We discuss this theoretical limitation in Section~\ref{sec:case-studies-conclusion} and mention how recent work on making 1-NN classifiers more robust \citep{wang_2017} can potentially resolve the issue.

Note that we reuse variable names across the applications, with the same variable name meaning something similar each time. We have already previewed some key recurring variables: $\numClusters$ is the number of clusters, $S$ (possibly with a superscript) is going to represent a separation between clusters of different labels, and $\sigma$ is a noise level (not always the standard deviation of a Gaussian).

\section{Time Series Forecasting}
\label{chap:time-series-classification}

In time series analysis, a recurring task is detecting anomalous events.  Our running example throughout this section is forecasting which news topics will go viral on Twitter \textit{before} Twitter declares the topic to go viral or not. To do so, we compare different news topics' time series of Tweet activity.  We assume we can collect such time series data for different historical news topics, where Twitter supplies ground truth labels for whether the news topics ever went viral or not (by whether they ever end up within the top 3 list of what Twitter calls ``trending topics'').  For this problem, \citet{nikolov_2012} have shown that a kernel classifier can predict whether a news topic will go viral in advance of Twitter 79\% of the time, with a mean early advantage of 1 hour and 26 minutes, a true positive rate of 95\%, and a false positive rate of~4\%. We summarize this result later in this section.

The success of nearest neighbor methods in time series classification is hardly limited to forecasting which news topics will go viral. Such methods have also been used, for example, to detect abnormal brain activity in EEG recordings \citep{time_series_abnormal_brain_activity}, classify protein sequences \citep{protein_sequence_classification}, and predict whether a cell phone subscriber will switch providers \citep{nn_time_series_classification}. In fact, while numerous standard classification methods have been tailored to classify time series, a simple nearest-neighbor approach has been found to be hard to beat in terms of classification performance on a variety of datasets \citep{xi_2006}, with results competitive to or better than various other more elaborate methods such as neural networks \citep{nanopoulos_2001}, decision trees \citep{rodriguez_2004}, and support vector machines \citep{wu_2004}. More recently, researchers have examined which feature space or distance to use with nearest-neighbor classification (\eg, \citealt{bagnall_2012,gustavo_2011,devlin_2015,ding_2008,weinberger_2009}).

Despite the plethora of empirical work showing that nearest neighbor methods work well in a variety of contexts for time series classification, nonasymptotic theoretical guarantees for such methods were not established until recently by \citet{georgehc_time_series_nips}, specifically for \mbox{1-NN} and kernel classifiers.  In this section, we walk through their results and some follow-up work by \citet[Chapter~3]{georgehc_thesis}.  We begin in Section~\ref{sec:time-series-inference} by stating the 1-NN and kernel time series classifiers to be analyzed. The theoretical guarantees rely on an underlying probabilistic model for time series that Chen \textit{et al.}~call a \textit{latent source model for time series classification}, presented in Section~\ref{sec:theory}. This model is guided by the hypothesis that there are $\numClusters$ canonical time series patterns that explain all plausible time series.  These are precisely the $\numClusters$ clusters. The main result of this section is as follows.

\begin{ftheorem}
[Informal statement of Theorem \ref{thm:time-series-main-result}]
\label{thm:time-series-main-result-informal}
Let $\delta\in(0, 1)$ be a probability tolerance.  Under the latent source model for time series classification, if the training time series with opposite labels ``viral'' and ``not viral'' are sufficiently different (depends on $\delta$), and if we have $n = \Theta(\numClusters \log \frac{\numClusters}{\delta})$ labeled training time series, then 1-NN and kernel classification each correctly classify a time series with probability at least $1-\delta$ after observing its first $T=\Omega(\log \frac{\numClusters}{\delta})$ time steps.
\end{ftheorem}
As the analysis accounts for how much of the test time series we observe, this result readily applies to the ``online'' setting in which a time series is to be classified while it streams in, as is the case for forecasting which ongoing news topics go viral, along with the ``offline'' setting where we have access to the entire time series.

Given that this result assumes there to be clusters, one could ask why not just estimate the cluster centers, which could more directly help us with prediction? In fact, if we knew what the true cluster centers were, then the optimal classifier is an ``oracle'' \textit{maximum a posteriori} (MAP) classifier, which we present in Section \ref{sec:time-series-oracle}.  This classifier is an oracle since it can access the true cluster centers, which we of course do not know in practice. If we had accurate estimates for the cluster centers, then we could just plug in these estimates in place of the unknown true cluster centers in the oracle MAP classifier to obtain an accurate classifier. It turns out though that accurately estimating the cluster centers is in some sense a harder problem than the classification task at hand.  In Section~\ref{sec:time-series-learn-latent-sources}, we show that existing approaches for estimating cluster centers either require more stringent conditions on the data or require more training data than what is sufficient for time series classification.

While 1-NN and kernel classifiers do not perform explicit clustering, they actually approximate the oracle MAP classifier that knows the $\numClusters$ true cluster centers.  We discuss this approximation in Section~\ref{sec:time-series-oracle}.  Naturally, we can't hope to outperform this oracle MAP classifier. However, if the number of training data grows sufficiently large, then we would expect 1-NN and kernel classifiers to better approximate this oracle MAP classifer and thus become closer to optimal.  To gauge how far 1-NN and kernel classifiers are from being optimal, in Section \ref{sec:time-series-learn-latent-sources}, we present a lower bound on the misclassification probability for \textit{any} classifier and compare this bound with the upper bounds on the misclassification probabilities of 1-NN and kernel classification.

We relate clustering structure to existing margin conditions that have been developed for classification in Section~\ref{sec:clustering-and-margin-conditions}. This provides a bridge to our earlier discussion of margin conditions for classification in Section~\ref{sec:classification-margin-conditions}.

Finally, we share some experimental results in Section \ref{sec:time-series-experiments}.  Using synthetic data, kernel classification outperforms \mbox{1-NN} classification early on when we observe very little of the time series to be classified. This suggests kernel classification to be better suited for forecasting anomalous events than 1-NN classification. Returning to our running example of forecasting viral news on Twitter, we summarize the kernel classification results by \citet{nikolov_2012}.

\subsection{Nearest Neighbor Time Series Classifiers}
\label{sec:time-series-inference}

We represent each time series as a function mapping $\mathbb{Z}$ (time indices) to~$\mathbb{R}$ (observed values, \eg, Tweet rate). Given time series $\obsVar$ observed at time steps $1, 2, \dots, T$, we want to classify it as having either label~$1$ (will at some point go viral) or~$0$ (will never go viral).  To do so, we have access to labeled training data consisting of time series $\obsVar_1, \dots, \obsVar_n$ with corresponding labels $\labelVar_1, \dots, \labelVar_n \in \{0, 1\}$.  We denote the value of time series $\obsVar$ at time $t$ as $\obsVar(t)$.

To come up with nearest neighbor and kernel classifiers, we specify distance and kernel functions. For the latter, we use a Gaussian kernel $K(s) = \exp(-\frac12 s^2)$. As for the distance function, the basic idea is that to measure how far apart two time series are, we first align them as well as possible, for simplicity by shifting one of them in time.  We denote the distance function between two time series $X$ and $X'$ by $\rho^{(T)}(X, X' \maxtimeshift{.85} \Delta_{\max})$, which has two parameters: $T$ is the number of time steps of $X$ that we observe, and $\Delta_{\max}$ is the maximum number of time steps we advance or delay~$X'$ by to align it with~$X$ (we assume we can access any time step of $X'$). In particular, $\rho^{(T)}(X, X' \maxtimeshift{.85} \Delta_{\max})$ is defined as follows.  We compute the Euclidean distance (looking only at time steps $1,2,\dots,T$) between $X$ and time-shifted versions of~$X'$, and use whichever Euclidean distance is smallest across the different time shifts:
\begin{align*}
\rho^{(T)}(X, X' \maxtimeshift{0.85} \Delta_{\max})
&=\min_{\Delta\in\{-\Delta_{\max},\dots,0,\dots,\Delta_{\max}\}}
    \| X
       -
       X'\,\scalebox{0.85}{\circled{\ensuremath{\leftarrow}}}\,\Delta \|^{(T)},
\end{align*}
where $X'\timeadvance{0.85}\Delta$ denotes time series $X'$ advanced by $\Delta$ time steps (\ie, $(X'\timeadvance{0.85}\Delta)(t) = X'(t+\Delta)$ for all $t\in\mathbb{Z}$), and $\|\cdot\|^{(T)}$ denotes Euclidean norm looking only at time steps $1,2,\dots,T$ (\ie, $\|Z\|^{(T)} = \sqrt{\sum_{t=1}^T (Z(t))^2}$ for any time series $Z$).  Note that $\rho^{(T)}(\cdot,\cdot\maxtimeshift{0.85} \Delta_{\max})$ is not a metric as it is not symmetric.  With the above choices for the distance and kernel functions, we obtain the following classifiers.

\vspace{.3em}
\noindent
\textbf{1-NN time series classifier.}
Letting $Y_{(1)}(x)$ be the label of the nearest training data point to~$x$ according to distance $\rho^{(T)}(\cdot, \cdot \maxtimeshift{0.85} \Delta_{\max})$, the \mbox{1-NN} classifier is given by
\begin{equation}
\widehat{Y}_{1\text{-NN}}^{(T)}(x;\Delta_{\max})
=
Y_{(1)}(x).
\label{eq:decision-rule-nn}
\end{equation}

\vspace{.3em}
\noindent
\textbf{Kernel time series classifier.}
The kernel time series classifier using the Gaussian kernel with bandwidth $h$ is given by
\begin{equation}
\widehat{\labelVar}_{\text{Gauss}}^{(T)}(\obsVar; \Delta_{\max}, h)
=\begin{cases}
   1 & \text{if }V_1^{(T)}(\obsVar; \Delta_{\max}, h)
        \ge V_0^{(T)}(\obsVar; \Delta_{\max}, h), \\
   0 & \text{otherwise},
\end{cases}
\label{eq:decision-rule-with-min}
\end{equation}
where $V_0^{(T)}(\obsVar; \Delta_{\max}, h)$ and $V_1^{(T)}(\obsVar; \Delta_{\max}, h)$ are the sum of weighted votes for labels 0 and 1:
\begin{align*}
V_0^{(T)}(\obsVar; \Delta_{\max}, h)
&= \sum_{i=1}^n
     \exp\Big(
       -\frac{[\rho^{(T)}(\obsVar, X_i \maxtimeshift{0.85} \Delta_{\max})]^2}
                {2 h^2}
     \Big)
     \ind\{Y_i = 0\}, \\
V_1^{(T)}(\obsVar; \Delta_{\max}, h)
&= \sum_{i=1}^n
     \exp\Big(
       -\frac{[\rho^{(T)}(\obsVar, X_i \maxtimeshift{0.85} \Delta_{\max})]^2}
                {2 h^2}
     \Big)
     \ind\{Y_i = 1\}.
\end{align*}
When $h \rightarrow 0$, we obtain 1-NN classification.  We remark that in practice, to account for either label class imbalance issues or to tune true and false positive classification rates, we may bias the weighted votes by, for instance, only declaring the label to be~1 if $V_1^{(T)}(\obsVar; \Delta_{\max}, h) \ge \tau V_0^{(T)}(\obsVar; \Delta_{\max}, h)$ for some factor $\tau>0$ that need not be 1. \citet{georgehc_thesis} refers to this as \textit{generalized weighted majority voting}. The theory to follow for kernel classification actually easily extends to generalized weighted majority voting (and is done so by \citet{georgehc_thesis}), but for ease of exposition we stick to the ``non-generalized'' case where $\tau=1$.

For both the 1-NN and kernel classifiers, using a larger time window size~$T$ corresponds to waiting longer before we make a prediction. We trade off how long we wait and how accurate we want our prediction.  Note that the theory to follow does critically make use of the kernel being Gaussian and the distance (modulo time series alignment issues) being Euclidean.

\subsection{A Probabilistic Model of Time Series Classification}
\label{sec:theory}

We assume there to be $\numClusters$ distinct canonical time series $\mu_1, \dots, \mu_{\numClusters}$ with corresponding labels $\lambda_1, \dots, \lambda_{\numClusters} \in \{0, 1\}$ that are not all the same. These labeled time series occur with strictly positive probabilities $\pi_1, \dots, \pi_{\numClusters}$ and are the cluster centers.  For prediction, we will not know what these cluster centers are, how many of them there are, or what probabilities they occur with. We model a new time series to be generated as follows:

\begin{enumerate}

\item Sample cluster index $G \in \{1, \dots, \numClusters\}$, where $\pi_g = \mathbb{P}(G=g)$.

\item Sample integer time shift $\Delta$ uniformly from $\{ 0,1,\dots,\Delta_{\max} \}$.\footnote{For a technical reason, we restrict to nonnegative shifts here, whereas in the distance function $\rho^{(T)}(\cdot, \cdot \maxtimeshift{0.85} \Delta_{\max})$ from before, we allow for negative shifts as well.  \citet[Section~3.7.1]{georgehc_thesis} discusses how the generative model could indeed allow for shifts from $\{-\Delta_{\max}, \dots, \Delta_{\max}\}$, in which case for the proof techniques to work, the distance function should instead look at shifts up to magnitude $2\Delta_{\max}$.}

\item Output time series $\obsVar$ to be cluster center~$\mu_G$ advanced by $\Delta$ time steps, followed by adding noise time series $W$, \ie, $\obsVar(t) = \mu_G(t + \Delta) + W(t)$.  Entries of noise $W$ are i.i.d.~zero-mean sub-Gaussian with parameter $\sigma$.  The true label for $\obsVar$ is assigned to be $\labelVar = \lambda_G$.

\end{enumerate}
In their original paper, \citet{georgehc_time_series_nips} refer to the cluster centers as \textit{latent sources} and the above probabilistic model as the \textit{latent source model for time series classification}.

\subsection{Approximating an Oracle Classifier}
\label{sec:time-series-oracle}

If we knew the cluster centers and if noise entries $W(t)$ were i.i.d.~$\mathcal{N}(0,h^2)$ across time indices $t$, then the MAP estimate for label~$\labelVar$ given the first $T$ time steps of time series $\obsVar$ is
\begin{equation}
\widehat{\labelVar}_{\text{MAP}}^{(T)}(\obsVar; \Delta_{\max}, h)
=\begin{cases}
   1 & \text{if }\ratioVar_{\text{MAP}}^{(T)}(\obsVar; \Delta_{\max}, h) \ge 1,\\
   0 & \text{otherwise},
 \end{cases}
\label{eq:decision-rule}
\end{equation}
where
\begin{align}
&\ratioVar_{\text{MAP}}^{(T)}(\obsVar; \Delta_{\max}, h) \nonumber \\
&\quad\triangleq
  \frac{\sum_{g=1}^\numClusters
        \pi_g
        \sum_{\Delta \in \mathcal{D}_+}
        \exp(- \frac{[\|\obsVar - \mu_g \timeadvance{0.7} \Delta\|^{(T)}]^2}{2h^2} ) \ind\{\lambda_g=1\}}
       {\sum_{g=1}^\numClusters
        \pi_g
        \sum_{\Delta \in \mathcal{D}_+}
        \exp( - \frac{[\|\obsVar - \mu_g \timeadvance{0.7} \Delta\|^{(T)}]^2}{2h^2} ) \ind\{\lambda_g=0\}},
\label{eq:likelihood-ratio-MAP}
\end{align}
and $\mathcal{D}_+ \triangleq \{0,\dots, \Delta_{\text{max}}\}$. Note that in the ratio above, the numerator is a sum of weighted votes for label 1, and the denominator is a sum of weighted votes for label 0.  For this problem setup, the MAP classifier is the same as the Bayes classifier, which we established optimality for in Proposition~\ref{prop:bayes-classifier-minimizes-prob-error}, in terms of minimizing probability of misclassification.

In practice, we do not know the cluster centers $\mu_1,\dots,\mu_{\numClusters}$.  We assume that we have access to $n$ training data sampled i.i.d.~from the latent source model for time series classification, where we have access to all time steps of each training time series, as well as every training time series' label.  Denote $\mathcal{D} \triangleq \{-\Delta_{\text{max}}, \dots, 0,\dots, \Delta_{\text{max}}\}$.  Then we approximate the MAP classifier by using training data as proxies for the cluster centers. Specifically, we take ratio~\eqref{eq:likelihood-ratio-MAP}, replace the inner sum by a minimum in the exponent, replace each cluster center time series with training time series, drop the proportions $\pi_1,\dots,\pi_{\numClusters}$ that we do not know, and replace $\mathcal{D}_+$ by $\mathcal{D}$ to obtain the ratio
\begin{align}
\label{eq:est}
\ratioVar^{(T)}(\obsVar; \Delta_{\max}, h)
&\triangleq
  \frac{\sum_{i=1}^n
          \exp\big(-\frac{
                      [\min_{\Delta\in \mathcal{D}}
                        \|\obsVar - \obsVar_i \timeadvance{0.7} \Delta\|^{(T)}]^2
                    }{2h^2}
              \big)
          \ind\{\labelVar_i = 1\}}
       {\sum_{i=1}^n
          \exp\big(-\frac{
                      [\min_{\Delta\in \mathcal{D}}
                        \|\obsVar - \obsVar_i \timeadvance{0.7} \Delta\|^{(T)}]^2
                    }{2h^2}
              \big)
          \ind\{\labelVar_i = 0\}} \nonumber \\
&= \frac{V_1^{(T)}(\obsVar; \Delta_{\max}, h)}{V_0^{(T)}(\obsVar; \Delta_{\max}, h)}.
\end{align}
Plugging $\ratioVar^{(T)}$ in place of $\ratioVar_{\text{MAP}}^{(T)}$ in classification rule~\eqref{eq:decision-rule} yields kernel time series classification~\eqref{eq:decision-rule-with-min}, which as we have already discussed becomes 1-NN classification when $h \rightarrow 0$.  That 1-NN and kernel classification approximate the oracle MAP classifier suggest that they should perform better when this approximation improves, which should happen with more training data and not too much noise as to muddle where the true decision boundaries are between labels 0 and 1. Also note that exponential decay in squared Euclidean distances naturally comes out of using Gaussian noise. Euclidean distance will thus appear in how we define the separation between time series of opposite labels.

As a technical remark, if we didn't replace the summations over time shifts with minimums in the exponent, then we have a kernel density estimate in the numerator and in the denominator \citep[Chapter~7]{fukunaga_1990}, where the kernel is Gaussian, and the theoretical guarantee for kernel classification to follow would still hold using the same proof. We use a minimum rather a summation over time shifts to make the method more similar to existing time series classification work (\eg, \citealt{xi_2006}), which minimize over nonlinear time warpings rather than simple shifts.

\subsection{Nonasymptotic Theoretical Performance Guarantees}
\label{sec:time-series-guarantees}

We now present nonasymptotic performance guarantees for 1-NN and kernel time series classifiers, accounting for the number of training data~$n$ and the number of time steps $T$ that we observe of the time series to be classified. This result depends on the following separation:
\begin{equation*}
\label{eq:gap}
\sepVar^{(T)}(\{\obsVar_i,\labelVar_i\}_{i=1}^n;\Delta_{\max})
\triangleq
  \min_{\substack{i, j \in \{1, \dots, n\}
                  \text{ s.t.~}\labelVar_i \ne \labelVar_j, \\
                  \Delta, \Delta' \in \{-\Delta_{\max},\dots,\Delta_{\max}\}}}
    \|\obsVar_i \timeadvance{0.85} \Delta
      - \obsVar_j \timeadvance{0.85} \Delta'\|^{(T)},
\end{equation*}
which we abbreviate by just writing $\sepVar^{(T)}$.  This quantity measures how far apart the two different label classes $0$ and $1$ are if we only look at length-$T$ chunks of each time series and allow all shifts of at most $\Delta_{\max}$ time steps in either direction.  There are different ways to change the separation, such as increasing how many time steps $T$ we get to observe of time series $\obsVar$, and changing what quantity the time series are tracking.  The main result is as follows.

\begin{ftheorem}[\citealt{georgehc_time_series_nips}, Theorems~1 and~2 slightly rephrased]
\label{thm:time-series-main-result}
Let $\pi_{\min} \triangleq \min\{\pi_1, \dots, \pi_{\numClusters}\}$, $\pi_1 \triangleq \mathbb{P}(\labelVar = 1) = \sum_{g=1}^{\numClusters} \pi_g \ind\{ \lambda_g = 1 \}$, and $\pi_0 \triangleq \mathbb{P}(\labelVar = 0) = \sum_{g=1}^{\numClusters} \pi_g {\ind\{ \lambda_g = 0 \}}$.  Under the latent source model for time series classification with $n$ training data points:
\begin{itemize}

\item[(a)] The probability that the 1-NN time series classifier~\eqref{eq:decision-rule-nn} misclassifies time series $\obsVar$ with label $\labelVar$ satisfies the bound
\begin{align}
&\mathbb{P}(\widehat{\labelVar}_{1\text{-NN}}^{(T)}(\obsVar;\Delta_{\max})\ne \labelVar) \nonumber \\
&\le
  \numClusters \exp\Big( -\frac{n\pi_{\min}}{8} \Big)
  +
  (2\Delta_{\max}+1)
  n
  \exp\Big(-\frac{(\sepVar^{(T)})^2}{16\sigma^2}
      \Big).
\label{eq:nn-main-bound}
\end{align}

\item[(b)] The probability that the kernel time series classifier~\eqref{eq:decision-rule-with-min} with bandwidth $h > \sqrt{2}\sigma$ misclassifies time series $\obsVar$ with label $\labelVar$ satisfies the bound
\begin{align}
&\mathbb{P}(\widehat{\labelVar}_\text{Gauss}^{(T)}(\obsVar; \Delta_{\max}, h)\ne \labelVar) \nonumber \\
&\le
   \numClusters \exp\Big( -\frac{n\pi_{\min}}{8} \Big) \nonumber \\
&\quad
   +
   (2\Delta_{\max}+1) n
   \exp\Big(
         -\frac{(h^2 - 2\sigma^2)(\sepVar^{(T)})^2}{2h^4} 
       \Big).
\label{eq:wmv-main-bound}
\end{align}

\end{itemize}
In particular, for any pre-specified probability tolerance $\delta\in(0,1)$, by choosing bandwidth $h=2\sigma$ for kernel classification, the two misclassification probability upper bounds~\eqref{eq:nn-main-bound} and~\eqref{eq:wmv-main-bound} match, and by choosing number of training data $n \ge \frac{8}{\pi_{\min}}\log(2\numClusters/\delta)$, and if the separation grows as $S^{(T)} \ge 4\sigma\sqrt{\log(2(2\Delta_{\max}+1)n/\delta)}$, then both of the upper bounds are at most $\delta$, \ie, 1-NN and kernel classification each correctly classify test data point~$X$ with probability at least $1-\delta$.
\end{ftheorem}
The proof is provided by \citet[Section~3.7.1]{georgehc_thesis}.

For the two misclassification probability upper bounds presented, the first term $\numClusters\exp(-\frac{n\pi_{\min}}8)$ can be thought of as the penalty for not having at least one training example time series from all $\numClusters$ clusters. The second term can be thought of as the penalty for the nearest training point found being from a cluster with the wrong label.

The linear dependence on $n$ in the second term for both upper bounds~\eqref{eq:nn-main-bound} and~\eqref{eq:wmv-main-bound} result from a worst-case analysis in which only one training time series comes from the same cluster as the test time series, and the other $n-1$ training time series have the wrong label. If we have some estimates or bounds on $\numClusters$, $\pi_{\min}$, $\Delta_{\max}$, and $\sigma^2$, then one way to prevent this linear scaling in $n$ is to randomly subsample the training data. Specifically, if we have access to a large enough pool of labeled time series, \ie, the pool has $\Omega(\frac{8}{\pi_{\min}} \log\frac{\numClusters}{\delta})$ time series, then we can subsample $n=\Theta(\frac{8}{\pi_{\min}} \log\frac{\numClusters}{\delta})$ of them to use as training data, in which case 1-NN classification \eqref{eq:decision-rule-nn} correctly classifies a new time series $\obsVar$ with probability at least $1 - \delta$ if the separation grows as
\begin{equation*}
\sepVar^{(T)}
= \Omega\Bigg(
          \sigma
          \sqrt{
          \log (2\Delta_{\max}+1)
		  +
          \log
          \bigg(
            \frac{1}{\delta \pi_{\min}}
            \log \frac{\numClusters}{\delta}
          \bigg)
          }
        \Bigg).
\end{equation*}
For example, consider when the clusters occur with equal probability, so $\pi_{\min} = 1/\numClusters$. Then so long as the separation grows as
\begin{align}
\sepVar^{(T)}
&= \Omega\Bigg(
           \sigma
           \sqrt{
           \log (2\Delta_{\max}+1)
           +
           \log
           \bigg(
             \frac{\numClusters}{\delta}
             \log \frac{\numClusters}{\delta}
           \bigg)
           }
         \Bigg) \nonumber \\
&= \Omega\Bigg(
           \sigma
           \sqrt{
           \log(2\Delta_{\max}+1)
           +
           \log
             \frac{\numClusters}{\delta}
           }
         \Bigg),
\label{eq:times-series-result-rather-confusing-explanation}
\end{align}
then 1-NN classification is correct with probability at least $1 - \delta$.  In particular, if the separation grows as $\Omega(\sigma \sqrt{T})$ (which is a reasonable growth rate since otherwise, the closest two training time series of opposite labels are within noise of each other\footnote{For example, consider an extreme case where there are only $\numClusters=2$ clusters, and the two cluster centers $\mu_1$ and $\mu_2$ associated with opposite labels are actually the same (which means that distinguishing between them is impossible!).  Furthermore, suppose there are no time shifts ($\Delta_{\max}=0$), the noise in the time series is i.i.d.~$\mathcal{N}(0,\sigma^2)$, and we have exactly two training points $X_1$ and $X_2$ that originate from the two different clusters. Then $X_1$ and $X_2$ can actually just be modeled as independent $T$-dimensional Gaussian random vectors each with mean $\mu_1=\mu_2$ and covariance $\sigma^2\mathbf{I}_{T\times T}$.  Then $(\sepVar^{(T)})^2 = \|X_1 - X_2\|^2$ has expectation $2 \sigma^2 T$, meaning that noise alone makes the squared separation roughly scale as $\sigma^2 T$.}), then observing the first $T=\Omega(\log (2\Delta_{\max}+1) + \log \frac{\numClusters}{\delta})$ time steps from the test time series~$X$ would ensure condition~\eqref{eq:times-series-result-rather-confusing-explanation} to hold. Thus, we classify $X$ correctly with probability at least $1-\delta$. This corresponds to the informal statement of Theorem \ref{thm:time-series-main-result-informal}.

Although the misclassification probability upper bounds for 1-NN and kernel time series classification are comparable, experimental results in Section \ref{sec:time-series-experiments} show that kernel classification outperforms 1-NN classification when $T$ is small, and then as $T$ grows large, the two methods exhibit similar performance in agreement with our theoretical analysis. For small $T$, it could still be fairly likely that the nearest neighbor found has the wrong label, dooming 1-NN classifier to failure. Kernel classification, on the other hand, can recover from this situation as there may be enough correctly labeled training time series close by that contribute to a higher overall vote for the correct class. This robustness of kernel classification makes it favorable in the online setting where we want to make a prediction as early as possible.

\subsection{Learning the Cluster Centers}
\label{sec:time-series-learn-latent-sources}

If we can estimate the cluster centers accurately, then we could plug these estimates in place of the true cluster centers in the MAP classifier and achieve classification performance close to optimal. If we restrict the noise to be Gaussian and assume $\Delta_{\max}=0$, then the latent source model for time series classification corresponds to a spherical Gaussian mixture model. To simplify discussion in this section, we assume clusters occur with equal probability $1/\numClusters$.  We could learn a spherical Gaussian mixture model using the modified EM algorithm by \citet{dasgupta_2007}. Their theoretical guarantee depends on the true separation between the closest two clusters, namely
\[
\sepVar^{(T)*}
\triangleq
  \min_{g,h \in \{1, \dots, \numClusters\}\text{ s.t.~}g\ne h} \|\mu_g - \mu_h\|^{(T)},
\]
which needs to satisfy $\sepVar^{(T)*} \gg \sigma T^{1/4}$. Then with number of training time series $n = \Omega( \max\{ 1, \frac{\sigma^2 T}{(\sepVar^{(T)*})^2} \} \numClusters\log\frac{\numClusters}{\delta} )$, separation $\sepVar^{(T)*} = \Omega( \sigma \sqrt{\log\frac{\numClusters}{\varepsilon}} )$, and number of initial time steps observed
\begin{equation*}
T
=
\Omega\bigg(
        \max\bigg\{ 1,
                    \frac{\sigma^{4}T^2}
                         {(\sepVar^{(T)*})^4}
            \bigg\}
        \log
        \bigg[\frac{\numClusters}{\delta}
              \max\bigg\{ 1,
                          \frac{\sigma^{4}T^2}
                               {(\sepVar^{(T)*})^4}
                  \bigg\}
        \bigg]
      \bigg),
\end{equation*}
their algorithm achieves, with probability at least $1 - \delta$, an additive $\varepsilon\sigma\sqrt{T}$ error (in Euclidean distance) close to optimal in estimating every cluster center. In contrast, the result by \citet{georgehc_time_series_nips} is in terms of separation $\sepVar^{(T)}$ that depends not on the true separation between two clusters but instead on the minimum observed separation in the training data between two time series of opposite labels. In fact, $\sepVar^{(T)}$ grows as $\Omega(\sigma \sqrt{T})$ even when $\sepVar^{(T)*}$ grows sublinear in~$\sqrt{T}$.

In particular, while Dasgupta and Schulman's result cannot handle the regime where $\mathcal{O}(\sigma\sqrt{\log\frac{\numClusters}{\delta}}) \le \sepVar^{(T)*} \le \sigma T^{1/4}$, the result by \citet{georgehc_time_series_nips} can, using $n=\Theta(\numClusters\log\frac{\numClusters}{\delta})$ training time series and observing the first $T=\Omega(\log\frac{\numClusters}{\delta})$ time steps to classify a time series correctly with probability at least $1 - \delta$. This follows from Theorem~\ref{thm:time-series-gaussian-result} below, which specializes Theorem \ref{thm:time-series-main-result} to the Gaussian setting with no time shifts and uses separation $\sepVar^{(T)*}$ instead of $\sepVar^{(T)}$. We also present an accompanying corollary (Corollary~\ref{cor:time-series-gaussian-result-interp}) to interpret the Theorem~\ref{thm:time-series-gaussian-result}.  Both Theorem~\ref{thm:time-series-gaussian-result} and Corollary~\ref{cor:time-series-gaussian-result-interp} still hold if separation $\sepVar^{(T)*}$ were instead replaced by the Euclidean distance between the closest two cluster centers with opposite labels:
\[
\sepVar^{(T)\dagger}
\triangleq
  \min_{g,h \in \{1, \dots, \numClusters\}\text{ s.t.~}
        \lambda_g \ne \lambda_h} \|\mu_g - \mu_h\|^{(T)}.
\]
Intuitively, if cluster centers of the same label are extremely close by (so that $\sepVar^{(T)*}$ is small) yet cluster centers of opposite labels are far away (so that $\sepVar^{(T)\dagger}$ is large), then we should expect the classification problem to be relatively easy compared to learning the cluster centers because the latter still needs to tease apart the different cluster centers that are extremely close by.

\begin{ftheorem}[\citealt{georgehc_thesis}, Theorem~3.5.1]
\label{thm:time-series-gaussian-result}
Let $s_1>0$ and $s_2>0$. Suppose that $\sepVar^{(T)*} \ge s_1$. Under the latent source model for time series with Gaussian noise, no time shifts, and $n$ training data points:
\begin{itemize}

\item[(a)] The probability that the 1-NN time series classifier~\eqref{eq:decision-rule-nn} misclassifies time series $\obsVar$ with label $\labelVar$ satisfies the bound
\begin{align*}
&\mathbb{P}(\widehat{\labelVar}_{1\text{-NN}}^{(T)}(\obsVar;\Delta_{\max})\ne \labelVar) \\
&\le
   \numClusters\exp\Big( -\frac{n\pi_{\min}}{8} \Big)
   + \frac{n^2}{4}\exp\Big( -\frac{s_1^2}{4\sigma^2} \Big)
   + \frac{n^2}{4}\exp(-s_2) \\
&\quad
   + n\exp\bigg(
        -\frac{
           (\sepVar^{(T)*})^2
           - 2s_1 \sepVar^{(T)*}
           + 2\sigma^2 T
           - 4\sigma^2\sqrt{T s_2}
		      }{16\sigma^2}
		 \bigg).
\end{align*}

\item[(b)] The probability that the kernel time series classifier~\eqref{eq:decision-rule-with-min} with bandwidth $h > \sqrt{2}\sigma$ misclassifies time series $\obsVar$ with label $\labelVar$ satisfies the bound
\begin{align*}
&\mathbb{P}(\widehat{\labelVar}_\text{Gauss}^{(T)}(\obsVar; \Delta_{\max}, h)\ne \labelVar) \nonumber \\
&\le
   \numClusters\exp\Big( -\frac{n\pi_{\min}}{8} \Big)
   + \frac{n^2}{4}\exp\Big( -\frac{s_1^2}{4\sigma^2} \Big)
   + \frac{n^2}{4}\exp(-s_2) \\
&\quad
   + n\exp\bigg(
		-\frac{(h^2 - 2\sigma^2)
		      }{2h^4} \\
&\quad\quad\quad\quad\;\,\;\times
			   \big(
			    (\sepVar^{(T)*})^2
			    - 2s_1 \sepVar^{(T)*}
			    + 2\sigma^2 T
			    - 4\sigma^2\sqrt{T s_2}
			   \big)
      \bigg).
\end{align*}

\end{itemize}
Both of these guarantees still hold with $\sepVar^{(T)*}$ replaced by $\sepVar^{(T)\dagger}$.
\end{ftheorem}
The proof is provided by \citet[Section~3.7.2]{georgehc_thesis}.

As with Theorem \ref{thm:time-series-main-result}, the two upper bounds are comparable and can be made to match by choosing bandwidth $h=2\sigma$ for kernel classification. To interpret this theorem, we choose specific values for $s_1$ and $s_2$ and consider the worst case tolerated by the theorem in which $\sepVar^{(T)*} = s_1$, arriving at the following corollary.

\begin{fcorollary}[\citealt{georgehc_thesis}, Corollary~3.5.1]
\label{cor:time-series-gaussian-result-interp}
Let $\delta \in (0,1)$.  Under the latent source model for time series classification with Gaussian noise, no time shifts, and $n \ge \frac{8}{\pi_{\min}}\log\frac{4\numClusters}{\delta}$ training data points,~if
\begin{align*}
\sepVar^{(T)*}
&\ge 2\sigma \sqrt{\log\frac{n^2}{\delta}}, \\
T
&\ge
   4\log\frac{n^2}{\delta}
   + 8\log\frac{4n}{\delta}
   + 2\sqrt{\Big(
              3\log\frac{n^2}{\delta}
              + 8\log\frac{4n}{\delta}
            \Big)\log\frac{n^2}{\delta}},
\end{align*}
then 1-NN and kernel classification (with bandwidth $h=2\sigma$) each classify a new time series correctly with probability at least $1-\delta$. This statement still holds if $\sepVar^{(T)*}$ is replaced by $\sepVar^{(T)\dagger}$.
\end{fcorollary}
The proof is provided by \citet[Section~3.7.3]{georgehc_thesis}.

When the clusters occur with equal probability (so $\pi_{\min}=1/\numClusters$), then with $n = \Theta(\sigma^2 \log\frac{\numClusters}{\delta})$ training data and so long as the separation grows as $\sepVar^{(T)*} = \Omega(\sigma \sqrt{\log\frac{\numClusters}{\delta}})$, after observing $T = \Omega(\log\frac{\numClusters}{\delta})$ time steps of $\obsVar$, the two classifiers correctly classify $\obsVar$ with probability at least $1 - \delta$.

\citet{vempala_wang} have a spectral method for learning Gaussian mixture models that can handle smaller $\sepVar^{(T)*}$ than Dasgupta and Schulman's approach but requires $n = \widetilde{\Omega}(T^3 \numClusters^2)$ training data, where we've hidden the dependence on $\sigma^2$ and other variables of interest for clarity of presentation. \citet{hsu_2013} have a moment-based estimator that doesn't use a separation condition but, under a different non-degeneracy condition, requires substantially more samples for this problem setup, \ie, $n = \Omega((\numClusters^{14} + T \numClusters^{11})/\varepsilon^2)$ to achieve an $\varepsilon$-approximation of the mixture components. These results need substantially more training data than what we've shown is sufficient for classification.

To fit a Gaussian mixture model to massive training datasets, in practice, using all the training data could be prohibitively expensive. In such scenarios, one could instead non-uniformly subsample $\mathcal{O}(T \numClusters^3/\varepsilon^2)$ time series from the training data using the procedure given by~\citet{feldman_2011} and then feed the resulting smaller dataset, referred to as a $(k,\varepsilon)$-{\em coreset}, to the EM algorithm for learning the cluster centers. This procedure still requires more training time series than needed for classification and lacks a guarantee that the estimated cluster centers will be close to the true ones.

We end this section by remarking that there is theory for learning considerably more complex time series models that use nonlinear time warps rather than just time shifts. For instance, some work in this direction includes that of \citet{kneip_1992,wang_1997,bigot_2010}. Extending theoretical guarantees for nearest neighbor and kernel time series classification to handle nonlinear time warps under these more elaborate time series models is an interesting direction for future research.  We suspect that even in these more complex time series models, the number of training data sufficient for successful time series classification is smaller than that of learning the time series cluster centers.

\subsection{Lower Bound on Misclassification Probability}
\label{sec:time-series-lower-bound}

To understand how good the misclassification probability upper bounds are for 1-NN and kernel classification in Theorem~\ref{thm:time-series-gaussian-result}, we present a lower bound on the misclassification probability for \textit{any} classifier under the Gaussian noise setting with no time shifts as in the previous section. This lower bound depends on the true separation between cluster centers of opposite labels, namely $\sepVar^{(T)\dagger}$.

\begin{ftheorem}[\citealt{georgehc_thesis}, Theorem~3.6.1]
\label{thm:time-series-lower-bound}
Under the latent source model for time series classification with Gaussian noise and no time shifts, the probability of misclassifying time series using any classifier satisfies the bound
\begin{align*}
&\mathbb{P}(\text{misclassify}) \\
&\ge
  \pi_0\pi_1\pi_{\min}^2
  \exp\Big(
    -\frac{(\sepVar^{(T)\dagger})^2}{2\sigma^2}
    - \frac{\Gamma((T+1)/2)\sqrt{2}}{\sigma\Gamma(T/2)}
      \sepVar^{(T)\dagger}
    - \frac{T}{2}
  \Big),
\end{align*}
where $\pi_0$ (probability of label 0 occurring), $\pi_1$ (probability of label~1 occurring), and $\pi_{\min}$ (minimum probability of a cluster occurring) are defined in the statement of Theorem~\ref{thm:time-series-main-result}, and $\Gamma$ is the Gamma function: $\Gamma(z)\triangleq\int_0^\infty x^{z-1}e^{-x} dx$ defined for $z > 0$.  Note that $ \frac{\Gamma((T+1)/2)}{\Gamma(T/2)} $ grows sublinear in $T$.
\end{ftheorem}
The proof is provided by \citet[Section~3.7.4]{georgehc_thesis}.

We can compare this result to the misclassification probability upper bound of 1-NN classification in Theorem \ref{thm:time-series-gaussian-result}. While this upper bound does not match the lower bound, its fourth and final term decays exponentially with separation $\sepVar^{(T)\dagger}$ as well as time horizon $T$, similar to the only term in the lower bound.  The other three terms in the upper bound could be made arbitrarily small but in doing so slows down how fast the fourth term decays.  We suspect the upper bound to be loose as the analysis used to derive it is worst-case. Even so, it's quite possible that 1-NN and kernel classification simply aren't optimal.

\subsection{Relation to Existing Margin Conditions for Classification}
\label{sec:clustering-and-margin-conditions}

We can relate separation $\sepVar^{(T)\dagger}$ in Sections \ref{sec:time-series-learn-latent-sources} and \ref{sec:time-series-lower-bound} to the margin condition of \citet{mammen_1999}, \citet{tsybakov_2004}, and \citet{audibert_2007} (note that separation $\sepVar^{(T)\dagger}$ between the true cluster centers can be related to the separation $\sepVar^{(T)}$ in the training data \citep[inequality~(3.18)]{georgehc_thesis}).  As discussed previously in Chapter~\ref{chap:classification}, classification should be challenging if for observed time series $\obsVar$, the conditional probability ${\mathbb{P}(\labelVar=1 \mid \obsVar)}$ is close to $1/2$. When this happens, it means that $\obsVar$ is close to the decision boundary and could plausibly be explained by both labels. Thus, if the probability that $\obsVar$ lands close to the decision boundary is sufficiently low, then a prediction algorithm that, either explicitly or implicitly, estimates the decision boundary well should achieve a low misclassification probability.

We reproduce the margin condition~\eqref{eq:mammen-tsybakov-first} below, slightly reworded:
\begin{equation}
\mathbb{P}\Big(\Big| \mathbb{P}(\labelVar=1 \mid \obsVar) - \frac{1}{2} \Big| \le s\Big)
\le C_{\text{margin}} s^\varphi,
\label{eq:mammen-tsybakov}
\end{equation}
for some finite $C_{\text{margin}} > 0$, $\varphi > 0$, and all $0 < s \le s^*$ for some $s^* \le 1/2$. Note that the randomness is over $\obsVar$.  With additional assumptions on the behavior of the decision boundary, \citet{tsybakov_2004} and \citet{audibert_2007} showed that nonparametric classifiers can have misclassification probabilities that exceed the optimal Bayes error rate by as low as $\mathcal{O}(n^{-1})$ or even lower under a far more restrictive assumption on how label $\labelVar$ relates to observation $\obsVar$.

To sketch how separation $\sepVar^{(T)\dagger}$ relates to the above margin condition, we consider the one-dimensional Gaussian case with no time shifts where we have two clusters: if $\obsVar$ has label $\labelVar=1$ then it is generated from $\mathcal{N}(\mu, \sigma^2)$ and if $\obsVar$ has label $\labelVar=0$ then it is generated from $\mathcal{N}(-\mu, \sigma^2)$ for constants $\mu > 0$, and $\sigma > 0$, and where ${\mathbb{P}(\labelVar=1)} = {\mathbb{P}(\labelVar=0)} = 1/2$.  For this example, an optimal MAP decision rule classifies~$\obsVar$ to have label 1 if $\obsVar \ge 0$, and to have label 0 otherwise. Thus, the decision boundary is at $\obsVar=0$. Meanwhile, the separation is given by $\sepVar^{(T)\dagger} = \mu - (-\mu) = 2\mu$. To relate to margin condition \eqref{eq:mammen-tsybakov}, note that for $s\in[0,1/2)$,
\[
\Big|\mathbb{P}(\labelVar=1\mid \obsVar)-\frac{1}{2}\Big|\le s
\;\Leftrightarrow
\;
\obsVar \in [-\ell, \ell],
\]
where
\[
\ell \triangleq
\frac{\sigma^{2}}{2\mu}\log\Big(\frac{1+2s}{1-2s}\Big).
\]
Note that $\log(\frac{1+2s}{1-2s})=0$ when $s=0$, and $\log(\frac{1+2s}{1-2s})\rightarrow\infty$ when~${s\rightarrow1/2}$.  Interval $[-\ell, \ell]$ corresponds to the decision boundary $\obsVar=0$ up to some closeness parameter~$s$. For this interval to be far away enough from the two cluster centers $-\mu$ and~$\mu$, henceforth, we assume that $s$ is small enough so that
\[
\ell < \mu.
\]
In other words, the interval doesn't contain the cluster centers $-\mu$ and~$\mu$.

Then
\begin{align*}
&\mathbb{P}\Big(
   \Big| \mathbb{P}(\labelVar=1\mid \obsVar)-\frac{1}{2} \Big|
   \le s
 \Big) \nonumber \\
&=\mathbb{P}(
    \obsVar \in [-\ell, \ell]
  ) \nonumber \\
&=\mathbb{P}(\labelVar=1)
  \mathbb{P}(
    \obsVar \in [-\ell, \ell]
    \mid
    \labelVar=1
  )
  +
  \mathbb{P}(\labelVar=0)
  \mathbb{P}(
    \obsVar \in [-\ell, \ell]
    \mid
    \labelVar=0
  ) \nonumber \\
&\overset{(i)}{=}\mathbb{P}(
    \obsVar \in [-\ell, \ell]
    \mid
    \labelVar=1
  ) \nonumber \\
\end{align*} 
\begin{align}
&=\mathbb{P}(
    \mathcal{N}(\mu, \sigma^2) \in [-\ell, \ell]
  ) \nonumber \\
&\overset{(ii)}{\le}
2\ell\cdot\mathcal{N}(\ell;\mu,\sigma^{2})\nonumber \\
&=\frac{\sigma}{\mu \sqrt{2\pi}}\log\Big(\frac{1+2s}{1-2s}\Big)\exp\bigg(-\frac{1}{\sigma^{2}}\Big(\mu-\frac{\sigma^{2}}{2\mu}\log\Big(\frac{1+2s}{1-2s}\Big)\Big)^{2}\bigg),\label{eq:mammen-tsybakov-noise-bound}
\end{align}
where step $(i)$ uses symmetry, and step $(ii)$ uses the fact that since $\ell<\mu$, the largest value of the density of $\mathcal{N}(\mu, \sigma^2)$ within interval $[-\ell, \ell]$ is $\mathcal{N}(\ell;\mu,\sigma^2)$.

To upper bound the last line in inequality \eqref{eq:mammen-tsybakov-noise-bound}, let us examine  function $f(s) = \log\big(\frac{1+2s}{1-2s}\big)$ in the neighborhood of $0$. Specifically, 
\begin{align}
f'(s) & = \frac{4}{1-4s^2}~\stackrel{\text{for}~s\in[0,\frac14]}{\leq} \frac{16}{3}.
\end{align}
Since $f(0) = 0$, it follows that for all $s \in [0, \frac14]$, 
\begin{align}
f(s) & \leq \frac{16 s}{3}
\end{align}
Let $s^* = \min\{\frac14, \frac{3\mu^2}{16 \sigma^2}\}$. Then, from inequality  \eqref{eq:mammen-tsybakov-noise-bound} it follows that for $s \in [0, s^*]$,
\begin{align}
\mathbb{P}\Big(
   \Big| \mathbb{P}(\labelVar=1\mid \obsVar)-\frac{1}{2} \Big|
   \le s
 \Big)
& \leq \frac{16\mu}{3\sigma\sqrt{2\pi}} \exp\bigg(-\frac{\mu^2}{4\sigma^{2}}\bigg) s.
\end{align}
That is, the margin condition \eqref{eq:mammen-tsybakov} is satisfied with constants $\varphi=1$, $C_{\text{margin}}=\frac{16\mu}{3\sigma\sqrt{2\pi}}\exp(-\frac{\mu^2}{4\sigma^2})$, and $s^*=\min\{\frac14, \frac{3\mu^2}{16\sigma^2}\}$.

We note that bound~\eqref{eq:mammen-tsybakov-noise-bound} decays exponentially with the separation $\sepVar^{(T)\dagger}=2\mu$. One way to intuit this result is that since the noise is sub-Gaussian, the probability that~$\obsVar$ deviates significantly from its generating cluster center decays exponentially as a function of how far~$\obsVar$ is from this cluster center. When the separation between cluster centers of opposite labels is large, then it means that to land close to the decision boundary, $\obsVar$ would have to be quite far from even the two closest cluster centers with opposite labels. This event's probability goes to 0 as the separation grows large.

\subsection{Experimental Results}
\label{sec:time-series-experiments}

\begin{figure}[!t]
\centering
\subfloat[][]{
\includegraphics[width=2.2in]{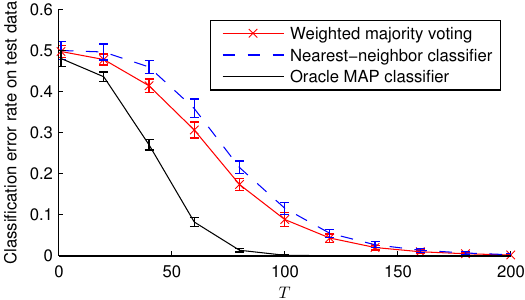}
\label{fig:error-vs-T}
}
\subfloat[][]{
\includegraphics[width=2.2in]{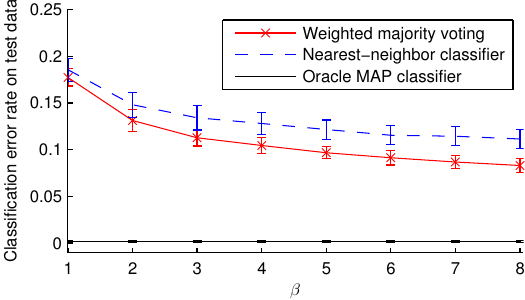}
\label{fig:error-vs-beta}
}
\caption[Time series classification results on synthetic data]{Plots for synthetic data reproduced from \citep{georgehc_time_series_nips}; note that they refer to kernel classification as weighted majority voting.  (a) Classification error rate vs.~number of initial time steps $T$ used; training set size: $n=\beta \numClusters\log \numClusters$ where $\beta=8$.  (b) Classification error rate at $T=100$ vs.~$\beta$.  All experiments were repeated 20 times with newly generated cluster centers, training data, and test data each time. Error bars denote one standard deviation above and below the mean value.}
\label{fig:synth-data}
\end{figure}

\textbf{Synthetic data.}
\citet{georgehc_time_series_nips} generate $\numClusters=200$ cluster centers that occur with equal probability, where each cluster center time series is constructed by first sampling i.i.d.~$\mathcal{N}(0,100)$ entries per time step and then applying a 1D Gaussian smoothing filter with scale parameter 30. Half of the clusters are labeled 1 and the other half 0. Then $n=\beta \numClusters\log \numClusters$ training time series are sampled, for various values of $\beta$, as per the latent source model where the noise added is i.i.d.~$\mathcal{N}(0,1)$, and the maximum time shift is $\Delta_{\max}=100$. Test data consisting of 1000 time series are similarly generated. Bandwidth $h=2$ is chosen for the Gaussian kernel.  For $\beta=8$, classification error rates on test data for 1-NN, kernel classification, and the MAP classifier with oracle access to the true cluster centers are shown in Figure~\ref{fig:synth-data}\subref{fig:error-vs-T}. We see that kernel classification outperforms 1-NN classification but as $T$ grows large, the two methods' error rates converge to that of the MAP classifier.  Fixing $T=100$, the classification error rates of the three methods using varying amounts of training data are shown in Figure~\ref{fig:synth-data}\subref{fig:error-vs-beta}; the oracle MAP classifier is also shown but does not actually depend on training data. We see that as $\beta$ increases, both 1-NN and kernel classification steadily improve in performance.

\begin{figure}[!t]
\centering
\subfloat[][]{
\includegraphics[width=3in]{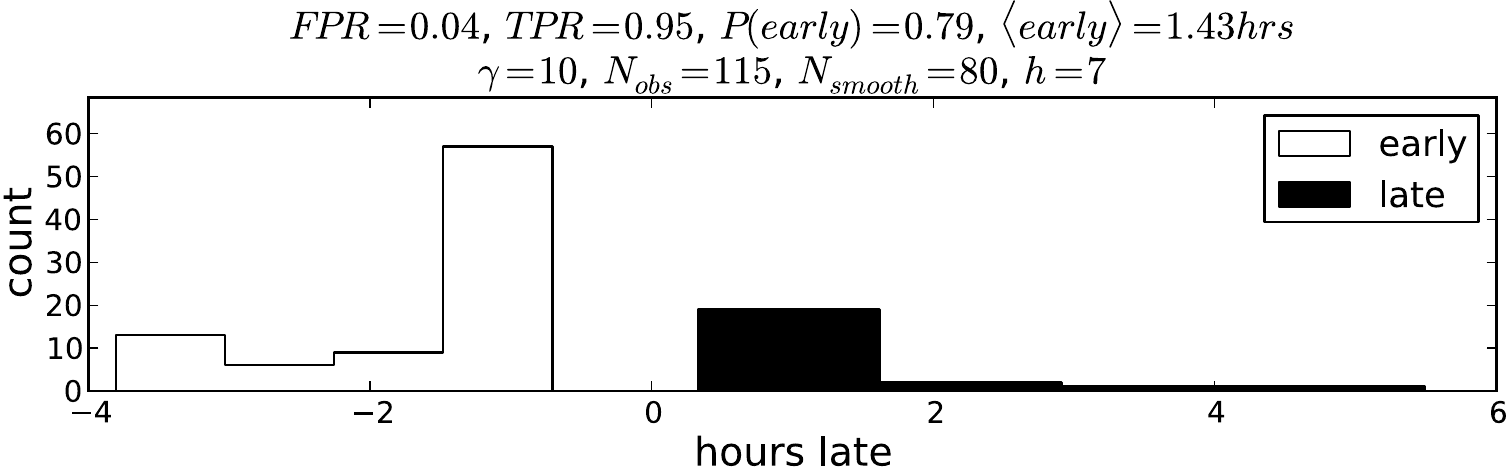}
\label{fig:early}
}~~
\subfloat[][]{
\includegraphics[height=1.2in]{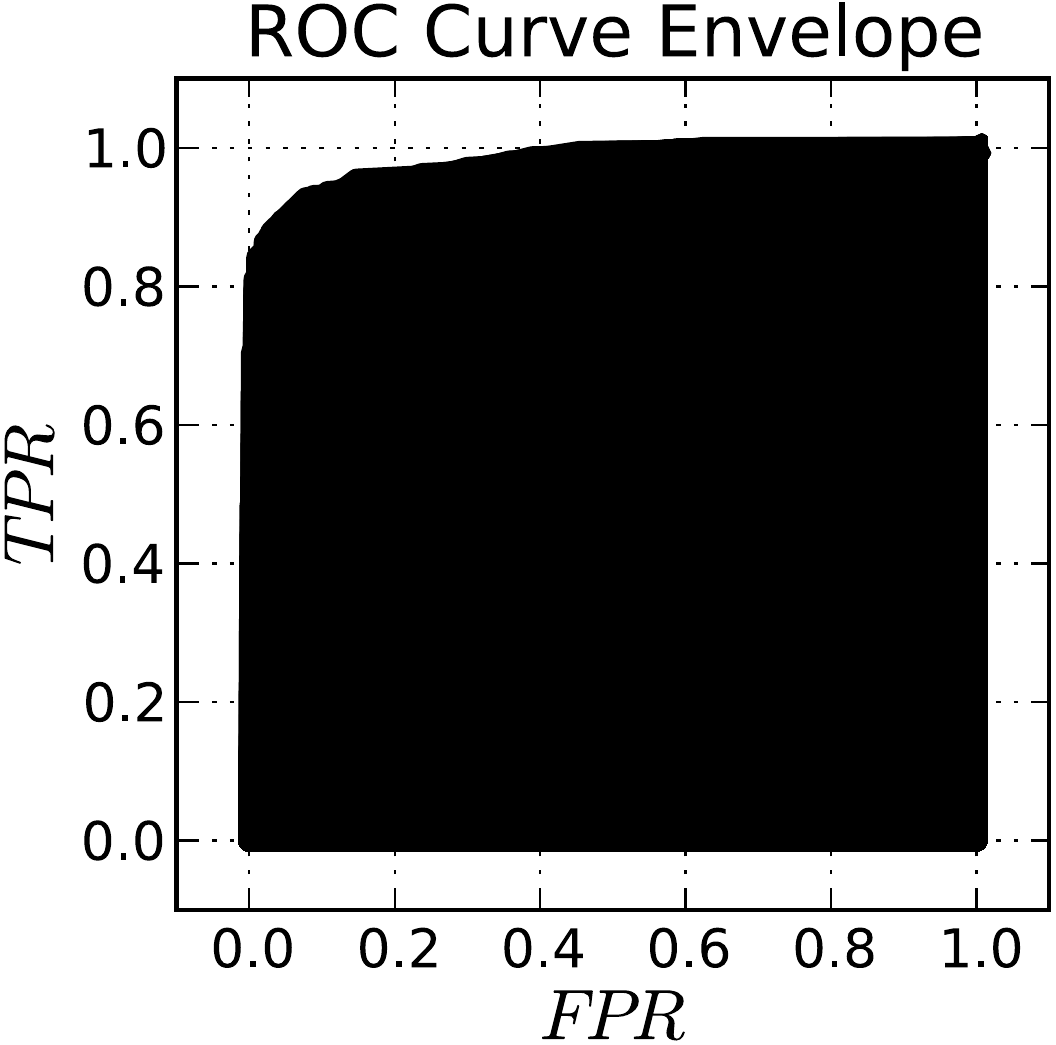}
\label{fig:roc}
}
\\
\subfloat[][]{
\hspace{-1.2em}
\includegraphics[width=2.43in, clip=true, trim=0 5.52in 0in 0in]{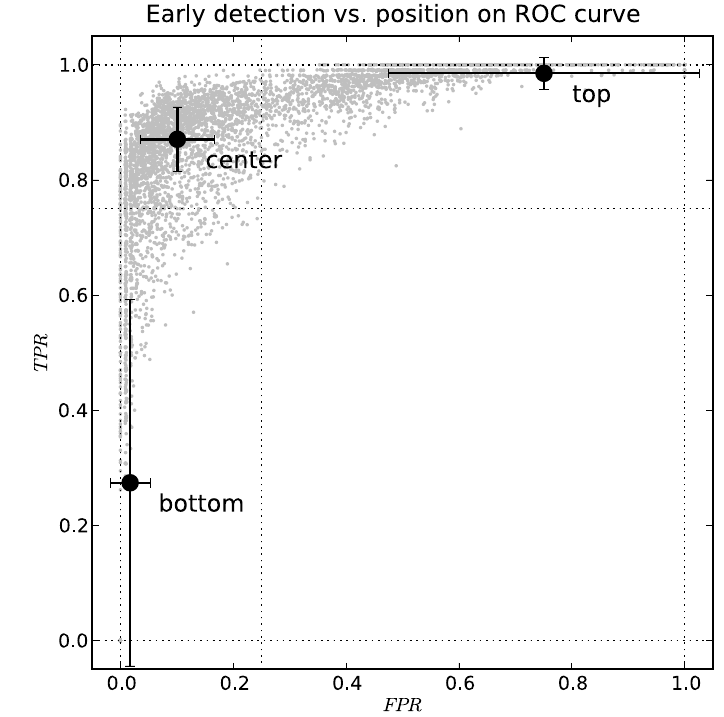}
\includegraphics[width=2.23in, clip=true, trim=0 0in 0in 5.14in]{time-series/roc_early_late-eps-converted-to}
\label{fig:roc-early-late}
}
\caption[Time series classification results by \citet{nikolov_2012} on Twitter data]{Results from \citet{nikolov_2012} on Twitter data.  (a) Kernel classification achieves a low error rate (FPR of~4\%, TPR of 95\%) and detects viral news topics in advance of Twitter 79\% of the time, with a mean of 1.43 hours when it does for a particular choice of parameters.  (b) Envelope of all ROC curves shows the tradeoff between TPR and FPR.  (c) Distribution of detection times for ``aggressive'' (top), ``conservative'' (bottom) and ``in-between'' (center) parameter settings.  \label{fig:twitter-main}}
\end{figure}

\vspace{.3em}
\noindent
\textbf{Forecasting viral news topics on twitter.}
We now summarize experimental results reported by \citet{nikolov_2012} and subsequently by \citet{georgehc_time_series_nips}.  From June 2012 user posts on Twitter, 500 examples of news topics that go viral (\ie, were ever within the top~3 on Twitter's trending topics list) were sampled at random from a list of news topics, and 500 examples of news topics that never go viral were sampled based on phrases appearing in posts.  As it's unknown to the public how Twitter chooses what phrases are considered as candidate phrases for trending topics, it's unclear what the size of the ``non-viral'' category is in comparison to the size of the trend category. Thus, for simplicity, the label class sizes were controlled to be equal, and kernel classification \eqref{eq:decision-rule-with-min} was used to classify time series, where $\Delta_{\max}$ is set to the maximum possible (all shifts are considered).

Per topic, the time series used for classification is based on a pre-processed version of the Tweet rate (how often the topic was shared by users on Twitter). The pre-processing pipeline is detailed in Appendix~E of \citep{georgehc_time_series_nips}.  Chen \textit{et al.}~empirically found that the time series for news topics that go viral tend to follow a finite number of patterns; a few examples of these patterns are shown in Figure \ref{fig:clusters}.  These few patterns could be thought of as the cluster centers. The viral and non-viral news topics were randomly divided into two halves, one to use as training data and one to use as test data.  \citet{nikolov_2012} applied kernel classification, sweeping over $h$, $T$, and data pre-processing parameters. As shown in Figure~\ref{fig:twitter-main}\subref{fig:early}, with one choice of parameters, kernel classification detected viral news topics in advance of Twitter 79\% of the time, and on average 1.43 hours earlier, with a true positive rate (TPR) of 95\% and a false positive rate (FPR) of 4\%. Naturally, there are tradeoffs between TPR, FPR, and how early one wants to make a prediction (\ie, how small time horizon $T$ is). As shown in Figure~\ref{fig:twitter-main}\subref{fig:roc-early-late}, an ``aggressive'' parameter setting yields early detection and high TPR but high FPR, and a ``conservative'' parameter setting yields low FPR but late detection and low TPR. An ``in-between'' setting can strike the right balance.

\subsection{Toward Better Guarantees: Meta Majority Voting}

Having to subsample the training data to keep the misclassification probability upper bounds from scaling with $n$ seems strange. Shouldn't more data only help us?  Or is it that by seeing more data, due to noise, as we get more and more samples, we're bound to get unlucky and encounter a training time series with the wrong label that is close to the time series we want to classify, causing 1-NN classification to get confused and buckle? In fact, later results in this chapter will also involve some training data subsampling, yet it's unclear how critical this is in practice for nearest neighbor methods.

When the number of training data $n$ is large, a more clever strategy that still involves subsampling but now uses all the training data is to randomly partition the training data into groups of size $\Theta(\frac{1}{\pi_{\min}} \log \frac{\numClusters}{\delta})$ each. Then we can apply kernel classification within each group. A final prediction can be made by a ``meta'' majority vote: choose the most popular label across the different groups' label predictions. This meta-voting strategy readily lends itself to analysis. In particular, since the training data in different groups are independent, kernel classification's predictions across the different groups are also independent and we know when we can ensure each of these predictions to be correct with probability at least $1-\delta$. Then among the $\Theta(\frac{n\pi_{\min}}{\log(\numClusters/\delta)})$ groups, the number of correct label predictions stochastically dominates a $\text{Binomial}\big(\Theta(\frac{n\pi_{\min}}{\log(\numClusters/\delta)}), 1-\delta\big)$ random variable. We can then apply a binomial concentration inequality to lower-bound this meta-voting strategy's probability of success.

\section{Online Collaborative Filtering}
\label{chap:online-collaborative-filtering}

Recommendation systems have become ubiquitous in our lives, helping us filter the vast expanse of information we encounter into small selections tailored to our personal tastes. Prominent examples include Amazon recommending items to buy, Netflix recommending movies, and LinkedIn recommending jobs.  In practice, recommendations are often made via collaborative filtering, which boils down to recommending an item to a user by considering items that other similar or ``nearby'' users liked. Collaborative filtering has been used extensively for decades now including in the \mbox{GroupLens} news recommendation system \citep{grouplens_1994}, Amazon's item recommendation system \citep{amazon_2003}, and the Netflix \$1 million grand prize winning algorithm by \mbox{BellKor's} Pragmatic Chaos \citep{bellkor_2009,bigchaos_2009,pragmatic_2009}.

Most such systems operate in the ``online'' setting, where items are constantly recommended to users over time. In many scenarios, it does not make sense to recommend an item that is already consumed. For example, once Alice watches a movie, there's little point to recommending the same movie to her again, at least not immediately, and one could argue that recommending unwatched movies and already watched movies could be handled as separate cases. Finally, what matters is whether a {\em likable} item is recommended to a user rather than an {\em unlikable} one. In short, a good online recommendation system should recommend different likable items continually over time.

Despite the success of collaborative filtering, there has been little theoretical development to justify its effectiveness in the online setting.  Instead, most work (\eg, \citealt{candes_recht_2009,cai_2010,candes2010matrix,keshavan_2010a,keshavan_2010b,recht_2011}) has been in the ``offline'' setting, where we freeze time, have access to all ``revealed'' ratings users have provided so far, and predict all ``missing'' ratings for items users haven't yet rated.  This offline problem setup perhaps gained enormous popularity among both academics and practitioners after Netflix offered a \$1 million dollar grand prize for a solution to the problem that outperformed theirs by a pre-specified performance metric. The setup itself is a matrix completion problem, where we consider a matrix of ratings where rows index users and columns index items (\eg, movies), and the $(u, i)$-th entry is either the rating that user $u$ gave item $i$, or marked as missing. The goal then is to fill in these missing ratings, typically by imposing a low-rank constraint on the ratings matrix.  The theoretical guarantees for such methods usually assume that the items that users view are uniform at random, which is not the case in reality and also doesn't account for the fact that real recommendation systems should and are biasing users into rating certain items, hopefully items that they like. Only recently has this assumption been lifted for theoretical analysis of the offline setting \citep{lee_2013}. Of course, none of these model the true online nature of recommendation systems with time marching forward and the system continuously providing recommendations and receiving user feedback.

Moving to this online setting, most work has been phrased in the context of the classical so-called \textit{multi-armed bandit} problem, first introduced by \citet{thompson_bandit_origin}. The name of the problem originates from the ``one-armed bandit'' slot machine found in casinos in which a gambler pulls the arm of the machine and receives a random reward. Naturally, the $m$-armed bandit problem refers to when the gambler has $m$ such machines to play and seeks to maximize her or his cumulative reward over time, playing one machine at each time step. Translated to the context of online recommendation systems, the $m$ machines are items, and playing a machine refers to recommending an item. We remark that in this standard setup of the multi-armed bandit problem, there is only a single user and hence no concept of collaboration between users, and only recently has there been work on incorporating a pool of users \citep{clustered_bandits,gentile_2014}.

Even so, justification for why collaborative filtering methods in particular work in this online setting was only recently established by \citet{georgehc_collaborative_filtering_nips}. In this section, we cover their main result, which relies on an underlying probabilistic model for an online recommendation system.  Note that real online recommendation systems are considerably complex. Users join and leave, items are added and removed, users' preferences change over time, certain items abruptly go viral, users consume items at drastically different rates, and so forth. To contain this unwieldy mess of possibilities, Bresler \textit{et al.}~introduce a toy model of an online recommendation system that, despite its simplicity, provides a clean baseline framework for theoretical analysis and already highlights the crucial role collaboration plays in recommendation. They refer to their model as the \textit{latent source model for online collaborative filtering}. We present this model and clearly state the recommendation objective in Section~\ref{sec:collaborative-filtering-model}. The main theoretical result is as follows.

\begin{ftheorem}
[Informal statement of Theorem \ref{thm:collaborative-filtering-main-result}]
\label{thm:collaborative-filtering-main-result-informal}
Under the latent source model for online collaborative filtering and a notion of separation between the $\numClusters$ user clusters, with number of users $n = \Theta(\numClusters m)$ where $m$ is the number of items, a variant of the so-called cosine-similarity collaborative filtering has an essentially optimal expected fraction of likable items recommended after an initial number of time steps scaling as nearly $\log(\numClusters m)$.
\end{ftheorem}
The collaborative filtering method achieving the above result, discussed in Section~\ref{sec:collaborative-filtering-main-result}, has a crucial algorithmic insight of using two types of exploration, one that explores the space of items to find likable items (ask each user to consume and rate a random unconsumed item), and one that explores the space of users to find similar users (ask users to consume a common item). When not exploring, the algorithm exploits by greedily choosing items to recommend based on fixed-radius NN regression estimates, where a cosine distance is used (hence the name \textit{cosine-similarity} collaborative filtering).  Experimental results by~\citet{georgehc_thesis}, which extend the results by \citet{georgehc_collaborative_filtering_nips}, are presented in Section~\ref{sec:collaborative-filtering-experiments}.

\subsection{A Model and a Learning Problem}
\label{sec:collaborative-filtering-model}

\noindent\textbf{Online recommendation system setup.}

We consider a recommendation system with $n$ users and $m$ items, both fixed constants, so users are neither joining nor leaving, and items are neither being added nor removed. We also assume user preferences to be static over time.  As for the actual dynamical process, at each time step $t=1,2,\dots$, exactly one item is recommended to each user, who immediately consumes and rates the item as either $+1$ (like) or $-1$ (dislike).\footnote{In practice, a user could ignore the recommendation. To keep our exposition simple, however, we stick to this setting that resembles song recommendation systems like Pandora that per user continually recommends a single item at a time. For example, if a user rates a song as ``thumbs down'' then we assign a rating of $-1$ (dislike), and any other action corresponds to $+1$ (like).} We reserve a rating of 0 to mean that the user has not rated an item yet. Once a user has consumed an item (\eg, watches a movie), the system is not allowed to recommend the item to that same user again.  Initially, none of the users have consumed any items, and so up to time $T$, each user will have consumed and rated exactly $T$ items. No structure is assumed over items in that the rating for each item gives no information about the rating for any other item a priori. So, for example, if Alice likes \textit{Star Wars: A New Hope}, we can't just recommend \textit{The Empire Strikes Back} as we don't know that these items are related. In Section~\ref{sec:collaborative-filtering-discussion}, we describe how knowledge of similar items helps us.

Note that the setup described thus far does not have a probability model associated with it yet, but it already draws out the importance of collaboration.  Consider when there's only a single user. Then to learn anything about an item, we have to ask the user to rate the item, but upon doing so, we can't recommend that item again!  Moreover, because the rating for an item is assumed to not provide us any information about other items (so we don't know which items are similar), we can't hope to find what the good items are except through exhaustive trial and error. However, with a pool of users, and with structure over users, we should be able to make good recommendations.

Probabilities come into play through user ratings being modeled as noisy.  Specifically, the latent item preferences for user $u$ are represented by a length-$m$ vector $p_u \in [0,1]^m$, where user $u$ likes item $i$ with probability $p_{ui}$, independently across items. For a user $u$, we say that item~$i$ is {\em likable} if $p_{ui} > 1/2$ and {\em unlikable} otherwise.

\vspace{.3em}
\noindent\textbf{Recommendation objective.}
The goal of the recommendation system is to maximize the expected number of likable items recommended:
\begin{equation}
\label{eq:reward+}
\mathcal{R}_+^{(T)}
\triangleq
  \sum_{t=1}^T\sum_{u=1}^n \E[\likableForUserTimeVar_{ut}]\,,
\end{equation}
where $\likableForUserTimeVar_{ut}$ is the indicator random variable for whether the item recommended to user $u$ at time $t$ is likable.  Recommending likable items for a user in an arbitrary order is sufficient for many real recommendation systems such as for movies and music. For example, we suspect that users wouldn't actually prefer to listen to music starting from the songs that their user cluster would like with highest probability to the ones their user cluster would like with lowest probability; instead, each user would listen to songs that she or he finds likable, ordered such that there is sufficient diversity in the playlist to keep the user experience interesting.

\vspace{.3em}
\noindent\textbf{Clustering structure.}
If the users have no similarity with each other, collaboration would not work very well! To this end, we assume a simple structure for shared user preferences.  We posit that there are $\numClusters \ll n$ different clusters of users, where users of the same cluster have identical item preference vectors. These are the $\numClusters$ clusters.  We denote the $\numClusters$ underlying item preference vectors as $\mu_1, \dots, \mu_{\numClusters} \in [0,1]^m$.  If users $u$ and $v$ are from the same cluster $g \in \{1, \dots, \numClusters\}$, then $p_u = p_v = \mu_g$. The number of user clusters~$\numClusters$ represents the heterogeneity in the population. For ease of exposition, in this section we assume that a user belongs to each user cluster with probability $1/\numClusters$. This sort of assumption on the ratings is a special case of low rank assumptions often used in matrix completion (\eg, \citealt{keshavan_2010a,keshavan_2010b}), where the rank corresponds to the number of users clusters. To see this, note that we are assuming each user's ratings vector to be a noisy version of one of $\numClusters$ possible cluster rating vectors. Low rank matrix completion is a bit more general since it can, for instance, encode item clusters as well.

We quantify how much noise there is within the user ratings as follows. We suppose there exists a constant $\sigma \in [0, 1/2)$ such that
\begin{equation}
\min\{ 1 - \mu_{gi}, \mu_{gi} \}
\le \sigma
\label{eq:cf-noise-level-def}
\end{equation}
for all user clusters $g \in \{1, \dots, \numClusters\}$ and items $i \in \{1, \dots, m\}$.  In particular, noise constant $\sigma$ measures how far the probabilities of liking items are from 0 or 1. If $\sigma=0$, then the probabilities are all 0 or 1, so user ratings are deterministic and there is no noise. If $\sigma=1/2$ (which we expressly disallow), then there is an item with probability $1/2$ of being liked, and we can't hope to predict whether a user will like this item better than chance.

\citet{georgehc_collaborative_filtering_nips} refer to the overall model for the online recommendation system, randomness in ratings, and clustering structure as the \emph{latent source model for online collaborative filtering}, where each user cluster corresponds to a latent source of users.

\vspace{.3em}
\noindent\textbf{Relation to bandit problems.}
This problem setup relates to some versions of the multi-armed bandit problem.  A fundamental difference between this setup and that of the standard stochastic multi-armed bandit problem \citep{thompson_bandit_origin,regret_analysis_book} is that the latter allows each item to be recommended an infinite number of times. Thus, the solution concept for the stochastic multi-armed bandit problem is to determine the best item (arm) and keep choosing it \citep{finite_time_stochastic_multiarmed_bandits}. This observation applies also to ``clustered bandits'' \citep{clustered_bandits}, which, however, does incorporate collaboration between users. On the other hand, ``sleeping bandits'' \citep{sleeping_bandits} allow for the available items at each time step to vary, but the analysis is worst-case in terms of which items are available over time. In the latent source model for online collaborative filtering, the sequence of items that are available is not adversarial. Thus, this model combines the collaborative aspect of clustered bandits with dynamic item availability of sleeping bandits, where there is a strict structure in how items become unavailable.

\vspace{.3em}
\noindent\textbf{Relation to learning mixture models.}
Similar to time series forecasting, we can relate the online collaborative filtering setup to that of learning mixture models (\cf, \citealt{chaudhuri_2008,belkin_2010,moitra_2010,anandkumar_2012}), where one observes samples from a mixture distribution and the goal is to learn the mixture components and weights. Existing results assume that one has access to the entire high-dimensional sample or that the samples are produced in an exogenous manner (not chosen by the algorithm). Neither assumption holds in online collaborative filtering, as we only see each user's revealed ratings thus far and not the user's entire preference vector, and the recommendation algorithm affects which samples are observed (by choosing which item ratings are revealed for each user). These two aspects make online collaborative filtering more challenging than the standard setting for learning mixture models. However, as with time series forecasting, the goal considered here is more modest than learning cluster components. Specifically, rather than learning the $\numClusters$ item preference vectors $\mu_1,\dots,\mu_{\numClusters}$, we merely classify items as likable or unlikable by each cluster.

\subsection{Collaborative Filtering with Two Exploration Types}
\label{sec:collaborative-filtering-two-explorations}

For clarity of presentation, we begin by describing a simpler recommendation algorithm \textsc{Simple-Collaborative-Greedy}. To make \textsc{Simple-Collaborative-Greedy} more amenable to analysis, we modify it slightly to obtain \textsc{Collaborative-Greedy}. Both algorithms are syntactically similar to an algorithm called $\varepsilon$\textsc{-Greedy} for the standard multi-armed bandit setting, which explores items with probability $\varepsilon$ and otherwise greedily chooses the best item seen so far based on a score function \citep{reinforcement_learning_textbook}.  The exploration probability~$\varepsilon$ is allowed to decay with time: as we learn more about the different bandit machines, or items in this online recommendation setting, we should be able to explore less and exploit more.

\vspace{.3em}
\noindent\textbf{Exploitation.}
During an exploitation step, \textsc{Simple-Collaborative-Greedy} chooses the best item for a user via cosine-similarity collaborative filtering. The basic idea of this approach is that at time~$t$, for each user~$u$, we first compute a score $\widehat{p}_{ui}^{(t)}$ for every item~$i$.  Then we recommend to user~$u$ whichever item has the highest score that user~$u$ has not already consumed, breaking ties randomly. As we explain next, the score $\widehat{p}_{ui}^{(t)}$ is defined using fixed-radius NN regression with cosine distance (which is just 1 minus cosine similarity).

The cosine distance $\rho_{\cos}$ between users is computed as follows.  Let $\labelVar_{u}^{(t)} \in \{+1, 0, -1\}^m$ be user $u$'s ratings up to (and including) time $t$, so $\labelVar_{ui}^{(t)}$ is the rating user $u$ has given item $i$ up to time $t$ (recall that~0 indicates that the user has not actually consumed and rated the item yet).  Then for two users $u$ and~$v$, let $\Psi_{uv}^{(t)} \triangleq \text{supp}(\labelVar_u^{(t)}) \cap \text{supp}(\labelVar_v^{(t)})$ be the support overlap of $\labelVar_u^{(t)}$ and $\labelVar_v^{(t)}$ (\ie, the items that both users have rated up to time $t$), and let $\langle \cdot, \cdot \rangle_{\Psi_{uv}^{(t)}}$ be the dot product restricted to entries in $\Psi_{uv}^{(t)}$. Then the cosine distance between users $u$ and $v$ at time $t$ is
\begin{align}
\rho_{\cos}(\labelVar_u^{(t)}, \labelVar_v^{(t)})
&=
1-
\frac{ \langle \labelVar_u^{(t)}, \labelVar_v^{(t)} \rangle_{\Psi_{uv}^{(t)}} }
     { |\Psi_{uv}^{(t)}| }\,.
\label{eq:cf-cosine-distance-def}
\end{align}
Assuming that the two users have rated at least one item in common, then the cosine distance ranges from 0 (the two users agree on all the items that they both rated) and 2 (the two users disagree on all the items that they both rated).

For user $u$, item $i$'s score $\widehat{p}_{ui}^{(t)}$ is defined to be the fraction of user $u$'s neighbors (up to a threshold distance $h$) that like item $i$ (among those who have consumed and rated item $i$):
\begin{equation}
\widehat{p}_{ui}^{(t)}
\triangleq
  \begin{cases}
    \frac{\textstyle
          \sum_{v=1}^n \ind\{\rho_{\cos}(\labelVar_u^{(t)}, \labelVar_v^{(t)}) \le h\}
            \ind\{\labelVar_{vi}^{(t)} = +1\}}
         {\textstyle
          \sum_{v=1}^n \ind\{\rho_{\cos}(\labelVar_u^{(t)}, \labelVar_v^{(t)}) \le h\}
            \ind\{\labelVar_{vi}^{(t)} \ne 0\}}
    & \text{if denom.}>0 \\
    1/2 & \text{otherwise}.
  \end{cases}
\label{eq:simple-collaborative-greedy-score}
\end{equation}
This score is a fixed-radius NN regression estimate of the true probability $p_{ui}$ of user~$u$ liking item~$i$.  If the estimate exceeds 1/2, then we can classify item~$i$ as likable by user $u$. This mirrors how regression function estimates can be turned into classifiers back in Section~\ref{sec:basic-classifiers}.  By recommending to user $u$ whichever item has the highest score $\widehat{p}_{ui}^{(t)}$, what we are doing is sorting these fixed-radius NN regression estimates and using whichever one hopefully exceeds 1/2 the most, which should be a likable item for user~$u$.

Similar to how in the time series forecasting setup, 1-NN and kernel classifiers approximate an oracle MAP classifier, in online collaborative filtering, cosine-similarity collaborative filtering approximates an oracle MAP recommender \citep[Section~4.2]{georgehc_thesis}. Once again, the approximation is bad when we do not have enough training data, which here means that we do not have enough revealed ratings, \ie, not enough time has elapsed. Thus, early on when we know very little about the users, we want to explore rather than exploit.

\vspace{.3em}
\noindent\textbf{Exploration.}
The standard multi-armed bandit setting does not have user collaboration, and one could interpret asking each user to randomly explore an item as probing the space of items. To explicitly encourage user collaboration, we ask users to all explore the same item, which probes the space of users.  Accounting for the constraint in our setting that an item can't be recommended to the same user more than once, we use the two following exploration types:
\begin{itemize}

\item \textit{Random exploration.}
For every user, recommend an item that she or he hasn't consumed yet uniformly at random.

\item \textit{Joint exploration.}
Ask every user to provide a rating for the next unseen item in a shared, randomly chosen sequence of the $m$ items.

\end{itemize}

\noindent\textbf{Two recommendation algorithms.}
The first algorithm \textsc{Simple-Collaborative-Greedy} does one of three actions at each time step~$t$: With probability $\varepsilon_{\text{R}}$, we do the above random exploration step. With probability $\varepsilon_{\text{J}}$, we do the above joint exploration step. Finally, if we do neither of these exploration steps, then we do a greedy exploitation step for every user: recommend whichever item $i$ user $u$ has not consumed yet that maximizes the score $\widehat{p}_{ui}^{(t)}$ given by equation \eqref{eq:simple-collaborative-greedy-score}, which relied on cosine distance to find nearby users.

We choose the exploration probabilities $\varepsilon_{\text{R}}$ and $\varepsilon_{\text{J}}$ as follows. For a pre-specified rate $\alpha \in (0,4/7]$, we set the probability of random exploration to be $\varepsilon_{\text{R}}(n)=1/n^\alpha$ (decaying with the number of users), and the probability of joint exploration to be $\varepsilon_{\text{J}}(t)=1/t^\alpha$ (decaying with time). For ease of presentation, the two explorations are set to have the same decay rate $\alpha$, but the proof easily extends to encompass different decay rates for the two exploration types. Furthermore, the constant $4/7 \ge \alpha$ is not special. It could be different and only affects another constant in the proof. The resulting algorithm is given in Algorithm \ref{alg:collaborative-greedy}.

\begin{algorithm}[!ht]
\DontPrintSemicolon
\caption{\textsc{Simple-Collaborative-Greedy}
         (and~\textsc{Collaborative-Greedy})
         \label{alg:collaborative-greedy}}
\KwIn{Parameters $h \in [0,1]$, $\alpha \in (0,4/7]$.}
Select a random ordering $\permutationVar$ of the items $\{1, \dots, m\}$. Define
\[
\varepsilon_{\text{R}}(n) = \frac{1}{n^\alpha},
\qquad
\text{and}
\qquad
\varepsilon_{\text{J}}(t) = \frac{1}{t^\alpha}.
\]
\For{\textnormal{time step} $t=1, 2, \dots, T$}{
With prob.~$\varepsilon_{\text{R}}(n)$: (\textbf{random exploration}) for each user, recommend a random item that the user has not rated.\;
With prob.~$\varepsilon_{\text{J}}(t)$: (\textbf{joint exploration}) for each user, recommend the first item in $\permutationVar$ that the user has not rated.\;
With prob.~$1 - \varepsilon_{\text{J}}(t) - \varepsilon_{\text{R}}(n)$: (\textbf{exploitation}) for each user~$u$, recommend an item $i$ that the user has not rated and that maximizes score $\widehat{p}_{ui}^{(t)}$ given by equation \eqref{eq:simple-collaborative-greedy-score}, which depends on distance $\rho_{\cos}$ given by equation~\eqref{eq:cf-cosine-distance-def} and threshold distance $h$. Break ties randomly. (For \textsc{Collaborative-Greedy}, use distance $\widetilde{\rho}_{\cos}^{(t)}$ given by equation~\eqref{eq:cf-random-cosine-distance-def} instead of $\rho_{\cos}$ in computing the score $\widehat{p}_{ui}^{(t)}$; note that $\widetilde{\rho}_{\cos}^{(t)}$ depends on the random ordering $\xi$).
}
\end{algorithm}

The main technical hurdle in analyzing \textsc{Simple-Collaborative-Greedy} is that it's not trivial reasoning about the items that two users have both rated, especially the items recommended by the cosine-similarity exploitation. In other words, which items have revealed ratings follows a nontrivial probability distribution. We can easily circumvent this issue by changing the definition of the neighborhood of a user $u$ to only consider items that have been jointly explored. Specifically, if we denote $t_{\text{J}}$ to be the number of joint exploration steps up to time $t$, then we're guaranteed that there's a subset of $t_{\text{J}}$ items chosen uniformly at random that all users have consumed and rated (this is the first $t_{\text{J}}$ items in random item sequence $\permutationVar$ in Algorithm \ref{alg:collaborative-greedy}). The algorithm \textsc{Collaborative-Greedy} results from this slight change.  Formally, letting $\Psi_{\text{J}}^{(t)}$ be the set of jointly explored items up to time~$t$ (so $|\Psi_{\text{J}}^{(t)}| = t_{\text{J}}$), we replace $\Psi_{uv}^{(t)}$ with $\Psi_{\text{J}}^{(t)}$ in the cosine distance equation~\eqref{eq:cf-cosine-distance-def} to come up with the following modified cosine distance:
\begin{align}
\widetilde{\rho}_{\cos}^{(t)}(\labelVar_u^{(t)}, \labelVar_v^{(t)})
&=
1-
\frac{ \langle \labelVar_u^{(t)}, \labelVar_v^{(t)} \rangle_{\Psi_{\text{J}}^{(t)}} }
     { |\Psi_{\text{J}}^{(t)}| }.
\label{eq:cf-random-cosine-distance-def}
\end{align}
Note that previously, since $\Psi_{uv}^{(t)}=\text{supp}(\labelVar_u^{(t)})\cap\text{supp}(\labelVar_v^{(t)})$ can directly be computed from $\labelVar_u^{(t)}$ and $\labelVar_v^{(t)}$, the distance function $\rho_{\cos}$ had no time dependence, whereas $\widetilde{\rho}_{\cos}^{(t)}$ does since it now depends on the jointly explored items $\Psi_{\text{J}}^{(t)}$ up to time~$t$, which in general we cannot just compute from $\labelVar_u^{(t)}$ and $\labelVar_v^{(t)}$.  We give \textsc{Collaborative-Greedy} in Algorithm~\ref{alg:collaborative-greedy}. The experimental results in Section~\ref{sec:collaborative-filtering-experiments} suggest that the two algorithms have similar performance.

\vspace{.3em}
\noindent\textbf{Noisy distances.}
The cosine distance $\widetilde{\rho}_{\cos}^{(t)}$ that the theory developed depends on is random: through joint exploration, we are uniformly at random selecting a subset $\Psi_{\text{J}}^{(t)}$ of the~$m$ items and computing the cosine similarity over this random subset. Note that this randomness in the choice of entries to use in comparing rating vectors is separate from the randomness in the ratings themselves (\ie, the randomness in which a user likes or dislikes an item); this latter source of randomness is not what makes the distance function random.

To provide some more detail for why the random subset of jointly explored items $\Psi_{\text{J}}^{(t)}$ is used, note that for any two users $u$ and $v$, what we actually would like to use is the cosine distance between \textit{fully revealed} rating vectors of the two users. Formally, by letting~$\labelVar_u^*$ and~$\labelVar_v^*$ denote fully revealed rating vectors of users $u$ and $v$, then the ``non-noisy'' distance between users $u$ and $v$ is $\rho_{\cos}(\labelVar_u^*, \labelVar_v^*)$. Again, even though there is label noise in $\labelVar_u^*$ and $\labelVar_v^*$, the distance function itself $\rho_{\cos}$ is not random.  In practice we will never obtain fully revealed rating vectors $\labelVar_u^*$ and $\labelVar_v^*$, especially as typically the number of items~$m$ is so large that a user could not possibly consume them all in a lifetime.  Thus, we cannot hope to compute $\rho_{\cos}(\labelVar_u^*, \labelVar_v^*)$.  Instead what we would like is a distance function that only uses partially revealed ratings $\labelVar_u^{(t)}$ and $\labelVar_v^{(t)}$, and that is an unbiased estimate of distance $\rho_{\cos}(\labelVar_u^*, \labelVar_v^*)$. By choosing a subset $\Psi_{\text{J}}^{(t)}$ uniformly at random out of the $m$ items in defining random (and thus ``noisy'') distance $\widetilde{\rho}_{\cos}^{(t)}$ in equation~\eqref{eq:cf-random-cosine-distance-def}, we ensure that $\widetilde{\rho}_{\cos}^{(t)}(\labelVar_u^{(t)}, \labelVar_v^{(t)})$ is an unbiased estimate of the true distance $\rho_{\cos}$.  The theoretical analysis crucially depends on this uniform randomness in $\Psi_{\text{J}}^{(t)}$ and, thus, does not easily extend to an analysis of \textsc{Simple-Collaborative-Greedy}, which uses the full subset $\Psi_{uv}^{(t)}$ of commonly rated items between users $u$ and $v$.

\subsection{A Theoretical Performance Guarantee}
\label{sec:collaborative-filtering-main-result}

We now present the main result of this section that characterizes the performance of \textsc{Collaborative-Greedy}, which depends on the following \textit{cosine separation} condition between the clusters: There exists a constant $\sepVar^* \in (0, 1]$ such that for two different user clusters $g$ and $h$,
\begin{equation}
\underbrace{1 - \frac{1}{m}\langle 2 \mu_g - \mathbf{1}, 2 \mu_h - \mathbf{1} \rangle}_{\text{expected cosine distance between user clusters~}g\text{~and~}h}
\!\!\!\!\!\!\!\!
\!\!\!\!\!\!\!\!
\!\!\!\!\!\!
  \ge 4 \Big( \sigma(1-\sigma) + \sepVar^* \Big(\frac12 - \sigma\Big)^2 \Big).
\label{eq:cf-cosine-distance-separation-condition}
\end{equation}
where $\mathbf{1}$ is the all ones vector, and noise level $\sigma\in[0,\frac12)$ is defined in equation~\eqref{eq:cf-noise-level-def}.  To see why the left-hand side is the expected cosine distance, let $\labelVar_u^*$ and $\labelVar_v^*$ be fully revealed rating vectors of users $u$ and $v$ from clusters $g$ and $h$ respectively. Then $\E[\rho_{\cos}(\labelVar_u^*, \labelVar_v^*)] = \E[1 - \frac{1}{m} \langle \labelVar_u^*, \labelVar_v^* \rangle] = 1 - \frac{1}{m}\langle 2 \mu_g - \mathbf{1}, 2 \mu_h - \mathbf{1} \rangle$, where the expectation is over the random ratings of items. (Note that for fully revealed rating vectors, the two distance functions discussed earlier coincide: $\rho_{\cos}(\labelVar_u^*, \labelVar_v^*) = \widetilde{\rho}_{\cos}^{(m)}(\labelVar_u^*, \labelVar_v^*)$.) Thus, the condition asks that the expected cosine distance be large enough to combat both noise and a notion of the true separation between user clusters encoded by the constant~$\sepVar^*$.

The cosine separation condition ensures that using cosine distance can tease apart users from different clusters over time. A worry may be that this condition is too stringent and might only hold when the expected cosine distance between users clusters is $1-o(1)$ (note that cosine distance is always bounded above by~1), which would mean that this condition is unlikely to ever hold.  We provide some examples after the statement of this section's main theoretical result for which the cosine separation condition holds with probability at least $1-\numClusters^2/m$.

Next, we assume that the number of users $n$ is neither too small nor too large.  As we make precise in the theorem statement to follow, we ask $n$ to at least scale as $\numClusters m$ (the precise condition is in the theorem statement to follow). As for $n$ being too large, we ask that $n = \mathcal{O}(m^C)$ for some constant $C > 1$.  This is without loss of generality since otherwise, we can randomly divide the $n$ users into separate population pools, each of size $\mathcal{O}(m^C)$ and run the recommendation algorithm independently for each pool to achieve the same overall performance guarantee. Importantly we would still need each pool of users to have size at least scaling as $\numClusters m$.

Finally, we define $\zeta$, the minimum proportion of likable items for any user (and thus any user cluster):
\[
\zeta
\triangleq
  \min_{g \in \{1, \dots, \numClusters \}}
    \frac{\sum_{i=1}^m \ind\{ \mu_{ui} > 1/2 \}}{m}.
\]
We're now ready to state this section's main theorem.
\begin{ftheorem}[\citealt{georgehc_collaborative_filtering_nips}, Theorem~1 slightly rephrased]
\label{thm:collaborative-filtering-main-result}
Let $\delta\in(0,1)$ be a pre-specified tolerance.  Under the latent source model for online collaborative filtering, suppose the cosine separation condition holds with parameter $\sepVar^*\in(0,1]$, the parameters to \textsc{Collaborative-Greedy} are set to be $h = 1 - 2(\frac12-\sigma)^2\sepVar^*$ and $\alpha \in (0,4/7]$, and the number of users~$n$ satisfies upper bound $n=\mathcal{O}(m^C)$ and lower bound
\begin{align*}
n
=
  \Omega
  \Big(\numClusters m \log \frac{1}{\delta} +
       \Big(\frac{4}{\delta}\Big)^{1/\alpha}\Big).
\end{align*}
Then for any $\tc \le T \le \zeta m$, the expected proportion of likable items recommended by \textsc{Collaborative-Greedy} up until time $T$ satisfies
\[
\frac{\mathcal{R}_+^{(T)}}{T n}
\ge
  \Big(1 - \frac{\tc}{T}\Big)(1-\delta),
\]
where
\begin{align*}
\tc &=
\Theta
\bigg(
  \bigg(
    \frac{\log \frac{\numClusters m}{(1 - 2\sigma) \delta}}{(1 - 2\sigma)^4 (\sepVar^*)^2}
  \bigg)^{1/(1-\alpha)}
  + \Big(\frac{4}{\delta}\Big)^{1/\alpha}
\bigg).
\end{align*}
\end{ftheorem}
The proof is provided by \citet[Section~4.5]{georgehc_thesis}, which includes precise conditions (without using big O notation) on the number of users~$n$ and learning duration $T_{\text{learn}}$ in the theorem.

The above theorem says that there are $\tc$ initial time steps for which \textsc{Collaborative-Greedy} may be giving poor recommendations. Afterward, for $\tc<T<\zeta m$, the algorithm becomes near-optimal, recommending a fraction of likable items $1-\delta$ close to what an optimal oracle algorithm (that recommends all likable items first) would achieve.  Then for time horizon $T>\zeta m$, we can no longer guarantee that there are likable items left to recommend. Indeed, if the user clusters each have the same fraction of likable items, then even an oracle recommender would use up the $\zeta m$ likable items by this time.  To give a sense of how long the learning period $\tc$ is, note that when $\alpha = 1/2$, we have $\tc$ scaling as $\log^2(\numClusters m)$, and if we choose $\alpha$ close to~0, then $\tc$ becomes nearly $\log (\numClusters m)$. In summary, after $\tc$ initial time steps, which could be made nearly $\log(\numClusters m)$, and with number of users scaling as $\numClusters m$, \textsc{Collaborative-Greedy} becomes essentially optimal in that it recommends a fraction of likable items that can be made arbitrarily close to what the oracle algorithm achieves. This reasoning recovers the informal statement of Theorem \ref{thm:collaborative-filtering-main-result-informal}.

To provide intuition for the cosine separation condition, we calculate parameter $\sepVar^*$ for three examples that build on top of each other.  In every case, the condition holds with probability at least $1-\numClusters^2/m$.  In reading the derivations in these examples, it may be helpful to note that the cosine separation condition~\eqref{eq:cf-cosine-distance-separation-condition} can also be equivalently expressed in terms of expected cosine similarity between every pair of user clusters~$g$ and~$h$:
\begin{equation*}
\underbrace{\frac{1}{m}\langle 2 \mu_g - \mathbf{1}, 2 \mu_h - \mathbf{1} \rangle}_{\text{expected cosine similarity between user clusters~}g\text{~and~}h}
\!\!\!\!\!\!\!\!
\!\!\!\!\!\!\!\!
\!\!\!\!\!\!\!\!
\!\!\!\!
  \le (1 - \sepVar^*)(1 - 2\sigma)^2.
\end{equation*}

\begin{fexample}
Consider when there is no noise, \ie, $\sigma = 0$. Then users' ratings are deterministic given their user cluster. We construct the true underlying item preference vectors $\mu_1, \dots, \mu_{\numClusters} \in [0, 1]^m$ by sampling every entry $\mu_{gi}$ ($g \in \{1, \dots, \numClusters\}$ and $i \in \{1, \dots, m\}$) to be i.i.d.~$\text{Bernoulli}(1/2)$. In this case, the cosine separation condition, with true separation $\sepVar^* = 1 - \sqrt{\frac{\log m}{m}}$, holds with probability at least $1 - \numClusters^2/m$.

To show this, note that for any item $i$ and pair of distinct user clusters $g$ and $h$, the product $(2\mu_{gi}-1)(2\mu_{hi}-1)$ is a Rademacher random variable ($+1$ or $-1$ each with probability $\frac12$), and thus the dot product $\langle 2\mu_g - \mathbf{1}, 2\mu_h - \mathbf{1} \rangle$ is equal to the sum of $m$ i.i.d.~Rademacher random variables, each of which is sub-Gaussian with parameter~1. Hence, the sum is zero-mean sub-Gaussian with parameter $\sqrt{m}$, implying that
\[
\mathbb{P}\big(
  \langle 2\mu_g - \mathbf{1}, 2\mu_h - \mathbf{1} \rangle
  \ge 
    s
\big)
\le
  \exp\Big( -\frac{s^2}{2m} \Big).
\]
Plugging in $s = m \sqrt{\frac{\log m}{m}}$, we see that
\[
\mathbb{P}\bigg(
  \frac{1}{m}
  \langle 2\mu_g - \mathbf{1}, 2\mu_h - \mathbf{1} \rangle
  \ge 
    \sqrt{\frac{\log m}{m}}
\bigg)
\le
  \frac{1}{m}.
\]
Union-bounding over all distinct pairs of user clusters,
\begin{align*}
&\mathbb{P}\Bigg(
   \bigcup_{g, h \in \{1, \dots, \numClusters\}\text{ s.t.~}g\ne h}
     \bigg\{
       \frac{1}{m}
       \langle 2\mu_g - \mathbf{1}, 2\mu_h - \mathbf{1} \rangle
       \ge 
         \sqrt{\frac{\log m}{m}}
     \bigg\}
 \Bigg) \\
&
\le
  \binom{\numClusters}{2}
  \frac{1}{m}
\le
  \frac{\numClusters^2}{m}.
\end{align*}
Hence, with probability at least $1 - \numClusters^2/m$, we have
\[
\frac{1}{m}
\langle 2\mu_g - \mathbf{1}, 2\mu_h - \mathbf{1} \rangle
< 
  \sqrt{\frac{\log m}{m}}
\]
for every distinct pair of user clusters $g$ and $h$.  Noting that $\sigma = 0$, the cosine separation condition holds with parameter~$\sepVar^*$ to be $1 - \sqrt{\frac{\log m}{m}}$.
\end{fexample}

\begin{fexample}
We expand on the previous example by introducing noise with parameter $\sigma \in (0, 1/2)$. Now let the item preference vectors $\mu_1, \dots, \mu_{\numClusters} \in [0,1]^m$ have i.i.d.~entries that are $1 - \sigma$ (likable) or $\sigma$ (unlikable) with probability $\frac12$ each. Then for a distinct pair of user clusters $g$ and $h$, if $\mu_{gi} = \mu_{hi}$ (which happens with probability 1/2), then $\mathbb{E}[(2\mu_g - 1)\cdot(2\mu_h - 1)] = (1 - \sigma)^2 + \sigma^2 - 2\sigma(1 - \sigma) = (1 - 2\sigma)^2$, and if $\mu_{gi} \ne \mu_{hi}$ (so $\mu_{gi} = 1 - \mu_{hi}$ in this example, also occurring with probability 1/2), then $\mathbb{E}[(2\mu_g - 1)\cdot(2\mu_h - 1)] = 2\sigma(1 - \sigma) - (1 - \sigma)^2 - \sigma^2 = -(1 - 2\sigma)^2$.  This means that $\langle 2\mu_g - \mathbf{1}, 2\mu_h - \mathbf{1} \rangle$ is again a sum of Rademacher random variables, except now scaled by $(1 - 2\sigma)^2$. This sum is sub-Gaussian with parameter $\sqrt{m} (1 - 2\sigma)^2$. By a similar calculation as the previous example, with probability at least $1 - \numClusters^2/m$,
\[
\frac{1}{m}
\langle 2\mu_g - \mathbf{1}, 2\mu_h - \mathbf{1} \rangle
< 
  (1 - 2\sigma)^2 \sqrt{\frac{\log m}{m}}
\]
for every distinct pair of user clusters $g$ and $h$. Thus, the cosine separation condition holds with parameter $\sepVar^* = 1 - \sqrt{\frac{\log m}{m}}$.
\end{fexample}

\begin{fexample}
Building off our second example, we now suppose that entries in the item preference vectors $\mu_1, \dots, \mu_{\numClusters} \in [0,1]^m$ have entries that are $1 - \sigma$ (likable) with probability $\zeta \in (0, 1/2)$, and $\sigma$ (unlikable) with probability $1 - \zeta$. Then for item $i$ and different user clusters $g$ and $h$, $\mu_{gi} = \mu_{hi}$ with probability $\zeta + (1 - \zeta)^2$. This implies that $ \mathbb{E}[\langle 2\mu_g - \mathbf{1}, 2\mu_h - \mathbf{1} \rangle] = m (1-2\sigma)^2 (1-2\zeta)^2 $, and one can verify that the dot product $\langle 2\mu_g - \mathbf{1}, 2\mu_h - \mathbf{1} \rangle$ is still sub-Gaussian with parameter $\sqrt{m} (1 - 2\sigma)^2$. Using a similar calculation as before but now accounting for the mean of the dot product no longer being~0, with probability at least $1 - \numClusters^2/m$,
\[
\frac{1}{m}
\langle 2\mu_g - \mathbf{1}, 2\mu_h - \mathbf{1} \rangle
< 
  (1 - 2\sigma)^2 \bigg((1 - 2\zeta)^2 + \sqrt{\frac{\log m}{m}}\bigg)
\]
for every distinct pair of user clusters $g$ and $h$. Then the cosine separation condition holds with parameter $\sepVar^* = 1 - (1 - 2\zeta)^2 - \sqrt{\frac{\log m}{m}}$.
\end{fexample}

\subsection{Experimental Results}
\label{sec:collaborative-filtering-experiments}

\citet{georgehc_thesis} demonstrates \textsc{Simple-Collaborative-Greedy} and \textsc{Collaborative}-\textsc{Greedy} on recommending movies, showing that the two algorithms have comparable performance and both outperform two existing recommendation algorithms Popularity Amongst Friends (PAF) \citep{paf} and Deshpande and Montanari's method (DM) \citep{deshpande_montanari}. At each time step, PAF finds nearest neighbors (``friends'') for every user and recommends to a user the ``most popular'' item, \ie, the one with the most number of $+1$ ratings, among the user's friends. DM doesn't do any collaboration beyond a preprocessing step that computes item feature vectors via matrix completion. Then during online recommendation, DM learns user feature vectors over time with the help of item feature vectors and recommends an item to each user based on whether it aligns well with the user's feature vector.

The experimental setup involves simulating an online recommendation system using real ratings from the MovieLens 10M \citep{movielens} and Netflix \citep{netflix} datasets, each of which provides a sparsely filled user-by-movie rating matrix with ratings out of 5 stars.  Unfortunately, existing collaborative filtering datasets such as the two movie rating datasets considered, do not offer the interactivity of a real online recommendation system, nor do they allow us to reveal the rating for an item that a user did not ever rate.  For simulating an online system, the former issue can be dealt with by simply revealing entries in the user-by-item rating matrix over time.  The latter issue can be addressed by only considering a dense ``top users vs.~top items'' subset of each dataset. In particular, by considering only the ``top'' users who have rated the most number of items, and the ``top'' items that have received the most number of ratings, we get a dense part of the dataset.  While this dense part is unrepresentative of the rest of the dataset, it does allow us to use actual ratings provided by users without synthesizing any ratings. A rigorous validation would require an implementation of an actual interactive online recommendation system.

An initial question to ask is whether these dense movie ratings matrices exhibit clustering behavior.  Automatically learning the structure of these matrices using the method by \citet{rgrosse_best_paper_2012} reveals that Bayesian clustered tensor factorization (BCTF) accurately models the data. This finding isn't surprising as BCTF has previously been used to model movie ratings data~\citep{bctf_2009}. BCTF effectively clusters both users and movies so that we get structure such as that shown in Figure \ref{fig:bctf} for the MovieLens 10M ``top users vs.~top items'' matrix.

Following the experimental setup of \citet{paf}, ratings of 4 stars or more are quantized to be $+1$ (likable), and ratings of 3 stars or less to be $-1$ (unlikable). The dense subsets considered still have missing entries.  If a user $u$ hasn't rated item $j$ in the dataset, then the corresponding true rating is set to 0, meaning that in the simulation, upon recommending item $j$ to user $u$, we receive 0 reward, but we still mark that user $u$ has consumed item $j$; thus, item $j$ can no longer be recommended to user $u$. For both MovieLens 10M and Netflix datasets, the top $n=200$ users and $m=500$ movies are considered. For MovieLens 10M, the resulting user-by-rating matrix has 80.7\% nonzero entries. For Netflix, the resulting matrix has 86.0\% nonzero entries.  For an algorithm that recommends item $\psi_{ut}$ to user $u$ at time $t$, the algorithm's average cumulative reward up to time $T$ is measured by
\begin{equation*}
\frac{1}{n} \sum_{t=1}^T\sum_{u=1}^n \labelVar_{u \psi_{ut}}^{(T)},
\end{equation*}
where we average over users.

For all methods, items are recommended until we reach time $T=500$, \ie, we make movie recommendations until each user has seen all $m=500$ movies.  The matrix completion step for DM is not allowed to see the users in the test set, but it is allowed to see the same items as what is in the simulated online recommendation system in order to compute these items' feature vectors (using the rest of the users in the dataset). Furthermore, when a rating is revealed, DM is provided with both the thresholded and non-thresholded ratings, the latter of which DM uses to estimate user feature vectors.

\begin{figure}
\noindent
\centering

\hspace{-.12em}
\includegraphics[width=5.8cm, clip=true, trim=.18in 0 .1in 0]{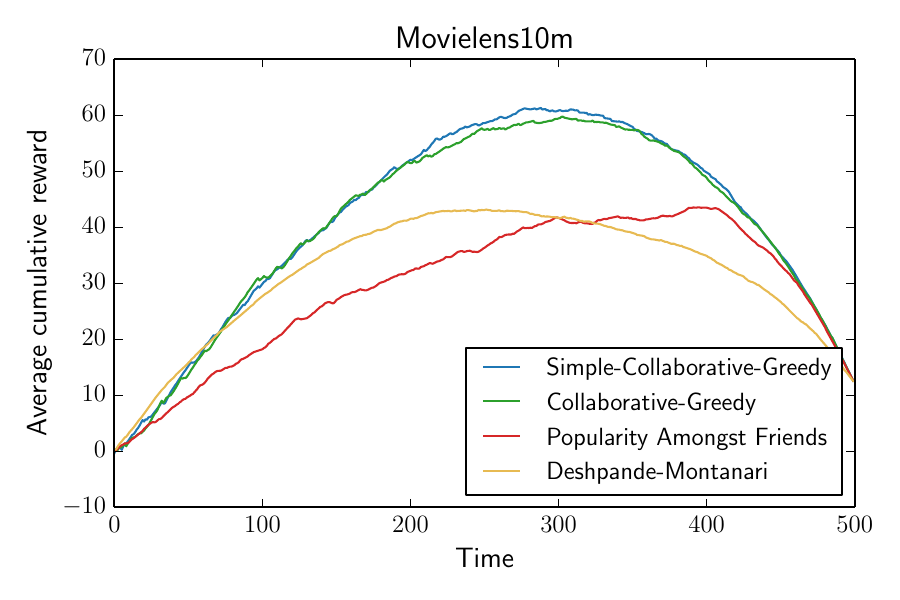}
\includegraphics[width=5.8cm, clip=true, trim=.18in 0 .1in 0]{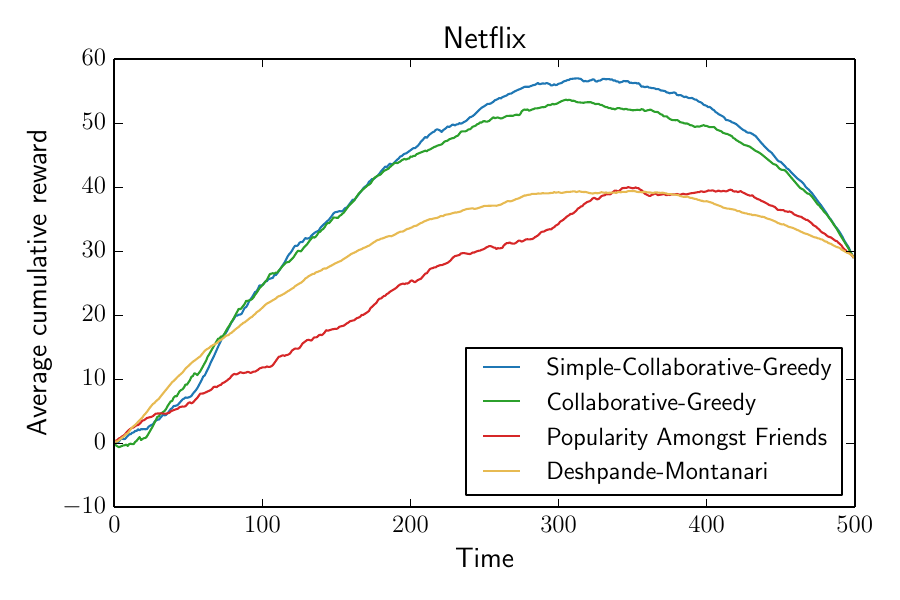}
\vspace*{-2em}
\caption{Average cumulative rewards over time for the MovieLens 10M and Netflix datasets (figure source: \citealt[Figure~4.2]{georgehc_thesis}).}
\label{fig:supp-results-top}
\end{figure}

Parameters $h$ and $\alpha$ for \textsc{Simple-Collaborative-Greedy} and \textsc{Collaborative-Greedy} are chosen by sweeping over the two parameters on training data consisting of 200 users that are the ``next top'' 200 users, \ie, ranked 201 to 400 in number of movie ratings they provided.  The search space is discretized to grids $h\in\{0.0, 0.1, \dots, 1.0\}$ and $\alpha\in\{0.1, 0.2, 0.3, 0.4, 0.5\}$.  The parameter setting achieving the highest area under the cumulative reward curve is chosen per algorithm. For both MovieLens 10M and Netflix datasets, this corresponded to setting $h=0.9$ and $\alpha=0.5$ for \textsc{Simple-Collaborative-Greedy}, and $h=1.0$ and $\alpha=0.5$ for \textsc{Collaborative-Greedy}. In contrast, the parameters for PAF and DM are chosen to be the best parameters for the \textit{test} data among a wide range of parameters.  The results are shown in Figure \ref{fig:supp-results-top}.  \textsc{Simple-Collaborative-Greedy} and \textsc{Collaborative-Greedy} have similar performance, and both outperform PAF and DM. We remark that the curves are roughly concave, which is expected since once the algorithms have finished recommending likable items (roughly around time step 300), they end up recommending mostly unlikable items until they have exhausted all the items.

\subsection{Accounting for Item Similarity}
\label{sec:collaborative-filtering-discussion}

So far, we did not assume knowledge of which items are similar. In fact, if we knew which items are similar, then a simple observation reveals that we can drastically improve recommendation performance. Specifically, suppose that we have $\numClusters_{\text{item}}$ item clusters, where within each item cluster, either all the items are likable or all the items are unlikable by each user cluster. Thus, an entire item cluster is either likable or unlikable by each user cluster. Moreover, suppose for simplicity that we actually knew what these item clusters are. For example, perhaps we estimated these from mining item descriptions (\eg, in a movie recommendation system, we can scrape information about movies such as their genres and cluster items based on genre).

Then we can use \textsc{Collaborative-Greedy} to explore item \textit{clusters} (basically we apply the same algorithm but instead track scores for the $\numClusters_{\text{item}}$ item clusters rather than the~$m$ items, where recommending item~$i$ to user~$u$ tells us what user~$u$ thinks of item~$i$'s cluster). Next, we can introduce a new exploitation step in which we recommend items from an item cluster that we already are confident that a user finds likable, assigning some probability to this exploitation. If the item clusters are large, then this strategy substantially increases the number of likable items we are able to recommend! This intuition not only suggests how to incorporate knowledge of item clusters but also why they can improve a recommendation system's performance.

Whereas our exposition has focused on online collaborative filtering by looking at which users are similar, recently, \citet{bresler_shah_voloch_2015} have established theoretical guarantees for online collaborative filtering by looking at which items are similar. \citet{karzand_bresler_2017} take a step further, extending the theory to handle simultaneously using both user and item structure.

\section{Patch-Based Image Segmentation}
\label{chap:image-segmentation}

Delineating objects in an image is a fundamental task that feeds into a flurry of analyses downstream, from parsing what is happening in a scene to discovering indicators for disease in the case of medical image analysis. To learn what a particular object looks like, whether a cat or a tumor, we rely on seeing examples of the object. Luckily, we now have access to enormous, ever-growing repositories of images of virtually everything, for example, on photo sharing websites like Flickr, contained within videos on YouTube, or privately stored in hospitals in the case of medical images. Often we could easily procure in excess of thousands of training images for a particular object, perhaps with the help of crowdsourcing manual labeling via Amazon Mechanical Turk.

At the outset though, thousands or even millions of images might seem small since the space of possible images is enormous. Consider a two-dimensional image of 100-by-100 pixels, where at each pixel, we store one of two values.  The number of such images is $2^{10000}$, which happens to be larger than the estimated number of atoms in the observable universe ($\approx 10^{80}$).  Nearly all of these possible images would be completely implausible to us as naturally occurring based on what we see with our eyes day-to-day.  What structure is present in an image that makes it ``plausible'', and how do we exploit this structure for prediction?

In this section, we focus on the simple setting of segmenting out where a foreground object of interest is from the background, a problem referred to as binary image segmentation. Our running example is the task of finding a human organ in a medical image. Specifically for medical image segmentation, nearest neighbor and kernel classifiers have been widely used, originally at the pixel (or voxel, for 3D images) level \citep{left_atrium_segmentation_2010,label_fusion_generative_model_2010} and more recently for image patches \citep{coupe_2011,rousseau_2011,bai_2013,patch_descriptors,deep_NAT,zandifar_2017}. Of course, in the extreme case, a patch can just be a single pixel, so patch-based approaches subsume pixel-based approaches.

We specifically study patch-based binary image segmentation, which sits at a middle ground between other patch-based prediction tasks and the earlier time series forecasting problem. To relate to patch-based prediction tasks beyond binary image segmentation, note that rather than predicting a binary label per pixel, we could predict real-valued labels or patches of such labels leading to patch-based methods for image denoising, reconstruction, restoration, and synthesis (\eg, \citealt{nonlocal_means,rousseau_2013,ender_patches_2013,eugenio_patches_2013}).  As these patch-based methods are all syntactically similar, analysis of the binary image segmentation setting could possibly be extended to handle these more sophisticated tasks as well. To relate to the earlier time series forecasting work, consider a myopic approach to patch-based segmentation, where we predict the label of each pixel using only the observed image patch centered at that pixel. This patch, when flattened out into a single dimension, could be thought of as a time series! Thus, the theoretical results for times series forecasting could help explain the performance of myopic nearest neighbor segmentation. In doing so, however, we want to account for image rather than temporal structure: pixels nearby, as well as their associated patches, are similar.

For medical image segmentation in particular, perhaps the primary reason for the popularity of nonparametric patch-based methods is that the main competing approach to the problem, called label fusion, requires robust alignment of images, whereas patch-based methods do not.  Specifically, standard label fusion techniques begin by first aligning all the images into a common coordinate frame and henceforth effectively assuming the alignment to be perfect. Because alignment quality substantially affects label fusion's segmentation performance, alignments are typically computed by so-called \textit{nonrigid registration} that estimates a highly nonlinear transformation between two images.  When nonrigid registration works well, then label fusion also works well.  The main issue is that nonrigid registration often fails when the images exhibit large anatomical variablity as in whole body CT scans, or when the images have other quality problems commonly found in clinical scans taken at hospitals, such as low resolution and insufficient contrast \citep{sridharan_2013}.  Unlike standard label fusion, patch-based methods do not require input images to be aligned perfectly, making them a promising alternative when the images we encounter present significant alignment challenges.

Of course, another reason for patch-based methods' growing popularity is their efficiency of computation, due to increasing availability of fast approximate nearest-neighbor search algorithms both for general high-dimensional spaces and tailored specifically for image patches (as discussed back in Chapter~\ref{chap:nn-survey-intro}). Thus, if the end goal is segmentation or a decision based on segmentation, for many problems solving numerous nonrigid registration subproblems required for standard label fusion could be a computationally expensive detour that, even if successful, might not produce better solutions than a patch-based approach.

Many patch-based image segmentation methods are precisely variants of nearest neighbor classification. In the simplest case, to determine whether a pixel in an input image should be labeled as the foreground object of interest or background, we consider the patch centered at that pixel. We compare this image patch to patches in a training database that are labeled labeled either foreground or background depending on the pixel at the center of the training patch. We transfer the label from the closest patch in the training database to the pixel of interest in the new image. A plethora of embellishments improve this algorithm, such as, but not limited to, using more than just the first nearest neighbor (\eg, $k$-NN for $k>1$, fixed-radius NN, kernel classification) \citep{coupe_2011,rousseau_2011,patch_descriptors}, incorporating hand-engineered or learned feature descriptors \citep{patch_descriptors,deep_NAT}, cleverly choosing the shape of a patch \citep{patch_descriptors}, and enforcing consistency among adjacent pixels by assigning each training intensity image patch to a label patch rather than a single label \citep{rousseau_2011,patch_descriptors} or by employing a Markov random field \citep{learning_low_level_vision}.

Despite the broad popularity and success of nonparametric patch-based image segmentation and the smorgasbord of tricks to enhance its performance over the years, nonasymptotic theory for such methods was only recently developed by \citet{georgehc_image_segmentation_miccai} for 1-NN and kernel patch-based segmentation.  In this section, we cover their main result.  As with the other applications, the theory here depends on an underlying probabilistic model, which once again assumes clustering structure.  In particular, each patch within an image is assumed to be accurately modeled by mixture model with at most $\numClusters$ components (note that different patches can be associated with different mixture models).  \citet{georgehc_image_segmentation_miccai} refer to their model as the \textit{latent source model for patch-based image segmentation}.

\begin{ftheorem}
[Informal statement of Theorem \ref{thm:image-segmentation-main-result}]
\label{thm:image-segmentation-main-result-informal}
Let $\delta\in(0, 1)$ be an error tolerance, and $|\mathcal{I}|$ be the number of pixels in the full image (and not just a patch).  Under a special case of the latent source model for patch-based image segmentation, if the training patches centered at nearby pixels with opposite labels ``foreground'' and ``background'' are sufficiently different (depends on~$\delta$), image patches satisfy a ``jigsaw condition'' that imposes local smoothness, and the number of labeled training images satisfies $n = \Theta(\numClusters \log (|\mathcal{I}|\numClusters/\delta))$, then \mbox{1-NN} and kernel patch-based classifiers each achieve average pixel mislabeling rate at most $\delta$.
\end{ftheorem}
Importantly, the training data are full images and not just the patches. For medical images, the number of training data $n$ corresponds to the number of human subjects in the training set.

We begin in Section \ref{sec:myopic} with a simple case of the model that corresponds to classifying each pixel's label separately from other pixels; this setup is referred to as \textit{pointwise segmentation}. The theoretical guarantees for 1-NN and kernel patch-based segmentation are for pointwise segmentation.  The analysis borrows heavily from the time series forecasting theory and crucially relies on a local structural property called the \textit{jigsaw condition}, which holds when neighboring patches are sufficiently similar.

The pointwise setup only incorporates local structure, and can be extended to incorporate global structure, resulting in the model presented in Section~\ref{sec:full}, which in turn leads to an iterative patch-based image segmentation algorithm that combines ideas from patch-based image restoration \citep{epll} and distributed optimization \citep{boyd_admm_2011}. This algorithm alternates between predicting label patches separately and merging these local estimates to form a globally consistent segmentation. Various existing patch-based algorithms can be viewed as special cases of this proposed algorithm, which, as shown in Section~\ref{sec:image-segmentation-experimental-results}, significantly outperforms the pointwise segmentation algorithms that theoretical guarantees are provided for.  However, the new iterative algorithm currently has no known theoretical guarantees.

We remark that in practice, patch-based image segmentation methods often do not achieve close-to-zero error. A corollary of the main theoretical result above is that if the error cannot be made arbitrarily small, then the assumptions made do not hold across the entire image, \eg, if the assumptions fail to hold for 10\% of the image, then the mislabeling rate will similarly not be guaranteed to go below 10\%.  Thus, segmentation algorithms should be designed to be robust even in parts of the image where the structural assumptions of~\citet{georgehc_image_segmentation_miccai} collapse.

\subsection{Pointwise Segmentation}
\label{sec:myopic}

For an image $A$, we use $A(i)$ to denote the value of image~$A$ at pixel~$i$, and $A[i]$ to denote the patch of image $A$ centered at pixel $i$ based on a pre-specified patch shape; $A[i]$ can include feature descriptors in addition to raw intensity values. Each pixel $i$ belongs to a finite, uniformly sampled lattice~$\mathcal{I}$.

Given an intensity image $\obsVar$, we aim to predict its label image~$\labelVar$ that delineates an object of interest in image $\obsVar$. In particular, for each pixel $i \in \mathcal{I}$, we predict label $\labelVar(i) \in \{0, 1\}$, where 1 corresponds to foreground (object of interest) and 0 to background. To make this prediction, we use patches of image $\obsVar$, each patch of dimensionality~$d$.  For example, for a 2D image, if we use 5x5 patches, then $d=25$, and for a 3D image, if we use 5x5x5 patches, then $d=125$.

We model the joint distribution $p(\labelVar(i),\obsVar[i])$ of label $\labelVar(i)$ and image patch $\obsVar[i]$ as a generalization of a Gaussian mixture model with diagonal covariances, where each mixture component corresponds to either $\labelVar(i)=1$ or $\labelVar(i)=0$. This generalization is called a \textit{diagonal sub-Gaussian mixture model}, to be described shortly.  First, we provide a concrete example where label $\labelVar(i)$ and patch $\obsVar[i]$ are related through a Gaussian mixture model with $\dsmmNumClusters_i$ mixture components. Mixture component $c \in \{1, \dots, \dsmmNumClusters_i\}$ occurs with probability $\pi_{ic} \in (0,1]$ and has mean vector $\mu_{ic} \in \mathbb{R}^d$ and label $\lambda_{ic} \in \{0,1\}$. In this example, we assume that all covariance matrices are $\sigma^2 \mathbf{I}_{d\times d}$, and that there exists constant $\pi_{\min}>0$ such that $\pi_{ic} \ge \pi_{\min}$ for all $i,c$. Thus, image patch $\obsVar[i]$ belongs to mixture component $c$ with probability $\pi_{ic}$, in which case $\obsVar[i] = \mu_{ic} + W_i$, where vector $W_i \in \mathbb{R}^d$ consists of white Gaussian noise with variance $\sigma^2$, and $\labelVar(i)=\lambda_{ic}$.  Formally,
\[
p(\labelVar(i),\obsVar[i])
=\sum_{c=1}^{\dsmmNumClusters_i}
   \pi_{ic}~\!
   \mathcal{N}(\obsVar[i];\mu_{ic},\sigma^2 \mathbf{I}_{d\times d})
   \ind\{\labelVar(i)=\lambda_{ic}\},
\]
where $\mathcal{N}(\cdot; \mu, \Sigma)$ is a Gaussian density with mean $\mu$ and covariance $\Sigma$.  Each pixel~$i$ has its own mixture model with parameters $(\pi_{ic}, \mu_{ic}, \lambda_{ic})$ for $c = 1,\dots,\dsmmNumClusters_i$.  The \textit{diagonal sub-Gaussian mixture model} refers to a simple generalization where noise vector $W_i$ consists of entries that are i.i.d.~zero-mean sub-Gaussian with parameter $\sigma$. This generalization precisely corresponds to the latent source model for time series classification where we disallow time shifts (so $\Delta_{\max}=0$) and the time horizon $T$ is replaced by the patch dimensionality $d$.

We assume that every pixel is associated with its own diagonal sub-Gaussian mixture model whose parameters are fixed but unknown. Similar to recent work on modeling natural imagery patches \citep{epll,patch_gmm}, the model only looks at each individual patch separately; no assumptions are made about how different overlapping patches behave jointly.  In natural imagery, image patches turn out to very accurately behave like samples from a Gaussian mixture model \citep{patch_gmm}.  We refer to this model as a pointwise latent source model for patch-based image segmentation.

As with time series forecasting and online collaborative filtering, rather than learning the mixture model components, we instead take a nonparametric approach, using available training data in nearest neighbor and kernel classifiers to predict label $\labelVar(i)$ from image patch $\obsVar[i]$. To this end, we assume we have access to~$n$ i.i.d.~training intensity-label image pairs $(\obsVar_1,\labelVar_1), \dots, (\obsVar_n,\labelVar_n)$ that obey our probabilistic model above. These are full images and not just the patches.

\subsubsection*{Algorithms}

We translate the 1-NN and kernel time series classifiers from Section~\ref{sec:time-series-inference} to the patch-based image segmentation setting.  Here, there are no time shifts, so the distance function used is just Euclidean distance, and the kernel function used is again the Gaussian kernel $K(s)=\exp(-\frac12 s^2)$. Unlike before, we now have to carefully keep track of which training intensity image a patch comes from, and also what the center pixel coordinate is for a patch. Similar to the online collaborative filtering setup, we index training data with dummy variables $u,v\in\{1,2,\dots,n\}$, which had referred to people in a recommendation system earlier, and now each training intensity image $\obsVar_u$ is associated with a person $u$ in the medical imaging context.

The following methods operate on each pixel $i$ separately, predicting label $\labelVar(i)$ only based on image patch~$\obsVar[i]$:

\vspace{.3em}
\noindent\textbf{Pointwise 1-NN segmentation.}
We first find which training intensity image $\obsVar_u$ has a patch centered at pixel $j$ that is closest to observed intensity patch $\obsVar[i]$. This amounts to solving
\[
(\widehat{u},\widehat{j})
=
 \underset{u\in\{1,2,\dots,n\}, j \in N(i)}
            {\text{arg}\,\text{min}}
   \|\obsVar_u[j]-\obsVar[i]\|^2,
\]
where $\|\cdot\|$ denotes Euclidean norm, and $N(i)$ refers to a user-specified finite set of pixels that are neighbors of pixel $i$. Label $\labelVar(i)$ is estimated to be the same as the closest training patch's label:
\[
\widehat{\labelVar}_{1\text{-NN}}(i|\obsVar[i])=
\labelVar_{\widehat{u}}(\,\widehat{j}\,).
\]
\textbf{Pointwise kernel segmentation.}
We use Euclidean distance and a Gaussian kernel with bandwidth~$h$. Then label $\labelVar(i)$ is estimated to be the label with the higher vote:
\[
\widehat{\labelVar}_{\text{Gauss}}(i|\obsVar[i];h)=
\begin{cases}
+1 & \text{if }V_{+1}(i|\obsVar[i];h)\ge V_{-1}(i|\obsVar[i];h),\\
-1 & \text{otherwise},
\end{cases}
\]
where the weighted votes for labels 0 and 1 are
\begin{align*}
V_{0}(i|\obsVar[i]; h)
&= \sum_{u=1}^n
     \sum_{j\in N(i)}
       \exp\Big(-\frac{\|\obsVar_u[j]-\obsVar[i]\|^2}{2h^2}\Big)
       \ind\{\labelVar_u(j)= 0\}, \\
V_{1}(i|\obsVar[i]; h)
&= \sum_{u=1}^n
     \sum_{j\in N(i)}
       \exp\Big(-\frac{\|\obsVar_u[j]-\obsVar[i]\|^2}{2h^2}\Big)
       \ind\{\labelVar_u(j)= 1\},
\end{align*}
and $N(i)$ again refers to user-specified neighboring pixels of pixel $i$. 

We obtain pointwise 1-NN segmentation with $h \rightarrow 0$. For an identical reason as in the time series forecasting (\cf, Section \ref{sec:time-series-oracle}), the prediction made at each pixel $i$ by pointwise kernel segmentation approximates an oracle MAP classifier that knows the parameters for the diagonal sub-Gaussian mixture model at pixel~$i$. This oracle MAP classifier is myopic, predicting label $\labelVar(i)$ only given patch $\obsVar[i]$.

Pointwise kernel segmentation has been used extensively for patch-based segmentation \citep{bai_2013,coupe_2011,rousseau_2011,patch_descriptors}, where we note that the formulation above readily allows for one to choose which training image patches are considered neighbors, what the patch shape is, and whether feature descriptors are part of the intensity patch vector $\obsVar[i]$. For example, a simple choice of feature at pixel $i$ is the coordinate for pixel $i$ itself. Thus, we can encode as part of the exponentially decaying weight $\exp(-\frac1{2h^2}\|\obsVar_u[j] - \obsVar[i]\|^2)$ how far apart pixels $i$ and $j \in N(i)$ are, yielding a segmentation algorithm previously derived from a Bayesian model that explicitly models this displacement \citep{bai_2013}.

\subsubsection*{Theoretical Guarantees}

The model above allows nearby pixels to be associated with dramatically different mixture models. However, real images are smooth, with patches centered at two adjacent pixels likely similar. We incorporate this smoothness via a structural property on the sub-Gaussian mixture model parameters associated with nearby pixels.

To build some intuition, we consider two extremes. First, it could be that the $|\mathcal{I}|$ mixture models (one per pixel) are actually all identical. This means that every intensity patch comes from the same distribution. If we know this, then when we do pointwise 1-NN or kernel segmentation, we could compare an observed intensity patch $\obsVar[i]$ with a training image patch centered \textit{anywhere} in the training image, since the patches all follow the same distribution. On the opposite extreme, the~$|\mathcal{I}|$ mixture models could have no commonalities, and so when we use pointwise 1-NN or kernel segmentation, it only makes sense to compare intensity patch $\obsVar[i]$ with training image patches also centered at pixel $i$. We can interpolate between these two extremes by saying how far away two pixels have to be while still sharing mixture component parameters, which we formalize as follows:

\begin{blockquote}
\textit{Jigsaw condition.}
For every mixture component $(\pi_{ic},\mu_{ic},\lambda_{ic})$ of the diagonal sub-Gaussian mixture model associated with pixel $i$, there exists a neighbor $j \in N^*(i)$ such that the diagonal sub-Gaussian mixture model associated with pixel $j$ also has a mixture component with mean $\mu_{ic}$, label $\lambda_{ic}$, and mixture weight at least $\pi_{\min}$; this weight need not be equal to $\pi_{ic}$.
\end{blockquote}

\noindent
The name of this structural property is inspired by a jigsaw puzzle, where the pieces of the puzzle somehow need to fit with nearby pieces to produce the final picture.

The shape and size of neighborhood $N^*(i)$, which is fixed and unknown like the mixture model parameters, control how similar the mixture models are across image pixels.  For the two extremes discussed previously, the true neighborhood $N^*(i)$ corresponds to the full space of pixels in the first case, and $N^*(i) = \{i\}$ in the second case. As we already hinted at, what $N^*(i)$ is affects how far from pixel $i$ we should look for training patches, \ie, how to choose neighborhood $N(i)$ in pointwise 1-NN and kernel segmentation, where ideally $N(i) = N^*(i)$.

As with the main result for time series forecasting, the main result here depends on the separation between training intensity image patches that correspond to the two different labels:
\[
\sepVar
\triangleq\min_{\substack{u,v\in\{1,\dots,n\},\\
i\in\mathcal{I},j\in N(i)\text{ s.t. }\labelVar_u(i)\ne \labelVar_v(j)}
}\|\obsVar_u[i]-\obsVar_v[j]\|.
\]
Intuitively, a small separation corresponds to the case of two training intensity image patches that are very similar but one corresponds to foreground and the other to background. In this case, a nearest neighbor approach may easily select a patch with the wrong label, resulting in an error.

We now state the main theoretical result of this section.

\begin{ftheorem}[\citealt{georgehc_image_segmentation_miccai}, Theorem~1 slightly rephrased]
\label{thm:image-segmentation-main-result}
Let $N(i)$ be the user-specified neighborhood of pixel $i$.  Denote $\dsmmNumClustersMax \triangleq \max_{i \in \mathcal{I}} \dsmmNumClusters_i$, $|N| \triangleq \max_{i \in \mathcal{I}} |N(i)|$, $\pi_{0}(i) \triangleq \mathbb{P}(\labelVar(i) = 0) = \sum_{c=1}^{\dsmmNumClusters_i} \pi_{ic} \ind\{\lambda_{ic} = 0\}$, and
\[
\pi_{1}(i) \triangleq \mathbb{P}(\labelVar(i) = 1)
 = \sum_{c=1}^{\dsmmNumClusters_i} \pi_{ic} \ind{\{\lambda_{ic} = 1\}}.
\]
Under the model above with $n$ training intensity-label image pairs and provided that the jigsaw condition holds with neighborhood $N^*$ such that $N^*(i) \subseteq N(i)$ for every pixel $i$:
\begin{itemize}

\item[(a)] Pointwise 1-NN segmentation has expected pixel labeling error rate
\begin{align}
&\mathbb{E}\bigg[
  \frac{1}{|\mathcal{I}|}
    \sum_{i\in\mathcal{I}}
      \ind\{ \widehat{\labelVar}_{1\text{-NN}}(i | \obsVar[i]) \ne \labelVar(i) \}
\bigg] \nonumber \\
&\le
   |\mathcal{I}|\dsmmNumClustersMax\exp\Big(-\frac{n\pi_{\min}}{8}\Big)
   +
   |N|n\exp\Big(-\frac{\sepVar^2}{16\sigma^{2}}\Big).
\label{eq:latent-patches-nn-main-bound}
\end{align}

\item[(b)] Pointwise kernel segmentation has expected pixel labeling error rate
\begin{align}
&\mathbb{E}\bigg[
   \frac{1}{|\mathcal{I}|}
     \sum_{i\in\mathcal{I}}
       \ind\{ \widehat{\labelVar}_{\text{Gauss}}(i | \obsVar[i]; h) \ne \labelVar(i) \}
 \bigg] \nonumber \\
&\le
   |\mathcal{I}|\dsmmNumClustersMax\exp\Big(-\frac{n\pi_{\min}}{8}\Big)
   +
   |N|n
   \exp\big(-\frac{(h^2 - 2\sigma^2)\sepVar^2}{2h^4}\big)
\label{eq:latent-patches-wmv-main-bound}
\end{align}

\end{itemize}
In particular, for any pre-specified error tolerance $\delta\in(0,1)$, by choosing bandwidth $h=2\sigma$ for pointwise kernel segmentation, the two expected pixel labeling error rate upper bounds~\eqref{eq:latent-patches-nn-main-bound} and~\eqref{eq:latent-patches-wmv-main-bound} match, and by choosing number of training data $n \ge \frac{8}{\pi_{\min}}\log(\frac{2|\mathcal{I}|\numClusters}{\delta})$, and if the separation grows as $S \ge 4\sigma\sqrt{\log(\frac{2|N|n}{\delta})}$, then each method achieves expected pixel label error rate at most~$\delta$.
\end{ftheorem}
The proof is provided by \citet[Section~5.4]{georgehc_thesis} and is a small modification of the proof for the main time series forecasting theorem (Theorem~\ref{thm:time-series-main-result}). The jigsaw condition plays a crucial role in enabling the time series forecasting proof ideas to be imported.

There are different ways to change the separation, such as changing the shape of the patch and including hand-engineered or learned patch features. For example, if the mixture models are all Gaussian mixture models, then provided that no true mean vectors of opposite labels are the same, then as shown in inequality~(3.18) of~\citep{georgehc_thesis} (where time horizon~$T$ corresponds to dimensionality~$d$ here), separation $\sepVar$ turns out to grow as $\Omega(\sigma^2 d)$.  Intuitively, using larger patches $d$ should widen the separation. But using larger patches also means that the (maximum) number of mixture components $\dsmmNumClustersMax$ needed to represent a patch increases, possibly quite dramatically.

As with the main time series forecasting result, to prevent the second terms in the upper bounds in Theorem \ref{thm:image-segmentation-main-result} from scaling linearly with $n$, we could subsample the training data so that $n$ is large enough to capture the diversity of mixture model components yet not so large that we start encountering cases of noise causing patches of one label to look like it is coming from another.  In other words, with estimates or bounds on $\dsmmNumClustersMax$, $\sigma^2$, and $\pi_{\min}$, we can collect $n = \Theta( \frac{1}{\pi_{\min}} \log(|\mathcal{I}|\dsmmNumClustersMax/\delta))$ training image pairs, and if the separation satisfies $\sepVar = \Omega\big( \sigma^2 \log\big( \frac{|N|}{\pi_{\min}\delta} \log\big(\frac{|\mathcal{I}|\dsmmNumClustersMax}{\delta}\big) \big) \big)$, then each algorithm achieves an expected pixel error rate of at most~$\delta$.  This recovers the informal statement of Theorem~\ref{thm:image-segmentation-main-result-informal}.

\subsection{Multipoint Segmentation}
\label{sec:full}

We generalize the basic model for pointwise segmentation to predict label patches $\labelVar[i]$ rather than just a single pixel's label $\labelVar(i)$. Every label patch $\labelVar[i]$ is assumed to have dimensionality $d'$, where $d$ and $d'$ need not be equal. For example, $\obsVar[i]$ could be a 5-by-5 patch, whereas $\labelVar[i]$ could be a 3-by-3 patch. When $d'>1$, estimates of label patches must be merged to arrive at a globally consistent estimate of label image~$\labelVar$.  This case is referred to as \textit{multipoint segmentation}.

In this general case, we assume there to be $\numBigClusters$ underlying latent label images $\Lambda_1, \dots, \Lambda_{\numBigClusters}$ that occur with probabilities $\Pi_1, \dots, \Pi_{\numBigClusters}$. To generate intensity image $\obsVar$, we first sample label image $\Lambda \in \{\Lambda_1, \dots, \Lambda_{\numBigClusters}\}$ according to probabilities $\Pi_1, \dots, \Pi_{\numBigClusters}$. Then we sample label image $\labelVar$ to be a perturbed version of $\Lambda$ such that $ p(\labelVar \mid \Lambda) \propto \exp(-\alpha\mathbf{d}(\labelVar, \Lambda)) $ for some decay factor $\alpha \ge 0$ and differentiable ``distance'' function $\mathbf{d}(\cdot, \cdot)$. For example, $\mathbf{d}(\labelVar, \Lambda)$ could relate to volume overlap between the segmentations represented by label images $\labelVar$ and $\Lambda$ with perfect overlap yielding distance 0.  Finally, intensity image~$\obsVar$ is generated so that for each pixel $i\in\mathcal{I}$, patch $\obsVar[i]$ is a sample from a mixture model patch prior $p(\obsVar[i]|\labelVar[i])$.  If $\alpha=0$, $d'=1$, and the mixture model is diagonal sub-Gaussian, we obtain the earlier model for pointwise segmentation.  We remark that this generative model describes, for every pixel $i$, the joint distribution between intensity image patch $\obsVar[i]$ and the full label image $\labelVar$. As with the pointwise segmentation model, we do not specify how overlapping intensity image patches are jointly distributed.

\citet{georgehc_image_segmentation_miccai} refer to this formulation as a \textit{latent source model for patch-based image segmentation} since the intensity image patches could be thought of as generated from the latent canonical label images $\Lambda_1, \dots, \Lambda_{\numBigClusters}$ combined with the latent mixture model clusters linking $\labelVar[i]$ to $\obsVar[i]$.  This hierarchical structure enables local appearances around a given pixel to be shared across the canonical label images. Put another way, there are two layers of clustering happening, one at the global level, and one at the local patch level.

\subsubsection{An Iterative Algorithm Combining Global and Local Constraints}

We derive an iterative algorithm based on the expected patch log-likelihood (EPLL) framework \citep{epll}.  First, note that the full latent source model that handles multipoint segmentation prescribes a joint distribution for label image $\labelVar$ and image patch $\obsVar[i]$. Thus, assuming that we know the model parameters, the MAP estimate for $\labelVar$ given $\obsVar[i]$ is
\[
\widehat{\labelVar}
=
\underset{\labelVar \in \{0, 1\}^{|\mathcal{I}|}}
         {\text{arg}\,\text{max}}
  \bigg\{
    \log\bigg(
          \sum_{g=1}^\numBigClusters
            \Pi_g
            \exp(-\alpha\mathbf{d}(\labelVar,\Lambda_g))
        \bigg)
    +
    \log p(\obsVar[i]|\labelVar[i])
  \bigg\}.
\]
If we average the objective function above across all pixels $i$, then we obtain the EPLL objective function, which we approximately maximize to segment an image:
\begin{equation*}
\widehat{\labelVar}
=\!\!
\underset{\labelVar \in \{0, 1\}^{|\mathcal{I}|}}
         {\text{arg}\,\text{max}}
  \bigg\{\!
    \log\bigg(
          \sum_{g=1}^{\numBigClusters}
            \Pi_g
            \exp(-\alpha\mathbf{d}(\labelVar,\Lambda_g))
        \bigg)
    +
    \frac{1}{|\mathcal{I}|}
    \sum_{i\in\mathcal{I}}\log p(\obsVar[i]|\labelVar[i])\!
  \bigg\}.
\end{equation*}
The first term in the objective function encourages label image $\labelVar$ to be close to the true label images $\Lambda_1, \dots, \Lambda_{\numBigClusters}$. The second term is the ``expected patch log-likelihood'', which favors solutions whose local label patches agree well on average with the local intensity patches according to the patch priors.

Since latent label images $\Lambda_1, \dots, \Lambda_{\numBigClusters}$ are unknown, we use training label images $\labelVar_1, \dots, \labelVar_n$ as proxies instead, replacing the first term in the objective function with
\[
 F(\labelVar;\alpha)
 \triangleq
 \log
 \Big(
 \frac{1}{n}
 \sum_{u=1}^n
   \exp(-\alpha\mathbf{d}(\labelVar,\labelVar_u))\Big).\]
Next, we approximate the unknown patch prior $p(\obsVar[i]|\labelVar[i])$ with a kernel density estimate
\[
\widetilde{p}(\obsVar[i]|\labelVar[i]; h)
\propto\sum_{u=1}^{n}\sum_{j \in N(i)}
\mathcal{N}\Big(
             \obsVar[i];
             \obsVar_u[j],
             h^2 \mathbf{I}_{d\times d}
             \Big)
             \,
             \ind\{\labelVar[i]=\labelVar_u[j]\},
\]
where the user specifies a neighborhood $N(i)$ of pixel $i$, and Gaussian kernel bandwidth $h>0$.  We group the pixels so that nearby pixels within a small block all share the same kernel density estimate. This approximation essentially assumes a stronger version of the jigsaw condition from Section \ref{sec:myopic} since the algorithm operates as if nearby pixels have the same mixture model as a patch prior.  Hence, we maximize objective $F(\labelVar;\alpha) + \frac{1}{|\mathcal{I}|} \sum_{i \in \mathcal{I}} \log \widetilde{p}(\obsVar[i]|\labelVar[i];h)$ to determine label image~$\labelVar$.

Similar to the original EPLL method~\citep{epll}, we introduce an auxiliary variable $\xi_{i}\in\mathbb{R}^{d'}$ for each patch $\labelVar[i]$, where $\xi_i$ acts as a local estimate for $\labelVar[i]$.  Whereas two patches $\labelVar[i]$ and $\labelVar[j]$ that overlap in label image~$\labelVar$ must be consistent across the overlapping pixels, there is no such requirement on their local estimates $\xi_i$ and $\xi_j$.  In summary, we solve
\begin{equation}
\widehat{\labelVar}
=\!\!\!\!\!\!\underset{\substack{\labelVar\in\{0, 1\}^{|\mathcal{I}|},\\(\xi_{i}\in\mathbb{R}^{d'},i\in\mathcal{I})\\
\text{s.t. }\labelVar[i]=\xi_{i}\text{ for }i\in\mathcal{I}
}
}{\text{arg}\,\text{min }}\!\!\!\!\!\!\!\!
\Big\{-F(\labelVar;\alpha)
-
\frac{1}{|\mathcal{I}|}
\sum_{i\in\mathcal{I}}\log \widetilde{p}(\obsVar[i]|\xi_i;h)+\frac{\beta}{2}\sum_{i\in\mathcal{I}}\|\labelVar[i]-\xi_{i}\|^{2}\Big\}, \label{eq:EPLL-with-aux}
\end{equation}
where $\beta>0$ is a user-specified constant.

The original EPLL method \citep{epll} progressively increases $\beta$ and, for each choice of $\beta$, alternates between updating the label image $\labelVar$ and the auxiliary variables $\xi_i$, ignoring the constraints $\labelVar[i]=\xi_i$ for $i\in\mathcal{I}$. The idea is that as $\beta$ grows large, these constraints will eventually be satisfied.  However, it is unclear how to increase $\beta$ in a principled manner. While heuristics could be used, an alternative approach is to fix $\beta$ and instead introduce a Lagrange multiplier $\eta_i$ for each constraint $\labelVar[i]=\xi_i$ and iteratively update these Lagrange multipliers. This can be achieved by the Alternating Direction Method of Multipliers (ADMM) for distributed optimization \citep{boyd_admm_2011}. Specifically, we form Lagrangian
\begin{align*}
\mathcal{L}_{\beta}(\labelVar,\xi,\eta)
&=-F(\labelVar;\alpha)
-\sum_{i\in\mathcal{I}}\log \widetilde{p}(\obsVar[i]|\xi_{i};h) \\
&\quad
+\frac{\beta}{2}\sum_{i\in\mathcal{I}}\|\labelVar[i]-\xi_{i}\|^{2}+\sum_{i\in\mathcal{I}}\eta_{i}^{T}(\labelVar[i]-\xi_{i}),
\end{align*}
where $\eta=(\eta_{i}\in\mathbb{R}^{d'}, i\in\mathcal{I})$ is the collection of Lagrange multipliers, and $\xi=(\xi_i, i\in\mathcal{I})$ is the collection of auxiliary variables. Indexing iterations with superscripts, the ADMM update equations are given by:
\begin{align*}
\xi^{t+1} & \leftarrow\underset{\xi}{\text{arg}\,\text{min }}\mathcal{L}_{\beta}(\labelVar^{t},\xi,\eta^{t}) &  & \!\!(\text{minimize Lagrangian in direction }\xi),\\
\labelVar^{t+1} & \leftarrow\underset{\labelVar}{\text{arg}\,\text{min }}\mathcal{L}_{\beta}(\labelVar,\xi^{t+1},\eta^{t}) &  & \!\!(\text{minimize Lagrangian in direction }\labelVar),\\
\eta^{t+1} & \leftarrow\eta^{t}+\beta(\xi^{t+1}-\labelVar^{t+1}) &  & \!\!(\text{update Lagrange multipliers }\eta).
\end{align*}
By looking at what terms matter for each update equation, we can rewrite the above three steps as follows:

\begin{enumerate}

\item \textit{Update label patch estimates}.
We update estimate $\xi_i$ for label patch $\labelVar[i]$ given observed image patch $\obsVar[i]$ in parallel across $i\in\mathcal{I}$:
\begin{align*}
\xi_{i}^{t+1}\leftarrow\underset{\xi_{i}}{\text{arg}\,\text{min }}
\Big\{
&
-
\frac{1}{|\mathcal{I}|}
\log \widetilde{p}(\obsVar[i]|\xi_{i};h) \\
&+\frac{\beta}{2}\|\labelVar^{t}[i]-\xi_{i}\|^{2}+(\eta_{i}^{t})^{T}(\labelVar^{t}[i]-\xi_{i})\Big\}.
\end{align*}
This minimization only considers $\xi_i$ among training label patches for which $\widetilde{p}$ is defined. Thus, this minimization effectively scores different nearest neighbor training label patches found and chooses the one with the best score.

\item \textit{Merge label patch estimates}.
Fixing $\xi_i$, we update label image~$\labelVar$:
\begin{align*}
\labelVar^{t+1}\leftarrow\underset{\labelVar}{\text{arg}\,\text{min }}\Big\{
&-F(\labelVar;\alpha)
+\frac{\beta}{2}\sum_{i\in\mathcal{I}}\|\labelVar[i]-\xi_{i}^{t+1}\|^{2}\\
&+\sum_{i\in\mathcal{I}}(\eta_{i}^{t})^{T}(\labelVar[i]-\xi_{i}^{t+1})\Big\}.
\end{align*}
With the assumption that $F$ is differentiable, gradient methods could be used to numerically solve this subproblem.

\item \textit{Update Lagrange multipliers}.
Set $\eta_i^{t+1} \leftarrow\eta_i^{t}+\beta(\xi_i^{t+1}-\labelVar^{t+1}[i]).$ This penalizes large discrepancies between $\xi_i$ and $\labelVar[i]$.

\end{enumerate}

\vspace{.4em}
\noindent
Parameters $\alpha$, $\beta$, and $h$ are chosen using held-out data or cross-validation.

Step 2 above corresponds to merging local patch estimates to form a globally consistent segmentation. This is the only step that involves expression $F(\labelVar;\alpha)$.  With $\alpha=0$ and forcing the Lagrange multipliers to always be zero, the merging becomes a simple averaging of overlapping label patch estimates $\xi_i$.  This algorithm corresponds to existing multipoint patch-based segmentation algorithms \citep{coupe_2011,rousseau_2011,patch_descriptors} and the in-painting technique achieved by the original EPLL method.  Setting $\alpha=\beta=0$ and $d'=1$ yields pointwise kernel segmentation.  When $\alpha > 0$, a global correction is applied, shifting the label image estimate closer to the training label images.  This should produce better estimates when the full training label images can, with small perturbations as measured by $\mathbf{d}(\cdot, \cdot)$, explain new intensity images.

\subsection{Experimental Results}
\label{sec:image-segmentation-experimental-results}

\citet{georgehc_image_segmentation_miccai} empirically explore the above iterative algorithm on 20 labeled thoracic-abdominal contrast-enhanced CT scans from the Visceral \textsc{anatomy3} dataset \citep{anatomy3}. The model is trained on 15 scans and tested on the remaining 5 scans. The training procedure amounted to using 10 of the 15 training scans to sweep over algorithm parameters, and the rest of the training scans to evaluate parameter settings.  Finally, the entire training dataset of 15 scans is used to segment the test dataset of 5 scans using the best parameters found during training. For each test scan, a fast affine registration is first run to roughly align each training scan to the test scan. Then four different algorithms are applied: a baseline majority voting algorithm (denoted ``MV'') that simply averages the training label images that are now roughly aligned to the test scan, pointwise 1-NN (denoted ``1NN'') and kernel (denoted ``WMV'' abbreviating weighted majority voting) segmentation that both use approximate nearest patches, and finally the proposed iterative algorithm (denoted ``ADMM''), setting distance $\mathbf{d}$ to one minus Dice overlap. Dice overlap measures volume overlap between the true and estimated pixels of an object, where 1 is perfect overlap and 0 is no overlap.  It can be written as a differentiable function by relaxing the optimization to allow each label to take on a value in $[-1,1]$, in which case the Dice overlap of label images $\labelVar$ and $\Lambda$ is given by $2\langle \widetilde{\labelVar}, \widetilde{\Lambda}\rangle/ (\langle \widetilde{\labelVar}, \widetilde{\labelVar} \rangle + \langle \widetilde{\Lambda}, \widetilde{\Lambda} \rangle)$, where $\widetilde{\labelVar} = (\labelVar+1)/2$ and $\widetilde{\Lambda} = (\Lambda+1)/2$.

\begin{figure}
\small
\centering
\begin{tabular}{|c|c|c|c|c|}
\cline{2-5} 
\multicolumn{1}{c|}{} & MV & 1NN & WMV & ADMM\tabularnewline
\hline 
Liver
& \raisebox{-.45\height}{\includegraphics[width=1.9cm]{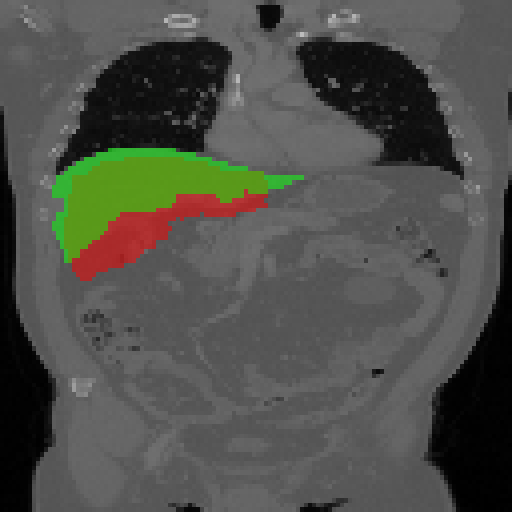}}
& \raisebox{-.45\height}{\includegraphics[width=1.9cm]{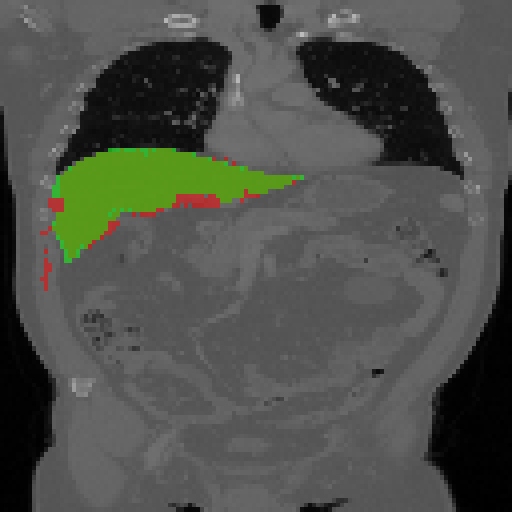}}
& \raisebox{-.45\height}{\includegraphics[width=1.9cm]{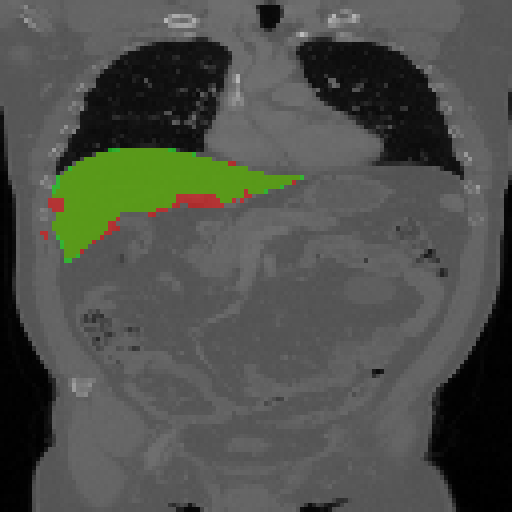}}
& \raisebox{-.45\height}{\includegraphics[width=1.9cm]{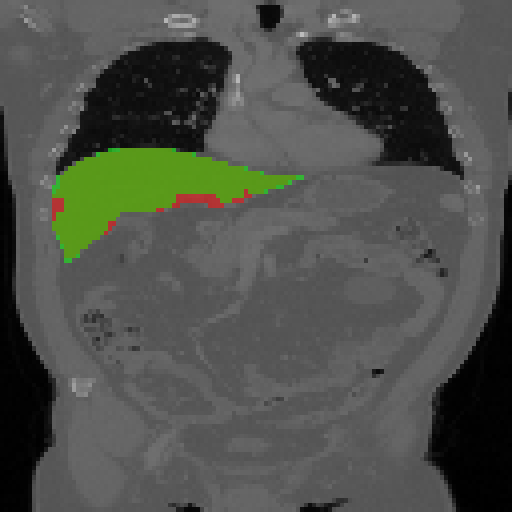}}
\tabularnewline
\hline 
Spleen
& \raisebox{-.45\height}{\includegraphics[width=1.9cm]{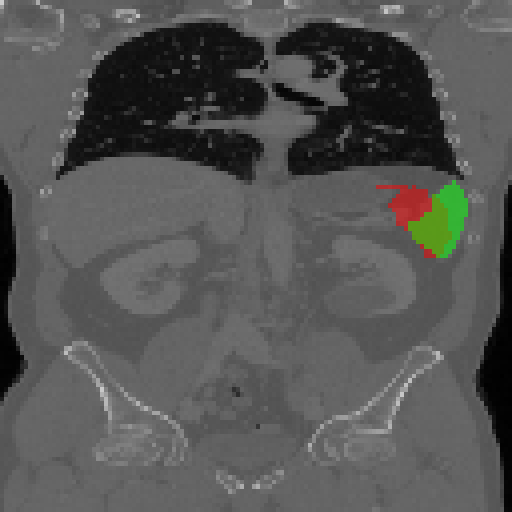}}
& \raisebox{-.45\height}{\includegraphics[width=1.9cm]{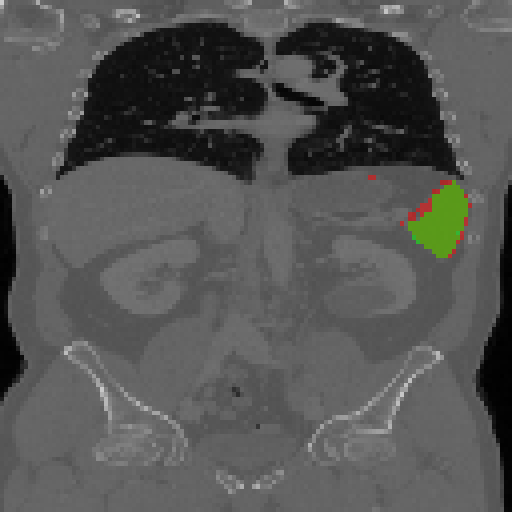}}
& \raisebox{-.45\height}{\includegraphics[width=1.9cm]{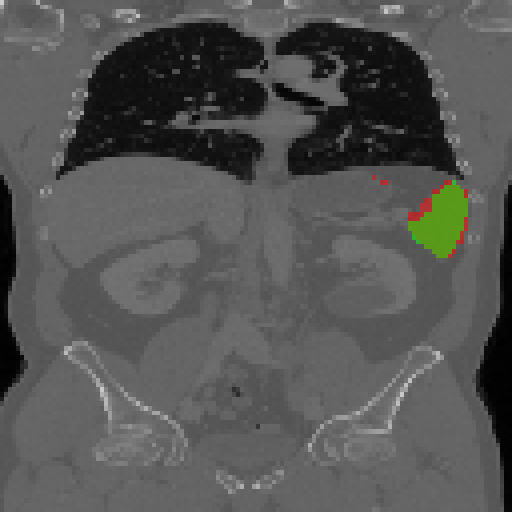}}
& \raisebox{-.45\height}{\includegraphics[width=1.9cm]{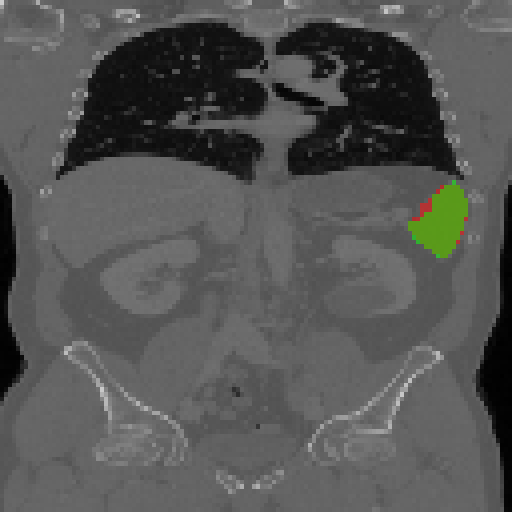}}
\tabularnewline
\hline 
Left kidney
& \raisebox{-.45\height}{\includegraphics[width=1.9cm]{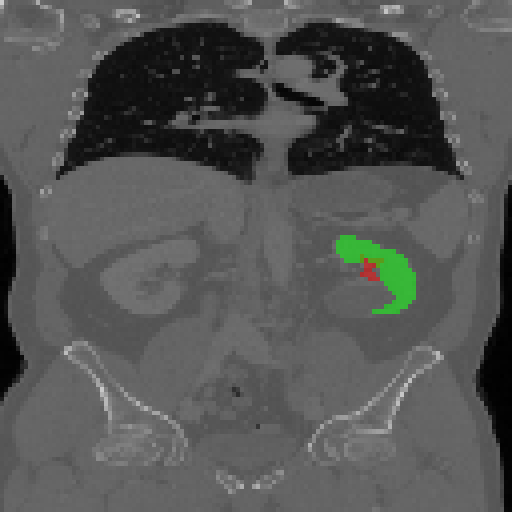}}
& \raisebox{-.45\height}{\includegraphics[width=1.9cm]{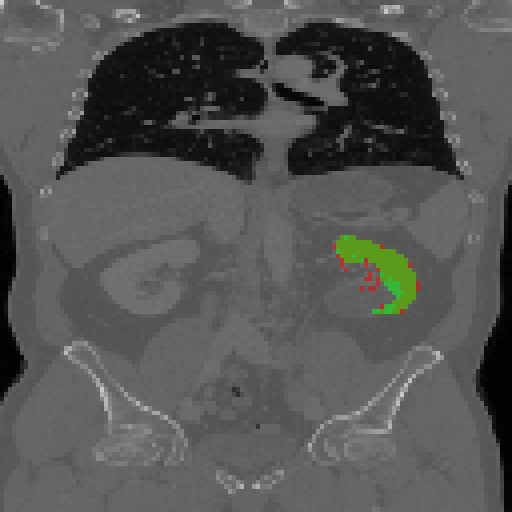}}
& \raisebox{-.45\height}{\includegraphics[width=1.9cm]{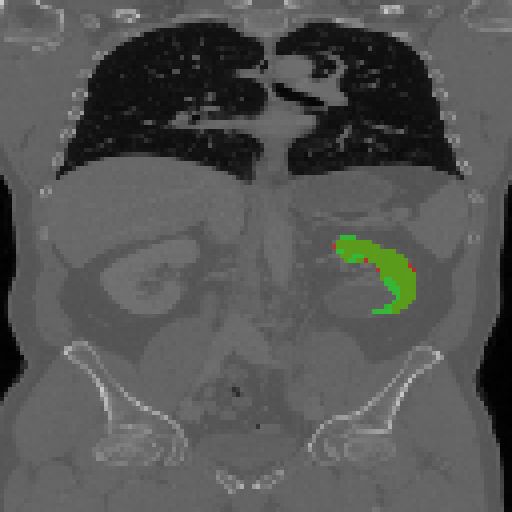}}
& \raisebox{-.45\height}{\includegraphics[width=1.9cm]{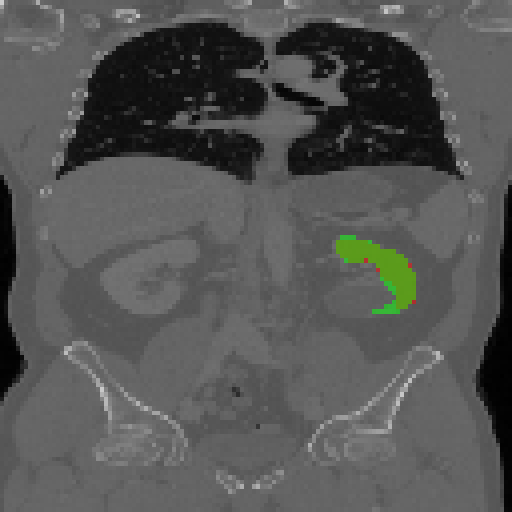}}
\tabularnewline
\hline 
Right kidney
& \raisebox{-.45\height}{\includegraphics[width=1.9cm]{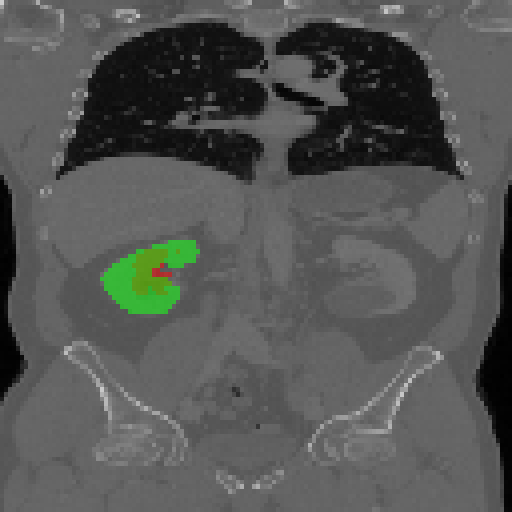}}
& \raisebox{-.45\height}{\includegraphics[width=1.9cm]{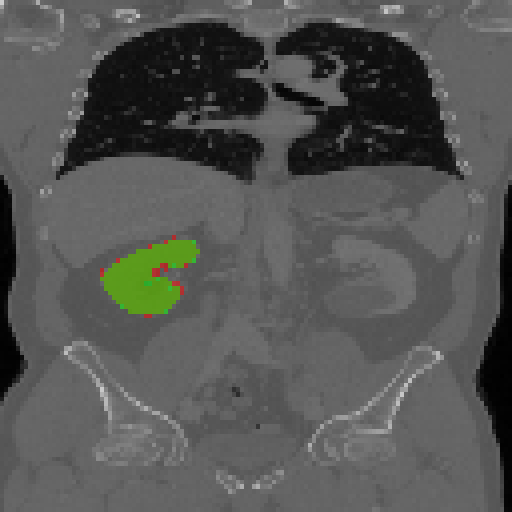}}
& \raisebox{-.45\height}{\includegraphics[width=1.9cm]{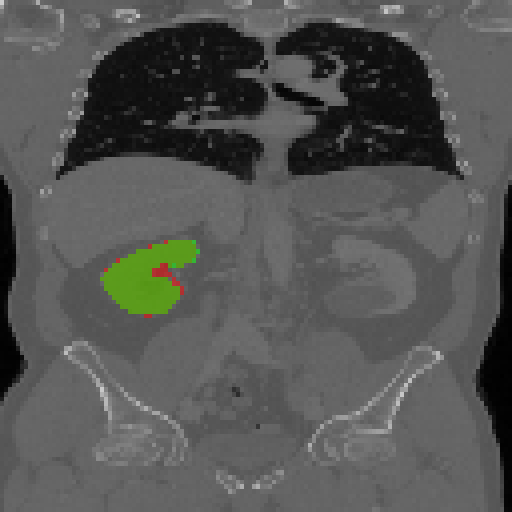}}
& \raisebox{-.45\height}{\includegraphics[width=1.9cm]{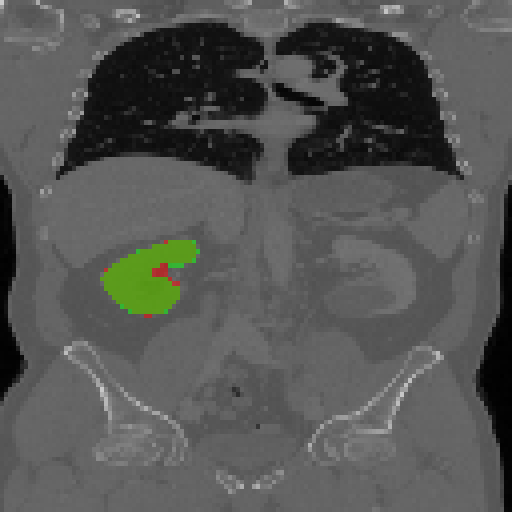}}
\tabularnewline
\hline 
\end{tabular}
\caption[Example image segmentation results]{Example image segmentation results (figure source: \citealt[Figure~5.1]{georgehc_thesis}). Green denotes the ground truth label and red denotes the estimated label, where a good segmentation result has the green and red regions perfectly overlap.}
\label{fig:example-segs}
\end{figure}

\begin{figure}
\captionsetup[subfloat]{captionskip=-0.3em}
\centering
\subfloat[][Liver]{
\includegraphics[height=3.8cm]{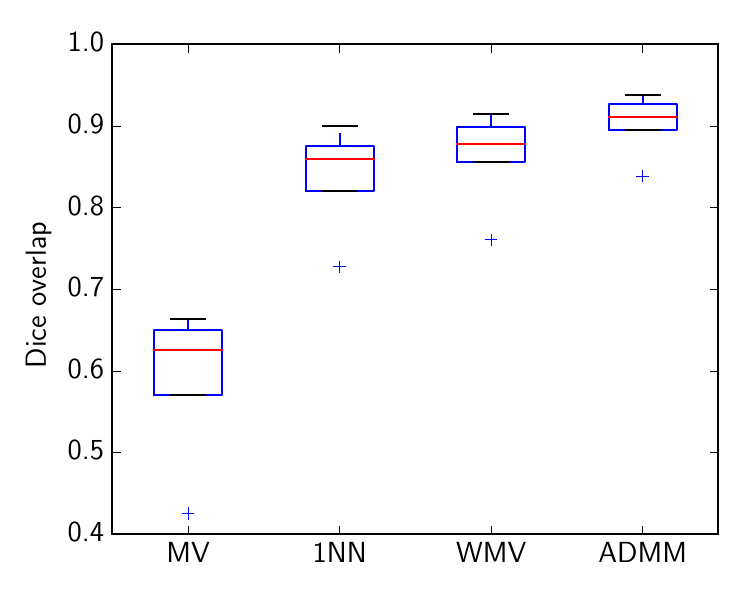}
}
\subfloat[][Spleen]{
\includegraphics[height=3.8cm]{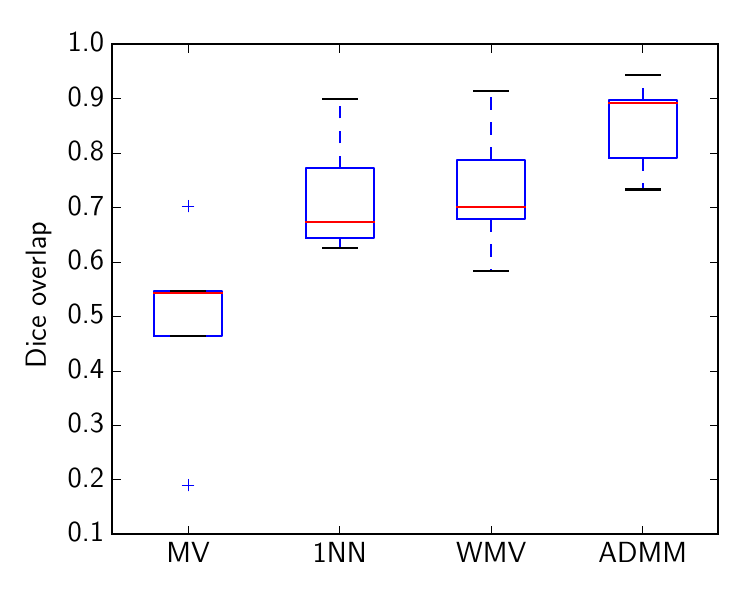}
}
\vspace{-1.1em}
\\
\subfloat[][Left kidney]{
\includegraphics[height=3.8cm]{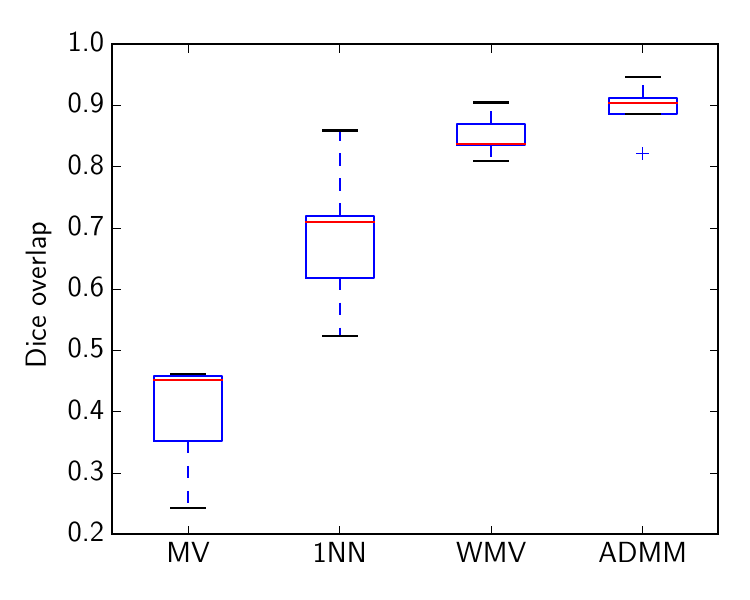}
}
\subfloat[][Right kidney]{
\includegraphics[height=3.8cm]{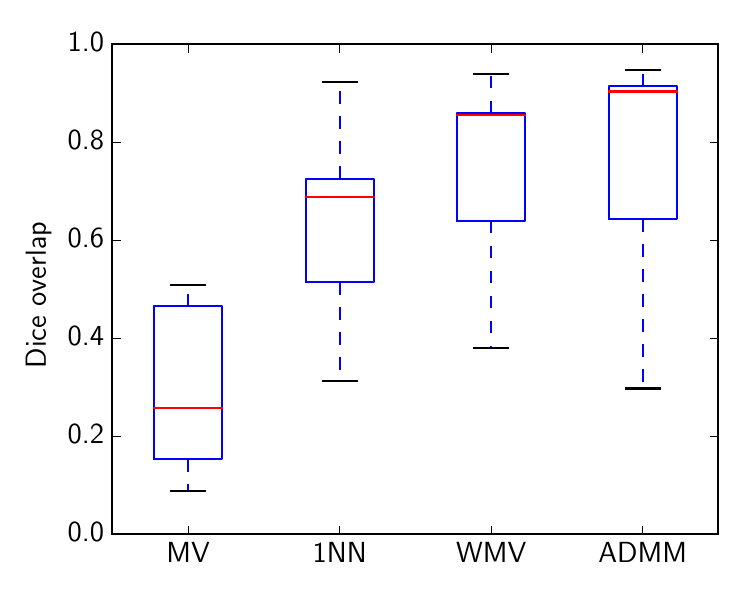}
}
\vspace{-0.5em}
\caption{Dice volume overlap scores (figure source:  \citealt[Figure~5.7]{georgehc_thesis}).}
\label{fig:dice}
\end{figure}

The liver, spleen, left kidney, and right kidney were segmented. Example segmentations are in Figure~\ref{fig:example-segs}, and Dice overlap scores are in Figure~\ref{fig:dice} using the four algorithms. In all cases, the proposed iterative algorithm outperforms pointwise kernel segmentation, which outperforms both pointwise 1-NN segmentation and the baseline majority voting.  For the organs segmented, there was little benefit to having $\alpha>0$, suggesting the local patch estimates to already be quite consistent and require no global correction.

\section{More Training Data, More Problems?}
\label{sec:case-studies-conclusion}

A limitation across the three main theoretical results of this chapter is that they become weak when the number of training data $n$ grows too large. The workaround each time was to subsample~$n$ to be smaller but in a way that depended on the number of clusters $\numClusters$.  Is this really necessary?  Is the worst-case analysis too pessimistic?  The main reason why this behavior appears is that the analysis techniques used to establish the guarantees in this chapter effectively do a 1-NN analysis. These techniques do not easily extend to looking at multiple nearest neighbors at once. Instead, when looking at $k>1$ nearest neighbors, the analysis would assume that the nearest neighbor found is potentially good, but that the 2nd, 3rd, $\dots$, $k$-th nearest neighbors are generated from the cluster with the wrong label!  In fact, the analysis in the time series classification case assumes that the $n-1$ training examples that are not the nearest neighbor come from a cluster with the wrong label \citep[Section~3.7.1]{georgehc_thesis}.

One way to see why larger $n$ can cause 1-NN classification to struggle is to consider a simple setup where data are generated i.i.d.~with equal probability from one of two Gaussians, either $\mathcal{N}(0, \sigma^2)$ or $\mathcal{N}(\mu, \sigma^2)$ for constants $\mu > 0,\sigma > 0$. The goal is to predict which of these two Gaussians a new data point is from with the help of training data $\obsVar_1, \dots, \obsVar_n$ drawn i.i.d.~from the same model with known labels $\labelVar_1, \dots, \labelVar_n \in \{0, 1\}$, where 1 corresponds to $\mathcal{N}(0, \sigma^2)$, and 0 corresponds to $\mathcal{N}(\mu, \sigma^2)$. This is the time series forecasting setting with $T=1$, no time shifts, and Gaussian noise. When $n$ grows large, with high probability, we will encounter training data generated from $\mathcal{N}(0, \sigma^2)$ that exceed $\mu$ and thus plausibly appear to come from the second Gaussian, and vice versa! This is disastrous as it means that with large amounts of training data, the separation $\sepVar^{(T)}$ could become vanishingly small.

To sketch why this mishap happens, first note that as per Lemma~2.2.2 of \citet{georgehc_thesis}, by collecting $n \ge 16\log(2/\delta)$ training data, then with probability at least $1 - \delta$, we have more than $n/4$ samples from each of the two Gaussians.  Assuming this event holds, we next rely on the fact that for random variables $X_1, \dots, X_d$ drawn i.i.d.~from $\mathcal{N}(0, \sigma^2)$ \citep{g_expectation_max_of_iid_gaussians},
\[
\sigma \sqrt{\frac{\log d}{\pi \log 2}}
\le
  \mathbb{E}\big[\max_{u=1,\dots,d} X_u\big]
\le
  \sigma \sqrt{2 \log d}.
\]
Hence, we know that the maximum of the training data generated from the first Gaussian $\mathcal{N}(0, \sigma^2)$ (for which there are more than $n/4$ such training data) has an expected value of at least $\sigma \sqrt{\frac{\log(n/4)}{\pi \log 2}}$, which for large enough $n$ exceeds $\mu$. One could then apply a concentration inequality to argue that as $n$ grows large, with high probability, there will be at least one training data point generated from $\mathcal{N}(0, \sigma^2)$ that is larger than $\mu$, and thus is more plausibly explained as being generated from the second Gaussian. A similar calculation could be used with the two Gaussians swapped. Increasing $n$ makes it more likely that more of these ``bad'' training data appear.

The key observation that provides a promising solution to the disaster scenario above is that the ``bad'' training data that stray far from their corresponding clusters are \textit{outliers}.  We could thus screen for and remove outliers and still have 1-NN classification work.  For example, one simple nonparametric way to do this is for a training data point to look around it to find its nearest neighbors (\eg, up to a certain number of them or within a ball of pre-specified radius), and then to ask whether its label agrees with the most popular label among its nearest neighbors; if not, we classify the point as an outlier and discard it \citep{wilson_1972}.

Making nearest neighbor methods robust to outliers or adversarial training examples is an active area of research (\eg, \citealt{sanchez_2003, smith_2011}).  \citet{wang_2017} recently defined robustness measures and proved that 1-NN classification is not robust. They modify 1-NN classification so that it is robust, resulting in a method called \textsc{robustNN}. We suspect that incorporating this sort of modification of 1-NN classification into the algorithms presented in this chapter should improve their theoretical performance guarantees. The training data would still be getting subsampled as outliers or adversarial examples are removed but the subsampling is targeted rather than completely random.

\chapter{Computation}
\label{chap:computation}

\newcommand{\randomVectorVar}{v}

Thus far, we have discussed nearest neighbor prediction methods and their statistical guarantees for a variety of problem setups. This chapter takes a look under the hood at the main workhorse underlying all of these prediction methods: nearest neighbor search. As this subroutine is executed repeatedly, often in parallel, the computational advances for nearest neighbor search have been vital to the popularity and success of nearest neighbor prediction. This chapter provides an overview of these advances, keeping the discussion high level and highlighting the recurring motifs. Thus, unlike other chapters in this monograph, we largely refrain from delving into specifics of any of the algorithms, deferring to references provided.  We also provide links to a variety of open-source nearest neighbor search software packages at the end of this chapter.

\section{Overview}

We focus on discussing $k$-NN search. Following our usual notation and setup, we assume we have access to prior observations or training data $(X_1, Y_1), (X_2, Y_2), \dots, (X_n, Y_n) \in \mathcal{X} \times \mathbb{R}$. The feature space $\mathcal{X}$ is endowed with a metric $\rho: \mathcal{X} \times \mathcal{X} \to \mathbb{R}_+$. Let $\PP_n$ denote the set of training feature vectors $\{X_1,\dots, X_n\}$.  For a data point $x \in \mathcal{X}$, our goal is to find the $k \in \{1,\dots,n\}$ nearest neighbors of $x$ within the $n$ training feature vectors~$\PP_n$. We denote $(X_{(i)}(x),Y_{(i)}(x))$ to be the $i$-th closest training data point among the training data $(X_{1},Y_{1}),\dots,(X_{n},Y_{n})$. Thus, the distance of each training data point to $x$ satisfies
\[
\rho(x,X_{(1)}(x))\le\rho(x,X_{(2)}(x))\le\cdots\le\rho(x,X_{(n)}(x)).
\]
As discussed earlier, ties are broken at random. 

A naive algorithm would compute distances of $x$ to all $n$ training data, sort the distances, and choose the $k$ smallest distances. Using this naive approach, a single search query has running time $\mathcal{O}(n\log n)$ if we use an efficient comparison sort such as merge sort.  A clever implementation that does not care about the ordering of smallest $k$ elements but simply finds them, \eg, using the Quickselect algorithm \citep{quickselect}, can find the $k$ nearest neighbors with effectively $\mathcal{O}(n)$ comparisons (on average). With a large number of training data $n$, as is common in modern applications (\eg, social networks, online recommendation systems, electronic health records), having each nearest neighbor search query take $\mathcal{O}(n)$ time can be prohibitively expensive, especially when we have to repeatedly execute these queries. Furthermore, modern applications regularly involve ever-growing (and sporadically shrinking) training data.

With the above discussion in mind, ideally we would like nearest neighbor data structures with the following properties:
\begin{itemize}

\item[1.] {\em Fast.} The cost of finding $k$ nearest neighbors (for constant $k$) should be sublinear in $n$, \ie, $o(n)$; the smaller the better.

\item[2.] {\em Low storage overhead.} The storage required for the data structure should be subquadratic in $n$, \ie, $o(n^2)$; the smaller the better.

\item[3.] {\em Low pre-processing.} The cost of pre-processing data to build the data structure should not require computing all pairwise distances and should thus be $o(n^2)$; the smaller the better.

\item[4.] {\em Incremental insertions.} It should be possible to add data incrementally to the data structure with insertion running time $o(n)$.

\item[5.] {\em Generic distances and spaces.} The data structure should be able to handle all forms of distances $\rho$ and all forms of spaces $\mathcal{X}$. 

\item[6.] {\em Incremental deletions.} The data structure should allow removal of data points from it with deletion running time $o(n)$.

\end{itemize}
In this chapter, we discuss various data structures that have been developed over more than five decades to precisely address the above desiderata. To date, there is {\em no} known data structure that can address {\em all} of the above simultaneously. However, substantial progress has been made to provide solutions that satisfy many of properties listed above, both in theory and in practice. In particular, many theoretically sound approaches have been developed for a variety---but not all---of distances primarily over Euclidean space (\ie, $\XX \subseteq {\mathbb R}^d$) that provide efficient running time, efficient storage, and low pre-processing time.  Meanwhile, many nearest neighbor data structures have been proposed that work well in practice and support all but the {\em incremental deletion} property, but they currently lack theoretical guarantees.

As the number of nearest neighbors developed over many decades is vast, we will not be able to do justice to covering all approaches in any detail. Instead, we focus on key representative approaches. We start in Section~\ref{sec:comp-exact-NN} by describing one of the earliest data structures that supports finding \textit{exact} nearest neighbors, the {\em k-d tree} (``k-dimensional'' tree) data structure \citep{kdtree}, which has remained a guiding principle for developing \textit{approximate} nearest neighbor search data structures in recent years.  A k-d tree operates in a Euclidean feature space $\XX \subseteq {\mathbb R}^d$ with Euclidean distance as the metric $\rho$.  It has search query time $\mathcal{O}(d 2^{\mathcal{O}(d)} + \log n)$, $\mathcal{O}(n)$ storage cost, and $\mathcal{O}(n \log n)$ pre-processing time---all of which are efficient provided that the number of dimensions~$d$ is sufficiently small, namely $d 2^{\mathcal{O}(d)} \ll n$. In general, it suffers from an exponential dependence on dimension $d$, also known as the {\em curse of dimensionality}.\footnote{Here, we talk about curse of dimensionality from a viewpoint of computational efficiency. The phrase ``curse of dimensionality'' also more generally encompasses other phenomena that arise when data reside in high-dimensional spaces. For example, in high-dimensional feature spaces, more training data may be required to see enough combinations of different feature values appearing.}

Considerable progress has been made on scaling exact nearest neighbor search to high dimensions such as with the cover tree data structure \citep{beygelzimer_2006}, which can remove the exponential dependence on $d$ completely.  Cover trees exploit the fact that even though the data might be high-dimensional, they often have low-dimensional structure and thus have low so-called ``intrinsic'' dimension.  The running time dependence on intrinsic dimension is nontrivial, however.  To date, exact nearest neighbor search can still be prohibitively expensive in practice.

Next, we turn to approximate nearest neighbor search in Section~\ref{sec:comp-approx-NN}. We begin with {\em locality-sensitive hashing}  (LSH) for designing approximate nearest neighbor data structures, pioneered by \citet{IndykMotwani}. LSH has been the primary approach in literature for systematically developing approximate nearest neighbor data structures with theoretical guarantees. The known results provide efficient query time,  storage, and pre-processing time for a Euclidean feature space $\XX \subseteq {\mathbb R}^d$ with various metrics including but not limited to $\ell_2$ (Euclidean) or more generally $\ell_p, p \in (1,2]$, Hamming, Jaccard, and cosine distance.

Finally, we turn to approaches that are different from LSH that have partial or no theoretical guarantees but have been successful in practice. Specifically, we describe the approach of random projection or partition trees \citep{randompartition1, randompartition2, randompartition3} and boundary trees \citep{mathy_2015}. Random projection trees can be viewed as a randomized variant of k-d trees.  It is suitable for a Euclidean feature space $\XX \subseteq {\mathbb R}^d$ and a Euclidean metric. Theoretical analysis shows that a single random projection or partition tree can find exact nearest neighbors with nontrivial probability if the underlying data has low intrinsic dimension  \citep{dasgupta_2015}.  However, in practice these trees are used for approximate nearest neighbor search. Extending the theory to this approximate case as well as to strategies that combine the results of multiple trees remains an open problem.  We also present the recently proposed boundary trees and forests that handle all of the desiderata stated above except for {\em incremental deletions}.  They work very well in practice but currently lack any sort of theoretical guarantees.

We conclude this chapter by pointing out some open source nearest neighbor search software packages in Section~\ref{sec:NN-open-source}.

\section{Exact Nearest Neighbors}
\label{sec:comp-exact-NN}

We consider the question of finding the exact nearest neighbor (so, \mbox{1-NN}) for any given query point $x$ with respect to set of $n$ points $\PP_n$ using a k-d tree \citep{kdtree}. The data structure works with a Euclidean feature space $\XX = {\mathbb R}^d$ for any number of dimensions $d \geq 1$ and the Euclidean distance as the metric $\rho$.

\begin{figure}
\centering
\subfloat[][]{
\includegraphics[scale=.66]{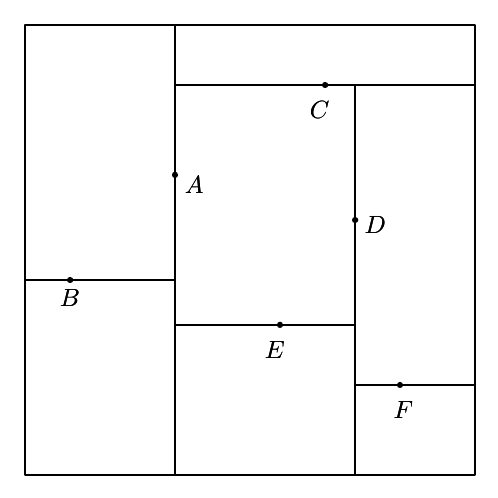}}
\subfloat[][]{
\includegraphics[scale=.66]{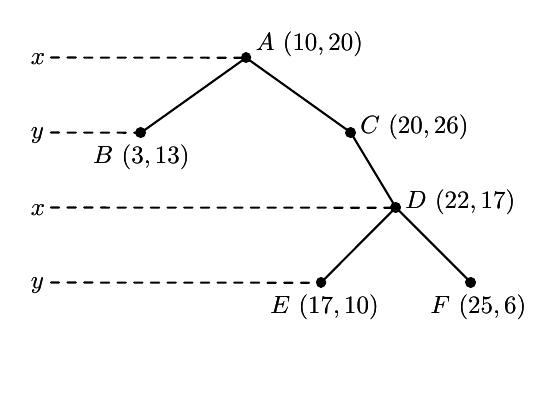}}
\caption{Example of a k-d tree in two dimensional space, where panel (a) shows the original feature space, and (b) shows the tree with the split points (labeled A through F). Each split is shown by a line, vertical for nodes with split along x axis, and horizontal for nodes with split along y axis. The root note splits the space into two parts; its children further divide the space into smaller parts.
\label{fig:kd-tree}}
\end{figure}

The k-d tree corresponding to $n$ points, $\PP_n$, is a binary tree-structured partition of ${\mathbb R}^d$ that depends on the data.  Each partition corresponds to hyper-rectangular cell based on the $n$ data points. The tree structure is constructed as follows. The root corresponds to the entire space. The two children of this root are created by first choosing one of the $d$ coordinates and then partitioning the space along this coordinate at the median value of this coordinate across all $n$ points. This way, the two partitions corresponding to left and right children have effectively half of the points. This procedure recurses within each partition, rotating through the $d$ coordinates along the depth of the tree to decide which dimension to split data along, until the number of points in a ``leaf'' partition is at most $n_0$, where $n_0$ is some pre-determined constant. Since the median of $n$ numbers can be found in $\mathcal{O}(n)$ time and the depth of the tree is $\mathcal{O}(\log n)$ (due to balanced partitioning at each stage), the cost of constructing k-d tree is $\mathcal{O}(n \log n)$.  An example k-d tree in $d=2$ dimensions is shown in Figure~\ref{fig:kd-tree}.

Now let's consider how to find the nearest neighbor for a given query point $x \in {\mathbb R}^d$ in $\PP_n$ using the k-d tree that is built from the $n$ data points $\PP_n$. To do this, we simply follow the path along the tree starting from the root to reach a leaf (partition) by determining at each stage which of the two branches one should take based on the coordinate value of $x$ and the median value along that coordinate that determined the partition in the k-d tree at that stage. Upon reaching the leaf partition, we do a brute-force search to find the nearest neighbor of $x$ among the points that belong to the leaf partition. As each leaf partition has at most $n_0$ points, this takes $\mathcal{O}(d n_0)$ time.  Now to make sure that this is indeed the nearest neighbor of $x$, we need to search for the nearest neighbor through the neighboring leaf partitions of the leaf partition to which $x$ belongs. In $d$ dimensional space, the number of such neighboring partitions can be $2^{\mathcal{O}(d)}$. Therefore, the effective query time scales as $\mathcal{O}(d 2^{\mathcal{O}(d)} n_0)$.  When  $d$ is small ($d 2^{\mathcal{O}(d)} = o(n)$), this results in an efficient algorithm ($o(n)$ query time).  However, the query cost grows exponentially in $d$, which can become prohibitively expensive even when $d$ is moderately large.

To escape this exponential dependence on $d$, more recent exact nearest neighbor search algorithms take advantage of the observation that in practice, data often have low intrinsic dimension (\eg, \citealt{DBD,beygelzimer_2006}). For example, the cover tree data structure by \citet{beygelzimer_2006} has query time $\mathcal{O}(\log n)$ and space usage $\mathcal{O}(n)$, while even supporting point insertions and deletions that each take time $\mathcal{O}(\log n)$---these depend on the intrinsic dimension rather than the original feature space dimension~$d$.  While these time and space complexities look quite attractive, they hide the fact that the dependence on intrinsic dimension is nontrival.  The data points currently in the cover tree data structure have an associated expansion constant $c$. Querying for nearest neighbors scales as $c^{12}$, which could be prohibitively large.  To date, while exact nearest neighbor search algorithms have made considerable progress since k-d trees, they can still be too slow for a variety of applications. Switching over to some of the latest approximate nearest neighbor search algorithms can lead to dramatic speedups.

\section{Approximate Nearest Neighbors}
\label{sec:comp-approx-NN}

We now present a few prominent approaches for approximate nearest neighbor search, some with strong theoretical guarantees and some  that are simply heuristics at this point.

\subsection{Locality-Sensitive Hashing}

The first approach we discuss is locality-sensitive hashing (LSH), which appears in numerous approximate nearest neighbor methods, especially ones with theoretical guarantees. LSH was first introduced by \citet{IndykMotwani} and works for a variety of feature spaces and metrics.  As the name suggests, LSH is about hash functions for which items that are similar produce similar hashes. The hashes themselves are short and can be easily compared. Thus, an approximate nearest neighbor search can be done by first hashing items and then comparing hashes.  As an aside, very recently evidence has emerged that fruit fly brains use a variant of LSH to find similar odors \citep{lsh_bio_evidence_2017}!

For ease of exposition, we only discuss a basic version of LSH for approximate nearest neighbor search where the feature space is Euclidean $\XX = {\mathbb R}^d$ and the metric $\rho$ is Euclidean distance.  For more detailed descriptions as well as recent state-of-the-art developments, we refer the reader to surveys by \citet{CACM} and \citet{HarPeledIndykMotwani}, and to recent work by \citet{andoni2015optimal} and \citet{andoni2015practical}.

\medskip
\noindent{\bf Key concepts.}
To begin with, we need to be precise about exactly what we mean by an ``approximate'' nearest neighbor. Two related commonly used definitions are below.
\begin{fdefinition}[$c$-approximate nearest neighbor]
Given $c > 1$, we call a point $x^\prime \in \PP_n = \{X_1,\dots, X_n\}$ a $c$-approximate nearest neighbor for a query point $x$ with respect to $\PP_n$ if $\rho(x, x^\prime) \leq c\rho(x, X_{(1)}(x))$. A randomized $c$-approximate nearest neighbor data structure outputs a $c$-approximate nearest neighbor with probability at least $1-\delta$ for any query point, for given parameter $\delta > 0$.
\end{fdefinition}
Next we define a related notion of a \textit{near} neighbor, which will be what LSH helps us find. At first, it seems weaker than the $c$-approximate nearest neighbor, but as we discuss shortly, the two notions of approximation are effectively equivalent.
\begin{fdefinition}[$c, R$-approximate near neighbor]
Given parameters $c > 1$, $R > 0$ and $\delta \in (0,1)$, a randomized $c, R$-approximate near neighbor data structure for an $n$ point set $\PP_n = \{X_1,\dots, X_n\}$ is such that for any query point $x$ it returns $x^\prime$ such that $\rho(x, x^\prime) \leq cR$ with probability $1-\delta$, if $\rho(x, X_{(1)}(x)) \leq R$.
\end{fdefinition}
While the $c, R$-approximate near neighbor definition seems weaker than that of the $c$-approximate nearest neighbor, a clever reduction of $c$-approximate nearest neighbor to $c, R$-approximate near neighbor shows that by incurring a factor of $\mathcal{O}(\log^2 n)$ in space and $\mathcal{O}(\log n)$ in query time, an approximate {\em near} neighbor data structure can help in finding approximate {\em nearest} neighbors \citep{HarPeledIndykMotwani}.

The basic algorithmic idea for this reduction is that it suffices to build $\mathcal{O}(\log n)$ near neighbor data structures. Then for a query point~$x$, we find a near neighbor for $x$ using each of the $\mathcal{O}(\log n)$ near neighbor data structures and then pick the closest one found.  First suppose that the  minimum distance of any point in $\PP_n$ to query point $x$ is at least $R_0 > 0$ and no more than ${\sf poly}(n) R_0$.  Now for values of $R$ that are of form $(1 + \gamma)^k R_0/c$ with $0\leq k \leq \mathcal{O}(\log n)$ for some $\gamma > 0$, we build $\mathcal{O}(\log n)$ different $c, R$-approximate near neighbor data structures. Using this, one can find a $c (1 + \gamma)$-approximate nearest neighbor for any query point by querying all of these $\mathcal{O}(\log n)$ data structure simultaneously and choosing the closest among the returned responses. The argument that considering such a setting is {\em sufficient} is clever and involved; we refer the interested reader to the paper by \citet{HarPeledIndykMotwani}. 

\medskip
\noindent{\bf Approximate near neighbor using LSH.}
The reduction of approximate nearest neighbor to approximate near neighbor suggests that it is sufficient to construct randomized approximate near neighbor data structures. A general recipe for doing so is given by LSH. Consider a family of hash functions, say $\HH$. Any hash function in $\HH$ takes as input a feature vector in the feature space $\mathcal{X}$ and outputs a hash in a space we call $U$.  In what follows, the randomness is in choosing a hash function uniformly at random from hash family $\HH$. What we want is that for two feature vectors that are close by, the probability that they produce the same hash value is sufficiently high, whereas if the two feature vectors are not close by, the probability that the produce the same hash value is sufficiently low.
\begin{fdefinition}[$(c, R, P_1, P_2)$-locality-sensitive hash]
A distribution over a family of hash functions $\HH$ is called $(c, R, P_1, P_2)$-locality-sensitive if for any pair of $x_1, x_2 \in \XX$ and randomly chosen $h$, 
\begin{itemize}

\item[$\circ$] if $\rho(x_1, x_2) \leq R$ then $\P(h(x_1) = h(x_2)) \geq P_1$, 

\item[$\circ$] if $\rho(x_1, x_2) \geq cR$ then $\P(h(x_1) = h(x_2)) \leq P_2$.

\end{itemize} 
\end{fdefinition}
Before proceeding, let us consider a simple example.
\begin{fexample}[LSH for Hamming distance.]\label{example:lsh}
Suppose $\XX = \{0,1\}^d$ and $\rho$ is Hamming distance, \ie, for $x, x^\prime \in \XX$
\begin{align*}
\rho(x, x^\prime) & = \sum_{i=1}^d \ind\{x_i \neq x^\prime_i\}.
\end{align*}
Consider a family of $d$ hash function $\HH = \{h_1,\dots, h_d\}$, where for any $i \in \{1, \dots, d\}$, we have $h_i: \XX \to \{0,1\}$ with 
\begin{align*}
h_i(x) & = x_i, 
\end{align*}
for any $x \in \XX$. Thus, a uniformly random chosen hash function, when applied to a point in $\XX$, effectively produces a random coordinate of the point. Given this, it can be easily checked that for any $x_1, x_2 \in \XX$, for a uniformly at random chosen hash function $h$ from $\HH$, 
\begin{align*}
\P(h(x_1) = h(x_2)) & = 1 - \frac{\rho(x_1, x_2)}{d}.
\end{align*}
Therefore, we can conclude that this family of hash functions $\HH$ is an LSH with $P_1 = 1-\frac{R}{d}$ and $P_2 = 1- \frac{c R}{d}$.  In particular, as long as $c > 1$ and $R \in [0, \frac{d}{c}]$, we have that $P_1 > P_2$.
\end{fexample}
Clearly, our interest will be in LSH where $P_1 > P_2$ just as in the above illustration. Ideally, we would like to have $P_1 \approx 1$ and $P_2 \approx 0$ so that any hash function $h$ from the family $\HH$ can separate relevant pairs of points (\ie, distance $\leq R$) from irrelevant ones (\ie, distance $\geq cR$). In practice, we might have values of $P_1$ and $P_2$ that are away from $1$ and $0$ respectively. Therefore, we need an ``amplification'' process that can convert a ``weak'' LSH into a ``strong'' LSH. But having a strong LSH alone is not sufficient. We also need a clever way to store an ``index'' using this LSH family $\HH$ so that we can query the data structure efficiently to solve the near neighbor search problem for any query point. 

\begin{figure}
\centering
\includegraphics[scale=.8]{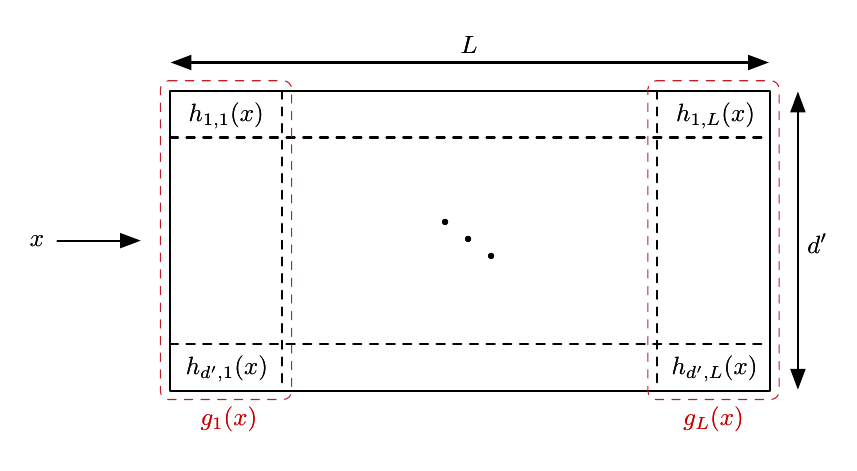}
\vspace{-1em}
\caption{A generic recipe to build a near neighbor data structure using weak LSH family. Here, each point $x$ is mapped to $d' \times L$ matrix with $(i, j)$-th entry of the matrix denoted $h_{i,j}(x)$. Here, $h_{i,j}$'s are drawn from a given LSH family for $i \in \{1,\dots,d'\}$, $j\in\{1,\dots,L\}$. Each column (red) of the matrix is denoted by $g_j(x)$ for $j\in\{1,\dots,L\}$. \label{fig:lsh}}
\end{figure}

To this end, we now describe a generic recipe that does both ``amplification'' and ``indexing'' \citep{CACM}.  Suppose that we have an LSH family $\HH$ with parameters $(c, R, P_1, P_2)$. We  create $L$ hash functions, where each hash function is actually the concatenation of $d'$ simpler hash functions from the family $\HH$ (\eg, for the Hamming distance example above, there would be $L$ hash functions that each take as input a feature vector and outputs a $d'$-bit string).  In particuar, we choose $d' L$ hash functions by drawing them independently and uniformly at random from the hash family $\HH$. We collect them into a a $d' \times L $ matrix denoted as $[h_{i, j}]_{i\in\{1,\dots d'\},  j\in\{1,\dots,L\}}$, where $h_{i,j}: \XX \to U$ for all $i, j$. We denote the $j$-th column of the matrix as a function $g_j: \XX \to U^{d'}$. Now any point $x \in \XX$ is mapped to $L$ different buckets where $j$-th bucket corresponds to $g_j(x)$. This process is depicted in Figure~\ref{fig:lsh}. The desired data structure is simply created by mapping each of the $n$ points in $\PP_n$ to $L$ such different buckets. Depending upon the size of the universe of hash values $U$, the number of possible buckets for each $j\in\{1,\dots,L\}$ may be much larger than $n$. Therefore, it may make sense to store only the nonempty buckets by further indexing the mapped value under $g_j$ for each of the $n$ points. This further indexing can be done by choosing another appropriate hash function. This way, the overall storage space is $\mathcal{O}(n L)$. 

To find a near neighbor for query point $x$, we hash it using each of the functions $g_1,\dots,g_L$. The $j$-th hash $g_j(x)$ corresponds to a bucket that has some subset of the $n$ training data points in it. Thus, across all the $L$ buckets that $x$ lands in, we look at the training points in those buckets and pick the closest one to $x$ to output as the near neighbor.

Why should this procedure work?  As it turns out, for any query point $x$, if there is a point $x'$ in training data $\mathcal{P}_n$ that is within distance~$R$, then the procedure will find $x'$ with high probability if we choose the number of buckets $L$ appropriately. Let's work out how we should set $L$.  Let point $x^\prime \in \PP_n$ be such that $\rho(x, x^\prime) \leq R$. Then by definition of $\HH$ being an LSH,
\[
\P(g_j(x) = g_j(x')) \geq P^{d'}_1, \quad\text{for }j\in\{1,\dots,L\}.
\]
This means that 
\[
\P\bigg(\bigcup_{j=1}^L \{g_j(x) = g_j(x') \}\bigg) \geq 1 - (1-P_1^{d'})^L.
\]
If we set the number of buckets to be
\[
L = \log \frac{\delta}{\log (1-P_1^{d'})}
 \approx P_1^{-{d'}} \log \frac{1}{\delta},
\]
then we have that $x^\prime$ is found as part of the above algorithm with probability at least $1-\delta$. Note that we have used the approximation $\log(1-z) \approx -z$ when real number $z$ is close to 0.

But what is the query time and storage cost?  These depend on how many points from $\PP_n$ are mapped to each of the $L$ buckets to which $x$ gets mapped to. Intuitively, we want to choose $d'$ large enough so that the number of such points is not too large (precisely $o(n/L)$), and $L$ is not too large so that $\mathcal{O}(n L) = o(n^2)$.  At some level, this boils down to making ratio $(P_1/P_2)^{d'}$  large enough. The optimized choice of parameters $d'$ and $L$ leads to query time $\Theta(d n^{\varphi})$ and storage cost $\Theta(n^{1+\varphi})$, where $\varphi = \frac{\log (1/P_1)}{\log (1/P_2)}$. 

In the context of Example~\ref{example:lsh}, for the Hamming distance with feature space $\XX = \{0,1\}^d$,
\begin{align*}
\varphi & = \frac{\log (1-\frac{R}{d})}{\log (1-\frac{cR}{d})} ~\approx~\frac{1}{c},
\end{align*}
again using the approximation $\log(1-z)\approx -z$ for $z$ close to 0. This means that for the Hamming distance, we have an approximate near neighbor data structure with  query time $\Theta(n^{1/c})$ and storage cost $\Theta(n^{1+1/c})$. Clearly, it makes sense that if  approximation ratio $c > 1$ is large (corresponding to tolerating near neighbors to be farther away from the true nearest neighbor), then the data structure becomes more efficient.

A few remarks are in order. To begin with, as long as $P_1 > P_2$ we have that $\varphi <1$. Therefore, the query time is sublinear in $n$ with a linear dependence on dimension $d$, and the storage cost is $o(n^2)$. But what metrics $\rho$ can we construct an LSH with, and what is the best $\varphi$ that we can achieve for these different metrics?  For Hamming distance, as discussed above, we have a simple solution. If metric $\rho$ is Euclidean distance over feature space $\XX = {\mathbb R}^d$, the answer is also reasonably well understood.  In particular, to achieve approximation ratio $c > 1$, the best~$\varphi$ is $\varphi \approx \frac{1}{c^2}$ \citep{datar_2004_good_intro_lsh,CACM}.  The matching lower bound was established by \citet{MNP07, OWZ14}. However, these guarantees assume that the algorithm is data agnostic and does not exploit structure specific to the given training feature vectors $\mathcal{P}_n$. In a recent development, it turns out that by using a {\em data dependent} construction, the optimal $\varphi$ for Euclidean distance turns out to be $\frac{1}{2c^2 - 1}$ for achieving  approximation ratio  $c > 1$. The construction of such an LSH is provided by \citet{andoni2015optimal}. The matching lower bound is provided by the same authors in a different paper \citep{AR15.2}.  For other metrics such as $\ell_p$ distances for $p \in [1,2)$, Jaccard, cosine, and $\ell_2$ distance on sphere, there are various LSH constructions known. We refer interested readers to the survey by \citet{CACM}.

To wrap up our discussion of LSH, we note that at a high level, LSH is reducing the dimensionality of the data, after which we execute exact search with the lower-dimensional representations (the hashes).  Other fast dimensionality reduction approaches are possible.  The challenge  is doing this reduction in a way where we can still ensure low approximation error. One such approach is given by the celebrated Johnson-Lindenstrauss lemma \citep{JL}, which reduces dimensionality while roughly preserving Euclidean distances.  For a user-specified approximation error tolerance $\varepsilon\in(0,1)$, the Johnson-Lindenstrauss lemma provides a way to transform any set of $n$ points in $d$-dimensional Euclidean space down to $\mathcal{O}(\log \frac{n}{\varepsilon^2})$ dimensions (of course, for this to be useful we need $d\gtrsim\log\frac{n}{\varepsilon^2}$). Then for any two points $x_1$ and $x_2$ among the original $n$ points, denoting their resulting low-dimensional representations as $x_1'$ and $x_2'$ respectively,
\[
(1-\varepsilon)\| x_1 - x_2 \|^2 \le \| x_1' - x_2' \|^2 \le (1+\varepsilon)\| x_1 - x_2 \|^2.
\]
A \textit{fast Johnson-Lindenstrauss transform} has been developed for doing this transformation that works for both $\ell_1$ and $\ell_2$ distances \citep{ailon_2009}.  Of course, if $n$ is massive, then the lower-dimensional space could still be too high-dimensional, in which case we could do an approximate---rather than exact---nearest neighbor search in the lower-dimensional space, such as using LSH.

\subsection{Random Projection Trees}

The k-d tree data structure is very efficient in terms of pre-processing data to build the data structure and in terms of the storage required. However, the query time can become prohibitively expensive due to its exponential dependence on the dimension of the data. As discussed, for the Euclidean metric, due to the Johnson-Lindenstraus lemma \citep{JL}, we can effectively assume that the $n$ high-dimensional data points lie in a $\mathcal{O}(\log n)$-dimensional space albeit incurring an approximation error in computing distances. Thus, the query time can be ${\sf poly}(n)$ with potentially a very large albeit constant exponent for $n$. However, what we would really like is for the query time to be $o(n)$.

The k-d tree readily offers a simple approximate nearest neighbor search  known as a {\em defeatist} search. Recall that to search for an {\em exact} nearest neighbor for a query point $x$, we first find the leaf node in the k-d tree that $x$ belongs to.  However, not only do we look for the closest training point within this leaf node, we also look at all the closest training point inside all the adjacent leaf nodes. We did this because within the leaf node that $x$ belongs to, which corresponds to a partition of the feature space, $x$ can be near the boundary of this partition. In particular, its nearest neighbor may be a training point just on the other side of this boundary in an adjacent leaf node.  The search over neighboring leaf nodes, however, is what results in the exponential explosion in dimension for query time.  The {\em defeatist} search takes an optimistic view: {\em let us assume that the nearest neighbor is within the same leaf tree node as that associated with the query point}. Therefore, we simply return the nearest neighbor found within the leaf node that $x$ belongs to and search no further.

\begin{figure}
\centering
\includegraphics[width=.38\linewidth]{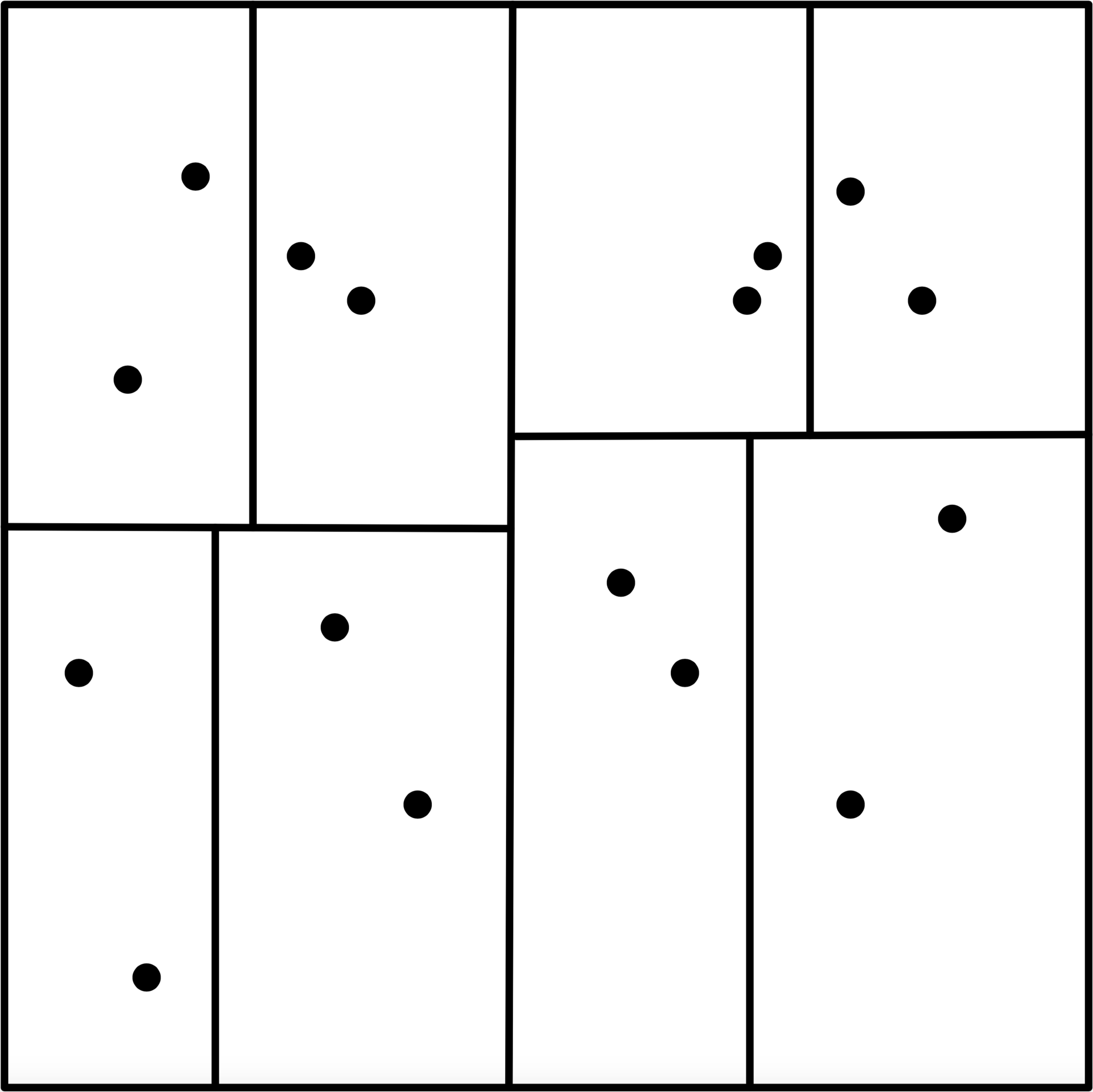}
\hspace{1.5em}
\includegraphics[width=.38\linewidth]{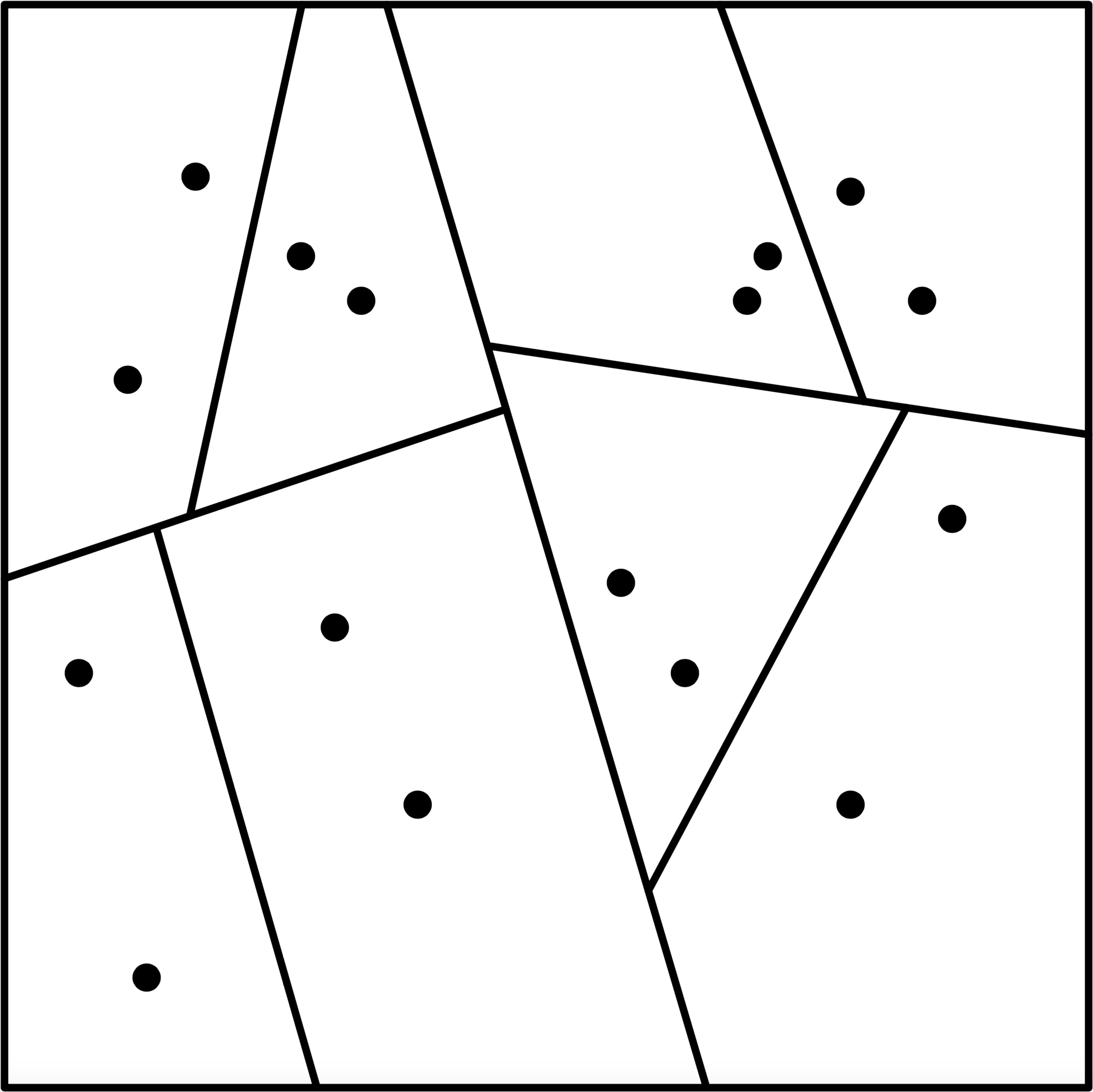}
\caption{A comparison between how a k-d tree and a randomized partition tree divide up the feature space for two dimensional data (figure source:  \citealt{dasgupta_2015}). Left: a k-d tree uses axis-aligned splits. Right: a randomized partition tree uses splits with directionality that is randomized (such as being drawn randomly from a unit ball).
\label{fig:rd-tree}}
\end{figure}

Clearly, this is a heuristic and suffers from the worst-case scenario where the true nearest neighbor is on the other side of the boundary. This can lead to an unbounded approximation error. However, the silver lining in such a method is that potentially such a worst-case scenario can be overcome via randomization. Specifically, instead of choosing the median along each coordinate to partition data in constructing the tree, it may make sense to do the partition using ``random projection''. We now consider one such heuristic to construct a {\em random projection tree} or a {\em randomized partition tree}. The high-level idea is that unlike in k-d trees where the feature space is divided by axis-aligned splits, in random projection trees, the split directions are randomized. This difference in split directions is shown in Figure~\ref{fig:rd-tree}.

Given $n$ data points $\PP_n$, as before the root corresponds to all the points (and also the entire feature space $\mathcal{X}$). To partition points into two groups, choose a random vector $\randomVectorVar \in {\mathbb R}^d$. The choice of random vector can be done as per various distributions. Here, we choose $\randomVectorVar$ so that each of its component is Gaussian with mean $0$ and variance $1/d$. We compute the median of $n$ values, $\randomVectorVar^T X_i$ for $i\in\{1,\dots,n\}$. If a point $X_i$ is such that $\randomVectorVar^T X_i$ is less than the median, then it goes to one partition (say left) and if it is greater or equal to the median than it goes to the other partition (say right). We use this method to recursively partition the training data until each leaf is left with at most $n_0$ points associated with it.

To find an approximate nearest neighbor, we use the defeatist search and hope that randomization would make sure that the query point of interest actual lies well within the interior of the leaf partition that it belongs to. In terms of computation, we traverse down the tree, which takes $\mathcal{O}(\log n)$ steps corresponding to the tree depth, where at each step of this traversal, we compute an inner product, which takes $\mathcal{O}(d)$ time. When we reach the leaf node, we compare distances with at most $n_0$ leaf data points. Treating $n_0$ as a constant, overall the query time is $\mathcal{O}(d \log n)$.

In summary, the above algorithm provides an efficient approximate nearest neighbor data structure in terms of query time, storage cost, and pre-processing time. For such a data structure, there are no crisp theoretical results known. However, \citet{dasgupta_2015} provide a nice theoretical analysis for variations of such an approach. Effectively, Dasgupta and Sinha argue that the probability of error in terms of finding an {\em exact} nearest neighbor scales as $\Phi \log (1/\Phi)$, where $\Phi$ is a function of the query point~$x$ and the~$n$ data points $\PP_n = \{X_1,\dots, X_n\}$, defined as 
\begin{align*}
\Phi(x, \PP_n) & \triangleq \frac{1}{n} \sum_{i=2}^n \frac{\|x - X_{(1)}(x) \| }{ \|x - X_{(i)}(x) \|},
\end{align*}
which can be thought of as an average ratio of the distance of $x$ to its nearest neighbors vs its distance to the $n-1$ other training points.

To understand the behavior of the above quantity, let's consider the following simple setting. Suppose the distance from the nearest neighbor to the query point is $1$, \ie, $\|x - X_{(1)}(x) \| = 1$. Moreover, let's assume that all the other points are at integer distances $> 1$ from the query point $x$ of interest. Let $N(z)$ be the number of points at distance $z \geq 2, z \in {\mathbb N}$ from the query point $x$ among $\PP_n$. Then 
\begin{align*}
\Phi(x, \PP_n) & = \frac{1}{n} \sum_{z \geq 2} \frac{N(z)}{z}.
\end{align*}
To get sense of the above quantity, further suppose that all the points in $\PP_n$ are laid on a regular $d$-dimensional grid around $x$. Then $N(z) \sim z^{d-1}$. This would suggest that 
\begin{align*}
\Phi(x, \PP_n) & \sim n^{-1/d}.
\end{align*}
That is, if indeed data $\PP_n$ were laid out nicely, then the error probability is likely to decay with $n$, \ie, the probability of finding the exact nearest neighbor goes to~1 for large $n$. Alas, such a guarantee does not hold in arbitrary point configurations for $\mathcal{P}_n$. However, as long as the underlying feature space corresponds to a {\em growth-restricted metric space} or has {\em bounded doubling dimension} \citep{DBD}, then the probability of success is nontrivially meaningful. We refer an interested reader to the paper by \citet{dasgupta_2015} for details.

A variation of the randomized partition tree is to simply use many such trees and merge their answers to produce a better answer than what any single tree can provide. This naturally lowers the error probability and has found reasonable success in practice.

\subsection{Boundary Trees}

The two approaches described above have a few limitations. The approach based on LSH, while theoretically sound in the most generality known in the literature, is restricted to specific metrics for which LSH constructions are known. In a similar manner, the randomized partition tree approach, because of the use of random projections with inner products, is fundamentally restricted to Euclidean distance (although the theory here could potentially extend to other inner-product induced norms). Moreover, both of the approaches have an inherent requirement that prevents them to be truly incremental in nature---all $n$ points need to be present before the data structure is constructed, and as new data points are added, one needs to re-start the construction or perform expensive adjustments. Recently, various heuristics have been suggested (\eg,  \citealt{pLSH}) which tries to exploit the inherent parallelism in LSH-based constructions while providing an incremental insertion property in an {\em amortized} sense. However, such approaches are inherently clunky workarounds and not truly incremental. They also do not support deletion. In that sense, out of all the desiderata stated in the beginning of this chapter, only the first three are satisfied by the approaches that we have discussed thus far. 

We now discuss the recently proposed boundary tree data structure by \citet{mathy_2015} that overcomes some of these limitations. Multiple such trees can be collectively used to form a boundary forest. Each boundary tree takes $\mathcal{O}(n)$ storage space, supports query time that is expected to be $o(n)$ (in an optimistic scenario $\mathcal{O}(\log n)$), supports incremental insertion, and works for {\em any} distance or metric. It does not support {\em incremental deletion}. Unfortunately, to date, boundary trees and forests lack theoretical guarantees. Initial empirical evidence suggests them to have outstanding performance \citep{mathy_2015, bf2}.

\begin{figure}
\centering
\includegraphics[width=.28\linewidth]{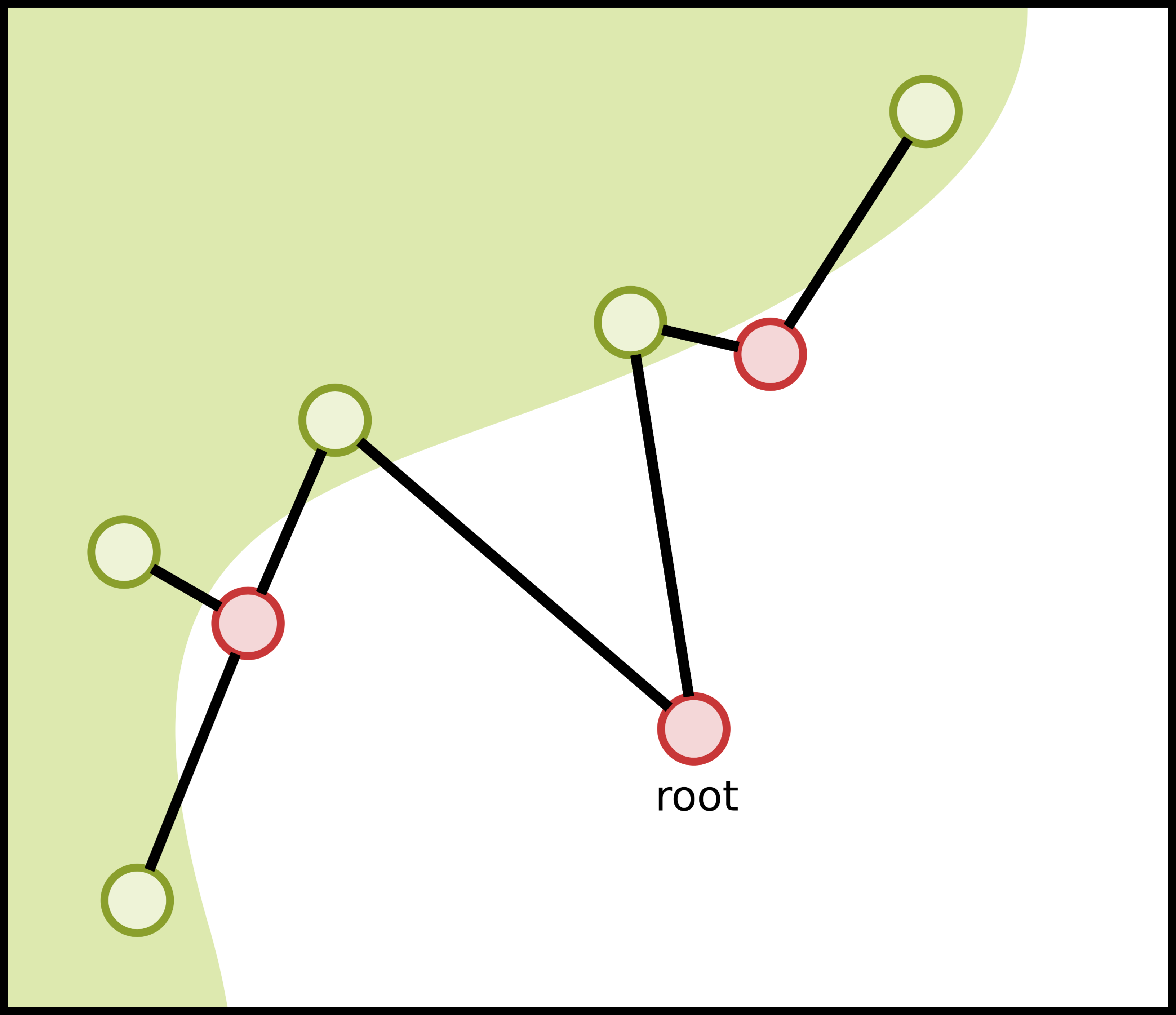}
\hspace{1em}
\includegraphics[width=.28\linewidth]{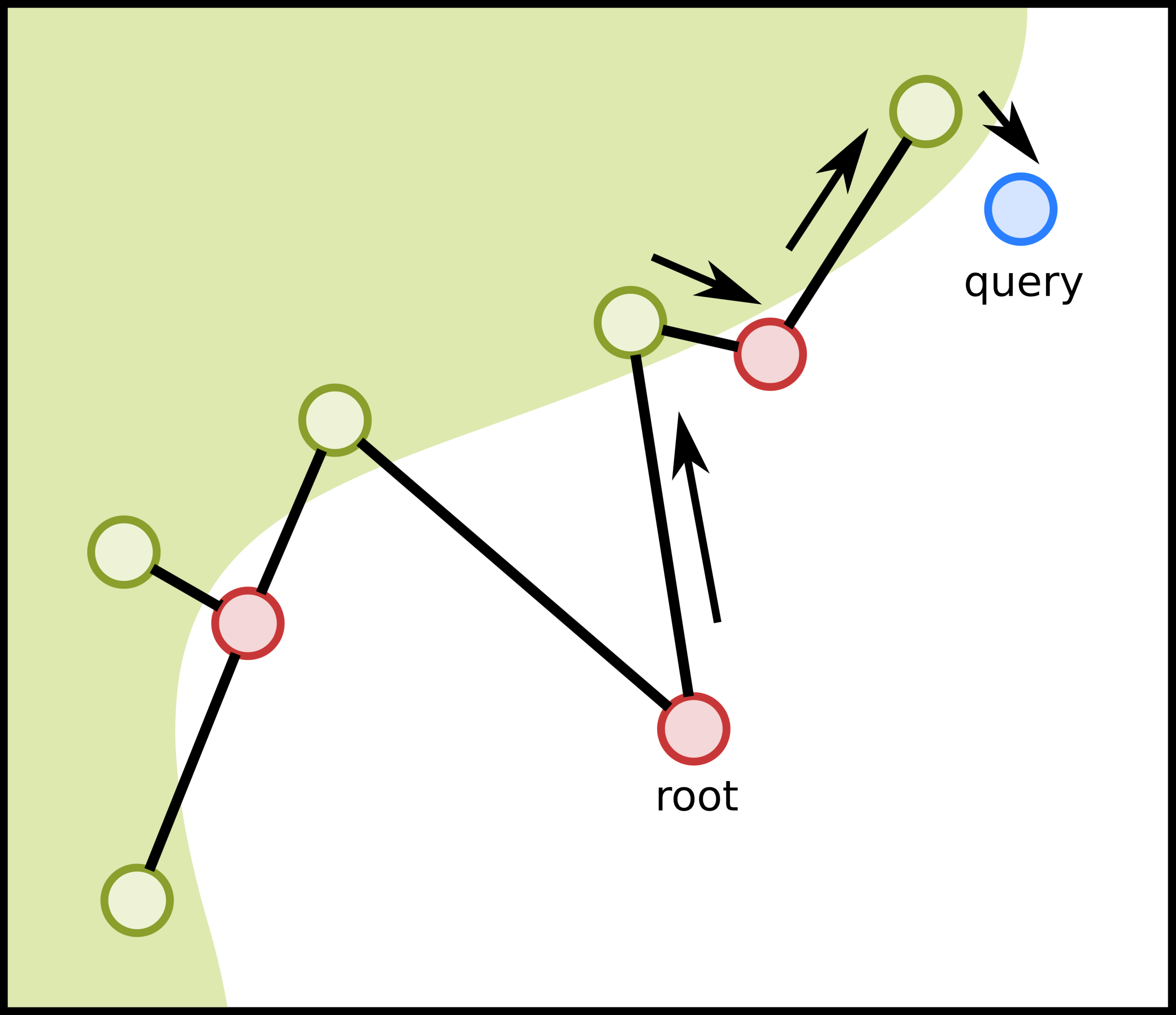}
\hspace{1em}
\includegraphics[width=.28\linewidth]{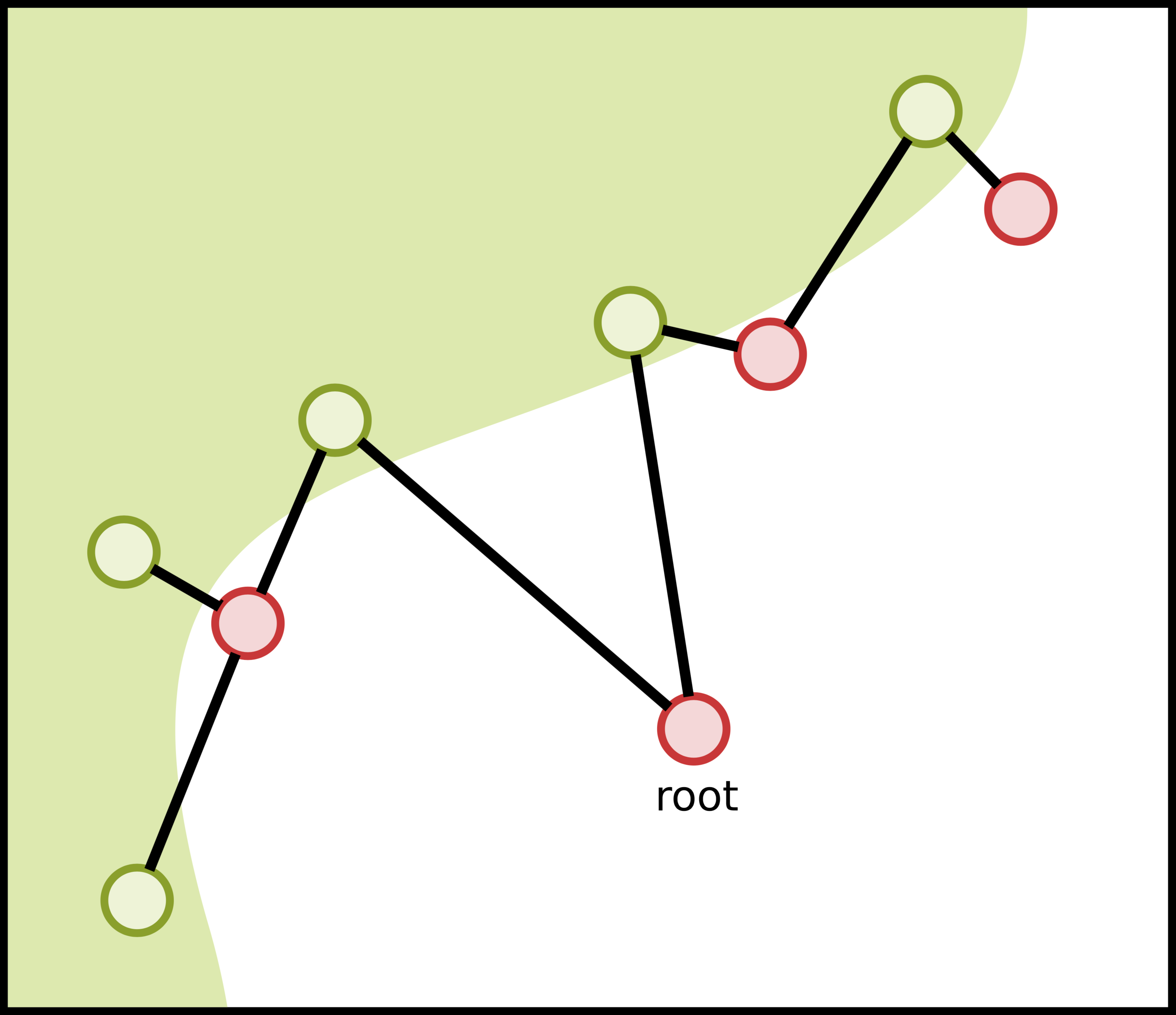}
\caption{Example of a boundary tree (figure source: \citealt{bf2}). Left panel: the current state of a boundary tree. Middle panel: an example approximate nearest neighbor search; we begin at the root and traverse the tree greedily looking for a nearest neighbor among the current nodes (in this example, the approximate nearest neighbor found is the top-right shaded white node, and the query point is denoted in blue). Right panel: the query point from the middle panel can be inserted into the boundary tree. For classification, a new data point is inserted into a tree only if its label differs from that of the approximate nearest neighbor found. Hence, neighboring nodes in the boundary tree have different labels (white vs.~red nodes across the three panels). Points added tend to be near the decision boundary. \label{fig:bf}}
\end{figure}

In a nutshell, a boundary tree is constructed as follows.  At the very beginning, there are no data points in the boundary tree.  The very first data point that gets added is just assigned as the root node.  Then for a subsequent data point $x$ to be added, we start a tree traversal from the root node. At each step of this traversal, we compute the distance between $x$ and the current node we are at, and also the distances between $x$ and all the children of the current node we are at (recall that each node corresponds to a previously inserted training data point). For whichever node is closest, we recurse and continue the traversal. We stop recursing when we either get to a leaf node or stay at the same node (\ie, this node is closest to the query point~$x$ than any of its children). Note that the node that we stop at, say $x'$, serves two purposes. First, it is the approximate nearest neighbor for~$x$. Second, it is where~$x$ gets inserted: node~$x$ gets added as a child of $x'$. Thus, if we are merely querying for an approximate nearest neighbor for~$x$ that is already in the boundary tree rather than doing an insertion, then we would output $x'$ without inserting $x$ into the tree.  An example boundary tree, along with a search query and subsequent insertion of the query point into the tree, is shown in Figure~\ref{fig:bf}.  Using multiple trees enables us to choose the closest point found across the different boundary trees.

We remark that there is a slight refinement on the above description to enforce that the maximum number of children per node is a pre-specified constant, which empirically improves the speed of the algorithm without much impact on the quality of approximate nearest neighbors found provided that $k$ is not too small. See the original boundary tree paper for details on this modification \citep{mathy_2015}.

Note that a boundary forest can readily be used for regression or classification.  For example, for classification, to predict the label of $x$, we pick the most popular label among the labels of the approximate nearest neighbors found across the boundary trees.  In this case of classification, an algorithmic edit to make the boundary tree data structure more efficient is that when a data point $x$ is to be added, if its approximate nearest neighbor has the same label, then $x$ is simply discarded rather than added to the tree. The boundary tree is named after the fact that in doing this algorithmic edit, the training points that are added tend to be points near the decision boundary.  This is depicted in all three panels of Figure~\ref{fig:bf}, where neighboring nodes have different colors because they have different class labels.

In the case of multiple boundary trees, it is worth noting that if each tree receives the data points for insertion in the same order then the trees will be identical. Thus, to obtain diversity which can improve the quality of approximate nearest neighbors found as well as regression or classification performance, randomizing the order of insertion helps. In practice, it may suffice to randomize the insertion order of an initial batch of points.

\section{Open Source Software}
\label{sec:NN-open-source}

We now point out a few open source nearest neighbor software packages that have been developed over the past decade or so. The packages below are provided in an arbitrary order.

Muja and Lowe's Fast Library for Approximate Nearest Neighbors (FLANN) provides access to a collection of approximate nearest neighbor search algorithms.\footnote{\url{https://www.cs.ubc.ca/research/flann/}} The system claims to use the best of the available options for approximate nearest neighbor search. Through experiments conducted on specific datasets, \citet{flannpaper} found that random partition trees described above and the priority K-means approach seem to work very well. The priority K-means effectively builds a tree data structure as follows. The root corresponds to the entire dataset. We partition the data into clusters using the K-means algorithm. The clusters become the children of the root and we recurse. The query process effectively boils down to doing a greedy search along the tree to find the right leaf. A variation of this search that involves some backtracking through ``priority'' leads to a refined search algorithm.  The FLANN library is implemented in C++ and is available with Python and MATLAB bindings.

Mount and Arya's library for Approximate Nearest Neighbor Searching (ANN) provides another alternative with a collection of approaches implemented in C++.\footnote{\url{https://www.cs.umd.edu/~mount/ANN/}}

The Python-based scikit-learn package provides an implementation of various exact and approximate nearest neighbor algorithms.\footnote{\url{http://scikit-learn.org/}} The package is quite popular and is very robust.

A recent release by Erik Bernhardsson starting from when he was still at Spotify and subsequently improved by an open source effort has resulted into the Annoy package (with the backronym ``Approximate Nearest Neighbors Oh Yeah'').\footnote{\url{https://github.com/spotify/annoy}} Annoy implements a variation of randomized projection trees and works well in practice. Annoy has also been implemented in Scala to work with Apache Spark in package called Ann4s.\footnote{\url{https://github.com/mskimm/ann4s}}

The FAst Lookup of Cosine and Other Nearest Neighbors (FALCONN) is based on LSH \citep{andoni2015practical}.\footnote{\url{https://github.com/FALCONN-LIB/FALCONN}} It is developed by Ilya Razenshteyn and Ludwig Schmidt. FALCONN has grown out of a research project with their collaborators Alexandr Andoni, Piotr Indyk, and Thijs Laarhoven. This open source project not only comes with theoretical guarantees but also works very well in practice. It is implemented in C++ with a Python wrapper.

Specifically for dealing with sparse data, Facebook Research has released a package called Approximate Nearest Neighbor Search for Sparse Data in Python (PySparNN).\footnote{\url{https://github.com/facebookresearch/pysparnn}}

To the best of our knowledge, there is currently no fast, highly reliable open source nearest neighbor library that can scale to massive amounts of data using a distributed implementation and that handles fast, real-time queries.

\chapter{Adaptive Nearest Neighbors and Far Away Neighbors}
\label{chap:discussion}

Up until now, the problem setups and algorithms considered each have some notion of distance that is treated as chosen independently of the training data. However, adaptively choosing distance functions based on training data can be immensely beneficial for nearest neighbor prediction. The literature on distance learning is vast and beyond the scope of this monograph.  For example, two popular approaches include Mahalanobis distance learning methods (see the survey by \citealt{metric_learning}) and, on the deep learning and neural network side, Siamese networks \citep{siamese1,siamese2}.

Rather than covering Mahalanobis distance learning and Siamese networks, we instead show in Section~\ref{sec:decision-trees-ensemble-learning} that decision trees and various ensemble learning methods based on bagging and boosting turn out to be adaptive nearest neighbor methods, where we can derive what the similarity functions being learned from training data are.  Examples include AdaBoost \citep{adaboost} with specifically chosen base predictors, the C4.5 \citep{C4_5} and CART \citep{CART} methods for learning decision trees, and random forests \citep{random_forests}.  Thus, for any test feature vector $x\in\mathcal{X}$, by computing the learned similarity function between $x$ and each of the training data points, we know exactly which training data have nonzero similarity with the point~$x$. These training points with nonzero similarity to $x$ are precisely the nearest neighbors of $x$.  Thus, the number of nearest neighbors $k=k(x)$ chosen depends on $x$.  However, not only are we automatically selecting $k(x)$, we are also learning which similarity function to use. This is in contrast to the earlier methods discussed in Section~\ref{sec:automatic-k-h} that adaptively select $k(x)$ or the bandwidth $h(x)$ for $k$-NN, fixed-radius~NN, and kernel regression but do not also learn a similarity or distance function.

\begin{figure}
\centering
\includegraphics[scale=.85]{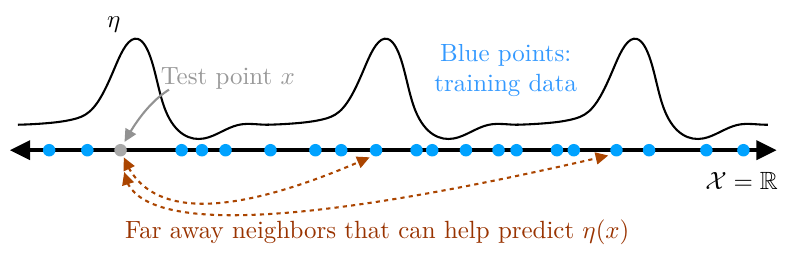}
\vspace{-1em}
\caption{An example regression problem where the regression function $\eta$ is periodic. For the test point $x\in\mathcal{X}$ in gray, to predict $\eta(x)$, we can look at training data integer multiples of the period away.}
\label{fig:far-away-neighbors}
\end{figure}

We conclude this monograph in Section~\ref{sec:final-section-beyond-nearest-neighbors} by looking at when potentially far away neighbors can help with prediction. For example, in the periodic regression function $\eta$ shown in Figure~\ref{fig:far-away-neighbors}, it is immediately apparent that for test point $x$ shown in gray, to predict $\eta(x)$, we can benefit from looking at labels of training data that are roughly at integer multiples of $\eta$'s period away.  This example is of course overly simplistic but raises the question of how we could automatically find training data from disparate regions of the feature space that can help each other with prediction.  We discuss how recently proposed \textit{blind regression} methods for collaborative filtering exploit this idea when only given access to labels of data and not their feature vectors \citep{LLSS16, BCLS17}.  By transferring ideas from blind regression back to the context of standard regression, we arrive at a two-layer nearest neighbor algorithm that takes the label output of one $k$-NN search query to feed as input to another. The latter search enables finding different parts of the feature space $\mathcal{X}$ with similar regression function values.

\section[Adaptive Nearest Neighbor Methods]{Adaptive Nearest Neighbor Methods: Decision Trees and Their Ensemble Learning Variants}
\label{sec:decision-trees-ensemble-learning}

We now discuss how decision trees and a wide class of ensemble methods turn out to all be nearest neighbor methods albeit where the distances are implicitly learned from training data. In particular, these methods actually learn nearest neighbor relationships and similarity functions, from which we can back out distance functions. For ease of exposition, we focus on regression. Of course, going from regression to binary classification can be done by thresholding. We talk about three main components frequently used in ensemble learning and how they relate to nearest neighbor or kernel regression: decision trees, bagging, and boosting. For example, decision trees combined with bagging results in random forests~\citep{random_forests}. Our coverage here builds on the observation made by \citet{lin_random_forests} that random forests can be viewed as adaptive nearest neighbor predictors. We effectively extend their observation to a much wider class of ensemble learning methods.

\subsection{Decision Trees for Regression}
\label{sec:decision-trees-NN}

A decision tree $\mathcal{T}$ partitions the feature space $\mathcal{X}$ into $J$ disjoint regions $\mathcal{X}_{1},\mathcal{X}_{2},\dots,\mathcal{X}_{J}$.  Each region $\mathcal{X}_{j}$ is associated with a leaf of the tree $\mathcal{T}$ and has an associated label $\lambda_{j}\in\mathbb{R}$.  Given a feature vector $x\in\mathcal{X}$, the tree $\mathcal{T}$ finds which region $\mathcal{X}_{j}$ contains $x$ and outputs $\lambda_{j}$. Formally,
\[
\mathcal{T}(x)=\sum_{j=1}^{J}\lambda_{j}\ind\{x\in\mathcal{X}_{j}\},
\]
where only one term in the summation is nonzero since the $\mathcal{X}_{j}$'s are disjoint.

Everything above is actually true for decision rules in general. What makes $\mathcal{T}$ a decision tree rather than just a general decision rule is that the disjoint regions exhibit tree structure. For example, if the tree is binary, the $J$ regions initially cover the entire space. Then there are two disjoint subsets of the $J$ regions that correspond to the two child nodes of the root node, and so forth.  While we could always assign some tree structure to the $J$ regions, this observation ignores a key computational aspect for decision trees.

Fundamentally, decision tree training procedures construct a tree and assigns every training data point to a leaf node of the tree.  At the start of training, there is a single leaf node corresponding to the entire feature space~$\mathcal{X}$ and that consists of all $n$ training data. Then we recursively split the feature space (\eg, the first split divides the single leaf node into two leaf nodes).  The recursive splitting happens until some termination criterion is reached, such as each leaf node having fewer than some pre-specified number of training data associated with~it.

We skirt discussing training details (\eg, choice of splitting rule, when to stop splitting); a variety of methods are available such as C4.5~\citep{C4_5} and CART~\citep{CART}.  For the purposes of this monograph, what matters is that after training, there are $J$ leaf nodes, where the $j$-th leaf node precisely corresponds to disjoint region~$\mathcal{X}_j$, has associated with it some subset $\mathcal{N}_j\in\{1,\dots,n\}$ of the training data, and also has a label $\lambda_j$ based on training data in~$\mathcal{N}_j$, which for regression we can take to be the average label:
\[
\lambda_{j}=\frac{1}{|\mathcal{N}_{j}|}\sum_{i\in\mathcal{N}_{j}}Y_{i}.
\]
Then to predict the label for a test feature vector~$x\in\mathcal{X}$, the decision tree quickly finds the leaf node~$j$ that~$x$ belongs to (so that $x\in\mathcal{X}_j$) and outputs the leaf node's label~$\lambda_j$.

\begin{figure}
\centering
\includegraphics[width=\linewidth]{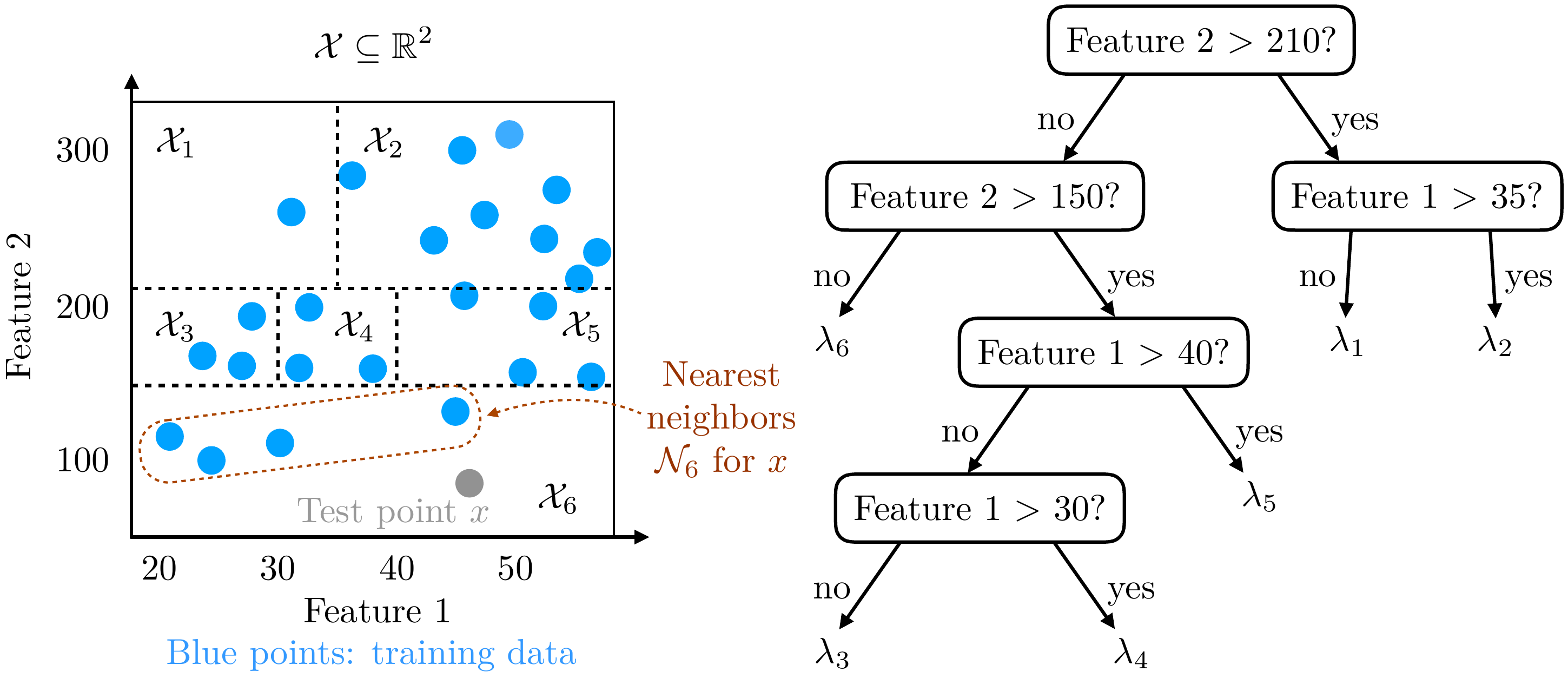}
\caption{Example of a decision tree with 6 leaf nodes corresponding to disjoint regions~$\mathcal{X}_1,\dots,\mathcal{X}_6$ that altogether make up the feature space~$\mathcal{X}$.  Each region $\mathcal{X}_j$ contains a subset $\mathcal{N}_j$ of the training data (blue points) and has an associated label~$\lambda_j$ (\eg, the average label among points in~$\mathcal{N}_j$).  Here, test point~$x$ (gray point) is in~$\mathcal{X}_6$, so its nearest neighbors are precisely the training data in~$\mathcal{N}_6$.  Left: an annotated picture of the feature space. Right: the decision tree with its branching decisions; at the leaves are the predicted values $\lambda_1,\dots,\lambda_6$.}
\label{fig:decision-tree-nn}
\end{figure}

The key observation is that we can think of nearest neighbors as feature vectors that belong to the same leaf node. In particular, for a test feature vector $x$, if it is inside $\mathcal{X}_{j}$, then the decision tree effectively tells us that the nearest neighbors of $x$ are the training points in $\mathcal{N}_{j}$, and the predicted value for $x$ is the average label among the nearest neighbors. This observation is shown for an example decision tree in Figure~\ref{fig:decision-tree-nn}.  Turning this observation into an equation, we have
\begin{align*}
\mathcal{T}(x) & =\sum_{j=1}^{J}\lambda_{j}\ind\{x\in\mathcal{X}_{j}\}
 =\sum_{j=1}^{J}\!\bigg[\frac{1}{|\mathcal{N}_{j}|}\!\sum_{i=1}^{n}Y_{i}\ind\{x\text{ and }X_{i}\text{ are both in }\mathcal{X}_{j}\}\bigg] \\
& =\sum_{i=1}^{n}Y_{i}\underbrace{\bigg[\sum_{j=1}^{J}\frac{1}{|\mathcal{N}_{j}|}\ind\{x\text{ and }X_{i}\text{ are both in }\mathcal{X}_{j}\}\bigg]}_{\text{similarity between }x\text{ and }X_{i}},
\end{align*}
which could be thought of as a kernel regression estimate. Training data point $X_{i}$ has weight $\widehat{\mathbb{K}}(x, X_i) \triangleq \sum_{j=1}^{J}\frac{1}{|\mathcal{N}_{j}|}\ind\{x\text{ and }X_{i}\text{ are both in }\mathcal{X}_{j}\}$.

Phrased another way, the nearest neighbors adaptively chosen for $x$ are precisely the training data points $X_i$ with nonzero learned weight $\widehat{\mathbb{K}}(x, X_i)$.  Thus, the number of nearest neighbors $k(x)$ chosen for test feature vector $x$ is
\[
k(x) = \sum_{i=1}^n \ind\{ \widehat{\mathbb{K}}(x, X_i) > 0 \}.
\]

\subsection{\texorpdfstring{\underline{B}ootstrap \underline{Agg}regat\underline{ing}}{Bootstrap Aggregating} (``Bagging'')}

For a specific supervised learning method that learns predictor $\mathcal{T}$ based on training data $(X_1,Y_1),\dots,(X_n,Y_n)\in\mathcal{X}\times\mathbb{R}$, where $\mathcal{T}(x)$ is the predicted label for a given feature vector~$x$, one way to improve prediction is to use what is called bootstrap aggregation (popularly called ``bagging''). Bagging yields a new predictor that can have lower variance than the original predictor $\mathcal{T}$.  Note that $\mathcal{T}$ need not be a decision tree although this notation foreshadows our discussion of random forests at the end of this section.  It turns out that bagging of nearest neighbor predictors (whether adaptive or not) results in adaptive nearest neighbor predictors.  Thus, a random forest, which we obtain by bagging many decision tree predictors (which are adaptive nearest neighbor predictors), is yet another adaptive nearest neighbor method.

Bagging works as follows.  We first produce $B$ sets of bootstrap samples $\mathcal{Z}^{1},\dots,\mathcal{Z}^{B}$, where each bootstrap sample $\mathcal{Z}^{b}$ comprises of~$n$ labeled data points drawn uniformly at random with replacement from the original $n$ training data. Then for each $b\in\{1,\dots,B\}$, we use the same supervised learning method to learn function $\mathcal{T}_{b}$ for each bootstrap sample $\mathcal{Z}^{b}$.  Finally, given feature vector $x$, the predicted label for $x$ is the average output across the $\mathcal{T}_{b}$'s:
\[
\mathcal{T}_{\text{bag}}(x)=\frac{1}{B}\sum_{b=1}^{B}\mathcal{T}_{b}(x).
\]
Using ensemble predictor $\mathcal{T}_{\text{bag}}$ instead of a single predictor $\mathcal{T}_{b}$ keeps the expected value of the prediction the same but can decrease the variance, depending on how correlated a randomly chosen pair of $\mathcal{T}_{b}$'s are~\citep[Sections~8.7 and~15.2]{ESL}. Computationally, bagging is also trivially parallelizable: we can run $B$ training procedures in parallel to obtain the decision rules $\mathcal{T}_{1},\dots,\mathcal{T}_{B}$.

The connection to nearest neighbor methods is that if each $\mathcal{T}_{b}$ has the form of a kernel regression estimate, namely
\[
\mathcal{T}_{b}(x)=\sum_{i=1}^{n}Y_{i}\mathbb{K}_{b}(x,X_{i}),
\]
where $\mathbb{K}_{b}(x,X_{i})$ is some sort of similarity between feature vectors $x$ and $X_{i}$ that predictor $\mathcal{T}_{b}$ uses (either manually specified or learned), then the ensemble predictor is still a kernel regression estimate:
\[
\mathcal{T}_{\text{bag}}(x)
=\frac{1}{B}\sum_{b=1}^{B}\underbrace{\bigg[\sum_{i=1}^{n}Y_{i}\mathbb{K}_{b}(x,X_{i})\bigg]}_{\mathcal{T}_b(x)}
=\sum_{i=1}^{n}Y_{i}\!\!\!\!\!\!\!\!\!\!\!\!\underbrace{\bigg[\frac{1}{B}\sum_{b=1}^{B}\mathbb{K}_{b}(x,X_{i})\bigg]}_{\text{new similarity between }x\text{ and }X_{i}}\!\!\!\!\!\!\!\!\!\!\!\!.
\]
This observation along with our observation in Section~\ref{sec:decision-trees-NN} explain why random forests, which are decision trees used with bagging, are actually doing kernel regression using a learned similarity function.  Again, training data points that have nonzero learned similarity with $x$ can be thought of as the adaptively chosen nearest neighbors of $x$.

\subsection{Boosting}

Boosting is another popular ensemble learning method for which using it with nearest neighbor predictors (adaptive or not) results in adaptive nearest neighbor predictors.  For a collection of predictors $\mathcal{T}_{1},\dots,\mathcal{T}_{B}$ where given a feature vector $x\in\mathcal{X}$, the $b$-th predictor outputs $\mathcal{T}_{b}(x)$, boosting at a high level learns a weighted sum of these predictors to form a stronger predictor over a course of $T$ iterations:
\[
\mathcal{T}_{\text{boost}}(x)=\sum_{t=1}^{T}\alpha_{t}\mathcal{T}_{(t)}(x),
\]
where at iteration $t$, we select a predictor $\mathcal{T}_{(t)}$ among the collection of predictors $\{\mathcal{T}_{1},\dots,\mathcal{T}_{B}\}$ and give it a weight $\alpha_{t}\ge0$. Thus, we start with using a single predictor and then keep adding predictors until we have $T$ predictors. We omit the exact details of how to select the next predictor to use at each iteration and how to weight it. A popular way for selecting these over the $T$ iterations is called AdaBoost~\citep{adaboost}.

For the purposes of this monograph, the key observation is that if each predictor $\mathcal{T}_{(t)}$ selected has the form of a kernel regression estimate, namely
\[
\mathcal{T}_{(t)}(x)=\sum_{i=1}^{n}Y_{i}\mathbb{K}_{(t)}(x,X_{i}),
\]
where $\mathbb{K}_{(t)}(x,X_{i})$ is some sort of similarity between feature vectors $x$ and $X_{i}$ that predictor $\mathcal{T}_{(t)}$ uses (either manually specified or learned), then the boosted predictor $\mathcal{T}_{\text{boost}}$ also has the form of a kernel regression estimate
\[
\mathcal{T}_{\text{boost}}(x)
=\sum_{t=1}^{T}\alpha_{t}
 \underbrace{\bigg[\sum_{i=1}^{n}Y_{i}\mathbb{K}_{(t)}(x,X_{i})\bigg]}_{\mathcal{T}_{(t)}(x)}
=\sum_{i=1}^{n}Y_{i}\!\!\!\!\!\!\!\!\!\!\underbrace{\bigg[\sum_{t=1}^{T}\alpha_{t}\mathbb{K}_{(t)}(x,X_{i})\bigg]}_{\text{new similarity between }x\text{ and }X_{i}}\!\!\!\!\!\!\!\!\!\!.
\]
In particular, this observation along with that of Section~\ref{sec:decision-trees-NN} explains why AdaBoost used with decision trees as the base predictors are actually doing kernel regression with a learned similarity function.  Once more, training data points that have nonzero learned similarity with $x$ can be thought of as the adaptively chosen nearest neighbors of $x$.

\subsection{Similarity Learning Aware of Nearest Neighbor Prediction Guarantees?}

A promising avenue of research is to understand how to modify similarity or distance learning methods like the above decision tree and ensemble methods to take better advantage of nearest neighbor prediction error bounds like those presented in Chapters~\ref{chap:regression}, \ref{chap:classification}, and \ref{chap:case-studies}. The goal here is to make the learned distance function particularly well-suited for downstream nearest neighbor prediction.  In the case of modifying decision tree learning or boosting algorithms, which are typically greedy, to account for prediction error bounds, a possible technical challenge is that we may want to enforce a constraint over the entire tree or boosted classifier to be learned.  Such a global constraint may be difficult to incorporate into the greedy local decisions made by the learning algorithms.  A framework that could enforce global constraints was recently proposed by \citet{bertsimas_dunn_2017}, who show that we can now train classification trees with mixed-integer optimization that constructs an entire decision tree all at once rather than greedily.

\section{Far Away Neighbors}
\label{sec:final-section-beyond-nearest-neighbors}

In this last part of the monograph, we consider nearest neighbor methods in which potentially far away neighbors can help us in prediction. Some straightforward ways to achieve this include cleverly choosing a distance function that makes specific parts of the feature space appear closer together, or expanding a data point's neighbors to include nearest neighbors of nearest neighbors and recursing. What isn't straightforward is how to do either of these in a way that has theoretical performance guarantees.  As it turns out, under reasonably general settings, these two strategies yield nearest neighbor methods that are guaranteed to have their prediction error vanish to 0 \citep{LLSS16, BCLS17}. These approaches solve a problem referred to as \textit{blind regression}.

Blind regression can be phrased in terms of standard ``offline'' collaborative filtering (unlike in online collaborative filtering from Section~\ref{chap:online-collaborative-filtering} that aims to make good predictions over time, the offline version aims to accurately predict all missing user ratings at a frozen snapshot in time).  Recall that in collaborative filtering, the partially revealed ratings matrix~$Y$ is also what we use to obtain ``feature vectors'': rows of~$Y$ index users and columns of~$Y$ index items, and for two users~$u$ and~$v$, we can compare the row vectors~$Y_u$ and~$Y_v$ to identify which users are nearest neighbors (\eg, using cosine distance).  Each user $u$ does not have some feature vector~$X_u$ separate from her or his revealed ratings~$Y_u$ to help us predict~$Y_u$.  This is in contrast to the standard regression setting (Section~\ref{sec:regression-setup}), where the $i$-th data point has both a feature vector~$X_i$ and a label~$Y_i$ that are different objects.  Blind regression posits that there are latent feature vectors~$X_u$'s for users (and more generally also latent feature vectors for items) that are related to revealed ratings~$Y_u$'s through an unknown function~$f$. However, we do not get to observe the latent feature vectors~$X_u$'s. In this sense, we are ``blind'' to what the true latent feature vectors are and instead make predictions (using regression) based on ratings revealed thus far (and hence the name blind regression).  The key idea is to exploit the structure of the unknown function~$f$ to identify nearest neighbors.

In Section~\ref{sec:matrix-est-cf}, we provide an overview of this blind regression setup along with the nearest neighbor blind regression algorithms by \citet{LLSS16, BCLS17}. These recently developed algorithms are promising, but they invite the question of whether we can use the same insights in the standard ``sighted'' regression setting, where we get to observe both the feature vectors and the labels. It seems like blind regression is solving a harder problem by hiding the feature vectors.  In Section~\ref{sec:far-away-nn}, we make an attempt at porting blind regression ideas to standard regression, which results in a two-layer nearest neighbor algorithm that we believe warrants further investigation.

\subsection{Collaborative Filtering using Blind Regression}
\label{sec:matrix-est-cf}

Consider an $n \times m$ matrix $A = [A_{ui}] \in {\mathbb R}^{n\times m}$ that can be thought of as the true albeit unknown expected ratings in the recommendation context, where rows $u\in\{1,\dots,n\}$ index the users, and columns $i\in\{1,\dots,m\}$ index the items.  We observe an $n\times m$ matrix $Y = [Y_{ui}]$ where all entries are independent.  Specifically, we assume there is a probability $p\in(0,1]$ such that each entry of $Y$ is revealed with probability~$p$, independently across entries.  If $Y_{ui}$ is unobserved (with probability~$1-p$), then it takes on the special value ``$\star$''. Otherwise (with probability~$p$), it is a random variable where $\E[Y_{ui}] = A_{ui}$.\footnote{For our online collaborative filtering coverage (Section~\ref{chap:online-collaborative-filtering}), we assumed that the observed $Y_{ui}$'s could only take on one of two values $+1$ or $-1$, and the unobserved entries took on value $0$. In contrast, the general setup here allows for the observed $Y_{ui}$'s to, for example, take on any value on the real line (\eg, if the observed $Y_{ui}$'s are Gaussian). As such, we now introduce the special value ``$\star$'' for unobserved values.} We assume $Y_{ui}=A_{ui}+\varepsilon_{ui}$ where $\varepsilon_{ui}$'s are noise random variables.  The goal is to recover matrix~$A$ from observed matrix~$Y$.

We remark that this setup extends beyond recommendation systems.  For example, in the problem of graphon estimation for learning the structure of graphs (\cf, \citealt{Lovasz12}), the rows and columns both index the same collection of nodes in a graph (so the matrix is symmetric), and each revealed entry $Y_{ij}\in\{0,1\}$ indicates whether an edge is present between nodes~$i$ and~$j$. Different models of randomness for $Y$ are possible. For example, in an Erd\H{o}s-R\'{e}nyi model \citep{erdos_renyi_1959}, each possible edge is present with some fixed probability $q\in[0,1]$, so the matrix $A$ has off-diagonal entries that are all $q$.

A nonparametric view of the general problem setup of recovering matrix $A$ is as follows. For each row~$u$ and column~$i$ of~$A$, there are row and column feature vectors $x^{\row}_u \in \XX^\row$ and $x^{\col}_i \in \XX^\col$, where $\XX^\row$ and $\XX^\col$ are row and column feature spaces. We assume that there is a function $f: \XX^\row \times \XX^\col \to {\mathbb R}$ such that
\begin{align*}
A_{ui} &= f(x^{\row}_u, x^{\col}_i). \label{eq:beyond-nn-first-appearance-f}
\end{align*}
Thus, for any observed entry $Y_{ui}$,
\begin{align*}
\E[Y_{ui}] & = A_{ui} = f(x^{\row}_u, x^{\col}_i).
\end{align*}
A typical assumption would be that $f$ satisfies some smoothness condition such as being Lipschitz. This assumption is reasonable as it says that if its inputs change slightly (\eg, consider users $u$ and $v$ with very similar row feature vectors $x_u^{\row}$ and $x_v^{\row}$, or items $i$ and $j$ with very similar column feature vectors $x_i^{\col}$ and $x_j^{\col}$), then the expected rating~$f$ should not change much.  Importantly, the algorithms to be presented do not know $f$.

The nearest neighbor approach would suggest that to estimate $A_{ui}$, we first find nearest row and column neighbors. For instance, row~$u$'s nearest neighbors can be found by looking for other rows with similar row feature vectors as that of row~$u$:
\begin{align*}
\NN^\row_u & = \big\{v \in \{1,\dots,n\}: ~x^\row_{u} \approx x^\row_v \big\}.
\end{align*}
Similarly, column~$i$'s nearest neighbors can be found by looking for other columns with similar column feature vectors as that of column~$i$:
\begin{align*}
\NN^\col_i  & = \big\{j \in \{1,\dots,m\}: ~x^\col_{i} \approx x^\col_j \big\}.
\end{align*}
Then, we produce estimate $\widehat{A}=[\widehat{A}_{ui}]$ of matrix~$A$ where
\begin{align*}
\widehat{A}_{ui} & = \frac{\sum_{v \in \NN^\row_u, j \in \NN^\col_i } \ind\{Y_{vj} ~\text{observed}\} Y_{vj} }{\sum_{v \in \NN^\row_u, j \in \NN^\col_i } \ind\{Y_{vj} ~\text{observed}\}}\,.
\end{align*}
The challenge is that the row and column feature vectors are unknown (and thus latent). This is why the above model is known as a {\em latent variable model}~\citep{Chatterjee15}.\footnote{Latent variable models arise as canonical representations for two-dimensional exchangeable distributions. There is a very rich literature on this topic starting with work by~\citet{definetti} for one-dimensional exchangeable distributions, followed by the generalization to higher dimensions by \citet{Aldous81, Hoover82}. A survey is provided by \citet{Austin12}. Latent variable models have also been connected to the asymptotic representation of graphs \citep{lovaszszegedy}.} The problem of blind regression refers to this task of recovering the matrix $A$ without observing any of the latent feature vectors \citep{LLSS16}.

In terms of solving this problem using a nearest neighbor approach, the issue is how to find the nearest row and column neighbors $\NN^\row_u$ and $\NN^\col_i$ to produce the estimate $\widehat{A}_{ui}$.  Since we do not know the latent row feature vectors $x_u^{\row}$'s, the standard collaborative approach (such as what was done in Section~\ref{chap:online-collaborative-filtering}) just uses rows of the revealed ratings matrix~$Y$ instead, and compares them using, say, cosine distance to identify nearest row neighbors.  Similarly, since we do not know the latent column feature vectors $x_i^{\col}$'s, the standard approach is to compare columns of~$Y$ to identify nearest column neighbors. However, the quality of nearest neighbors we find is degraded not only by label noise for revealed entries in $Y$ but also by the fact that only a random subset of items is revealed per user.

Could exploiting some structure in~$f$ improve the quality of nearest neighbors we find?  The key observation is that what we really want is to find nearest row or column neighbors in the expected ratings matrix~$A$ rather than the revealed ratings matrix~$Y$ (if~$Y$ were completely noiseless and all entries of it were revealed, then it would equal~$A$).  Put another way, since $A_{ui} = f(x^{\row}_u, x^{\col}_i)$, what matters is finding nearest row or column neighbors with similar $f$ value rather than nearest row or column neighbors of the revealed ratings values $Y$, or even the nearest row or column latent feature vectors that we can't observe.  Hence, we want to somehow obtain the following sets of nearest neighbors for row~$u$ and column~$i$:
\begin{align*}
\NN^\row_u & = \big\{v \in \{1,\dots,n\}: \underbrace{f(x^\row_{u},\cdot)}_{u\text{-th row of }A} \approx \underbrace{f(x^\row_v,\cdot)}_{v\text{-th row of }A} \big\}, \\
\NN^\col_i  & = \big\{j \in \{1,\dots,m\}: \underbrace{f(\cdot,x^\col_{i})}_{i\text{-th column of }A} \approx \underbrace{f(\cdot,x^\col_j)}_{j\text{-th column of }A} \big\}.
\end{align*}
As we do not know $f$, some approximation is of course needed but we can do better than simply taking the raw values that we find in revealed ratings matrix~$Y$.

To exploit the structure of $f$, \citet{LLSS16} present two key twists on the standard collaborative filtering approach.  We state these twists in terms of user-user collaborative filtering (so we only find similar users and not similar items), although these ideas easily translate over to item-item collaborative filtering.  Suppose that user $u$ has not yet rated item $i$, meaning that $Y_{ui}=\star$. We aim to predict $A_{ui}$ as to figure out whether we should recommend item $i$ to user $u$. The two twists are as follows:
\begin{enumerate}

\item Let's say that user $v$ has rated item~$i$ and is ``similar'' to user $u$, and that both users have rated item $j\ne i$ (we comment on how to define ``similar'' momentarily). The standard approach would be to guess that perhaps $A_{ui}\approx Y_{vi}$ (we are only considering a single similar user $v$; of course if there are multiple similar users then we could average their revealed ratings for item $i$). However, Lee \textit{et al.}~instead show that we should use the estimate $A_{ui}\approx Y_{vi}+Y_{uj}-Y_{vj}$ based on a Taylor approximation of $f$. In other words, if $f$ is appropriately smooth, then we should add a correction term $Y_{uj}-Y_{vj}$.

\item We can actually bound how bad the new suggested estimate $Y_{vi}+Y_{uj}-Y_{vj}$ is at estimating $A_{ui}=f(x_u^{\row}, x_i^{\col})$ by looking at the variance of the difference in rows~$Y_{u}$ and~$Y_{v}$; details for precisely how to compute this variance is in equation (7) of \citet{LLSS16}. In particular, to estimate~$A_{ui}$, we can first identify which user $v\ne u$ has the lowest error bound in estimating $A_{ui}$ and declare this user to be the nearest neighbor of user $u$. Then we use the newly suggested estimate $Y_{vi}+Y_{uj}-Y_{vj}$ for $A_{ui}$.  Thus, the distance function used to identify nearest neighbors is now based on an estimation error bound for $f$ rather than a heuristic such as cosine distance.

\end{enumerate}
These changes to the standard collaborative filtering approach reduce error in estimating $A_{ui}$ by accounting for smoothness in $f$.  As for what smoothness structure is enough, \citet[Theorem~1]{LLSS16}~provide a theoretical guarantee on their algorithm's prediction accuracy for which the only assumption on $f$ is Lipschitz continuity.

To empirically validate their algorithm, Lee \textit{et al.}~predict movie ratings in the MovieLens \citep{movielens} and Netflix \citep{netflix} datasets.  Their algorithm outperforms standard collaborative filtering (that finds nearest neighbors based on comparing rows or columns of $Y$ using cosine distance), and also performs as well as iterative soft-thresholded SVD-based matrix completion~\citep{softimpute} that exploits low rank structure in $A$ (whereas Lee \textit{et al.}'s algorithm does not assume $A$ to be low rank).

By assuming that $f$ is Lipschitz and $A$ has low rank $\numClusters$, \citet{BCLS17} present a collaborative filtering method that again exploits structure in $f$ and, now, also the rank of $A$.  Their method looks at nearest neighbors of nearest neighbors and recurses. Two users can be declared as neighbors even if they have no commonly rated items!  Borgs \text{et al.}~provide analyses to explain when to stop the neighborhood expansion, and how to aggregate the now larger set of neighbors' revealed ratings appropriately to yield an algorithm with a theoretical guarantee on prediction accuracy (Theorems~4.1 and 4.2 in their paper; they have two theorems since they analyze their algorithm using two difference distance functions). Their algorithm is particularly well-suited to scenarios in which the number of revealed ratings per user is extremely small, where traditional collaborative filtering approaches may simply not find enough or any nearest neighbors as the users have extremely rare commonly rated items.  Specifically, when $A$ and $Y$ are both $n\times n$, and the probability of whether each entry in $Y$ is revealed satisfies $p=\omega(\numClusters^5 n)$, Borgs \textit{et al.}~can guarantee that their algorithm's mean squared prediction error goes to zero. In contrast, the theoretical guarantee for the algorithm by \citet{LLSS16} requires $p=\Omega(n^{3/2})$, \ie, dramatically more revealed ratings are needed when $n \gg \numClusters$.

\subsection{A Two-Layer Nearest Neighbor Algorithm}
\label{sec:far-away-nn}

The nearest neighbor blind regression algorithms suggest that potentially far away neighbors in the latent row or column feature spaces but that are close in $f$ value could be useful for prediction.  Could this same idea be applied to the original ``sighted'' regression setting of Section~\ref{sec:regression-setup}, where the feature vectors are directly seen?  In what follows, we provide an attempt at crafting a ``nearest'' neighbor regression algorithm in which nearest neighbors could be in distant parts of the feature space but are declared to be close by since they have similar regression function values.

Recall that in the standard regression setting, our goal is to estimate the regression function $\eta(x)=\mathbb{E}[Y\,|\,X=x]$ given training data $(X_1,Y_1),\dots(X_n,Y_n)\in\mathcal{X}\times\mathbb{R}$. Here, $\eta$ corresponds to function~$f$ in the previous section.  Then $k$-NN regression estimates $\widehat{\eta}_{k\text{-NN}}(x)$ as:
\[
\widehat{\eta}_{k\text{-NN}}(x)=\frac{1}{k}\sum_{i=1}^{k}Y_{(i)}(x),
\]
where $(X_{(i)}(x),Y_{(i)}(x))$ denotes the $i$-th closest training data pair to point $x$ among the training data, where the distance function used between feature vectors is $\rho$.

Note that for a point $x$, the labels of its $k$ nearest neighbors $(Y_{(1)}(x),Y_{(2)}(x),\dots,Y_{(k)}(x))$ altogether can be treated as a proxy for function~$\eta$ at~$x$, capturing some neighborhood information about~$\eta$ around~$x$ (the size of this neighborhood depends on the choice of~$k$). Thus, we construct the following $k$-dimensional proxy feature vector
\[
\eta^{\text{prx}}_{k}(x)\triangleq(Y_{(1)}(x),Y_{(2)}(x),\dots,Y_{(k)}(x)).
\]
Then instead of using feature vectors in the feature space~$\mathcal{X}$ to determine the nearest neighbors, we can instead use proxies for the $\eta$ values to determine the nearest neighbors. In particular, for a point~$x$, we find its $k'$ (which need not equal $k$) nearest training points by using their proxy $\eta$ vectors instead of their feature vectors, where we compare the proxy $\eta$ vectors with, say, Euclidean distance, \ie, the distance used between $x$ and training data point $X_i$ is
\[
\rho_{k}^{\text{prx}}(x,X_i) = \| \eta^{\text{prx}}_{k}(x) - \eta^{\text{prx}}_{k}(X_i) \|.
\]
Let's denote the training feature vectors of the $k'$ nearest neighbors found for $x$ using the above distance as $\mathcal{N}_{k\rightarrow k'}^{\text{prx}}(x)$. Then the resulting $k'$-NN regression estimate for $x$ using these proxy $\eta$ nearest neighbors is
\[
\widehat{\eta}_{k'\text{-NN}_{k}^{\text{prx}}}(x)
\triangleq \frac{1}{k'}\sum_{x' \in \NN^\proxy_{k\rightarrow k'}(x)}Y(x'),
\]
where by $Y(x')$ we mean the label associated with training feature vector $x'$.

This two-layer nearest neighbor algorithm has a few seemingly attractive properties. To start with, choosing a distance over proxy~$\eta$ vectors is straightforward since they consist of actual label values, so if, for example, the goal is to minimize squared label prediction error, then one should use Euclidean distance for proxy vectors. However, the choice of distance function $\rho$ specified for the feature space $\mathcal{X}$ remains more of an art.  Next, the method can potentially have better performance compared to standard $k$-NN regression because parts of the feature space $\mathcal{X}$ that lead to similar function values $\eta$ can now help each other with prediction. In particular, in searching for nearest neighbors, this algorithm should be able to automatically jump across periods in the example in Figure~\ref{fig:far-away-neighbors}. But to what extent can the algorithm achieve this for much more elaborate regression functions, and when does it bridge together parts of the feature space that should be left disconnected?  A proper understanding of when and why this algorithm may work better than standard $k$-NN regression remains an open question.

\newpage

\bibliographystyle{abbrvnat}
\bibliography{nn_survey}

\end{document}